\documentclass[10pt,twocolumn,letterpaper]{article}

\usepackage{iccv}
\usepackage{times}
\usepackage{epsfig}
\usepackage{graphicx}
\usepackage{amsmath}
\usepackage{amssymb}
\usepackage{stfloats}
\usepackage{epstopdf}
\usepackage{subfig}
\usepackage{float}
\usepackage{algorithm}
\usepackage{graphics}

\usepackage{graphbox}
\usepackage{setspace}
\usepackage{makecell}
\usepackage{array}
\usepackage{threeparttable}
\usepackage{multirow}
\usepackage{booktabs}
\usepackage{algorithm}
\usepackage{algpseudocode}
\usepackage{color}
\usepackage{authblk}
\usepackage[T1]{fontenc}
\usepackage[utf8]{inputenc}
\usepackage{fancyhdr}

\newcommand*\samethanks[1][\value{footnote}]{\footnotemark[#1]}
\makeatletter
\renewcommand\AB@affilsepx{\protect\Affilfont}

\makeatother
\usepackage{epstopdf}
\usepackage{algpseudocode}
\usepackage{float}
\usepackage{bbm} 
\usepackage{comment}
\usepackage[font=footnotesize,margin=2pt,skip=2pt]{caption}
\usepackage{bm}
\usepackage{diagbox}

\usepackage[pagebackref=true,breaklinks=true,colorlinks,bookmarks=false]{hyperref}
\def\iccvPaperID{3438}

\iccvfinalcopy

\begin{document}

\title{Dual Path Learning for Domain Adaptation of Semantic Segmentation\vspace{-0.5cm}}
\author[1]{Yiting Cheng}
\author[2]{\hspace{0.3cm}Fangyun Wei\thanks{Corresponding author.}}
\author[2]{\hspace{0.3cm}Jianmin Bao}
\author[2]{\hspace{0.3cm}Dong Chen}
\author[2]{\hspace{0.3cm}Fang Wen}
\author[1]{\hspace{0.3cm}Wenqiang Zhang\samethanks[1]\vspace{-0.3cm}}
\affil[1]{Fudan University \hspace{0.5cm} }
\affil[2]{Microsoft Research Asia
\authorcr
{\tt\small \{ytcheng18, wqzhang\}@fudan.edu.cn}~~ 
{\tt\small \{fawe, jianbao, doch, fangwen\}@microsoft.com}}
\maketitle

\begin{abstract}
 Domain adaptation for semantic segmentation enables to alleviate the need for large-scale pixel-wise annotations. Recently, self-supervised learning (SSL) with a combination of image-to-image translation shows great effectiveness in adaptive segmentation. The most common practice is to perform SSL along with image translation to well align a single domain (the source or target). However, in this single-domain paradigm, unavoidable visual inconsistency raised by image translation may affect subsequent learning. In this paper, based on the observation that domain adaptation frameworks performed in the source and target domain are almost complementary in terms of image translation and SSL, we propose a novel dual path learning (DPL) framework to alleviate visual inconsistency. Concretely, DPL contains two complementary and interactive single-domain adaptation pipelines aligned in source and target domain respectively. The inference of DPL is extremely simple, only one segmentation model in the target domain is employed. Novel technologies such as dual path image translation and dual path adaptive segmentation are proposed to make two paths promote each other in an interactive manner. Experiments on GTA5$\rightarrow$Cityscapes and SYNTHIA$\rightarrow$Cityscapes scenarios demonstrate the superiority of our DPL model over the state-of-the-art methods. The code and models are available at: \url{https://github.com/royee182/DPL}.
\end{abstract}

\section{Introduction \label{section:intro}}
In the past decades, significant progress \cite{zhao2017pspnet, chen2018deeplab,xiao2018unified,WangSCJDZLMTWLX19,YuanCW20,tao2020hierarchical,mohan2020efficientps} in semantic segmentation has been achieved with Deep Convolutional Neural Networks. The empirical observation \cite{raffel2019exploring, xie2019self} demonstrates that the leading performance is partially attributed to a large volume of training data, thus dense pixel-level annotations are required in supervised learning, which is laborious and time-consuming. To avoid this painstaking task, researchers resort to train segmentation models on synthetic but photo-realistic large-scale datasets such as GTA5~\cite{richter2016playing} and SYNTHIA~\cite{ros2016synthia} with computer-generated annotations. However, due to the cross-domain differences, these well-trained models usually undergo significant performance drops when tested on realistic datasets (e.g., Cityscapes~\cite{cordts2016cityscapes}). Therefore, unsupervised domain adaptation (UDA) methods have been widely adopted to align the domain shift between the rich-labeled source data (synthetic images) and the unlabeled target data (real images). 
\begin{figure}[t]
    \centering 
    \subfloat[Illustration of adaptation in domain-$\mathcal{T}$.]{\includegraphics[align=c,width=3.1in]{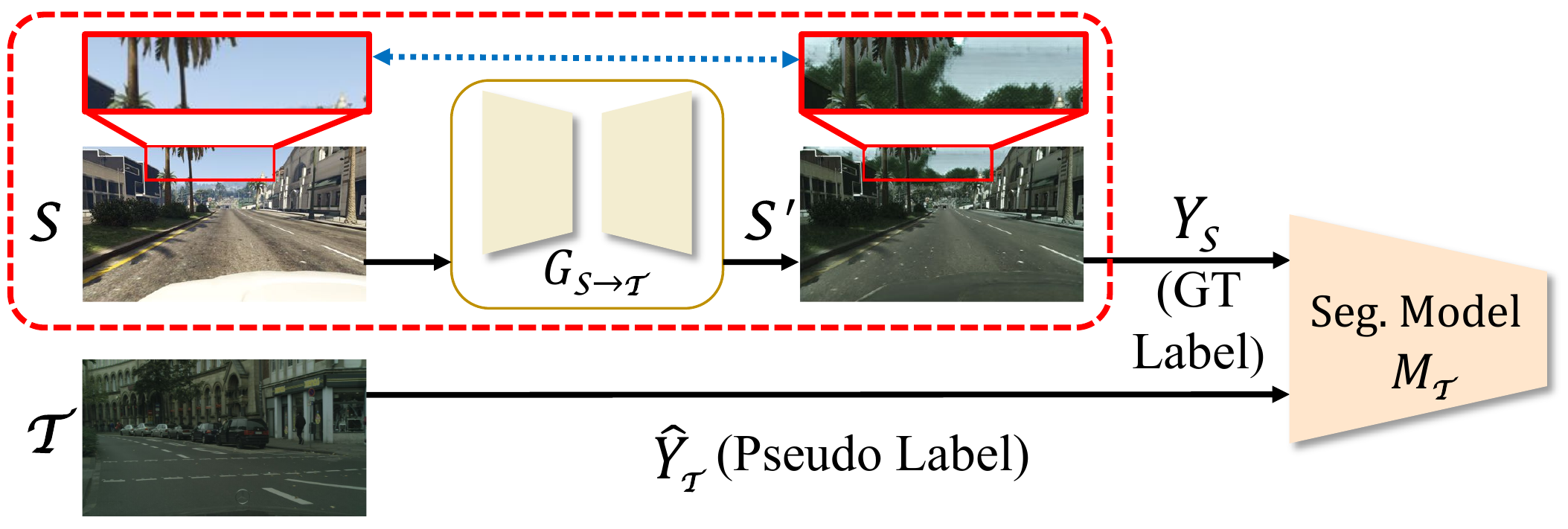}}
    \vspace{-0.1cm}
  \subfloat[Illustration of adaptation in domain-$\mathcal{S}$.]{\includegraphics[align=c,width=3.1in]{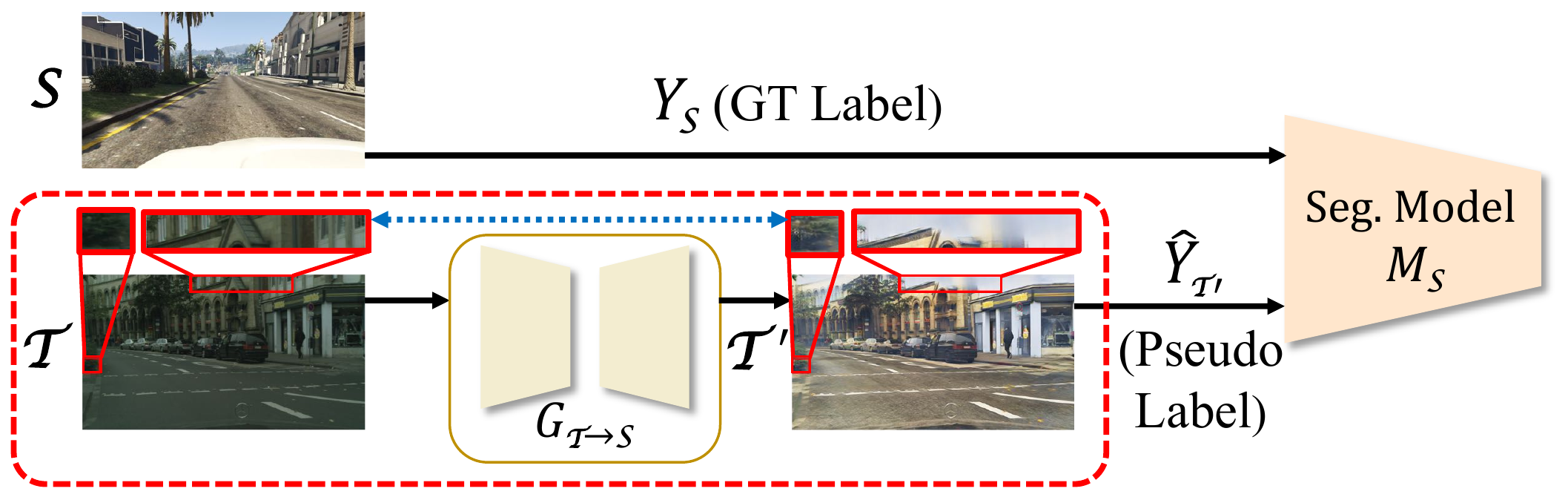}}
  
\caption{Illustration of single-domain adaptation pipelines. $\mathcal{S}$ is source image with ground-truth label $Y_{\mathcal{S}}$, and $\mathcal{T}$ is target image. $G_{\mathcal{S}\rightarrow\mathcal{T}}$ represents image translation from domain-$\mathcal{S}$ to domain-$\mathcal{T}$ and vice versa. $\mathcal{S}' = G_{\mathcal{S}\rightarrow\mathcal{T}}(\mathcal{S})$ and $\mathcal{T}' = G_{\mathcal{T}\rightarrow\mathcal{S}}(\mathcal{T})$ are translated images in the corresponding domain. $M_{\mathcal{S}}$ and $M_{\mathcal{T}}$ are semantic segmentation models in domain-${\mathcal{S}}$ and domain-${\mathcal{T}}$, respectively. $\hat{Y}_{\mathcal{T}}$ and $\hat{Y}_{\mathcal{T}'}$ represent the corresponding pseudo labels of $\mathcal{T}$ and $\mathcal{T}'$. Red dash rectangles denote that visual inconsistency raised by image translations disturbs domain adaptation learning in either supervised part or SSL part.}
    \label{fig:flowchat}
    \vspace{-0.7cm}
\end{figure}

Two commonly used paradigms in unsupervised domain adaptive segmentation are image-to-image translation based methods~\cite{murez2018image, hoffman2018cycada} and self-supervised learning (SSL) based methods~\cite{zou2018unsupervised, zou2019confidence, zhang2019category,Two-phase}. The most common practice for image-to-image translation based methods is to translate synthetic data from source domain (denote as domain-$\mathcal{S}$) to target domain (denote as domain-$\mathcal{T}$)~\cite{hoffman2018cycada,chang2019all} to reduce the visual gap between different domains. Then adaptive segmentation is trained on translated synthetic data. However, by only applying the image-to-image translation to domain adaptation task, the results are always unsatisfying. One of the leading factors is that image-to-image translation may change the image content involuntarily and introduce \emph{visual inconsistency} between raw images and translated images. Training on translated images with uncorrected ground-truth labels of source images introduces noise which disturbs the domain adaptation learning.

A combination of SSL and image-to-image translation~\cite{li2019bidirectional, yang2020label, kim2020learning} has been demonstrated great effectiveness in the UDA field. SSL utilizes a well-trained segmentation model to generate a set of pseudo labels with high confidence for unlabeled target data, then the adaptive segmentation training can be divided into two parallel parts, namely supervised part (training is performed on source data with ground-truth labels) and SSL part (training is performed on target data with pseudo labels). In this paradigm, the most prevalent practice is to perform adaptation to well align a single domain, i.e., either source domain (named domain-$\mathcal{S}$ adaptation)~\cite{li2019bidirectional, kim2020learning} or target domain (named domain-$\mathcal{T}$ adaptation)~\cite{yang2020label}.
However, both domain-$\mathcal{S}$ and domain-$\mathcal{T}$ adaptation heavily rely on the quality of image-to-image translation models, where visual inconsistency is always unavoidable. For domain-$\mathcal{T}$ adaptation (as shown in Figure~\ref{fig:flowchat}.(a)), visual inconsistency brings in misalignment between translated source images and uncorrected ground-truth labels, which disturbs the supervised part. In contrast, domain-$\mathcal{S}$ adaptation (as shown in Figure~\ref{fig:flowchat}.(b)) avoids image translation on source images, but simultaneously introduces visual inconsistency between target images and the corresponding translated images. Defective pseudo labels generated by unaligned images disturb the SSL part.

Notice the above single-domain adaptation pipelines are almost complementary in terms of the two training parts, i.e., visual inconsistency caused by image translation disturbs the training of supervised part in domain-$\mathcal{T}$ adaptation and SSL part in domain-$\mathcal{S}$ adaptation. In contrast, SSL part in domain-$\mathcal{T}$ adaptation and supervised part in domain-$\mathcal{S}$ adaptation are unaffected. It is natural to raise a question: \emph{could we combine these two complementary adaptation pipelines into a single framework to make good use of each strength and make them promote each other?} Based on this idea, we propose the \emph{dual path learning} framework which considers two pipelines from opposite domains to alleviate unavoidable visual inconsistency raised by image translations. We name two paths used in our framework as path-$\mathcal{T}$ (adaption is performed in domain-$\mathcal{T}$) and path-$\mathcal{S}$ (adaption is performed in domain-$\mathcal{S}$), respectively. Path-$\mathcal{S}$ assists path-$\mathcal{T}$ to learn precise supervision from source data. Meanwhile, path-$\mathcal{T}$ guides path-$\mathcal{S}$ to generate high-quality pseudo labels which are important for SSL in return. It is worth noting that path-$\mathcal{S}$ and path-$\mathcal{T}$ are not two separated pipelines in our framework, interactions between two paths are performed throughout the training, which is demonstrated to be effective in our experiments. The whole system forms a closed-loop learning. Once the training has finished, we only retain a single segmentation model well aligned in target domain for testing, no extra computation is required. The main contributions of this work are summarized as:
\begin{itemize}
\item We present a novel dual path learning (DPL) framework for domain adaptation of semantic segmentation. DPL employs two complementary and interactive single-domain pipelines (namely path-$\mathcal{T}$ and path-$\mathcal{S}$) in the training phase. In the testing, only a single segmentation model well aligned in target domain is used. The proposed DPL framework surpasses state-of-the-art methods on representative scenarios.
\item We present two interactive modules to make two paths promote each other, namely dual path image translation and dual path adaptive segmentation.
\item We introduce a novel warm-up strategy for the segmentation models which helps adaptive segmentation in the early training stage.
\end{itemize}

\section{Related Work}
{\noindent \textbf{Domain Adaptation.}}\hspace{3pt}
Domain adaptation is a broadly studied topic in computer vision. It aims to rectify the mismatch in cross-domains and tune the models toward better generalization at testing~\cite{patel2015visual}. A variety of domain adaptation methods for image classification~\cite{saito2017maximum,Chen_2019_CVPR,tzeng2017adversarial,kang2019contrastive} and object detection~\cite{chen2018domain,bhattacharjee2020dunit} have been proposed. In this paper, we focus on the unsupervised domain adaptation of semantic segmentation.

\begin{figure*}[t]
    \centering 
    \subfloat[Training pipeline of DPL.]{\includegraphics[align=c,width=0.85\linewidth]{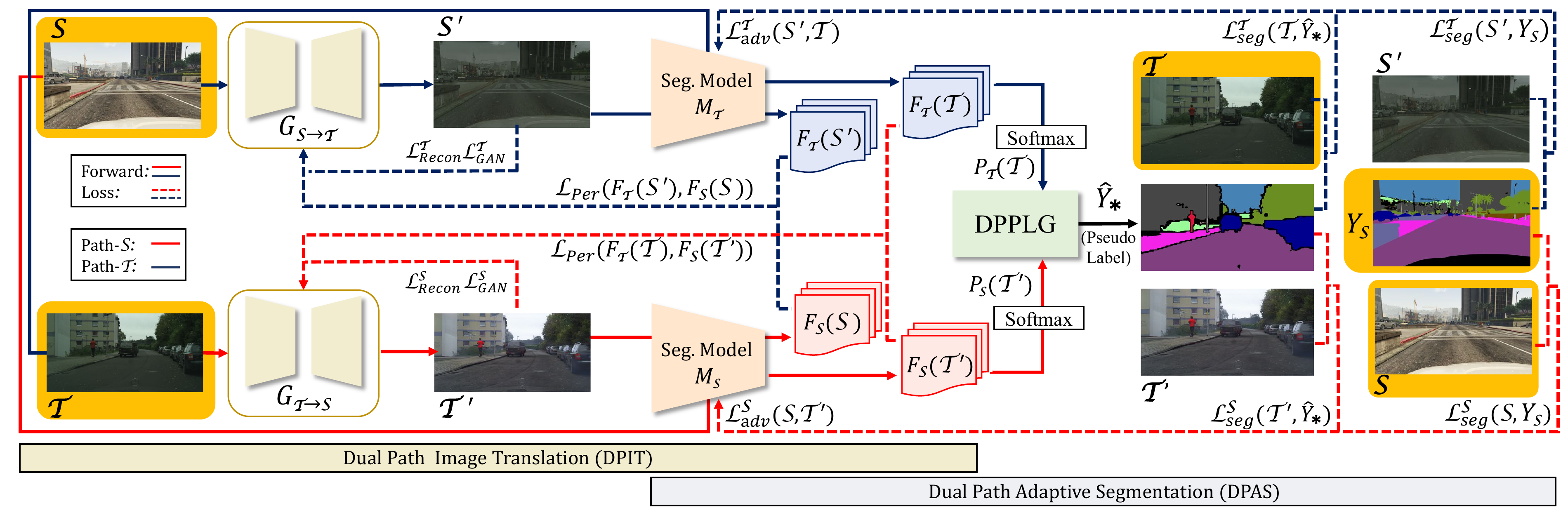}}
    \hspace{0.2cm}
  \subfloat[Testing.]{\includegraphics[align=c,width=0.1\linewidth]{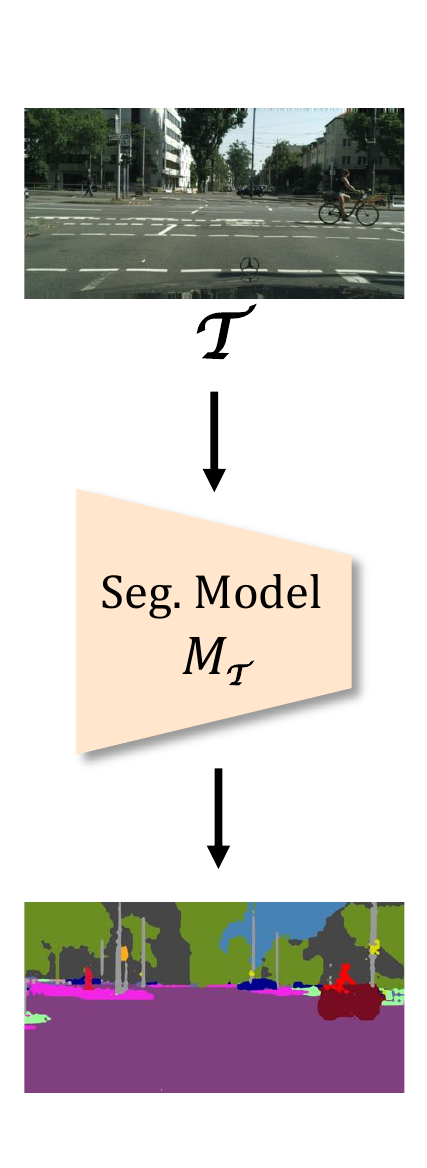} }
\caption{(a) Overview of DPL framework. Inputs are highlighted by orange rectangles. DPL consists of two complementary single-domain paths: path-$\mathcal{S}$ (learning is performed in \emph{source} domain) and path-$\mathcal{T}$ (learning is performed in \emph{target} domain). Dual path image translation (DPIT) and dual path adaptive segmentation (DPAS) are proposed to make two paths interactive and promote each other. In DPIT, unpaired image translation models ($G_{\mathcal{T}\rightarrow\mathcal{S}}$ and $G_{\mathcal{S}\rightarrow\mathcal{T}}$) are supervised by general GAN loss and cross-domain perceptual loss. DPAS employs the proposed dual path pseudo label generation (DPPLG) module to produce pseudo labels $\hat{Y}_{*}$ of target images, then segmentation models ($M_{\mathcal{S}}$ and $M_{\mathcal{T}}$) are trained on both source images (or translated source images) with ground-truth labels and target images (or translated target images) with pseudo labels. (b) Testing of DPL. Only $M_{\mathcal{T}}$ is used for inference.}
    \label{fig:DPL_flowchat}
    \vspace{-0.5cm}
\end{figure*}

{\noindent \textbf{Domain Adaptation for Semantic Segmentation.}}\hspace{3pt}
Semantic segmentation needs a large volume of pixel-level labeled training data, which is laborious and time-consuming in annotation. A promising solution to reduce the labeling cost is to train segmentation networks on synthetic dataset (e.g., GTA5~\cite{richter2016playing} and SYNTHIA~\cite{ros2016synthia}) with computer-generated annotations before testing on realistic dataset (e.g., Cityscapes~\cite{cordts2016cityscapes}). Although synthetic images have similar appearance to real images, there still exist domain discrepancies in terms of layouts, colors and illumination conditions, which always cripples the models' performance. Domain adaptation is necessary to align the synthetic and the real dataset~\cite{wu2018dcan,zou2018unsupervised,zhao2019madan,kang2020pixel}. 

Adversarial-based methods~\cite{hoffman2016fcns, long2018transferable,tsai2018learning} are broadly explored in unsupervised domain adaptation, which align different domains at image-level~\cite{murez2018image,hoffman2018cycada,wu2018dcan} or feature-level~\cite{tsai2018learning,huang2020contextual}. The image-level adaptation regards domain adaptation as an image synthesis problem, and aims to reduce visual discrepancy (e.g., lighting and object texture) in cross-domains with unpaired image-to-image translation models~\cite{CycleGAN2017,liu2017unsupervised,park2020contrastive}. However, the performance is always unsatisfactory by simply applying image translation to domain adaptation task. One reason is that image-to-image translation may change the image content involuntarily and further disturb the following segmentation training~\cite{li2019bidirectional}.

In recent years, self-supervised learning (SSL)~\cite{grandvalet2005semi,zhu2007semi} shows tremendous potential in adaptive segmentation~\cite{zou2018unsupervised,zou2019confidence,subhani2020learning,Two-phase}. The key principle for these methods is to generate a set of pseudo labels for target images as the approximation to the ground-truth labels, then segmentation model is updated by leveraging target domain data with pseudo labels. CRST~\cite{zou2018unsupervised} is the first work to introduce self-training into adaptive segmentation, it also alleviates category imbalance issue by controlling the proportion of selected pseudo labels in each category. Recent TPLD~\cite{Two-phase} proposes a two-phase pseudo label densification strategy to obtain dense pseudo labels for SSL. 

Two works~\cite{li2019bidirectional, yang2020label} which explore the combination of image translation and SSL are closely related to ours. Label-Driven~\cite{yang2020label} performs a target-to-source translation and a label-driven reconstruction module is used to reconstruct source and target images from the corresponding predicted labels. In contrast, BDL~\cite{li2019bidirectional} represents a bidirectional learning framework which alternately trains the image translation and the adaptive segmentation in target domain. Meanwhile, BDL utilizes a single-domain perceptual loss to maintain visual consistency. We will demonstrate this kind of design is suboptimal compared with the proposed dual path image translation module in Section~\ref{section:DPIT}. These two works demonstrate the combination of image translation and SSL can promote adaptive learning. Different from these single-domain adaptation methods, the proposed dual path learning framework integrates two complementary single-domain pipelines in an interactive manner to address visual inconsistency problem by: 1) utilizing segmentation models aligned in different domains to provide cross-domain perceptual supervision for image translation; 2) combining knowledge from both source and target domain for self-supervised learning.

\section{Method}
Given the source dataset $\mathcal{S}$ (synthetic data) with pixel-level segmentation labels $Y_\mathcal{S}$, and the target dataset $\mathcal{T}$ (real data) with no labels. The goal of unsupervised domain adaptation (UDA) is that by only using $\mathcal{S}$, $Y_\mathcal{S}$ and $\mathcal{T}$, the segmentation performance can be on par with the model trained on $\mathcal{T}$ with corresponding ground-truth labels $Y_\mathcal{T}$. Domain gap between $\mathcal{S}$ and $\mathcal{T}$ makes the task difficult for the network to learn transferable knowledge at once.

To address this problem, we propose a novel dual path learning framework named DPL. As shown in Figure~\ref{fig:DPL_flowchat}.(a), DPL consists of two complementary and interactive paths: path-$\mathcal{S}$ (adaptive learning is performed in \emph{source} domain) and path-$\mathcal{T}$ (adaptive learning is performed in \emph{target} domain). How to allow one of both paths provide positive feedbacks to the other is the key to success. To achieve this goal, we propose two modules, namely dual path image translation (DPIT) and dual path adaptive segmentation (DPAS). DPIT aims to reduce the visual gap between different domains without introducing visual inconsistency. In our design, DPIT unites general unpaired image translation models with dual perceptual supervision from two single-domain segmentation models. Notice any unpaired image translation models can be used in DPIT, we use CycleGAN~\cite{CycleGAN2017} as our default model due to its popularity and it provides bidirectional image translation inherently. We use
$\mathcal{T}'=G_{\mathcal{T}\rightarrow\mathcal{S}}(\mathcal{T})$ and $\mathcal{S}'=G_{\mathcal{S}\rightarrow\mathcal{T}}(\mathcal{S})$ to denote translated images in path-$\mathcal{S}$ and path-$\mathcal{T}$ respectively, where $G_{\mathcal{T}\rightarrow\mathcal{S}}$ and $G_{\mathcal{S}\rightarrow\mathcal{T}}$ are image translation models in the corresponding path. DPAS utilizes translated images from DPIT and the proposed dual path pseudo label generation (DPPLG) module to generate high-quality pseudo labels for target images, then segmentation models $M_{\mathcal{S}}$ (in path-$\mathcal{S}$) and $M_{\mathcal{T}}$ (in path-$\mathcal{T}$) are trained with both transferred knowledge in source domain and implicit supervision in target domain. The testing of DPL is extremely simple, we only retain $M_{\mathcal{T}}$ for inference as shown in Figure~\ref{fig:DPL_flowchat}.(b).

The training process of DPL consists of two phases: single-path warm-up and DPL training. DPL benefits from well-initialized $M_{\mathcal{S}}$ and $M_{\mathcal{T}}$, since both DPIT and DPAS rely on the quality of segmentation models. A simple but efficient warm-up strategy can accelerate the convergence of DPL. Once the warm-up finishes, DPIT and DPAS are trained sequentially in DPL training phase. 

In this section, we first describe our warm-up strategy in Section~\ref{section:warm_phase}. Then, we introduce the key components of DPL: DPIT in Section~\ref{section:DPIT} and DPAS in Section~\ref{section:DPSA}. Next, we revisit and summarize the whole training process in Section~\ref{section:trainpipeline}. Finally, testing pipeline of DPL is presented in Section~\ref{section:testpipeline}.

\subsection{Single Path Warm-up\label{section:warm_phase}}
Perceptual supervision in DPIT and pseudo label generation in DPAS rely on the quality of segmentation models. To accelerate convergence of DPL, a warm-up process for segmentation models $M_{\mathcal{S}}$ and $M_{\mathcal{T}}$ is required. 

\noindent\textbf{$\boldsymbol{M_{\mathcal{S}}}$ Warm-up.} 
The warm-up for ${M_{\mathcal{S}}}$ is easily conducted in a fully supervised way by using source dataset $\mathcal{S}$ with ground-truth labels $Y_\mathcal{S}$. 

\noindent\textbf{$\boldsymbol{M_{\mathcal{T}}}$ Warm-up.}
It is difficult to directly train $M_{\mathcal{T}}$ in a supervised manner since no labels can be accessed in target dataset $\mathcal{T}$. A straightforward idea is to translate source images $\mathcal{S}$ to target domain by using naive CycleGAN, and then $M_{\mathcal{T}}$ is trained on translated images $\mathcal{S}'$ with approximate ground-truth labels $Y_{\mathcal{S}}$. Unfortunately, naive CycleGAN does not apply any constrains to preserve visual consistency between $\mathcal{S}$ and $\mathcal{S'}$, i.e., visual content may be changed when $\mathcal{S}$ is translated to $\mathcal{S'}$. Misalignment between $\mathcal{S'}$ and $Y_{\mathcal{S}}$ can disturb the training of $M_{\mathcal{T}}$.

\begin{figure}[t]
	
    \centering 
  \includegraphics[width=2.2in]{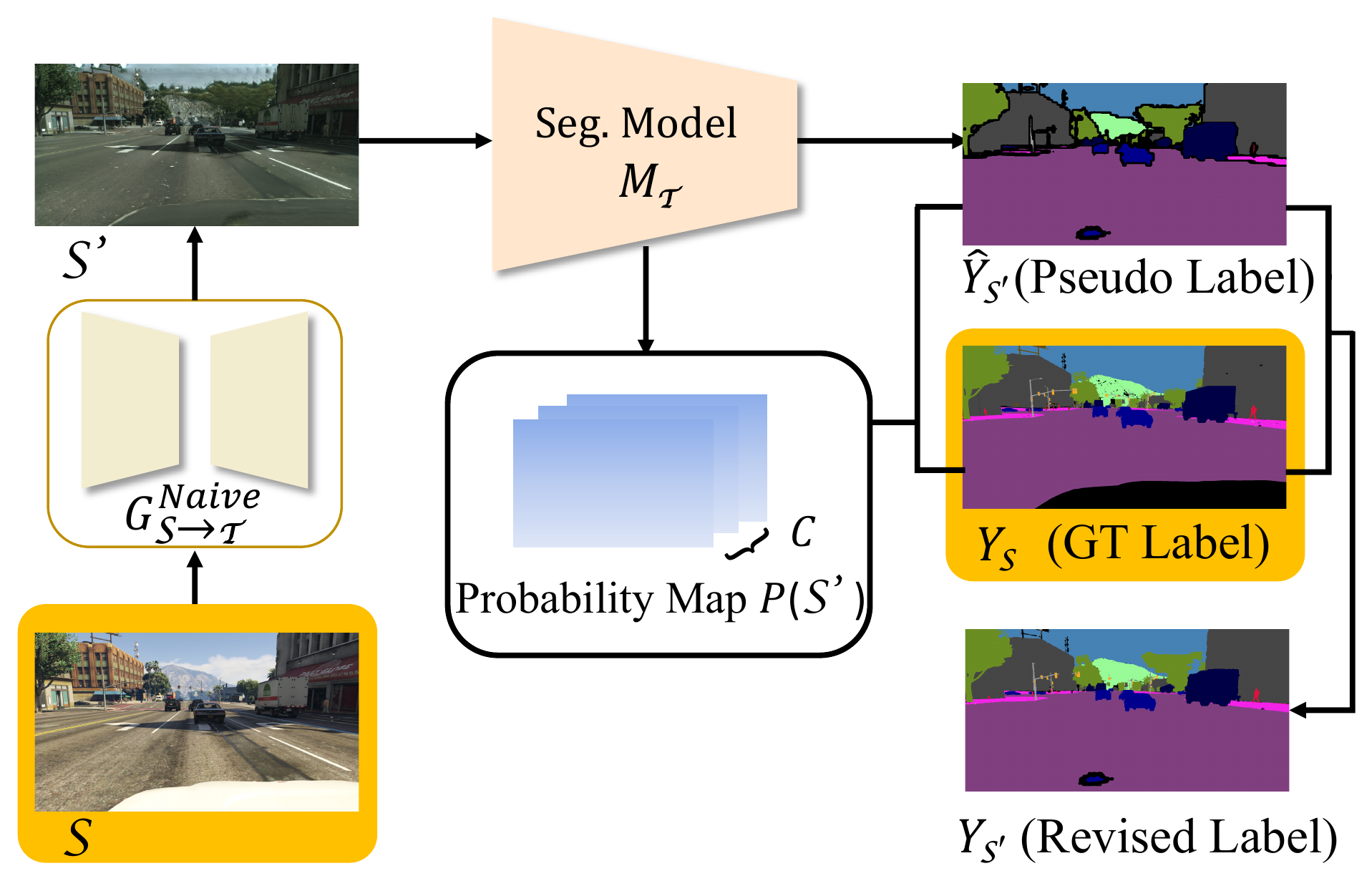}
  
    \caption{Illustration of label correction strategy. Inputs are highlighted by orange rectangles.}

 \label{fig:initialazion_T}
 \setlength{\belowcaptionskip}{-2cm}
 \vspace{-0.7cm}
\end{figure}

To address this issue, we propose a novel label correction strategy as shown in Figure~\ref{fig:initialazion_T}. The core principle is to find a revised label $Y_{\mathcal{S}'}$ for ${\mathcal{S}'}$ by considering both ground-truth labels $Y_{\mathcal{S}}$ and segmentation predictions of ${\mathcal{S}'}$. Specially, we feed $\mathcal{S'}$ into $M_{\mathcal{T}}$ (which is initialized as $M_{\mathcal{S}}$ at the beginning) to generate pseudo labels $\hat{Y}_{\mathcal{S}'}$. Then label correction module revises raw ground-truth labels $Y_{\mathcal{S}}$ by replacing pixel-wise labels in $Y_{\mathcal{S}}$ with high-confidence pixel-wise labels in $\hat{Y}_{\mathcal{S}'}$, which means the labels of content-changed areas have been approximately corrected by reliable predictions. Formally, define revised labels $Y_{\mathcal{S}'} = \{ Y_\mathcal{S'}^{(i,j)}\}~(1 \leq i \leq H, 1 \leq j \leq W)$ as:
\begin{equation}
\label{equ:correction}
 Y_\mathcal{S'}^{(i,j)}=\left\{
\begin{array}{ll}
\hat{Y}^{(i,j)}_{\mathcal{S}'}, & \mbox{if} \ P^{(i,j,\hat{c})}(\mathcal{S}')  - P^{(i,j,c)}(\mathcal{S}') > \delta\\
Y^{(i,j)}_\mathcal{S},&  \mbox{else,}
\end{array}
\right.
\end{equation}
where $H$ and $W$ denote the height and width of the input image respectively, $P(\cdot)$ is probability map predicted by segmentation model, $\hat{c}$ and $c$ denote the category index of $\hat{Y}^{(i,j)}_{\mathcal{S}'}$ and $Y^{(i,j)}_{\mathcal{S}}$ respectively, $\delta$ controls correction rate, we set $\delta=0.3$ empirically.

In addition, we also use $M_{\mathcal{T}}$ to generate pseudo labels $\hat{Y}_{\mathcal{T}}$ for $\mathcal{T}$. Now we have paired training data $(\mathcal{S}', Y_{\mathcal{S}'})$ and $(\mathcal{T}, \hat{Y}_{\mathcal{T}})$ which approximately lie in target domain for $M_{\mathcal{T}}$ training. The overall loss is defined as:
\begin{equation}
 \label{loss:init_seg_t}
 \begin{aligned}
 \mathcal{L}_{M_{\mathcal{T}}} &=\mathcal{L}_{seg}(\mathcal{S}', Y_\mathcal{S'}) + \mathcal{L}_{seg}(\mathcal{T}, \hat{Y}_{\mathcal{T}})\\
 &+\lambda_{adv} \mathcal{L}_{adv}(\mathcal{S'},\mathcal{T}),
 \end{aligned}
 \end{equation}
where $\mathcal{L}_{adv}$ represents typical adversarial loss as used in~\cite{tsai2018learning,li2019bidirectional,yang2020label} to further align target domain, $\mathcal{L}_{seg}$ indicates the commonly used per-pixel segmentation loss: 
\begin{equation}
  \label{loss:seg}
\mathcal{L}_{seg}(I,Y)=-\frac{1}{HW}\sum_{i=1}^{H}\sum_{j=1}^{W}\sum_{c=1}^{C}Y^{(i,j,c)}\log P^{(i,j,c)}(I),
 \end{equation}
 where $I$ and $Y$ denote input image (raw image or translated image) and corresponding labels (ground-truth labels or pseudo labels), respectively.
 
Once warm-up procedure is finished, we obtain preliminary segmentation models which are approximately aligned in the corresponding domain. These well-initialized models facilitate the training of DPIT and DPAS, which will be described in next sections.

\subsection{Dual Path Image Translation\label{section:DPIT}}
  Image-to-image translation aims to reduce the gap in visual appearance (e.g., object textures and lighting) between source and target domain. As discussed in Section~\ref{section:intro}, unavoidable visual inconsistency caused by image translation may mislead the subsequent adaptive segmentation learning, and thus extra constraints to maintain visual consistency are required. 
  
  BDL~\cite{li2019bidirectional} introduces a perceptual loss to maintain visual consistency between paired images (i.e., raw images and corresponding translated images). Perceptual loss measures distance of perceptual features\footnote{Perceptual feature denotes the probability map before softmax layer of segmentation model.} extracted from a well-trained segmentation model. In BDL, domain adaptation is only performed in target domain, as a result, the perceptual loss of paired images ($\mathcal{S}$, $\mathcal{S}'$) and ($\mathcal{T}$, $\mathcal{T}'$) is computed with identical segmentation model. Notice paired images are from two different domains ($\mathcal{S}$ and $\mathcal{T}'$ are in source domain while $\mathcal{T}$ and $\mathcal{S}'$ are in target domain), using segmentation model aligned in a single domain to extract features for perceptual loss computation may be suboptimal. 
  
 \begin{figure}[t]
  \centering
  \includegraphics[width=\linewidth]{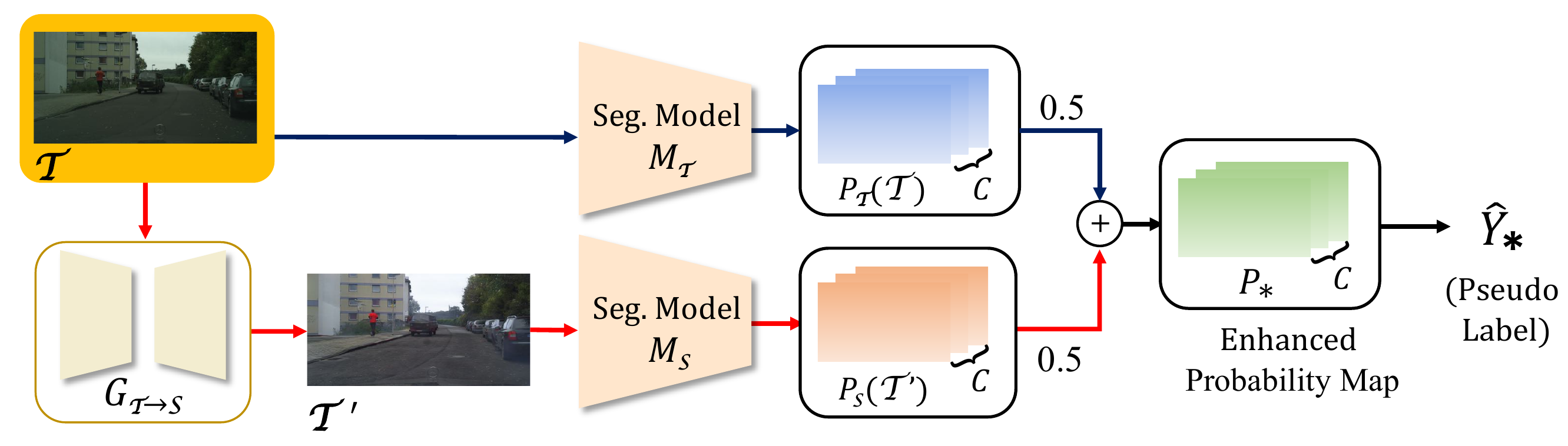}
\caption{Illustration of dual path pseudo label generation (DPPLG). Input is highlighted by orange rectangle.}
  \label{fig:DPPLG}
  \vspace{-4mm}
\end{figure}

Now we introduce our dual path image translation (DPIT) as illustrated in Figure~\ref{fig:DPL_flowchat}.(a). DPIT is an bidirectional image translation model with cross-domain perceptual supervision. We use ${G_{\mathcal{S}\rightarrow\mathcal{T}}}$ and ${G_{\mathcal{T}\rightarrow\mathcal{S}}}$ to denote image translation in Path-$\mathcal{T}$ and Path-$\mathcal{S}$ respectively. CycleGAN is served as our default model since it provides bidirectional image translation inherently, however, any unpaired image translation algorithms can be used in DPIT. Different from BDL, DPIT makes use of two paths aligned in opposite domains and extracts perceptual features for paired images from their corresponding path to better maintain visual consistency. Concretely, DPIT utilizes $M_{\mathcal{S}}$ to extract perceptual features for $\mathcal{S}$ and $\mathcal{T}'$, and $M_{\mathcal{T}}$ to extract perceptual features for $\mathcal{T}$ and $\mathcal{S}'$, respectively. Then we can formulate our dual perceptual loss $\mathcal{L}_{DualPer}$ as:
\begin{equation}
  \label{loss:tranlation_loss}
  \begin{aligned}
 \mathcal{L}_{DualPer}(\mathcal{S}, \mathcal{S}', \mathcal{T}, \mathcal{T}')&=  \mathcal{L}_{Per}(F_{\mathcal{T}}(\mathcal{S}'), F_{\mathcal{S}}(\mathcal{S}))\\
 &+\mathcal{L}_{Per}(F_{\mathcal{T}}(\mathcal{T}), F_{\mathcal{S}}(\mathcal{T}')),
  \end{aligned}
  \vspace{-0.1cm}
 \end{equation}
 where $\mathcal{L}_{Per}$ is perceptual loss as in ~\cite{li2019bidirectional}, $F_{\mathcal{S}}(\cdot)$ and $F_{\mathcal{T}}(\cdot)$ represent perceptual feature extracted by $M_{\mathcal{S}}$ and $M_{\mathcal{T}}$ respectively. 

Besides the supervision of dual perceptual loss, DPIT is also supervised by general adversarial and reconstruction loss. The overall loss of DPIT can be formulated as:
\begin{equation}
\label{loss:dualGAN}
\begin{aligned}
\mathcal{L}_{DPIT}&=  \mathcal{L}^{\mathcal{S}}_{GAN}(\mathcal{S},\mathcal{T}') + \mathcal{L}^{\mathcal{T}}_{GAN}( \mathcal{S}',\mathcal{T})\\
 &+\lambda_{Recon}\mathcal{L}^{\mathcal{S}}_{Recon}(\mathcal{S}, G_{\mathcal{T}\rightarrow\mathcal{S}}(\mathcal{S}'))\\
 &+\lambda_{Recon}\mathcal{L}_{Recon}^{\mathcal{T}}(\mathcal{T}, G_{\mathcal{S}\rightarrow\mathcal{T}}(\mathcal{T}'))\\
 &+{\lambda}_{DualPer} \mathcal{L}_{DualPer}(\mathcal{S}, \mathcal{S}', \mathcal{T}, \mathcal{T}'),
 \end{aligned}
 \vspace{-0.1cm}
\end{equation}
where $\mathcal{L}^{\mathcal{S}}_{GAN}$ ($\mathcal{L}^{\mathcal{T}}_{GAN}$) and $\mathcal{L}^{\mathcal{S}}_{Recon}$ ($\mathcal{L}^{\mathcal{T}}_{Recon}$) are GAN loss and reconstruction loss as in~\cite{CycleGAN2017}, $\lambda_{Recon}$ and ${\lambda}_{DualPer}$ denote the weights of reconstruction loss and dual perceptual loss respectively. We set $\lambda_{Recon}=10$ and $\lambda_{DualPer}=0.1$ by default.

\subsection{Dual Path Adaptive Segmentation \label{section:DPSA}} 

 Once DPIT is symmetrically trained, translated images $\mathcal{S}'= {G_{\mathcal{S}\rightarrow\mathcal{T}}}(\mathcal{S})$ and $\mathcal{T}'= {G_{\mathcal{T}\rightarrow\mathcal{S}}}(\mathcal{T})$ are fed into dual path adaptive segmentation (DPAS) module for subsequent learning. As shown in figure~\ref{fig:DPL_flowchat}.(a), DPAS utilizes self-supervised learning with combination of well-trained image translation for adaptive segmentation learning, i.e., segmentation models are trained on both source images (or translated source images) with ground-truth labels and target images (or translated target images) with pseudo labels. The core of DPAS is to generate high-quality pseudo labels of target images by combining predicted results from two paths. The training process of DPAS can be formulated as two alternative steps: 1) dual path pseudo label generation; 2) dual path segmentation training.

{\noindent \textbf{Dual Path Pseudo Label Generation.}}\hspace{3pt}
The labels of target dataset are unavailable in unsupervised domain adaptation tasks. Self-supervised learning (SSL) has been demonstrated great success when the labels of dataset are insufficient or noisy. The way to generate pseudo labels plays an important role in SSL. As described in Section~\ref{section:intro}, in path-$\mathcal{T}$, visual inconsistency brings in misalignment between translated source images $\mathcal{S}'$ and uncorrected ground-truth labels $Y_\mathcal{S}$, which disturbs the training of $M_\mathcal{T}$. Similar issue exists in  path-$\mathcal{S}$ (see Figure~\ref{fig:flowchat}). Inspired by the observation that two paths from opposite domains are almost complementary, we take full advantages of two paths and present a novel dual path pseudo label generation (DPPLG) strategy to generate high-quality pseudo labels as shown in Figure~\ref{fig:DPPLG}.

Concretely, let $P_{\mathcal{S}}(\cdot) = \mbox{Softmax}(F_{\mathcal{S}}(\cdot))$ and $P_{\mathcal{T}}(\cdot) = \mbox{Softmax}(F_{\mathcal{T}}(\cdot))$ denote probability map predicted by $M_\mathcal{S}$ and $M_\mathcal{T}$, respectively. In path-$\mathcal{T}$, target images can be directly fed into $M_\mathcal{T}$ to generate $P_{\mathcal{T}}(\mathcal{T})$. In contrast, path-$\mathcal{S}$ requires image translation to generate $\mathcal{T}'= {G_{\mathcal{T}\rightarrow\mathcal{S}}}(\mathcal{T})$, then $P_{\mathcal{S}}(\mathcal{T}')$ can be obtained by feeding $\mathcal{T}'$ into $M_\mathcal{S}$. Finally, enhanced probability map $P_*$ which is used for generating pseudo labels of target images can be obtained by a weighted sum of two separate probability maps $P_{\mathcal{T}}(\mathcal{T})$ and $P_{\mathcal{S}}(\mathcal{T}')$:
\vspace{-0.2cm}
\begin{equation}
    {P_*}=\frac{1}{2} P_{\mathcal{T}}(\mathcal{T}) + \frac{1}{2} P_{\mathcal{S}}(\mathcal{T'}),
	\vspace{-0.2cm}
\end{equation}
Following common practice~\cite{li2019bidirectional, Two-phase}, we use max probability threshold (MPT) to select the pixels with higher confidence in $P_*$ as pseudo labels of unlabeled target images. Concretely, define pseudo labels $\hat{Y}_{*} = \{ \hat{Y}_{*}^{(i,j,c)}\}~(1 \leq i \leq H, 1 \leq j \leq W, 1 \leq c \leq C)$ as:
\vspace{-0.2cm}
\begin{equation}
\label{eq:sharepseudolabel}
 {\hat{Y}_*^{(i,j,c)}}=\left\{
 \begin{array}{ll}
 1, & {\rm if} \quad c=\mathop{argmax}\limits_{c} {({P_*^{(i,j,c)}})} \\
 & {\rm and} \ {P_*^{(i,j,c)}}>{\lambda} \\
0,& {\rm else},
\end{array}\\
\right.
\vspace{-0.1cm}
\end{equation}
where $\lambda$ denotes threshold to filter pixels with low prediction confidence. We set $\lambda=0.9$ as default according to ~\cite{li2019bidirectional}.

Though path-$\mathcal{S}$ and path-$\mathcal{T}$ can use respective pseudo labels generated by themselves, we will demonstrate the benefits by using shared pseudo label $\hat{Y}_*$ in Section~\ref{section:Experiments}.

{\noindent \textbf{Dual Path Segmentation Training.}}
Now we introduce the process of dual path segmentation training. Concretely, for path-$\mathcal{T}$, the objective is to train a well generalized segmentation model ${M_{\mathcal{T}}}$ in target domain. Training data for ${M_{\mathcal{T}}}$ includes two part, translated source images $\mathcal{S}' = {G_{\mathcal{S}\rightarrow\mathcal{T}}}(\mathcal{S})$ with ground-truth labels $Y_{\mathcal{S}}$, and raw target images $\mathcal{T}$ with pseudo labels $\hat{Y}_*$ generated by DPPLG. In contrast, path-$\mathcal{S}$ requires good generalization in source domain. Similarly, ${M_{\mathcal{S}}}$ is trained on source images $\mathcal{S}$ with ground-truth labels $Y_{\mathcal{S}}$ and translated images $\mathcal{T}'={G_{\mathcal{T}\rightarrow\mathcal{S}}}(\mathcal{T})$ with shared pseudo labels $\hat{Y}_*$. Besides the supervision from segmentation loss, we also utilize a discriminator on top of the features of the segmentation model to further decrease the domain gap as in~\cite{hoffman2018cycada, li2019bidirectional}. The overall loss function of dual path segmentation can be defined as:
\begin{equation}
 \label{loss:seg_t}
 \vspace{-0.1cm}
 \begin{aligned}
 \mathcal{L}_{DualSeg} &= \mathcal{L}_{seg}^{\mathcal{T}}(\mathcal{S'}, Y_\mathcal{S}) + \mathcal{L}_{seg}^{\mathcal{T}}(\mathcal{T}, \hat{Y}_*)\\
  &+\mathcal{L}_{seg}^{\mathcal{S}}(\mathcal{S}, Y_\mathcal{S}) + \mathcal{L}_{seg}^{\mathcal{S}}(\mathcal{T'}, \hat{Y}_*)\\
 &+\lambda_{adv} (\mathcal{L}_{adv}^{\mathcal{T}}(\mathcal{S'}, \mathcal{T}) +  \mathcal{L}_{adv}^{\mathcal{S}}(\mathcal{S}, \mathcal{T'})),
 \end{aligned}
 \end{equation}
where $\mathcal{L}_{adv}^{\mathcal{S}}$ and $\mathcal{L}_{adv}^{\mathcal{T}}$ denote typical adversarial loss, $\mathcal{L}_{seg}^{\mathcal{S}}$ and $\mathcal{L}_{seg}^{\mathcal{T}}$ are per-pixel segmentation loss as defined in Equation~\ref{loss:seg}, $\lambda_{adv}$ controls contribution of adversarial loss.
\vspace{-0.1cm}
\begin{algorithm}[b]
	\vspace{-0.1cm}
\caption{Training process of DPL}\label{algr:train_proc}
\begin{algorithmic}
    \Require ${\mathcal{S}}$, ${Y_\mathcal{S}}$, ${\mathcal{T}}$
\Ensure{${M_{\mathcal{T}}^{(N)}}$, ${M_{\mathcal{S}}^{(N)}}$}
\State {warm-up $M^{(0)}_{\mathcal{S}}$, $M^{(0)}_{\mathcal{T}}$}
\State {train DPIT with Equation~\ref{loss:dualGAN}}
\For{$n \gets 1$ to $N$} DPAS\\
    \hspace{\algorithmicindent}generate $\hat{Y}_*^{(n)}$ with Equation~\ref{eq:sharepseudolabel} \\
        \hspace{\algorithmicindent}train $M_{\mathcal{T}}^{(n)}$ and $M_{\mathcal{S}}^{(n)}$ with Equation \ref{loss:seg_t} 
\EndFor 
\end{algorithmic}
\end{algorithm}

\subsection{Training pipeline}
\label{section:trainpipeline}
Algorithm~\ref{algr:train_proc} summarizes the whole training process of DPL. First, ${M_{\mathcal{S}}}$ and ${M_{\mathcal{T}}}$ are initialized by the proposed warm-up strategy. Next, we train DPIT to provide well-translated images for subsequent learning. At last, following the common practice that self-supervised learning is conducted in an iterative way~\cite{li2019bidirectional,zou2019confidence,Two-phase}, DPAS is trained $N$ times for domain adaptation. We use superscript $(n)$ to refer to the $n$-th iteration.

\subsection{Testing Pipeline}
\label{section:testpipeline}
As shown in Figure~\ref{fig:DPL_flowchat}.(b), the inference of DPL is extremely simple, we only retain $M_{\mathcal{T}}$ when testing on target images. Though DPL already shows the superiority over the state-of-the-art methods, we explore an optional dual path testing pipeline named DPL-Dual to boost performance by considering predictions from two paths. Concretely, we first generate probability map $P_{\mathcal{T}}(\mathcal{T})$ and $P_{\mathcal{S}}(\mathcal{T}')$ from two well-trained segmentation models $M_{\mathcal{T}}$ and $M_{\mathcal{S}}$ respectively, then an average function is used to generate final probability map $P_F = (P_{\mathcal{S}}(\mathcal{T}') + P_{\mathcal{T}}(\mathcal{T}))/2$. Though DPL-Dual promotes the performance, extra computation is introduced. We recommend DPL-Dual as an optional inference pipeline when computation cost is secondary.

\section{Experiments}
\label{section:exp}
\subsection{Datasets}
Following common practice, We evaluate our framework in two common scenarios, GTA5~\cite{richter2016playing}$\rightarrow$Cityscapes~\cite{cordts2016cityscapes} and SYNTHIA~\cite{ros2016synthia}$\rightarrow$Cityscapes. GTA5 consists of 24,996 images with the resolution of $1914\times1052$ and we use the 19 common categories between GTA5 and Cityscapes for training and testing. For SYNTHIA dataset, we use the SYTHIA-RAND-CITYSCAPES set which contains 9,400 images with resolution $1280\times760$ and 16 common categories with Cityscapes. Cityscapes is split into training set, validation set and testing set. Training set contains 2,975 images with resolution $2048\times1024$. Following common practice, we report the results on the validation set which contains 500 images with same resolution. All ablation studies are performed on GTA5$\rightarrow$Cityscapes, and comparison with state-of-the-art is performed on both GTA5$\rightarrow$Cityscapes and SYNTHIA$\rightarrow$Cityscapes. We use category-wise IoU and mIoU to evaluate the performance.

\subsection{Network Architecture}

Following common practice, we use DeepLab-V2~\cite{chen2018deeplab} with ResNet-101~\cite{he2016deep} and FCN-8s~\cite{long2015fully} with VGG16~\cite{simonyan2014very} as our semantic segmentation models. The discriminator used in adversarial learning is similar to~\cite{radford2015unsupervised}, which has 5 convolutional layers with kernel size $4 \times 4$ with channel number \{64, 128, 256, 512, 1\} and stride of 2. For each of convolutional layer except the last one, a leaky ReLU~\cite{xu2015empirical} layer parameterized by 0.2 is followed. The discriminator is implemented over the softmax output of segmentation model. For DPIT, following \cite{li2019bidirectional}, we adopt the architecture of CycleGAN with 9 blocks and use the proposed dual perceptual loss to maintain visual consistency.

\subsection{Implementation Details}
When training DPIT, the input image is randomly cropped to the size $512 \times 256$ and it is trained for 40 epochs. The learning rate of first 20 epochs is 0.0002 and decreases to 0 linearly after 20 epochs. Following \cite{CycleGAN2017, li2019bidirectional}, in Equation \ref{loss:dualGAN}, $\lambda_{Recon}$ is set to 10, $\lambda_{DualPer}$ is set to 0.1, respectively. For DPAS training, the input images are resized to the size $1024\times512$ with batch size 4. For DeepLab-V2 with ResNet-101, we adopt SGD as optimizer and set initial learning rate with $5 \times 10^{-4}$, which is decreased with `poly' learning rate policy with power as 0.9. For FCN-8s with VGG16, we use Adam optimizer with momentum $\{0.9,0.99\}$ and initial learning rate is set to $2 \times 10^{-5}$. The learning rate is decreased with `step' policy with step size 50000 and drop factor 0.1. For adversarial learning, $\lambda_{adv}$ is set to $1 \times 10^{-3}$ for DeepLab-V2 and $1 \times 10^{-4}$ for FCN-8s in Equation~\ref{loss:init_seg_t} and \ref{loss:seg_t}. The discriminator is trained with Adam optimizer with the initial learning rate $2 \times 10^{-4}$. The momentum parameters are set as 0.9 and 0.99. All ablation studies are conducted on the first iteration ($N=1$). We set $N=4$ when comparing with state-of-the-art methods.

\begin{table}[t]  
  \centering  
  \caption{Comparison of different image translation models.}  
  \small
  \label{tab:translation_choice}
        \begin{tabular}{ccc}
        \toprule[1.0pt]
        \makecell{Image translation module}& mIoU(${M^{(1)}_{\mathcal{S}}}$) & mIoU(${M^{(1)}_{\mathcal{T}}}$)\\
        \hline
       CycleGAN &41.4 &48.5\\
       SPIT & 48.6&51.1\\
    DPIT &\textbf{49.6}&\textbf{51.8}\\
        \bottomrule
        \end{tabular}
		\vspace{-0.2cm}
\end{table}  

\begin{table}[t]  
  \centering  
  \small
  \caption{Comparison of different pseudo label generation strategies.}  
  \label{tab:Ablation_SSl}
  \setlength{\tabcolsep}{2.5pt}
        \begin{tabular}{cccc}
        \toprule[1.0pt]
        \makecell{Pseudo label generation strategy}& mIoU(${M^{(1)}_{\mathcal{S}}}$)& mIoU(${M^{(1)}_{\mathcal{T}}}$)\\
        \hline
       SPPLG&46.0&50.0\\
      \hline
       DPPLG-Max & 49.2&50.6\\
       DPPLG-Joint&49.1 &50.3\\
       DPPLG-Weighted&\textbf{49.6}& \textbf{51.8}\\
        \bottomrule
        \end{tabular}
        \vspace{-0.5cm}
\end{table}

\subsection{Experiments}
{\noindent \textbf{Dual Path Image Translation Improves Translation Quality.}}\hspace{3pt}
\label{section:Experiments}
DPIT encourages visual consistency through dual perceptual loss computed by segmentation models $M_\mathcal{S}$ and $M_\mathcal{T}$. To demonstrate the effectiveness of DPIT, we compare it with: 1) naive CycleGAN, in which no perceptual loss is used to maintain visual consistency; 2) Single Path Image Translation (SPIT) used in BDL~\cite{li2019bidirectional}, which applies CycleGAN and perceptual loss computed by single segmentation model aligned in target domain. Notice the only difference in this ablation study is that different image translation methods are used in DPL. Table~\ref{tab:translation_choice} shows the comparison. By using perceptual loss to maintain visual consistency, both SPIT and DPIT can significantly improve the adaptation performance compared with naive CycleGAN. Our DPIT surpasses SPIT in both segmentation models ($M_\mathcal{S}$ and $M_\mathcal{T}$) demonstrates that extracting aligned perceptual features can further alleviate visual inconsistency caused by image translation.

{\noindent \textbf{The Effectiveness of Dual Path Pseudo Label Generation.}}\hspace{3pt}
In our proposed DPPLG module, predictions from two paths jointly participate in the generation of pseudo labels. We compare DPPLG with single path pseudo label generation (SPPLG) method, i.e., path-$\mathcal{S}$ and path-$\mathcal{T}$ generate respective pseudo labels by themselves. Meanwhile, we study three different strategies of DPPLG: 1) DPPLG-Max, which selects the prediction with maximum probability of two paths; 2) DPPLG-Joint, in which two paths generate pseudo labels separately and intersections are selected as final pseudo labels; 3) DPPLG-Weighted, which is the default strategy as described in Section~\ref{section:DPSA}. Table \ref{tab:Ablation_SSl} shows the results. All of the DPPLG strategies have better performance than SPPLG, which means the joint decision of two complementary paths can improve the quality of pseudo labels. We use DPPLG-Weighted as our pseudo label generation strategy due to the preeminent experimental result.

\begin{figure}[t!]
  \begin{minipage}[c]{0.45\linewidth}
    \small
    \centering
    \makeatletter\def\@captype{table}\makeatother\caption{Ablation study on stage-wise DPAS.}
    \label{tab:Ablation_performance}
    \setlength{\tabcolsep}{4pt}
      \begin{tabular}{cccc}
        \toprule[1.0pt]
        $M_{\mathcal{S}}$ &mIoU& $M_{\mathcal{T}}$ &mIoU\\
        \hline
        ${M_{\mathcal{S}}^{(0)}}$ & 43.7 & ${M_{\mathcal{T}}^{(0)}}$ &48.5\\
               
        ${M_{\mathcal{S}}^{(1)}}$ & 49.6 & ${M_{\mathcal{T}}^{(1)}}$ &51.8\\
        
        ${M_{\mathcal{S}}^{(2)}}$ & 50.6 & ${M_{\mathcal{T}}^{(2)}}$&52.4\\
        
         ${M_{\mathcal{S}}^{(3)}}$ & \textbf{50.7} & ${M_{\mathcal{T}}^{(3)}}$ &52.6\\
        ${M_{\mathcal{S}}^{(4)}}$ & \textbf{50.7} & ${M_{\mathcal{T}}^{(4)}}$ &\textbf{52.8}\\
        \bottomrule
        \end{tabular}
   \end{minipage}
   \hspace{.15in}
 \begin{minipage}[c]{0.45\linewidth}
\small
\centering
     \makeatletter\def\@captype{table}\makeatother\caption{Ablation study on $M_{\mathcal{T}}$ warm-up.}
         \setlength{\tabcolsep}{4pt}
        \label{tab:Ablation_init}
        \begin{tabular}{ccc}
        
        \toprule[1.0pt]
        Model & $\delta$ &mIoU\\
        \hline
        $M_{\mathcal{T}}$ &0.2 &47.4\\
        $M_{\mathcal{T}}$ &0.3 &\textbf{48.5}\\
       $M_{\mathcal{T}}$ &0.5 & 47.3\\
        \hline
        $M_{\mathcal{T}}$ w/ $Y_{\mathcal{S}}$& - &46.2\\
        $M_{\mathcal{T}}$ w/ $\hat{Y}_{\mathcal{S}'}$ & - &44.3\\
        \bottomrule
        
        \end{tabular}
  \end{minipage}
\vspace{-0.7cm}
\end{figure}

\renewcommand\arraystretch{1.1}
\begin{table*}[tp]
\scriptsize
\centering
\caption{Comparison with state-of-the-art methods on GTA5$\rightarrow$Cityscapes scenario. \textcolor{red}{Red}: best result. \textcolor{blue}{Blue}: second best result.}
\label{tab:comparison_gta5}
\setlength{\tabcolsep}{2.5pt}
\begin{tabular}{ccccccccccccccccccccccc}
\hline
 \shortstack{Segmentation \\Model}&{Method}  & \rotatebox{90}{road}  & \rotatebox{90}{sidewalk} &\rotatebox{90}{building} & \rotatebox{90}{wall} & \rotatebox{90}{fence} & \rotatebox{90}{pole} & \rotatebox{90}{t-light} & \rotatebox{90}{t-sign} & \rotatebox{90}{vegetation } & \rotatebox{90}{terrain} & \rotatebox{90}{sky} & \rotatebox{90}{person} & \rotatebox{90}{rider} & \rotatebox{90}{car} & \rotatebox{90}{truck} & \rotatebox{90}{bus} & \rotatebox{90}{train} & \rotatebox{90}{motorbike} & \rotatebox{90}{bicycle} & mIoU\\
\hline

\multirow{9}{*}{ResNet101\cite{he2016deep}}
&BDL\cite{li2019bidirectional} & {{91.0}} & {{44.7}} & {{84.2}} & {{34.6}} & {{27.6}} & 30.2 & 36.0 & 36.0 & {{85.0}} & \textcolor{blue}{{43.6}} & {{83.0}} & 58.6 & {{31.6}} & {{83.3}} & {{35.3}} & {{49.7}} & 3.3 & 28.8 & 35.6 & {{48.5}} \\

&SIM \cite{wang2020differential}  &
90.6 & 44.7 & 84.8 & 34.3 & 28.7 & 31.6 & 35.0 & 
\textcolor{blue}{37.6} & 84.7 & 43.3 & 85.3 & 57.0 & 31.5 & 83.8 & 
\textcolor{blue}{42.6} & 48.5 & 1.9 & 30.4 & 39.0 & 49.2 \\
&FADA\cite{wang2020classes} & 
92.5& 47.5 &85.1 &37.6 &\textcolor{red}{32.8}& 33.4 &33.8 &18.4 &85.3 &37.7& 83.5 &\textcolor{red}{63.2}& \textcolor{red}{39.7} &\textcolor{red}{87.5} &32.9 &47.8& 1.6& 34.9& 39.5& 49.2\\
&Label-Driven\cite{yang2020label} &
				90.8 & 41.4 & 84.7 &  35.1 &27.5&31.2&38.0&32.8&\textcolor{blue}{85.6}&42.1&84.9&59.6&
        \textcolor{blue}{34.4}&85.0& \textcolor{red}{42.8}&\textcolor{blue}{52.7}&3.4&30.9&38.1&49.5 \\
&Kim et al. \cite{kim2020learning}  &
\textcolor{blue}{92.9} & \textcolor{blue}{55.0} &   85.3 & 34.2 &   31.1 & 34.9 &   40.7 & 
34.0 & 85.2 & 40.1 &   \textcolor{red}{87.1} & 61.0 & 31.1 & 82.5 & 
32.3 & 42.9 & 0.3 &   36.4 & 46.1 & 50.2 \\

&FDA-MBT \cite{yang2020fda}  &
92.5 & 53.3 & 82.4 & 26.5 & 27.6 &   \textcolor{blue}{36.4} & 40.6 & 
\textcolor{red}{38.9} & 82.3 & 39.8 & 78.0 &   62.6 &   \textcolor{blue}{34.4} & 84.9 &
34.1 &   \textcolor{red}{53.1} &   16.9 & 27.7 &   \textcolor{blue}{46.4} &   50.5 \\

&TPLD \cite{Two-phase}&
\textcolor{red}{94.2} &\textcolor{red}{60.5} &82.8 &36.6 &16.6& \textcolor{red}{39.3}& 29.0 &25.5& \textcolor{blue}{85.6} &\textcolor{red}{44.9} &84.4& 60.6 &27.4& 84.1 &37.0& 47.0 &\textcolor{red}{31.2} &36.1& \textcolor{red}{50.3} &51.2 \\
\cline{2-22}
&DPL &92.5	&52.8	&\textcolor{blue}{86.0}	&\textcolor{blue}{38.5}	&31.7	&36.2	&\textcolor{blue}{47.3}	&34.9	&85.5	&39.9	&85.2	&\textcolor{blue}{62.9}	&33.9	&86.8	&37.2	&45.3	&\textcolor{blue}{20.1}	&\textcolor{red}{44.1}	&42.4& \textcolor{blue}{52.8}\\

&DPL-Dual& 92.8	&54.4	&\textcolor{red}{86.2}	&\textcolor{red}{41.6}	&\textcolor{blue}{32.7}	&\textcolor{blue}{36.4}	&\textcolor{red}{49.0}	&34.0	&\textcolor{red}{85.8}	&41.3	&\textcolor{blue}{86.0}	&\textcolor{red}{63.2}	&34.2	&\textcolor{blue}{87.2}	&39.3	&44.5	&18.7	&\textcolor{blue}{42.6}	&43.1& \textcolor{red}{53.3}
\\
\hline
\multirow{9}{*}{VGG16\cite{simonyan2014very}}
&TPLD \cite{Two-phase}&
83.5& 49.9& 72.3& 17.6& 10.7& \textcolor{blue}{29.6}& 28.3& 9.0& 78.2& 20.1& 25.7& 47.4& 13.3& 79.6& 3.3& 19.3& 1.3& 14.3& \textcolor{red}{33.5}& 34.1 \\

&BDL \cite{li2019bidirectional} &
 89.2& 40.9& 81.2& 29.1& 19.2& 14.2& 29.0& 19.6& 83.7& 35.9& 80.7& 54.7& 23.3& 82.7& 25.8& 28.0& 2.3& 25.7& 19.9& 41.3 \\ 

&FDA-MBT \cite{yang2020fda}  &
 86.1& 35.1& 80.6& 30.8& 20.4& 27.5& 30.0& 26.0& 82.1& 30.3& 73.6& 52.5& 21.7& 81.7& 24.0& 30.5& \textcolor{red}{29.9}& 14.6& 24.0& 42.2 \\
&Kim et al. \cite{kim2020learning}  &
 
\textcolor{red}{ 92.5}&  \textcolor{red}{54.5}& \textcolor{red}{83.9}& \textcolor{blue}{34.5}& \textcolor{blue}{25.5}& \textcolor{red}{31.0}& 30.4& 18.0& \textcolor{blue}{84.1}& \textcolor{blue}{39.6}& \textcolor{blue}{83.9}& 53.6& 19.3& 81.7& 21.1& 13.6& \textcolor{blue}{17.7}& 12.3& 6.5& 42.3 \\

&SIM \cite{wang2020differential}  &
 88.1& 35.8& 83.1& 25.8& 23.9& 29.2& 28.8& \textcolor{red}{28.6}& 83.0& 36.7& 82.3& 53.7& 22.8& 82.3& 26.4& 38.6& 0.0& 19.6& 17.1& 42.4 \\
	
&Label-Driven\cite{yang2020label} &
 90.1& 41.2& 82.2& 30.3& 21.3& 18.3& 33.5& 23.0& \textcolor{blue}{84.1}& 37.5& 81.4& 54.2& 24.3& 83.0& 27.6& 32.0& 8.1& \textcolor{blue}{29.7}& 26.9& 43.6 \\
&FADA\cite{wang2020classes} & \textcolor{blue}{92.3}& \textcolor{blue}{51.1}& \textcolor{blue}{83.7}& 33.1& \textcolor{red}{29.1}& 28.5& 28.0& 21.0& 82.6& 32.6&  \textcolor{red}{85.3}& \textcolor{red}{55.2}& \textcolor{red}{28.8}& \textcolor{red}{83.5}& 24.4& 37.4& 0.0& 21.1& 15.2& 43.8\\

\cline{2-22}
&DPL&
 88.9& 43.6& 83.4& 33.8& {24.7}& 28.0& \textcolor{blue}{37.6}& \textcolor{blue}{26.2}& \textcolor{blue}{84.1}& \textcolor{red}{40.3}& 81.5& \textcolor{blue}{54.9}& 25.0& 83.0& \textcolor{blue}{27.7}& \textcolor{blue}{48.6}& 4.8& 29.1& 32.0& \textcolor{blue}{46.2}
\\
&DPL-Dual&  
89.2& 44.0& 83.5& \textcolor{red}{35.0}& {24.7}& 27.8& \textcolor{red}{38.3}& 25.3&\textcolor{red}{84.2}& 39.5& 81.6& 54.7& \textcolor{blue}{25.8}& \textcolor{blue}{83.3}& \textcolor{red}{29.3}& \textcolor{red}{49.0}& 5.2& \textcolor{red}{30.2}& \textcolor{blue}{32.6}& \textcolor{red}{46.5}
\\
\hline

\end{tabular}
\end{table*}
\begin{table*}[t!p]
\scriptsize
\centering
\caption{Comparison with state-of-the-art methods on SYNTHIA$\rightarrow$Cityscapes scenario. \textcolor{red}{Red}: best result. \textcolor{blue}{Blue}: second best result.}
\setlength{\tabcolsep}{3.5pt}
\label{tab:comparison_synthia}
\begin{tabular}{cccccccccccccccccccc}
\hline
\shortstack{Segmentation \\Model}&{Method}  & \rotatebox{90}{road}  & \rotatebox{90}{sidewalk} &\rotatebox{90}{building}&\rotatebox{90}{wall} & \rotatebox{90}{fence} & \rotatebox{90}{pole} & \rotatebox{90}{t-light} & \rotatebox{90}{t-sign} & \rotatebox{90}{vegetation } & \rotatebox{90}{sky} & \rotatebox{90}{person} & \rotatebox{90}{rider} & \rotatebox{90}{car} & \rotatebox{90}{bus} & \rotatebox{90}{motorbike} & \rotatebox{90}{bicycle} &\makecell[b]{ mIoU \\ (16)}&\makecell[b]{ mIoU \\ (13)}\\
\hline

\multirow{9}{*}{ResNet101\cite{he2016deep}}
&Kim et al. \cite{kim2020learning}  &
\textcolor{red}{92.6} & \textcolor{red}{53.2} & 79.2&-&-&-&   1.6 & 7.5 & 78.6 & 84.4 & 52.6 & 
		20.0 & 82.1 & 34.8 & 14.6 & 39.4 &-& 49.3 \\
  
    &BDL\cite{li2019bidirectional} & {{86.0}} & \textcolor{blue}{{46.7}} & {{80.3}} &-&-&- & 14.1 & 11.6 & {{79.2}}	 & 81.3 & {{54.1}} & {{27.9}} & {{73.7}} & \textcolor{blue}{{42.2}} & {{25.7}} & {{45.3}} &-& {{51.4}} \\

    &SIM \cite{wang2020differential}  &
				83.0 & 44.0 & 80.3 &-&-&-& 17.1 & 15.8 & 80.5 & 81.8 & 
				59.9 &  \textcolor{red}{33.1} & 70.2 & 37.3 & 28.5 & \textcolor{blue}{45.8} & -&52.1 \\
				
				&FDA-MBT \cite{yang2020fda}   &
				79.3 & 35.0 & 73.2 &-&-&- &  19.9 &  24.0 & 61.7 & 82.6 &  \textcolor{red}{61.4} & 
				{31.1} &  83.9 & 40.8 &  \textcolor{red}{38.4} &  \textcolor{red}{51.1} &-& 52.5 \\
				
        &FADA\cite{wang2020classes} & 84.5&40.1&\textcolor{red}{83.1}&4.8&0.0&\textcolor{blue}{34.3}&{20.1}&\textcolor{blue}{27.2}&\textcolor{red}{84.8}&84.0&53.5&22.6&\textcolor{red}{85.4}&\textcolor{red}{43.7}&{26.8}&27.8&45.2&{52.5}\\	
				
        &Label-Driven\cite{yang2020label}& 85.1&44.5&81.0&-&-&-&16.4&15.2&80.1&84.8&59.4&\textcolor{blue}{31.9}&73.2&41.0&\textcolor{blue}{32.6}&44.7&-&53.1 \\

&TPLD \cite{Two-phase}&80.9 &44.3 &82.2 &\textcolor{red}{19.9}&0.3&\textcolor{red}{40.6}& 20.5& \textcolor{red}{30.1}& 77.2 &80.9& \textcolor{blue}{60.6}& 25.5& \textcolor{blue}{84.8}& 41.1& 24.7 &43.7&\textcolor{red}{47.3}& 53.5\\
\cline{2-20}
&DPL & 87.4&45.5&82.7&\textcolor{blue}{14.8}&\textcolor{red}{0.7}&33.0&\textcolor{blue}{21.9}&20.0&82.9&\textcolor{blue}{85.1}&56.4&21.7&82.1&39.5&30.8&45.2&46.9&\textcolor{blue}{53.9}

\\
&DPL-Dual&\textcolor{blue}{87.5}&45.7&\textcolor{blue}{82.8}&13.3&\textcolor{blue}{0.6}&33.2&\textcolor{red}{22.0}&20.1&\textcolor{blue}{83.1}&\textcolor{red}{86.0}&56.6&21.9&83.1&40.3&29.8&45.7&\textcolor{blue}{47.0}&\textcolor{red}{54.2}\\

\hline
\multirow{9}{*}{VGG16~\cite{simonyan2014very}}
&CrCDA \cite{huang2020contextual}        &74.5& 30.5& 78.6& \textcolor{blue}{6.6}& 0.7& 21.2& 2.3& 8.4& 77.4& 79.1& 45.9& 16.5& 73.1& 24.1& 9.6& 14.2& 35.2& 41.1\\

&TPLD \cite{Two-phase}&        81.3& 34.5& 73.3& \textcolor{red}{11.9}&0.0& 26.9& 0.2& 6.3& 79.9& 71.2& 55.1& 14.2& 73.6& 5.7& 0.5& 41.7& 36.0& 41.3  \\

&Kim et al. \cite{kim2020learning}  & \textcolor{red}{89.8}& \textcolor{red}{48.6}& 78.9&-&-&-&0.0& 4.7& 80.6& 81.7& 36.2& 13.0& 74.4& 22.5& 6.5& 32.8&-& 43.8  \\
&BDL \cite{li2019bidirectional}         & 72.0& 30.3& 74.5& 0.1& 0.3& 24.6& 10.2& 25.2& 80.5& 80.0& 54.7& \textcolor{blue}{23.2}& 72.7& 24.0& 7.5& 44.9& 39.0& 46.1 \\
&FADA \cite{wang2020classes}         & 80.4& 35.9& \textcolor{red}{80.9}& 2.5& 0.3& \textcolor{red}{30.4}& 7.9& 22.3& \textcolor{red}{81.8}& \textcolor{red}{83.6}& 48.9& 16.8& 77.7& \textcolor{red}{31.1}& 13.5& 17.9& 39.5& 46.1  \\
&FDA-MBT \cite{yang2020fda}  & \textcolor{blue}{84.2}& 35.1& 78.0& 6.1& 0.4& 27.0& 8.5& 22.1& 77.2& 79.6& 55.5& 19.9& 74.8& 24.9& \textcolor{red}{14.3}& 40.7& 40.5& 47.3  \\
&Label-Driven \cite{yang2020label}& 73.7& 29.6& 77.6& 1.0& 0.4& 26.0& 14.7& 26.6& 80.6& 81.8& \textcolor{red}{57.2}& \textcolor{red}{24.5}& 76.1& \textcolor{blue}{27.6}& \textcolor{blue}{13.6}& \textcolor{red}{46.6}& 41.1& 48.5  \\ 
\cline{2-20}
&DPL& 
82.7& 37.3& 80.1& 1.6& \textcolor{blue}{0.9}& \textcolor{blue}{29.5}& \textcolor{red}{20.5}& \textcolor{red}{33.1}& \textcolor{blue}{81.7}& \textcolor{blue}{82.9}& 55.6& 20.2& \textcolor{blue}{79.2}& 26.3& 6.8& 45.5& \textcolor{blue}{42.7}& \textcolor{blue}{50.2}
\\
&DPL-Dual &83.5& \textcolor{blue}{38.2}& \textcolor{blue}{80.4}& 1.3& \textcolor{red}{1.1}& 29.1& \textcolor{blue}{20.2}& \textcolor{blue}{32.7}& \textcolor{red}{81.8}& \textcolor{red}{83.6}& \textcolor{blue}{55.9}& 20.3& \textcolor{red}{79.4}& 26.6& 7.4& \textcolor{blue}{46.2}& \textcolor{red}{43.0}& \textcolor{red}{50.5}

\\
\bottomrule

\end{tabular}
\vspace{-1em}
\end{table*}

{\noindent \textbf{The Effectiveness of Dual Path Adaptive Segmentation.}}\hspace{3pt}
We show the stage-wise results of DPAS in Table~\ref{tab:Ablation_performance}. When warm-up is finished, $M_\mathcal{S}^{(0)}$ and $M_\mathcal{T}^{(0)}$ achieve mIoU of 43.7 and 48.5, respectively. After first iteration, $M_\mathcal{S}^{(1)}$ achieves 49.6 (+13.5{\%} improvement), and $M_\mathcal{T}^{(1)}$ achieves 51.8 (+6.8{\%} improvement). The big improvements on two segmentation models demonstrates that the interactions between two complementary paths facilitate the adaptive learning mutually. Though subsequent iterations ($M_\mathcal{S}^{(2)}$-$M_\mathcal{S}^{(4)}$ and $M_\mathcal{T}^{(2)}$-$M_\mathcal{T}^{(4)}$) can still promote the performance, the improvement is limited.

{\noindent \textbf{Ablation Study on Label Correction Strategy.}}\hspace{3pt}
In Section~\ref{section:warm_phase}, we propose a label correction strategy for $M_{\mathcal{T}}$ warm-up. Now we study different warm-up strategies as well as hyper parameters in Table~\ref{tab:Ablation_init}. Recall that label correction is used to find a revised label $Y_{\mathcal{S}'}$ by considering both ground-truth labels $Y_\mathcal{S}$ and pseudo labels $\hat{Y}_{\mathcal{S}'}$ (see Equation~\ref{equ:correction}). We ablate two extreme cases: 1) directly leverage ground-truth labels $Y_\mathcal{S}$ without label correction; 2) directly leverage pseudo labels $\hat{Y}_{\mathcal{S}'}$ without label correction. Results in Table~\ref{tab:Ablation_init} shows the superiority of our label correction module. We also study different $\delta$ which controls correction rate, from the table, we find $\delta$ is a less-sensitive hyper parameter which can be set as 0.3 by default.

{\noindent \textbf{Comparison with State-of-the-art Methods.}}\hspace{3pt}
\label{sec:experiments}
We evaluate DPL and DPL-Dual with state-of-the-art methods on two common scenarios, GTA5$\rightarrow$Cityscapes and SYNTHIA$\rightarrow$ Cityscapes. For each scenario, we report the results on two segmentation models, ResNet101 and VGG16. Table~\ref{tab:comparison_gta5} shows the results on the scenario GTA5$\rightarrow$Cityscapes, DPL achieves state-of-the-art performance on both models (with mIoU of 52.8 on ResNet101 and 46.2 on VGG16). DPL-Dual further achieves mIoU of 53.3 on ResNet101 and 46.5 on VGG16. Domain gap between SYNTHIA and Cityscapes is much larger than that of GTA5 and Cityscapes, and their categories are not fully overlapped. We list both of the results for the 13-category and 16-category for a fair comparison with state-of-the-art methods. Results are shown in Table~\ref{tab:comparison_synthia}, mIoU (13) and mIoU (16) represent adaptation methods are evaluated on 13 common categories and 16 common categories, respectively. Once again, under 13-category metric, DPL achieves state-of-the-art result on both ResNet101 and VGG16, DPL-Dual further boosts performance. For 16-categories metric, the performance of DPL with ResNet101 is slightly worse since the domain shift is much larger in \{\textit{wall, fence, pole}\} categories, and DPL with VGG16 still surpasses state-of-the-art with mIoU 42.7, DPL-Dual further promotes the performance to 43.0.

\section{Conclusion}
In this paper, we propose a novel dual path learning framework named DPL, which utilizes two complementary and interactive paths for domain adaptation of segmentation. Novel technologies such as dual path image translation and dual path adaptive segmentation are presented to make two paths interactive and promote each other. Meanwhile, a novel label correction strategy is proposed in the warm-up stage. The inference of DPL is extremely simple, only one segmentation model well aligned target domain is used. Experiments on common scenarios GTA5$\rightarrow$Cityscapes and SYNTHIA$\rightarrow$Cityscapes demonstrate the superiority of our DPL over the state-of-the-art methods.

{\small
\bibliographystyle{ieee_fullname}
\bibliography{DPL_arxiv}
}
\cleardoublepage
\onecolumn
\renewcommand\thesubsection{Appendix \Alph{subsection}}

\subsection{More Experiments}

{\noindent \textbf{Category-Wise IoU for Ablation Studies.}}\hspace{3pt}
We use mIoU as evaluation metric for ablation studies in our paper. Now we report the detailed category-wise IoU for these ablation studies. Table~\ref{tab:Cate_image} and Table~\ref{tab:Cate_PLG} present category-wise IoU of different image translation methods and pseudo label generation strategies, respectively. We can observe that our dual path image translation (DPIT) module and dual path pseudo label generation (DPPLG) strategy achieve best performance on most categories, which is consistent with the reported mIoU results in our paper. To further demonstrate the effectiveness of the proposed label correction strategy used in $M_{\mathcal{T}}$ warm-up, Table~\ref{tab:label_correction} reports detailed results. Comparing with strategies without utilizing label correction (i.e., $M_{\mathcal{T}}$ w/ $Y_{\mathcal{S}}$ and $M_{\mathcal{T}}$ w/ $\hat{Y}_{\mathcal{S}'}$), almost all categories have a significant improvement by using our proposed label correction strategy. From the table, we also conclude that $\delta$ is a less-sensitive hyper parameter.

\renewcommand{\arraystretch}{1.8}
\begin{table*}[bh]

\scriptsize
\centering
\vspace{0.5em}
\caption{Category-wise IoU evaluation for different image translation methods.\label{tab:Cate_image}}
\vspace{1em}
\setlength{\tabcolsep}{3pt}
\begin{tabular}{cccccccccccccccccccccc}
\toprule
 {Method} &Model & \rotatebox{90}{road}  & \rotatebox{90}{sidewalk} &\rotatebox{90}{building} & \rotatebox{90}{wall} & \rotatebox{90}{fence} & \rotatebox{90}{pole} & \rotatebox{90}{t-light} & \rotatebox{90}{t-sign} & \rotatebox{90}{vegetation} & \rotatebox{90}{terrain} & \rotatebox{90}{sky} & \rotatebox{90}{person} & \rotatebox{90}{rider} & \rotatebox{90}{car} & \rotatebox{90}{truck} & \rotatebox{90}{bus} & \rotatebox{90}{train} & \rotatebox{90}{motorbike} & \rotatebox{90}{bicycle} &mIoU\\
\hline

 CycleGAN &${M^{(1)}_{\mathcal{S}}}$ &
84.1& 30.9& 79.8& 24.3& 26.0& 32.3& 37.1& \bf 32.0& 80.5& 13.7& 73.9& 55.9& 25.3& 69.4& 31.3& 26.6& 0.5& 24.3& 39.3&41.4\\
       SPIT &${M^{(1)}_{\mathcal{S}}}$&
 91.7& \bf 50.9& 84.5& \bf 37.0& 25.9& 32.7& 42.4& 28.5& \bf 84.2& \bf 39.6& \bf 82.2& 60.9& 31.9& 84.4& 26.9& 37.3& 13.5& 29.8& 38.1&48.6\\

    DPIT &${M^{(1)}_{\mathcal{S}}}$&
 \bf 92.3& 49.0& \bf 84.7& 35.9& \bf 27.5& \bf 33.5& \bf 42.6& 28.9& 83.9& 38.3& 82.0& \bf 61.1& \bf 32.5& \bf 85.3& \bf 32.5& \bf 44.7& \bf 13.9& \bf 33.5& \bf 39.7&\bf49.6\\

\hline
       CycleGAN &${M^{(1)}_{\mathcal{T}}}$&91.3&	49.1&	84.3&	33.8&	24.2&	34.5&	41.7&	33.0	&82.7	&35.5	&79.2&	60.2&	30.2&	84.3&	34.1&	40.9&	14.4	&31.6	&36.9&48.5
 \\
       SPIT &${M^{(1)}_{\mathcal{T}}}$& \bf92.4	&\bf51.3&	85.4&	39.9	&\bf29.3&	35.4&	44.5&	33.3&	\bf85.1&	\bf41.1&	\bf82.6	&61.6&	31.2&	86.1&	32.3&	41.8&	\bf19.4&	35.2&	\bf42.2&51.1
\\

   DPIT &${M^{(1)}_{\mathcal{T}}}$ &91.7&	49.3&	\bf85.7	&\bf40.4&	28.4	&\bf36.0&	\bf45.7	&\bf38.0& 84.7	&39.5&	81.4&	\bf62.5&	\bf31.8 	&\bf86.7&	\bf39.4&	\bf48.1&	15.7&	\bf38.9	&39.9& \bf 51.8

\\
\bottomrule
\end{tabular}
\end{table*}

\begin{table*}[h]
\scriptsize
\centering
\caption{Category-wise IoU evaluation for different pseudo label generation strategies.\label{tab:Cate_PLG}}
\setlength{\tabcolsep}{3pt}
\begin{tabular}{cccccccccccccccccccccc}
\toprule
 {Method}  & Model&\rotatebox{90}{road}  & \rotatebox{90}{sidewalk} &\rotatebox{90}{building} & \rotatebox{90}{wall} & \rotatebox{90}{fence} & \rotatebox{90}{pole} & \rotatebox{90}{t-light} & \rotatebox{90}{t-sign} & \rotatebox{90}{vegetation} & \rotatebox{90}{terrain} & \rotatebox{90}{sky} & \rotatebox{90}{person} & \rotatebox{90}{rider} & \rotatebox{90}{car} & \rotatebox{90}{truck} & \rotatebox{90}{bus} & \rotatebox{90}{train} & \rotatebox{90}{motorbike} & \rotatebox{90}{bicycle} &mIoU \\
\hline

       SPPLG &${M^{(1)}_{\mathcal{S}}}$&
       90.0& 32.4& 83.1& 34.2& \bf 28.5& 31.2& 39.0& 19.1& 83.8& \bf 39.9& \bf 82.0& 59.3& 30.0& 83.9& 28.1& 38.7& 1.2& 30.0& \bf 39.8 &46.0\\
       DPPLG-Max &${M^{(1)}_{\mathcal{S}}}$&
       91.2& 46.6& 84.2& 34.7& 28.1& 33.5& 40.1& 27.5& 83.7& 38.2& 79.0& 60.3& 31.3& 85.3& \bf 34.8& 42.2& 22.5& 33.3& 38.0&49.2\\
       DPPLG-Joint&${M^{(1)}_{\mathcal{S}}}$&
       91.8& 47.5& 84.5& \bf 37.9& 27.4& \bf 34.7& 42.3& 27.0& 83.7& 38.9& 79.3& 60.5& 32.0& \bf 85.5& 30.7& 35.0& \bf 24.3& 32.6& 36.9&49.1\\
       DPPLG-Weighted&${M^{(1)}_{\mathcal{S}}}$&
       \bf 92.3& \bf 49.0& \bf 84.7& 35.9& 27.5& 33.5& \bf 42.6& \bf 28.9& \bf 83.9& 38.3& 82.0& \bf 61.1& \bf 32.5& 85.3& 32.5& \bf 44.7& 13.9& \bf 33.5& 39.7& \bf49.6\\
      \hline
       SPPLG &${M^{(1)}_{\mathcal{T}}}$&
        90.9& 48.3& 85.1& \bf 42.4& 27.4& 33.4& 40.1& 35.6& 84.3& 39.4& 80.4& 60.6& 30.9& 86.1& 35.3& 38.1& 16.7& 35.2& 39.4&50.0\\
       DPPLG-Max &${M^{(1)}_{\mathcal{T}}}$&
        91.4& 45.7& 84.9& 38.5& 27.6& 35.6& 44.2& 30.5& 84.0& 37.9& 79.2& 60.7& 31.2& 86.3& 36.5& \bf 48.8& \bf 22.0& 36.1& \bf 41.2&50.6\\
       DPPLG-Joint&${M^{(1)}_{\mathcal{T}}}$&
         91.2& 47.3& 85.1& 38.5& 27.3& 35.1& 43.5& 32.6& 84.3& \bf 40.6& 78.3& 61.2& 31.2& 85.8& 36.1& 47.3& 13.7& 35.4& 40.2&50.3\\
       DPPLG-Weighted&${M^{(1)}_{\mathcal{T}}}$&
        \bf 91.7& \bf 49.3& \bf 85.7& 40.4& \bf 28.4& \bf 35.9& \bf 45.7& \bf 38.0& \bf 84.7& 39.5& \bf 81.4& \bf 62.5& \bf 31.8& \bf 86.7& \bf 39.4& 48.1& 15.7& \bf 38.9& 39.9& \bf51.8\\
      \bottomrule
\end{tabular}
\end{table*}

\begin{table*}[h]
\scriptsize
\centering
\vspace{1em}
\caption{Category-wise IoU evaluation for ablation study on $M_{\mathcal{T}}$ warm-up. \label{tab:label_correction}}
\vspace{1em}
\setlength{\tabcolsep}{3pt}
\begin{tabular}{cccccccccccccccccccccc}
\toprule
Model  & $\delta$&\rotatebox{90}{road}  & \rotatebox{90}{sidewalk} &\rotatebox{90}{building} & \rotatebox{90}{wall} & \rotatebox{90}{fence} & \rotatebox{90}{pole} & \rotatebox{90}{t-light} & \rotatebox{90}{t-sign} & \rotatebox{90}{vegetation} & \rotatebox{90}{terrain} & \rotatebox{90}{sky} & \rotatebox{90}{person} & \rotatebox{90}{rider} & \rotatebox{90}{car} & \rotatebox{90}{truck} & \rotatebox{90}{bus} & \rotatebox{90}{train} & \rotatebox{90}{motorbike} & \rotatebox{90}{bicycle} &mIoU \\
\hline

       ${M_{\mathcal{T}}}$ &0.2&89.6&	46.1&	83.4&	34.4&	22.4&	34.4&	36.8&	28.2&	\bf82.8&	\bf37.5&	 \bf76.7&	56.7&	25.2&	83.6&	36.3	&\bf46.0	&18.0&	26.9&	34.9&	47.4
       \\
     ${M_{\mathcal{T}}}$ &  0.3&  \bf90.4&46.5&\bf84.1&\bf37.7&23.2&34.2&\bf39.1&\bf32.1&82.0&36.6&72.9&\bf58.9&\bf28.0&\bf84.8&35.1&41.8&\bf24.9&\bf30.6&\bf38.2 &\bf48.5 
     \\
        ${M_{\mathcal{T}}}$ &   0.5&    89.8&	\bf47.3&	 84.0&	32.9&	\bf25.6&	\bf34.5&	36.1&	30.6&	82.5&	36.5&	76.0&	58.3&	 26.7&	84.3&	31.8&	41.6&	20.8&	25.1&	35.0&	47.3\\
       \hline
       $M_{\mathcal{T}}$ w/ $Y_{\mathcal{S}}$&-&
        89.6&	42.6&	83.4&	34.5&	23.7&	34.4&	37.2&	27.5&	81.8&	34.7&	75.3&	56.9&	25.5&	84.3&	35.0&	39.7&	11.8&	24.2&	35.0&	46.2\\
       $M_{\mathcal{T}}$ w/ $\hat{Y}_{\mathcal{S}'}$&-&89.7&	42.9&	82.4&	26.0&	21.0&	31.7&	28.2&	24.1&	79.7&	29.6&	70.2&	52.1&	23.8&	82.1&	\bf39.0&	42.3&	21.6&	22.2&	32.8&	44.3\\
       \bottomrule
      
\end{tabular}
\end{table*}

\clearpage

{\noindent \textbf{Analysis of Different Pseudo Label Generation Strategies.}}\hspace{3pt}
Effective pseudo label generation strategies aim to generate sufficient amount of pseudo labels with high-quality. To evaluate the quality of pseudo labels generated by different methods (i.e., SPPLG, DPPLG-Max, DPPLG-Joint and DPPLG-Weighted), we investigate the per-class accuracy, mean accuracy and selected-pixel ratio. Table~\ref{tab:pseudo_analyze} shows the comparison. We can observe that limited by the capability of single path model, SPPLG generates unreliable pseudo labels with the lowest mean accuracy. Among all dual path pseudo label generation strategies, DPPLG-Joint achieves the best mean accuracy but generates less amount of pseudo labels (lowest selected-pixel ratio). In contrast, DPPLG-Max has the highest selected-pixel ratio, but in the mean time, the low mean accuracy introduces more noise which disturbs the subsequent learning. Our default DPPLG-Weighted which achieves best mIoU performance makes a reasonable trade-off between mean accuracy and selected-pixel ratio and thus guarantees both quality and quantity of generated pseudo labels.

{\noindent \textbf{The Effectiveness of \boldsymbol{$M_{\mathcal{T}}$} Warm-Up.}}\hspace{3pt}
As described in Section 3.1, cross-domain perceptual supervision in DPIT and pseudo label generation in DPAS rely on the quality of segmentation models. Now we show more details to demonstrate the effectiveness of $M_{\mathcal{T}}$ warm-up. Again, the warm-up of ${M_{\mathcal{S}}}$ is simple, while the warm-up of ${M_{\mathcal{T}}}$ is exclusive in our paper. A simpler way to avoid the particular warm-up procedure of ${M_{\mathcal{T}}}$ it to directly initialize $M_{\mathcal{T}}^{(0)}$ with $M_{\mathcal{S}}^{(0)}$. In Table~\ref{tab:M_T_warmup}, we show the importance of our warm-up strategy. Compared with the model achieving state-of-the-art performance (52.8 mIoU), the model trained without ${M_{\mathcal{T}}}$ warm-up strategy only achieves mIoU of 49.0 and has worse performance in most categories.

\begin{table*}[h]
	\scriptsize
	\vspace{1em}
	\centering
	\caption{Per-class accuracy, mean accuracy and selected-pixel ratio comparison among different pseudo label generation strategies.\label{tab:pseudo_analyze}}
	
	\setlength{\tabcolsep}{3pt}
	\vspace{1em}
	\resizebox{1\textwidth}{!}{
	\begin{tabular}{ccccccccccccccccccccccccc}
	\toprule
	 {Method}  & \rotatebox{90}{road}  & \rotatebox{90}{sidewalk} &\rotatebox{90}{building} & \rotatebox{90}{wall} & \rotatebox{90}{fence} & \rotatebox{90}{pole} & \rotatebox{90}{t-light} & \rotatebox{90}{t-sign} & \rotatebox{90}{vegetation} & \rotatebox{90}{terrain} & \rotatebox{90}{sky} & \rotatebox{90}{person} & \rotatebox{90}{rider} & \rotatebox{90}{car} & \rotatebox{90}{truck} & \rotatebox{90}{bus} & \rotatebox{90}{train} & \rotatebox{90}{motorbike} & \rotatebox{90}{bicycle} &\makecell[b]{ mean\\ acc }&\makecell[b]{ pixel\\ ratio }&\makecell[b]{ mIoU\\ ($M_{\mathcal{S}}^{(1)}$ })&\makecell[b]{ mIoU\\ ($M_{\mathcal{T}}^{(1)}$ })\\
	\hline

		   SPPLG&96.6	&62.2&	96.6&	69.0&	48.9&	57.6&	62.8	&35.0&	90.3&	85.3&	99.6&	88.3&	51.3&	96.2&	82.3&	51.4	&44.9	&68.3&	61.1&	92.3&  \bf77.0&46.0&50.0\\
	
		   DPPLG-Max &97.2	&61.6	&96.8	&68.1	&49.2	&55.9	&62.9	&32.9	&90.7	&85.3	&99.6	&88.4	&52.8	&96.3	&82.3	&52.8	&44.3	&69.1	&62.3	&92.7		& 76.8&49.2&50.6
	 \\
		   DPPLG-Joint&\bf99.7&	49.9&	\bf98.8&	\bf81.4	&\bf63.7	&\bf68.8&	\bf80.0&	25.5&	\bf96.9	&\bf93.5	&\bf99.8	&\bf95.6	&\bf65.69	&\bf98.2	&\bf93.1	&\bf67.5	&10.8	&\bf91.5	&\bf85.4	&\bf96.9	&64.5&49.1&50.3
	\\
		   DPPLG-Weighted&98.7	&\bf67.6	&97.5	&78.9	&58.6	&66.4	&76.7	&\bf40.9	&94.4	&86.7	&99.6	&92.8	&59.9	&96.8	&89.9	&60.9	&\bf46.7	&87.7	&76.6	&94.9 	&71.0&\bf49.6&\bf51.8
	\\
	\bottomrule
	\end{tabular}}
	\vspace{1em}
	\end{table*}

	\begin{table*}[h]
	\scriptsize
	\centering
	 \vspace{1em}
	\caption{Evaluation of the necessity of $M_{\mathcal{T}}$ warm-up. \label{tab:M_T_warmup}}
	 \vspace{1em}
	\setlength{\tabcolsep}{3pt}
	\begin{tabular}{cccccccccccccccccccccc}
	\toprule
	\shortstack{$M_{\mathcal{T}}$\\ warm-up} &\rotatebox{90}{road}  & \rotatebox{90}{sidewalk} &\rotatebox{90}{building} & \rotatebox{90}{wall} & \rotatebox{90}{fence} & \rotatebox{90}{pole} & \rotatebox{90}{t-light} & \rotatebox{90}{t-sign} & \rotatebox{90}{vegetation} & \rotatebox{90}{terrain} & \rotatebox{90}{sky} & \rotatebox{90}{person} & \rotatebox{90}{rider} & \rotatebox{90}{car} & \rotatebox{90}{truck} & \rotatebox{90}{bus} & \rotatebox{90}{train} & \rotatebox{90}{motorbike} & \rotatebox{90}{bicycle} &mIoU \\
	\hline
			 & 91.5 & 47.8 & 83.8 & \bf40.7 & 22.5 & 34.1 & 40.7 & 27.2 & 82.4 & 35.6 & 81.4 & 59.3 & 28.9 & 84.8 & 31.8 & 39.7 & \bf25.7 & 35.4 & 38.2 & 49.0 \\ 
		   $\surd$& \bf92.5	&\bf52.8	&\bf86.0	&38.5	&\bf31.7	&\bf36.2	&\bf47.3	&\bf34.9	&\bf85.5	& \bf39.9	&\bf85.2	&\bf62.9	&\bf33.9	&\bf86.8	&\bf37.2	&\bf45.3	&20.1	&\bf44.1	&\bf42.4& \bf52.8 \\  
	\bottomrule
	\end{tabular}
	 \vspace{1em}
	\end{table*}

\clearpage

\noindent\textbf{Network designs of segmentation models.} In our proposed dual path adaptive segmentation(DPAS) module, two segmentation models $M_\mathcal{S}$ and $M_\mathcal{T}$ are trained separately. We compare this `Seperate' choice with two weight sharing designs: 1) Share-Part, where the first 4 blocks of ResNet101 are shared; 2) Share-All, where the whole segmentation model is shared. Weight sharing designs can reduce the number of parameters, while performance declined is observed in Table~\ref{tab:Comp_diff_seg_choice}.

\renewcommand{\arraystretch}{1.2}

\noindent\textbf{Weight coefficient  in Eq.6.} We re-define Eq.6 as ${P_*}=\alpha P_{\mathcal{T}}(\mathcal{T})+ (1-\alpha)P_{\mathcal{S}}(\mathcal{T'})$ and ablate $\alpha$ in Table~\ref{tab:Ablation_weights}.
\begin{table}[h]  
  \centering  
  \small
  \vspace{1em}
  \caption{Comparison of different network designs.}  

  \label{tab:Comp_diff_seg_choice}
        \begin{tabular}{ccc}
        \toprule
        Method& mIoU (${M^{(1)}_{\mathcal{S}}}$) & mIoU (${M^{(1)}_{\mathcal{T}}}$)\\
        \hline
       Share-Part &49.3  & 49.6\\
       Share-All &48.6  &49.2\\
       \hline
     \bf{Separate (default)}& \bf{49.6} & \bf{51.8}\\
        \bottomrule
        \end{tabular}
        \vspace{1em}
\end{table}  

\begin{table}[h]
	\centering  
	\small
	\caption{Ablation study on weight coefficient $\alpha$.}  
	\label{tab:Ablation_weights}
	\begin{tabular}{ccc}
		\toprule
		\makecell{$\alpha$} & mIoU (${M^{(1)}_{\mathcal{S}}}$) & mIoU (${M^{(1)}_{\mathcal{T}}}$)\\
		\hline
	0 & 46.0 & 49.1 \\ 
	0.3 & 47.9 & 50.1 \\  
	0.4 & 49.2 & 51.1 \\  
	\textbf{0.5 (default)} & \textbf{49.6} & \textbf{51.8}  \\
	0.6 & 49.4 & 51.2 \\ 
	0.7 & 49.0 & 50.0 \\  
	1 & 48.2 & 50.0 \\ 
	\bottomrule
	\end{tabular}
	\end{table}

\clearpage
 
\subsection{Qualitative Results}
\begin{figure*}[h]
\renewcommand{\baselinestretch}{1}
	\centering
	\vspace{1cm}
	\scalebox{1}{
		\begin{tabular}{cccccc}

			\subfloat{\includegraphics[width=0.16\linewidth]{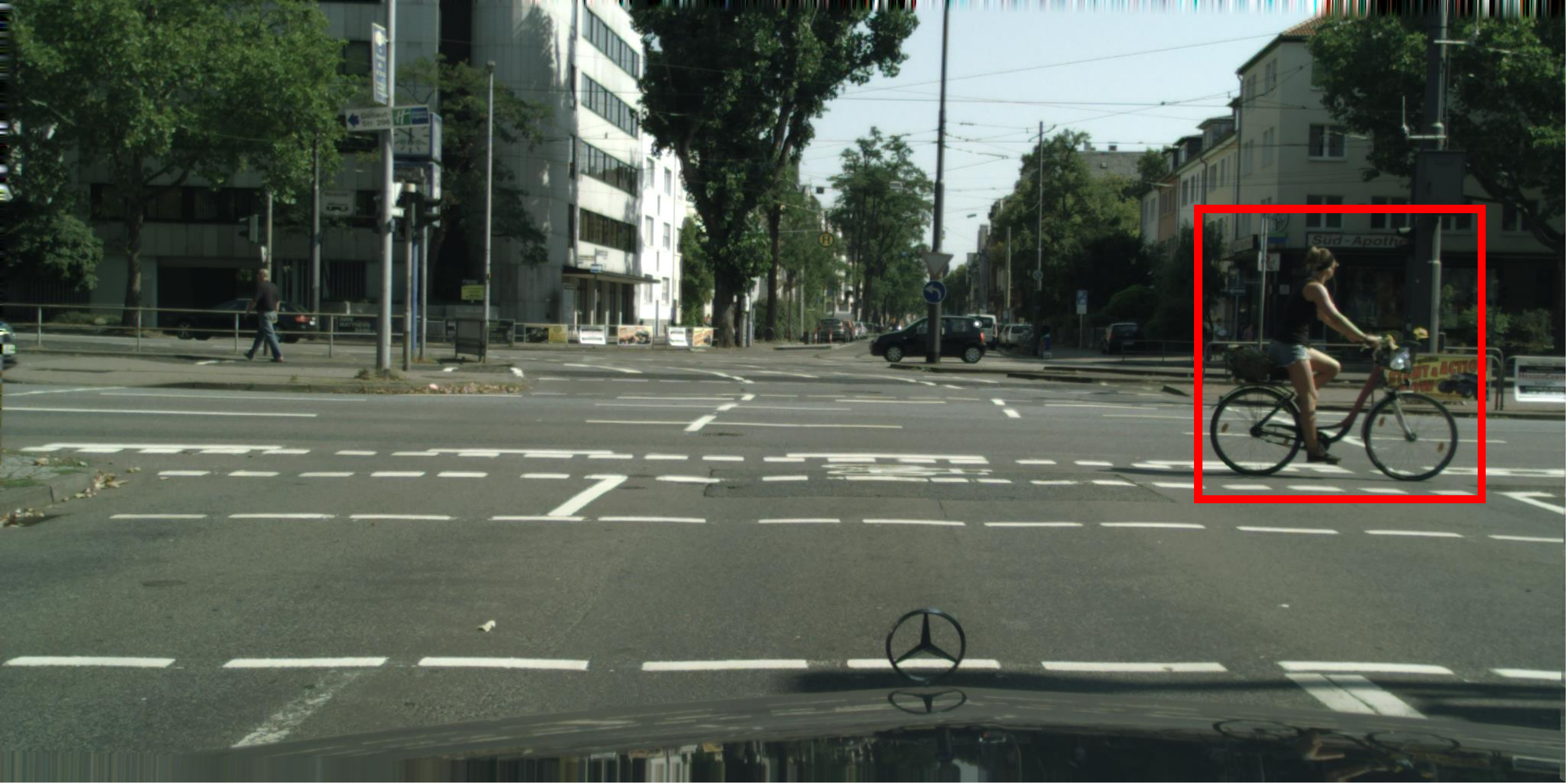}} &\hspace{-4.5mm}		
			\subfloat{\includegraphics[width=0.16\linewidth]{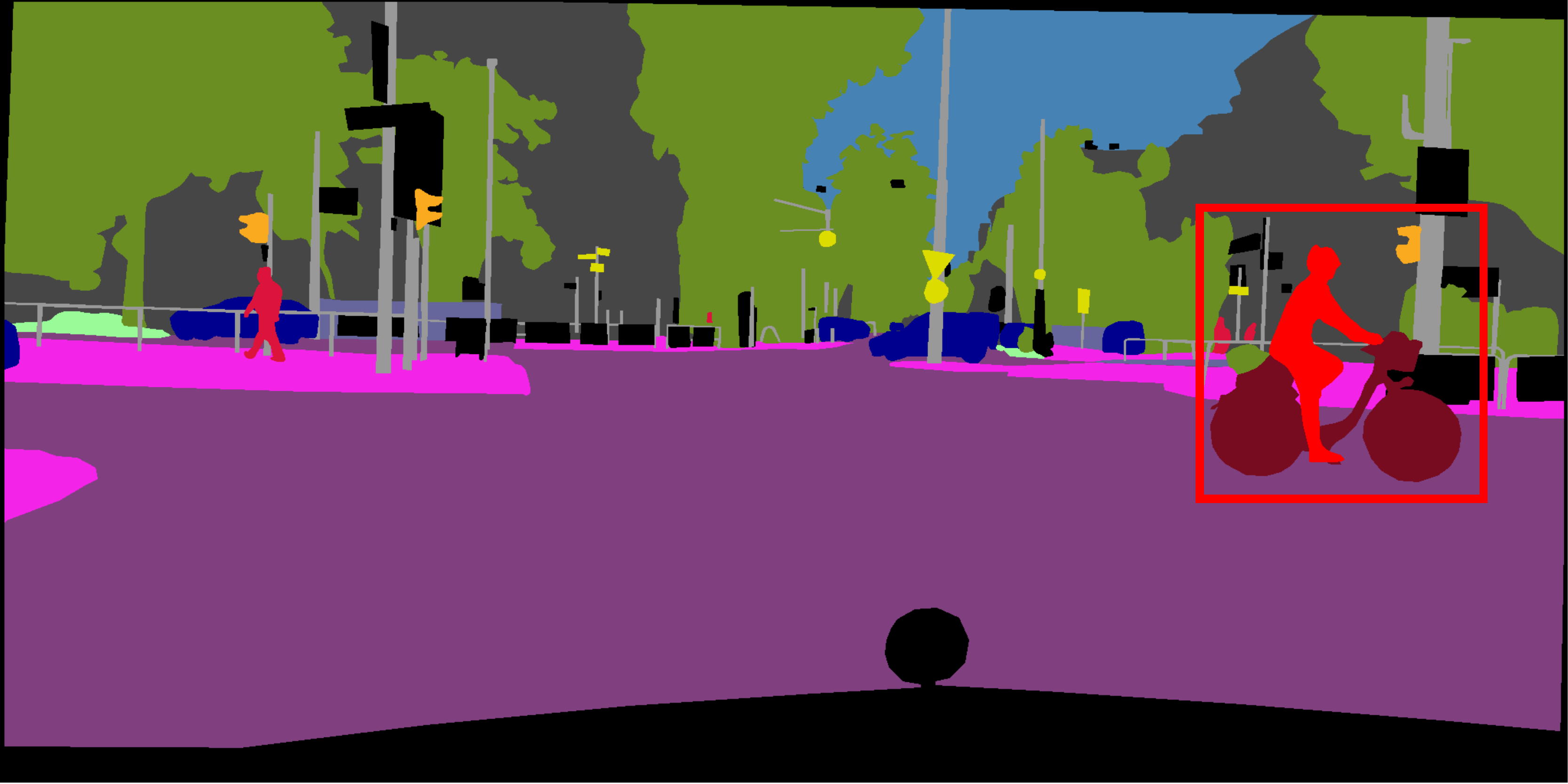}} &\hspace{-4.5mm}
			\subfloat{\includegraphics[width=0.16\linewidth]{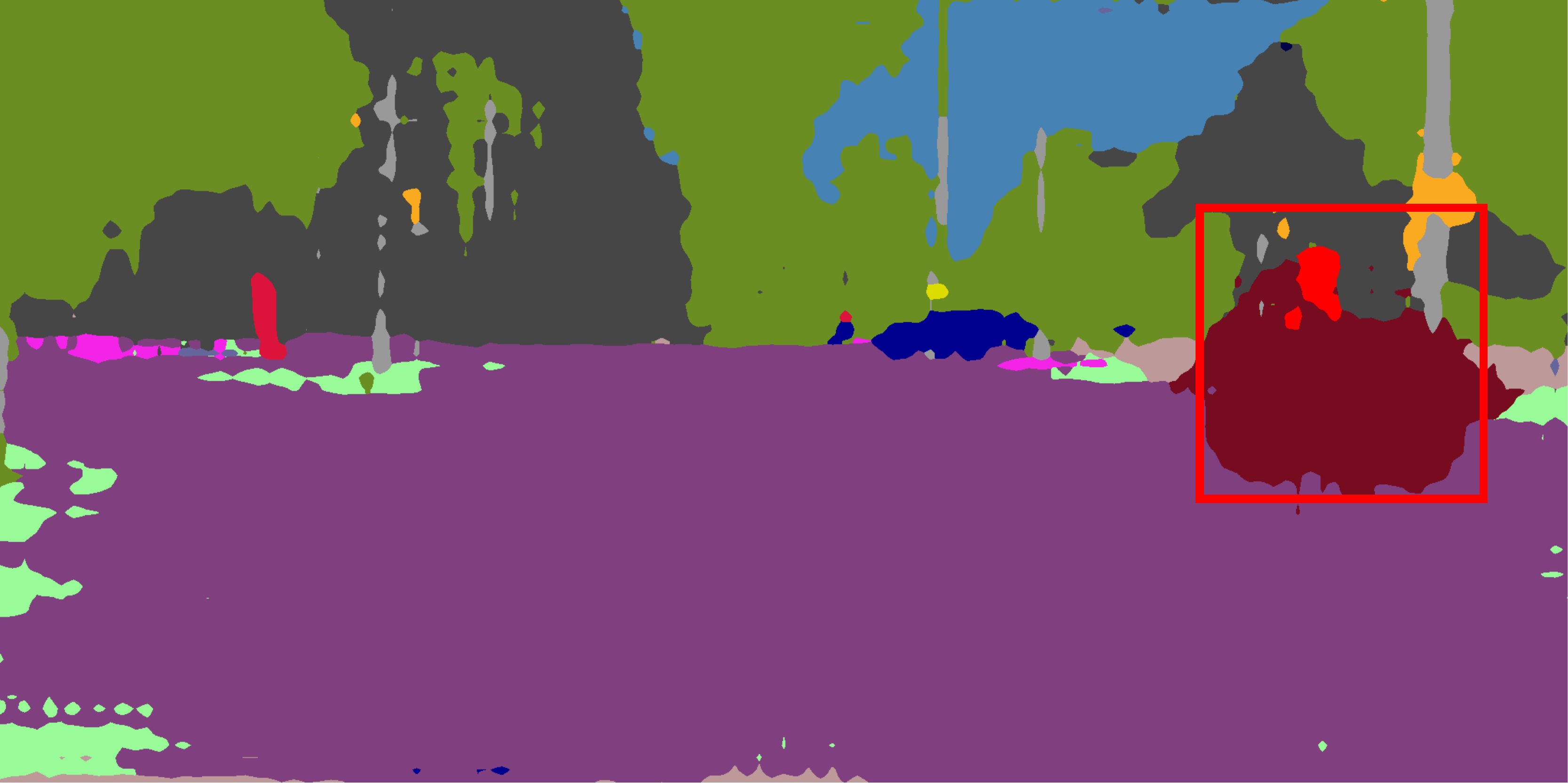}} &\hspace{-4.5mm}
			\subfloat{\includegraphics[width=0.16\linewidth]{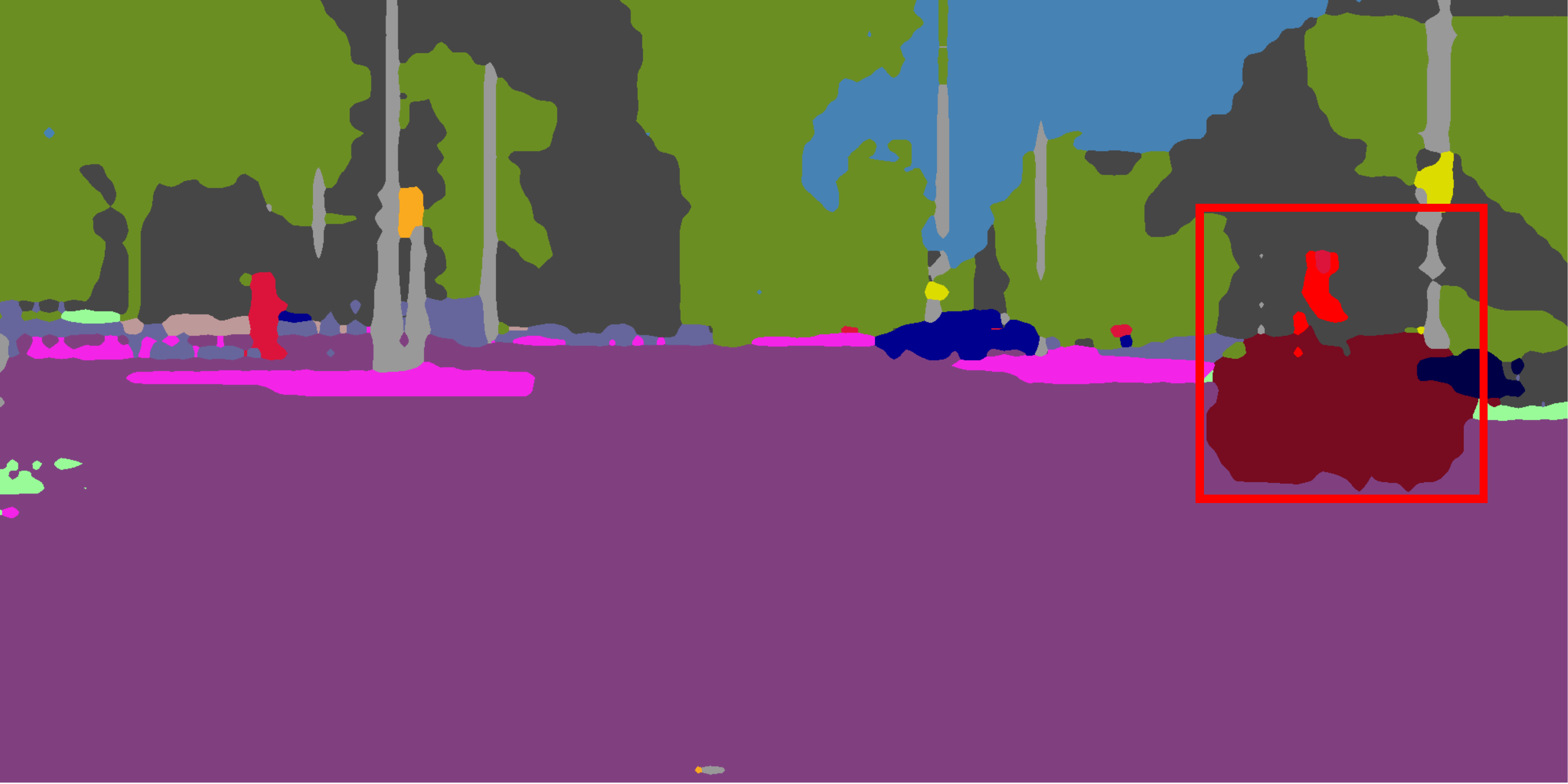}} &\hspace{-4.5mm}
			\subfloat{\includegraphics[width=0.16\linewidth]{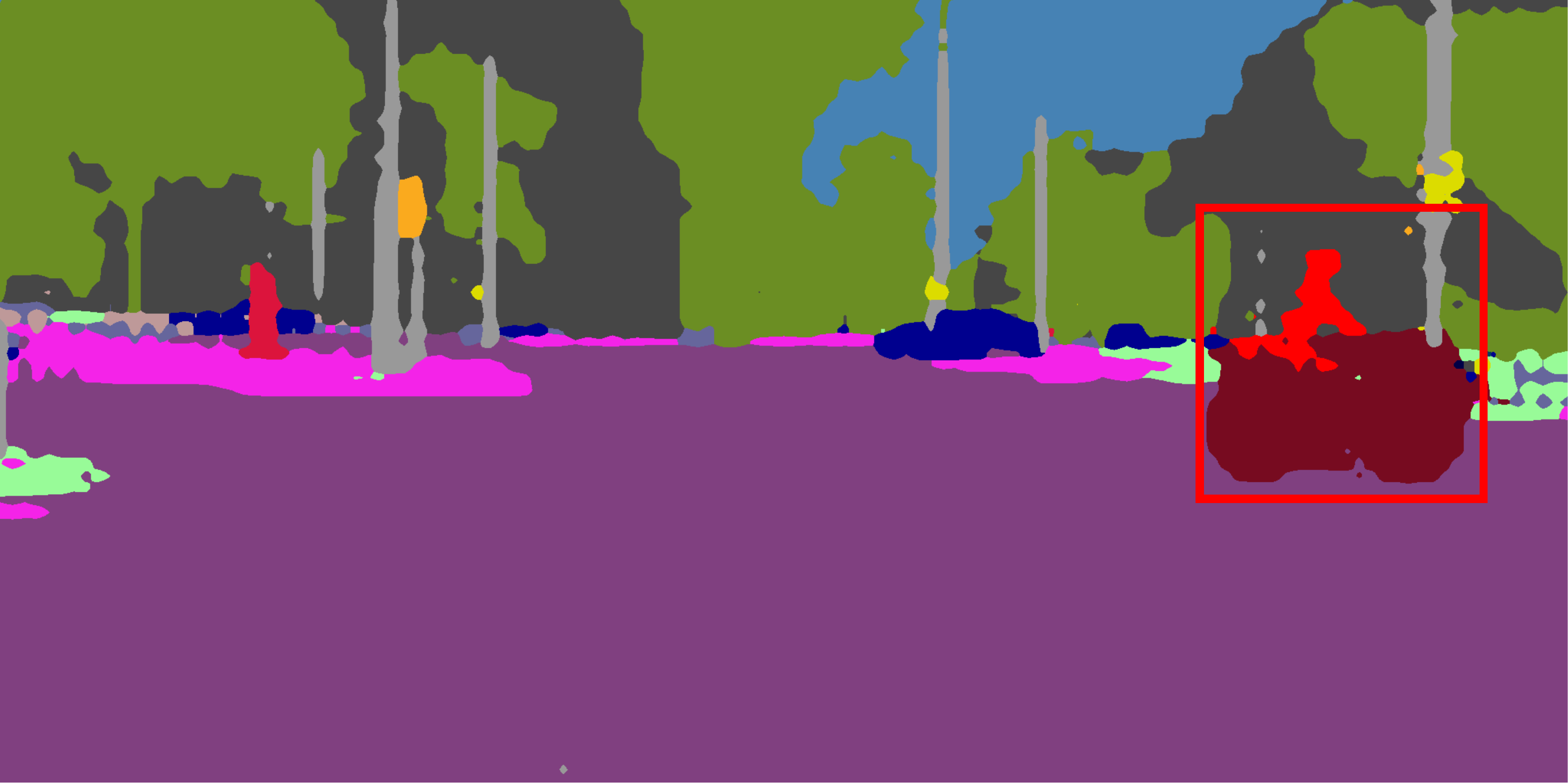}} &\hspace{-4.5mm}
			\subfloat{\includegraphics[width=0.16\linewidth]{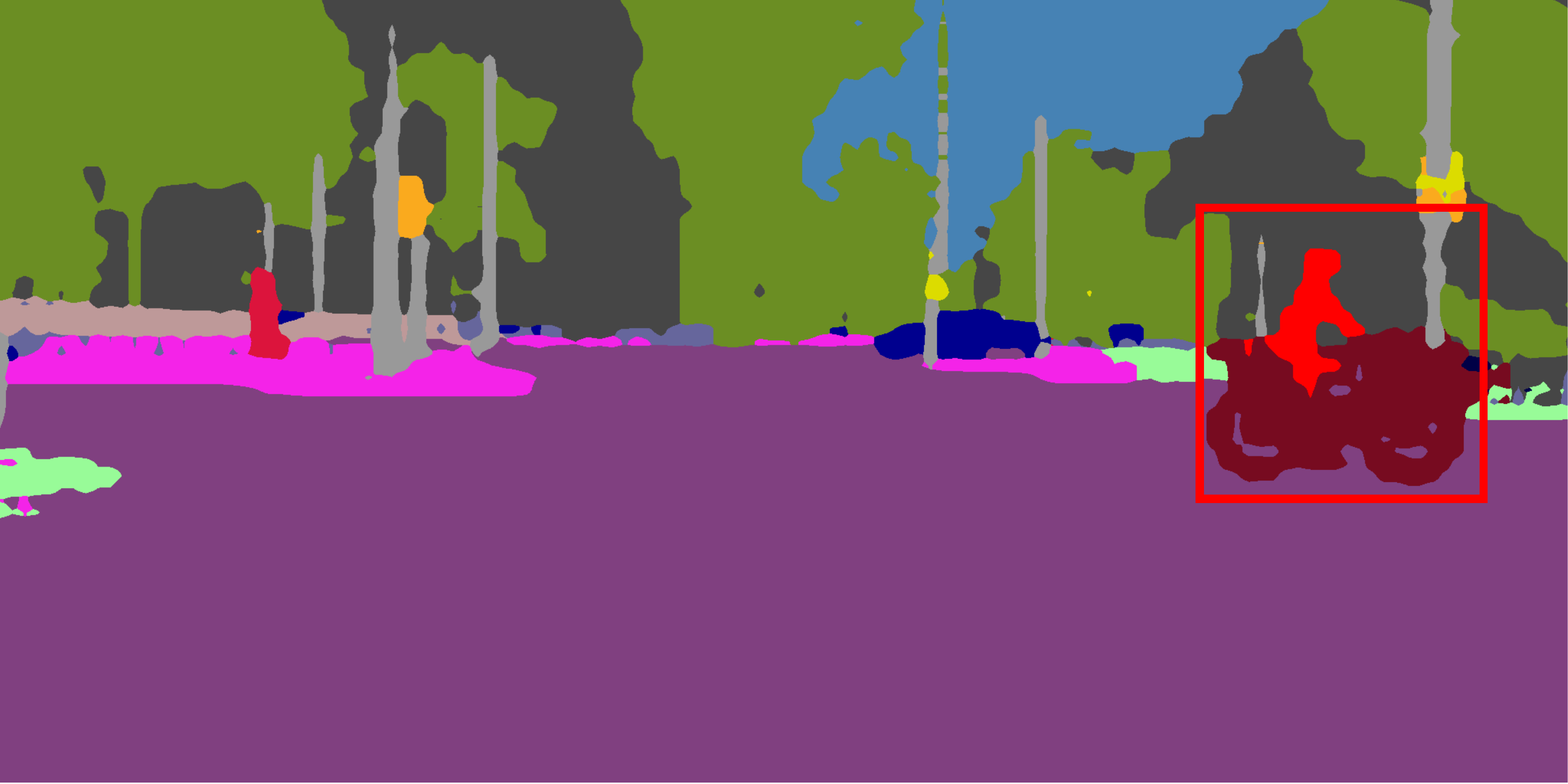}} 
			\\
			\subfloat{\includegraphics[width=0.16\linewidth]{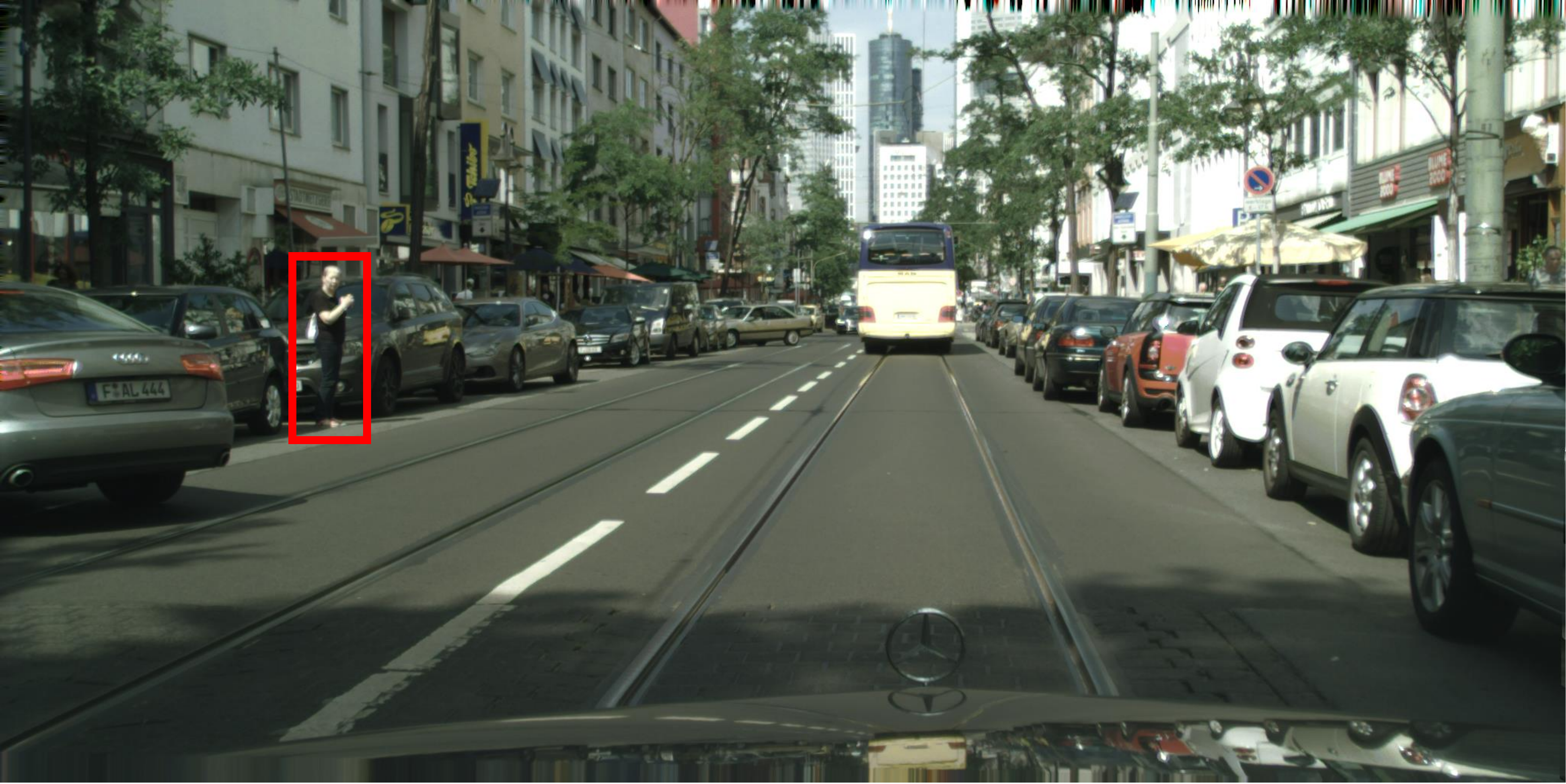}} &\hspace{-4.5mm}			
			\subfloat{\includegraphics[width=0.16\linewidth]{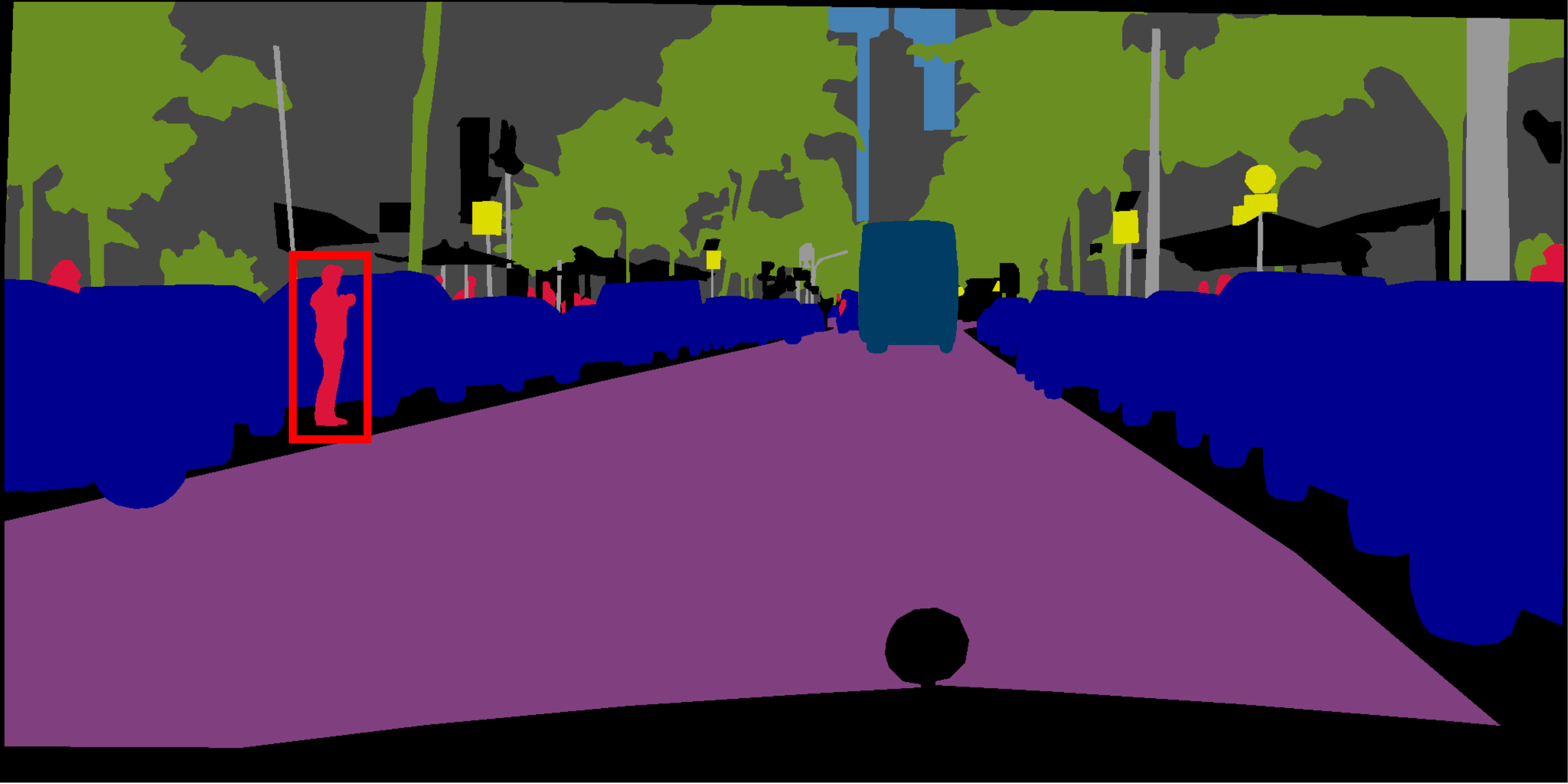}} &\hspace{-4.5mm}
			\subfloat{\includegraphics[width=0.16\linewidth]{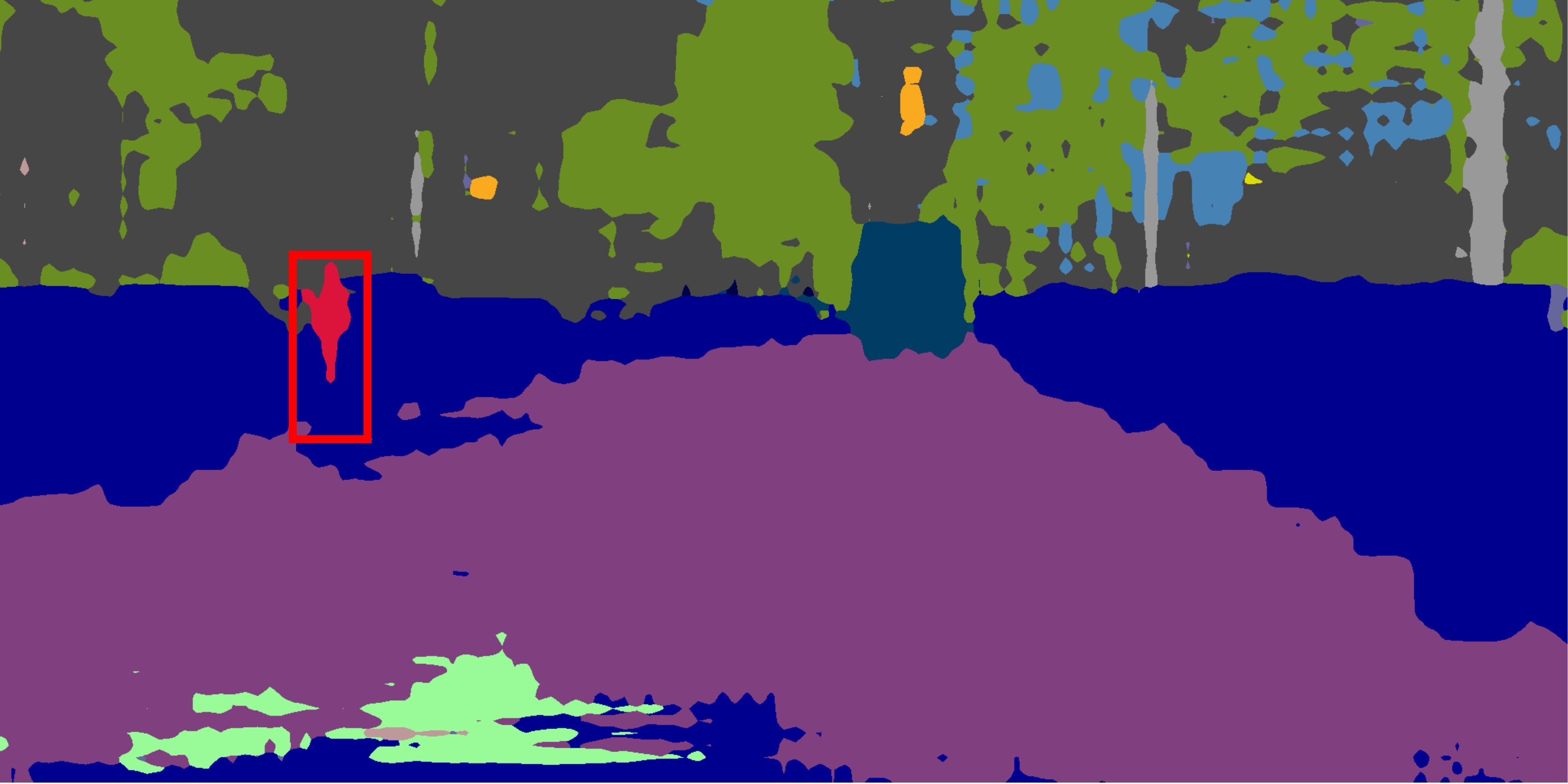}} &\hspace{-4.5mm}
			\subfloat{\includegraphics[width=0.16\linewidth]{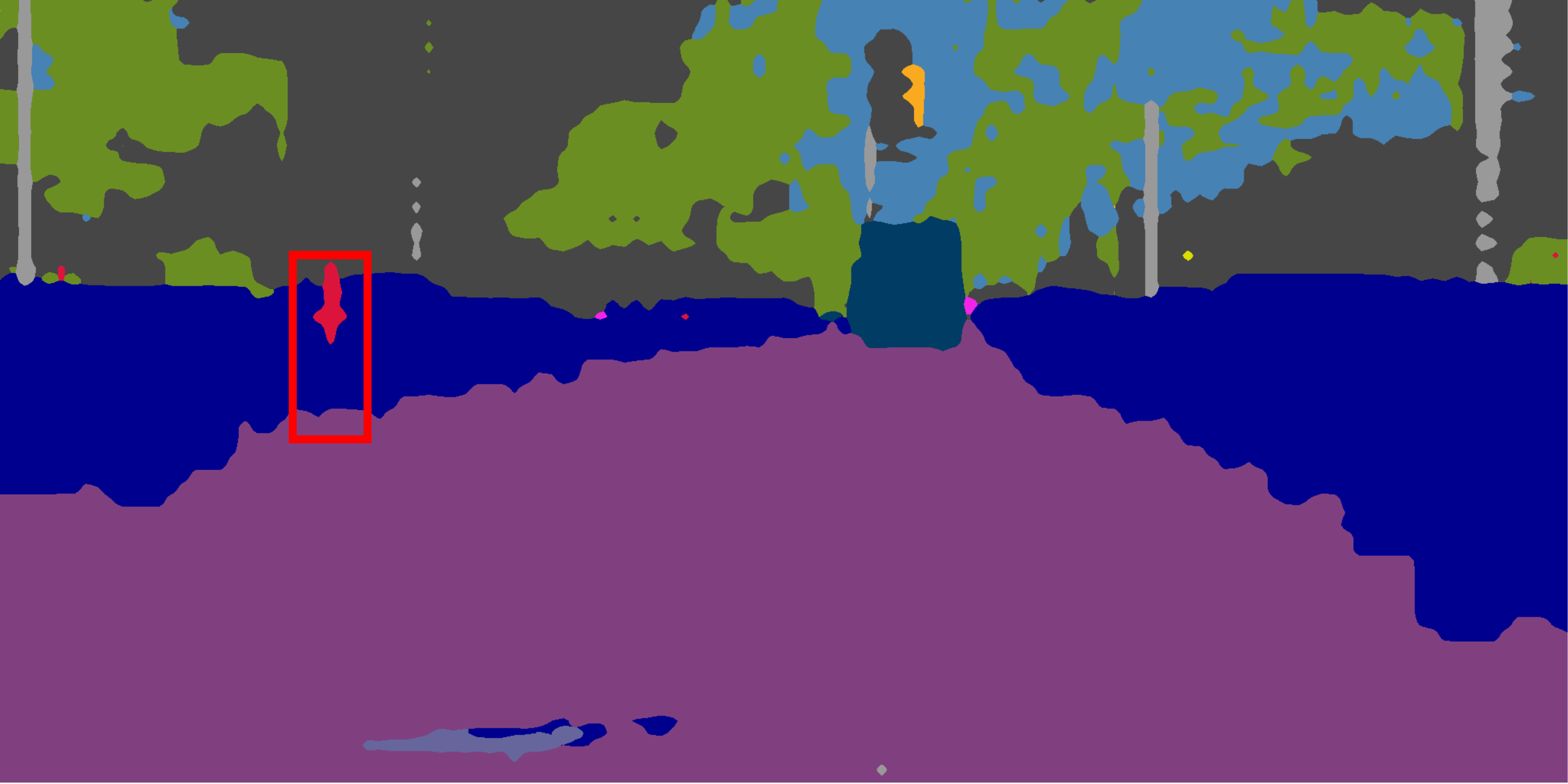}} &\hspace{-4.5mm}
			\subfloat{\includegraphics[width=0.16\linewidth]{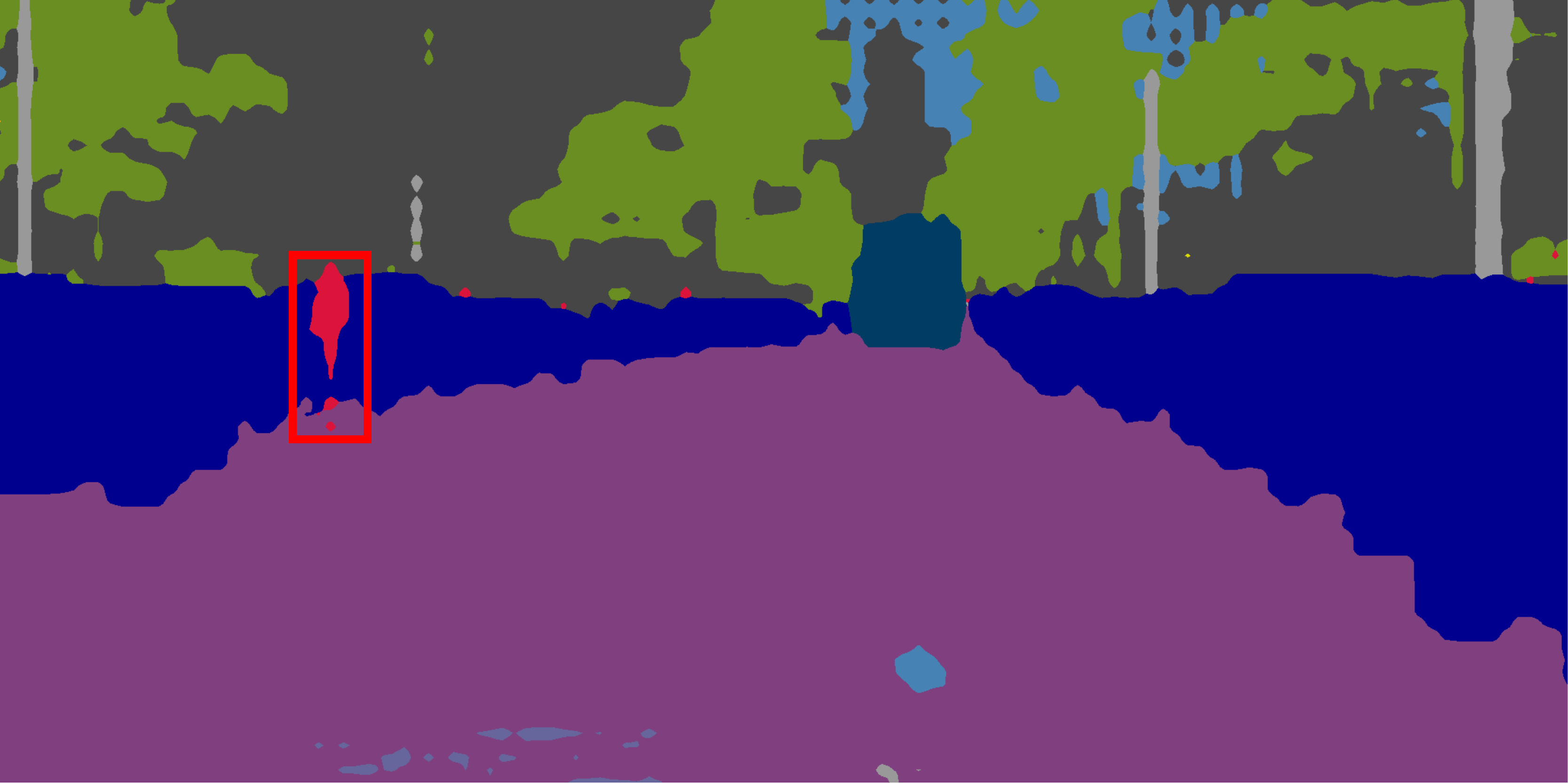}} &\hspace{-4.5mm}
			\subfloat{\includegraphics[width=0.16\linewidth]{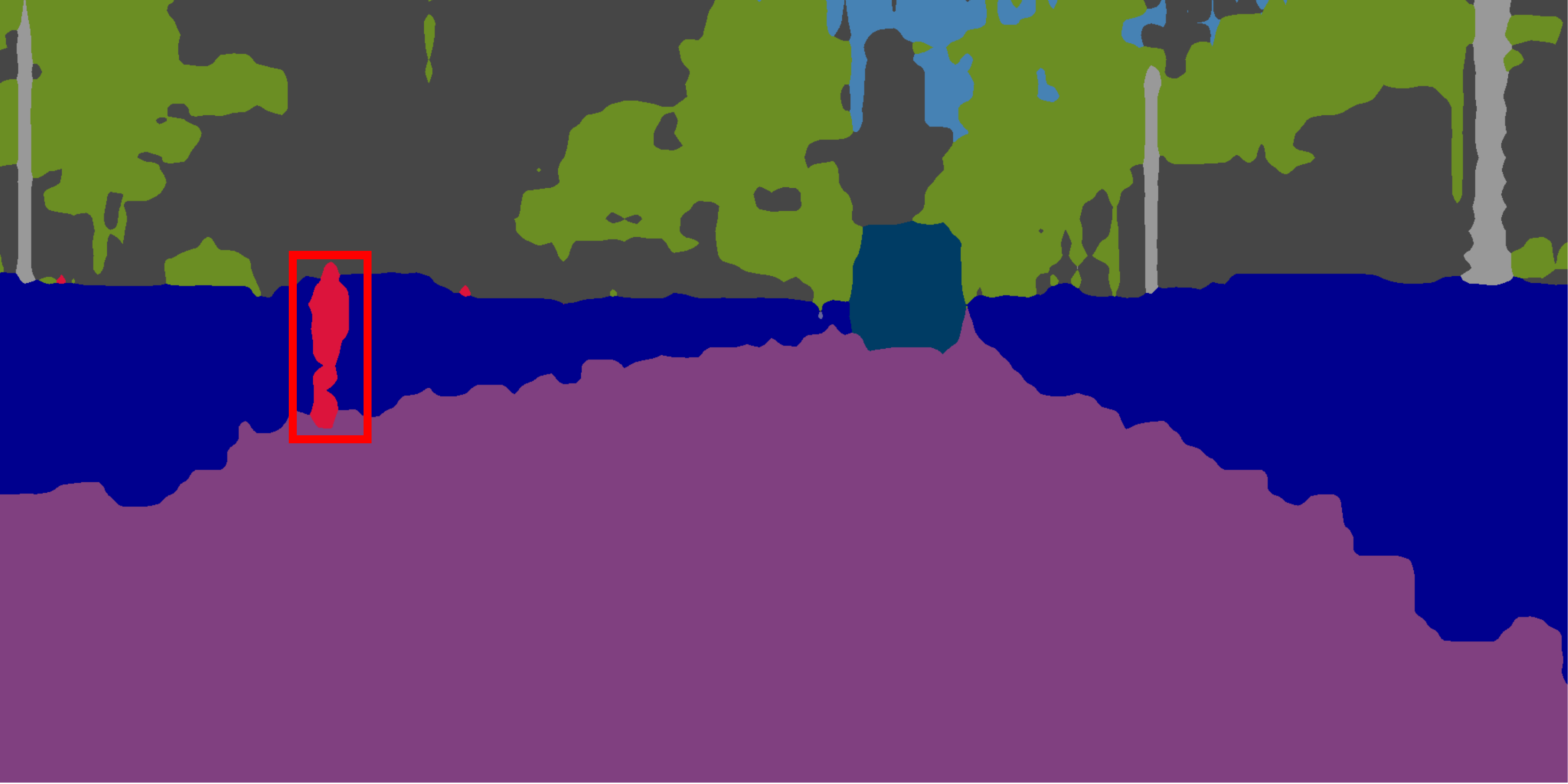}} 
			\\
			\subfloat{\includegraphics[width=0.16\linewidth]{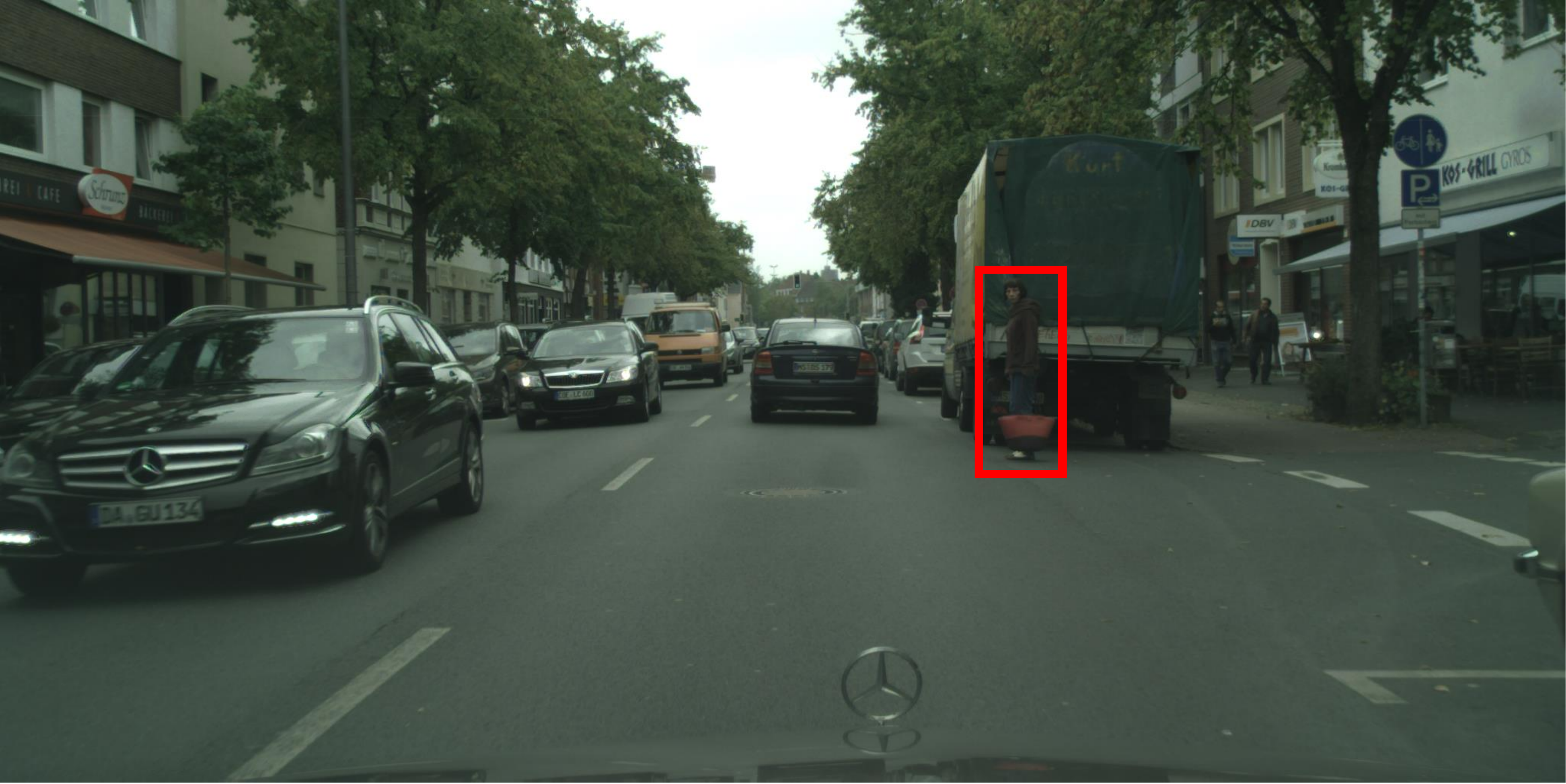}} &\hspace{-4.5mm}
			\subfloat{\includegraphics[width=0.16\linewidth]{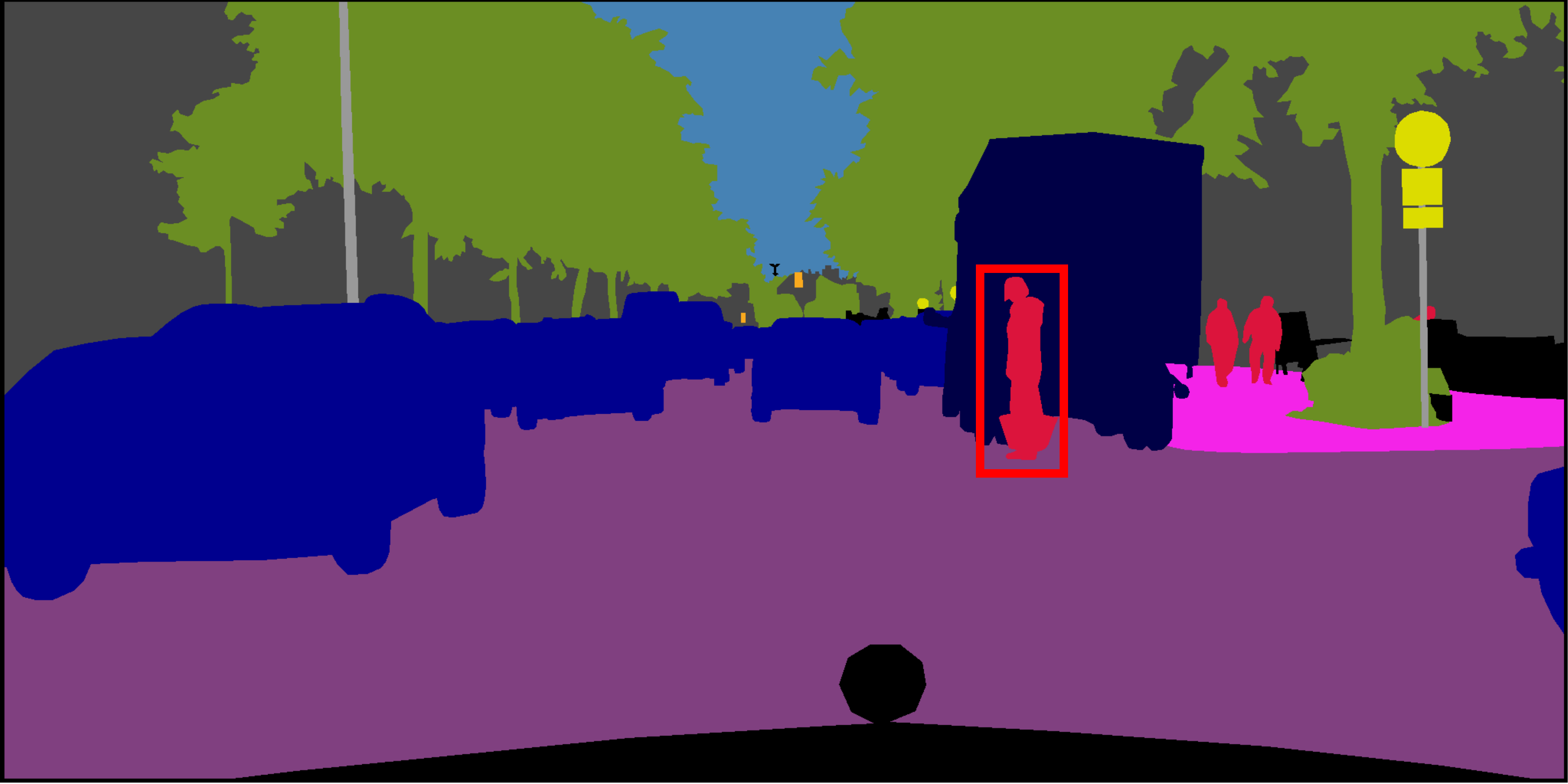}} &\hspace{-4.5mm}
			\subfloat{\includegraphics[width=0.16\linewidth]{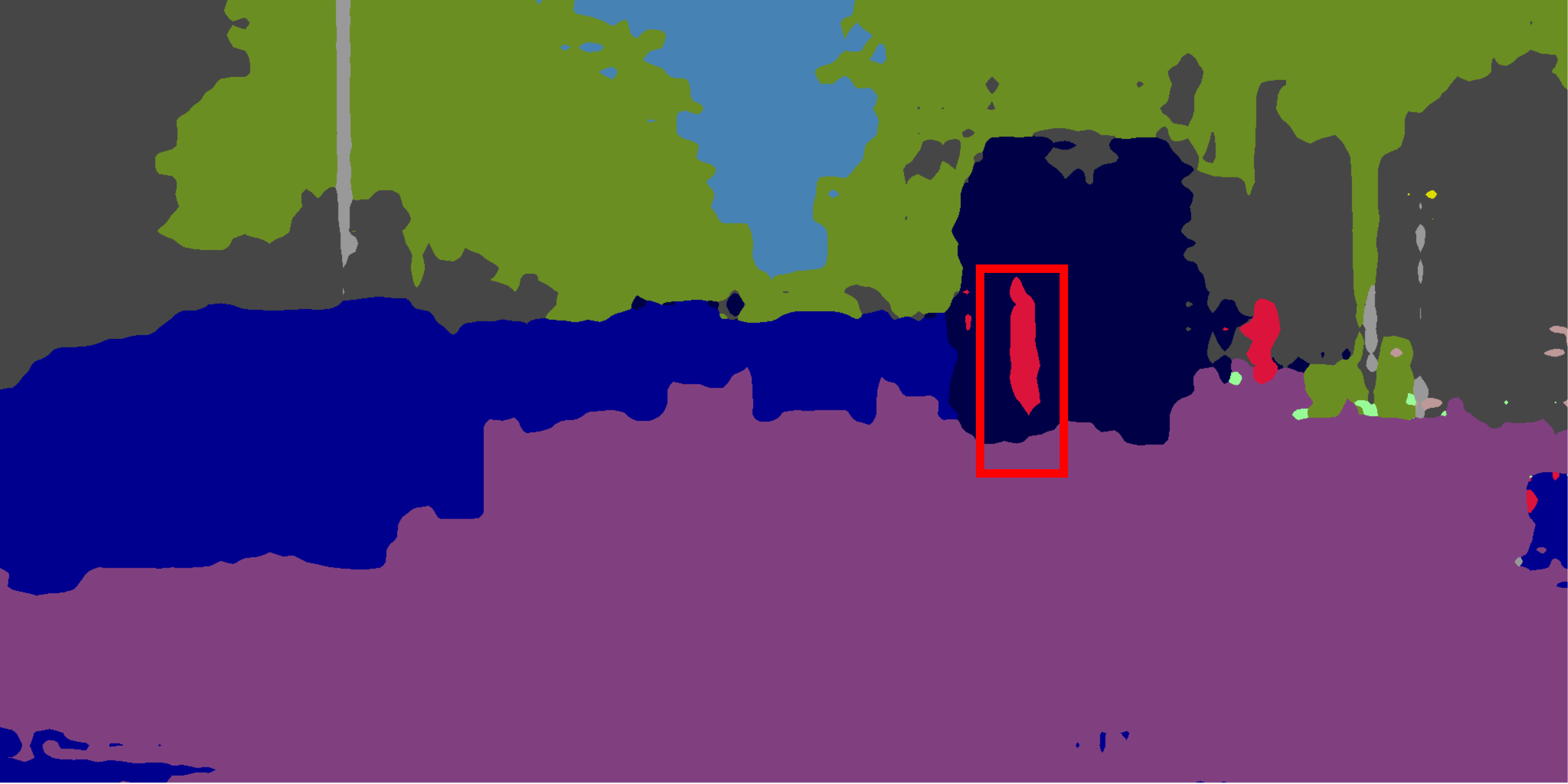}} &\hspace{-4.5mm}
			\subfloat{\includegraphics[width=0.16\linewidth]{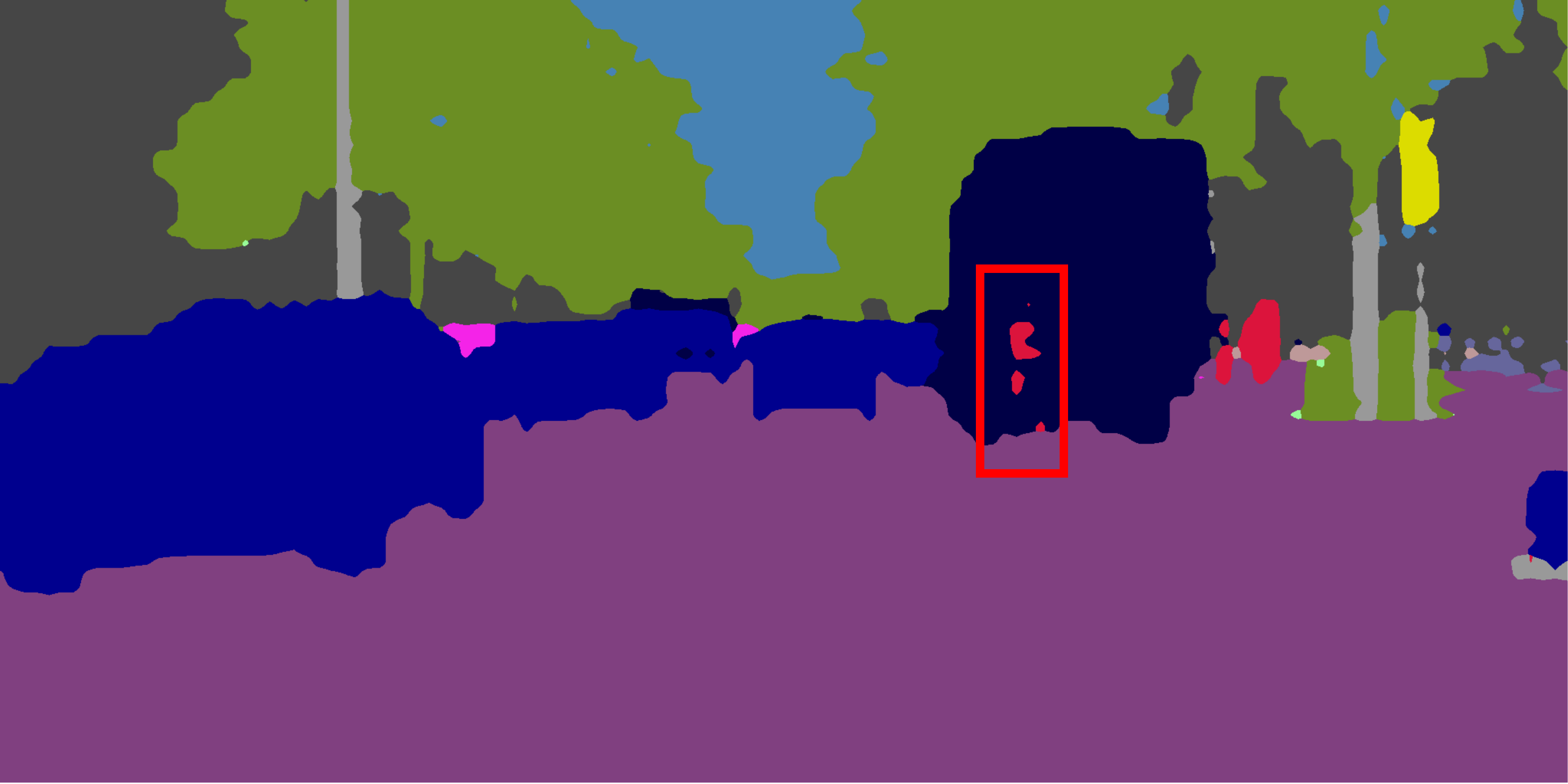}} &\hspace{-4.5mm}
			\subfloat{\includegraphics[width=0.16\linewidth]{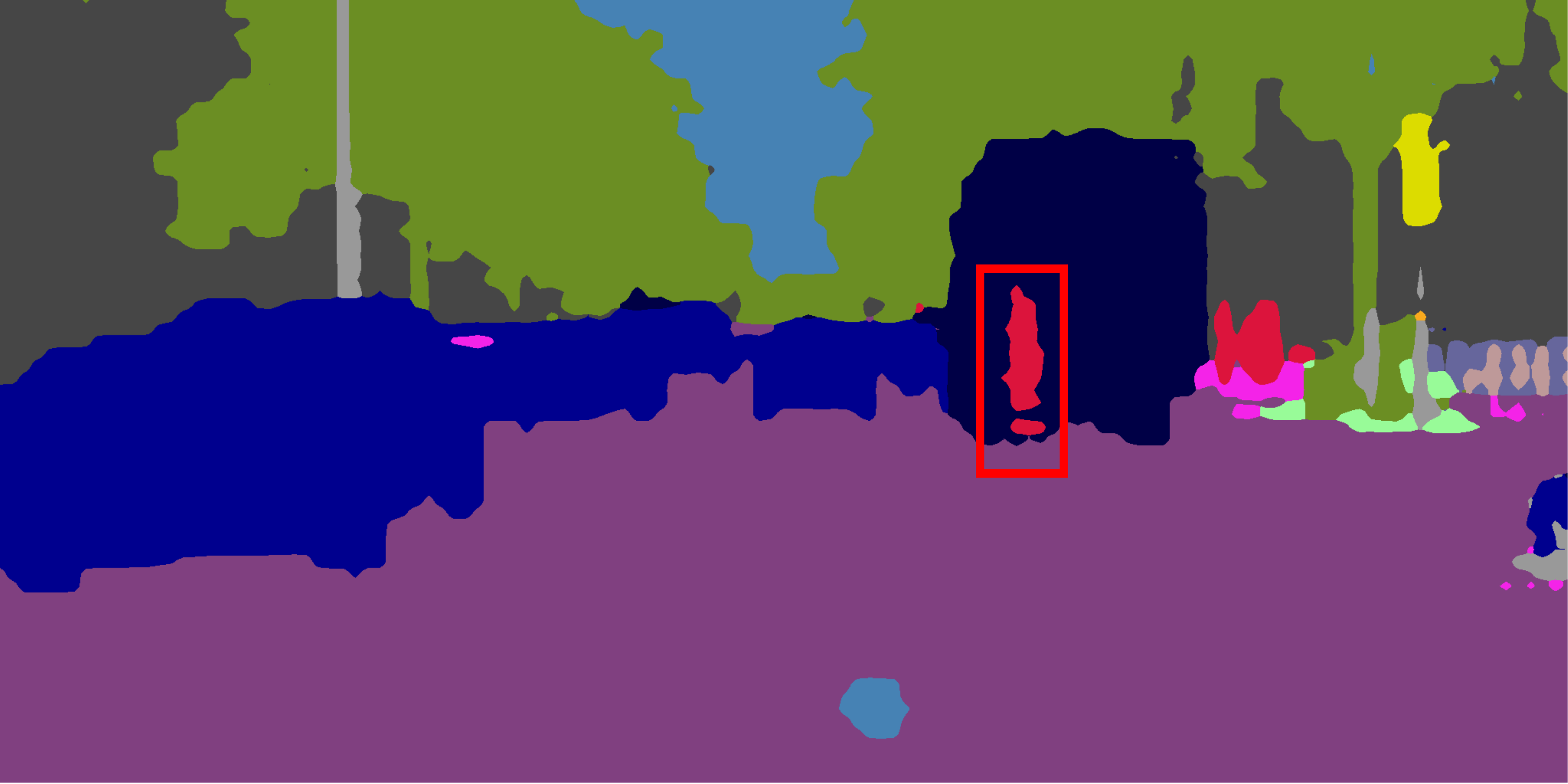}} &\hspace{-4.5mm}
			\subfloat{\includegraphics[width=0.16\linewidth]{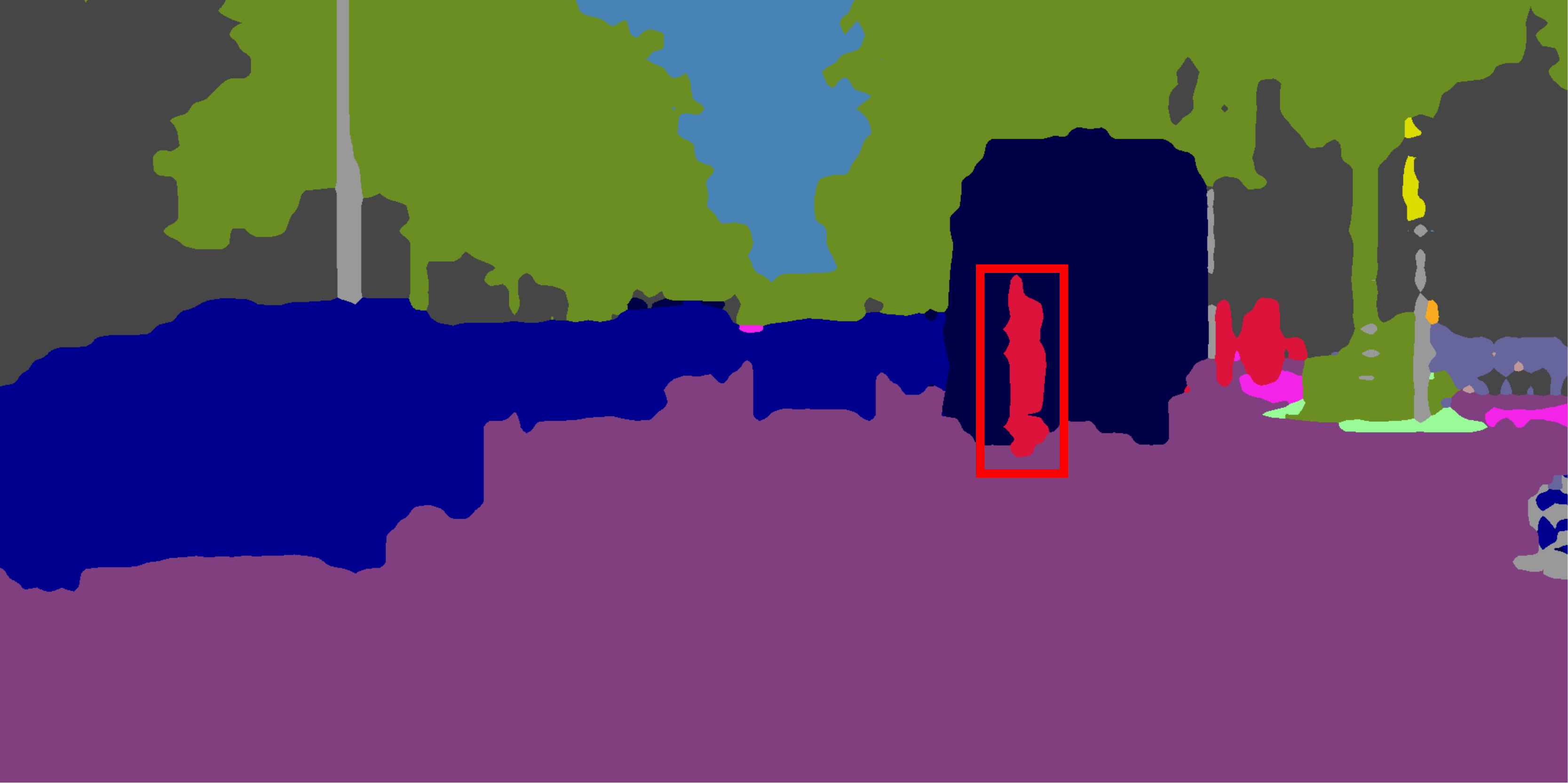}} 
			\\
			\subfloat{\includegraphics[width=0.16\linewidth]{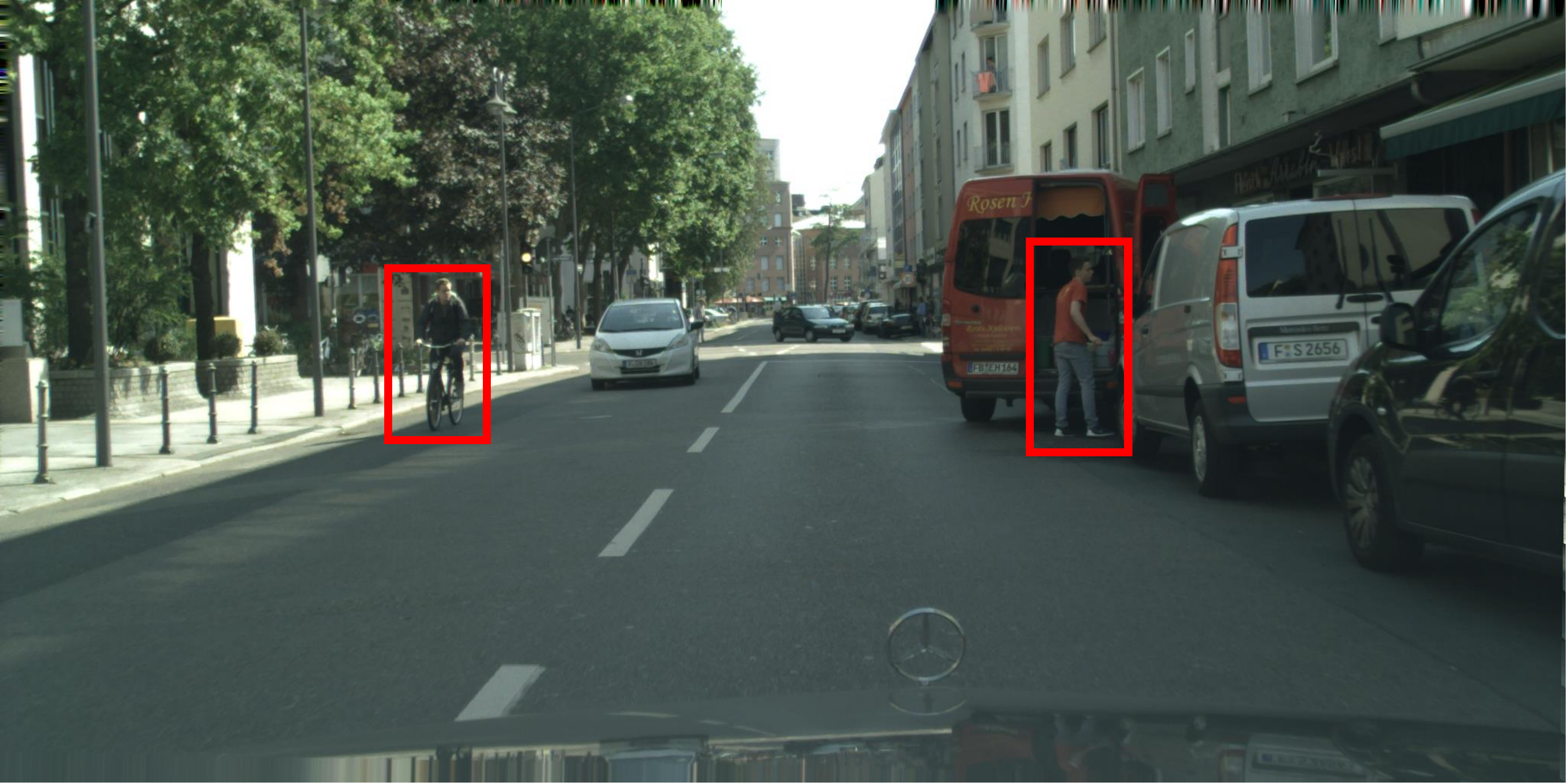}} &\hspace{-4.5mm}
			\subfloat{\includegraphics[width=0.16\linewidth]{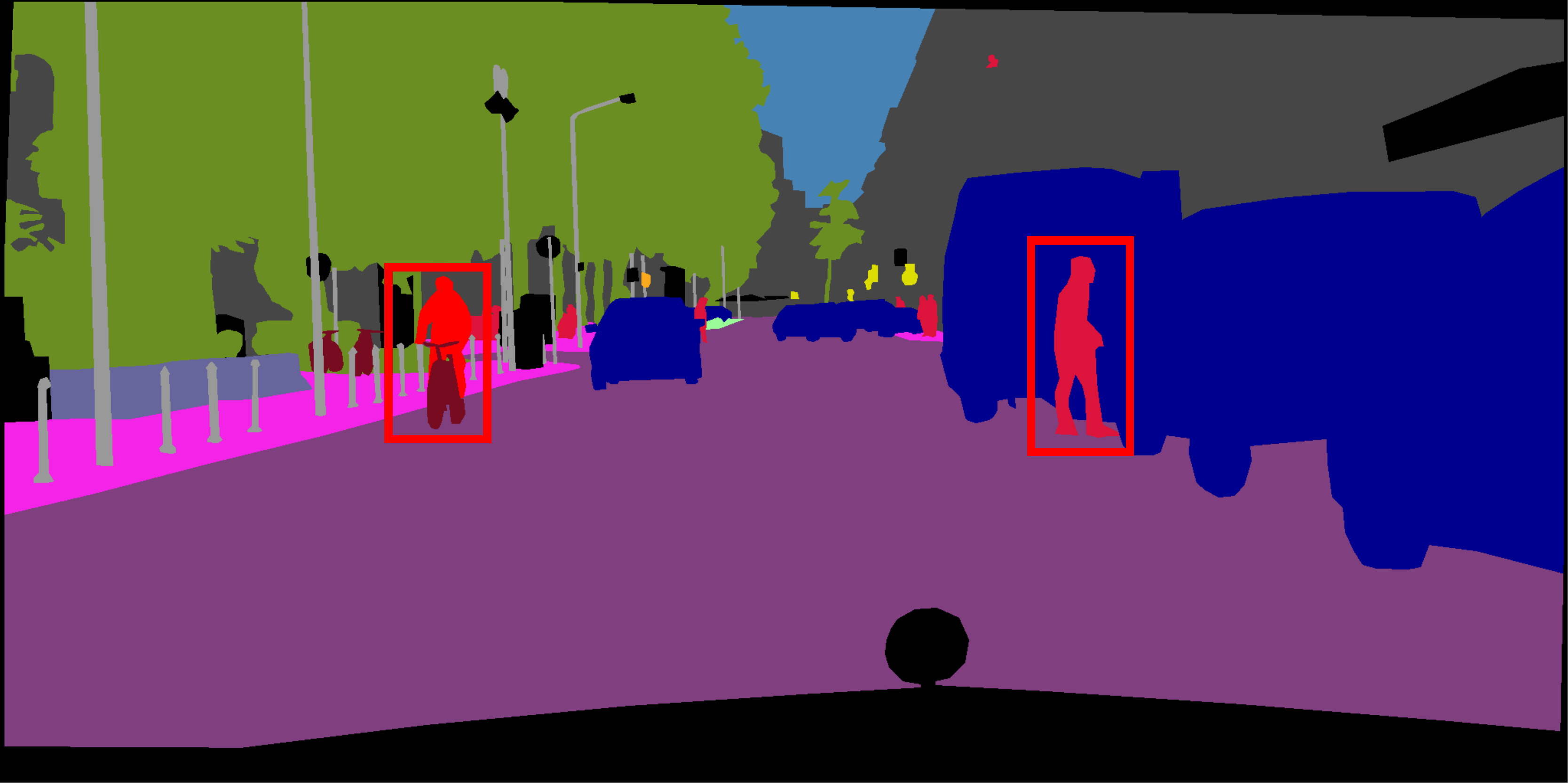}} &\hspace{-4.5mm}
			\subfloat{\includegraphics[width=0.16\linewidth]{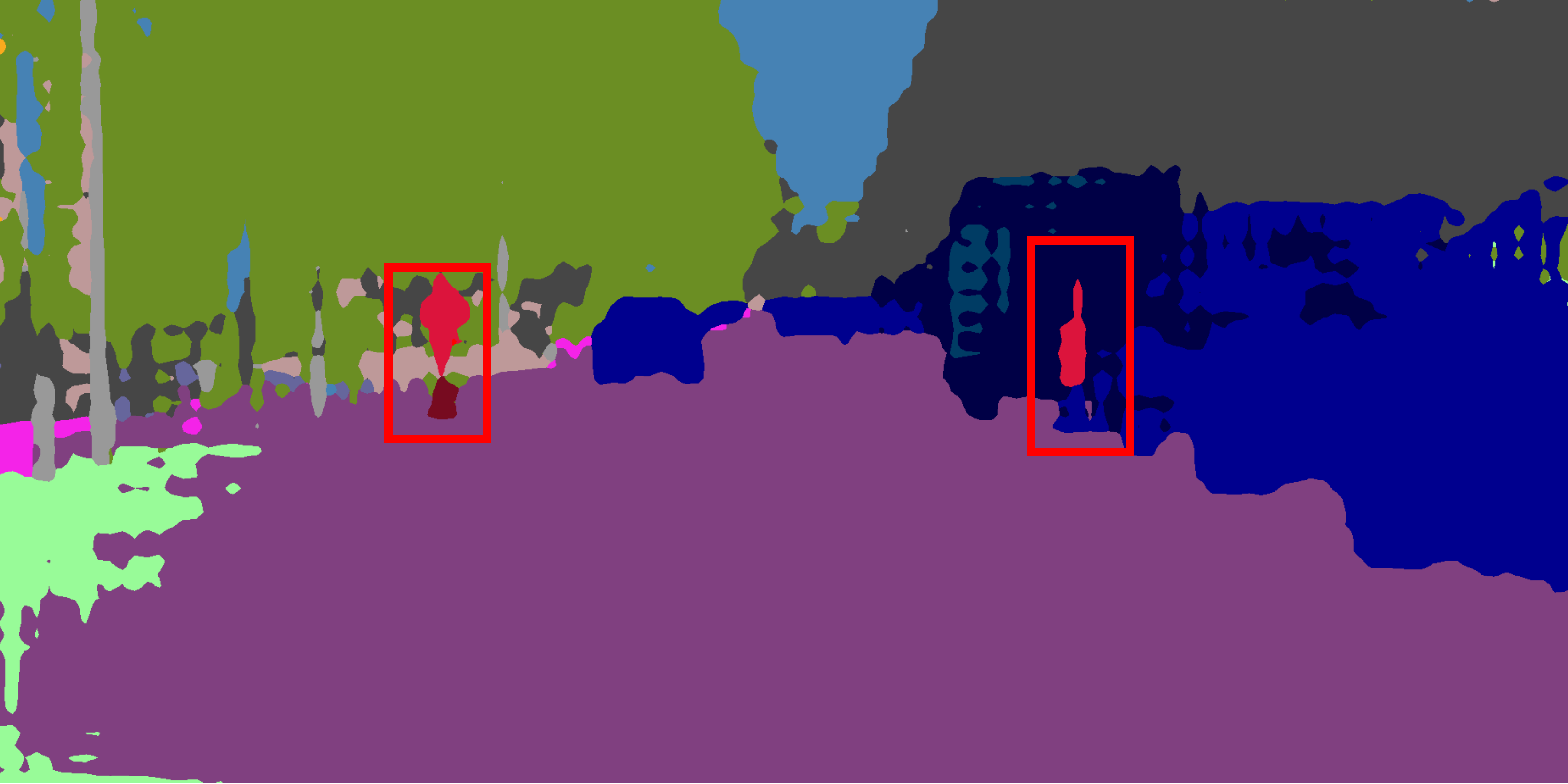}} &\hspace{-4.5mm}
			\subfloat{\includegraphics[width=0.16\linewidth]{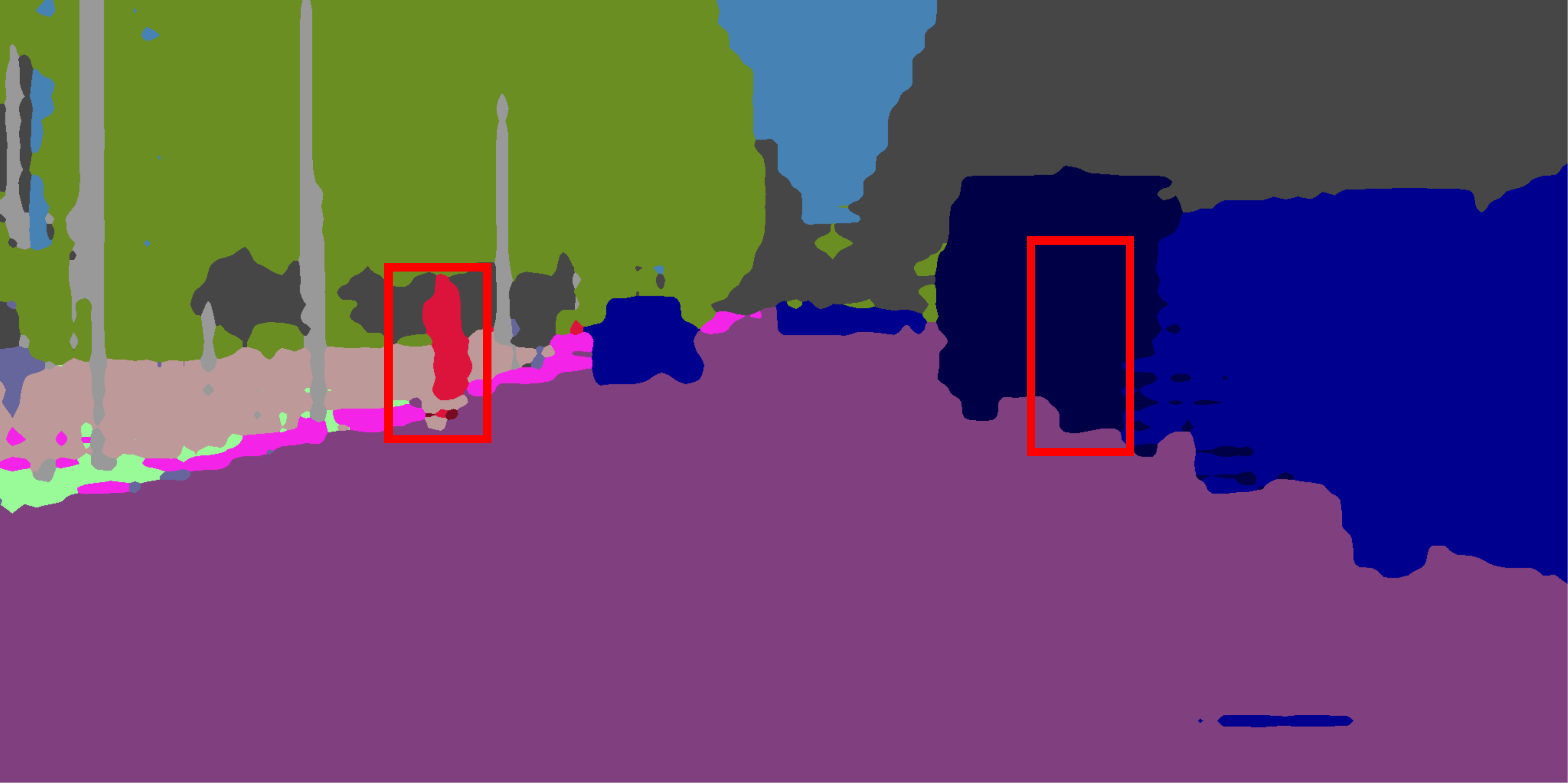}} &\hspace{-4.5mm}
			\subfloat{\includegraphics[width=0.16\linewidth]{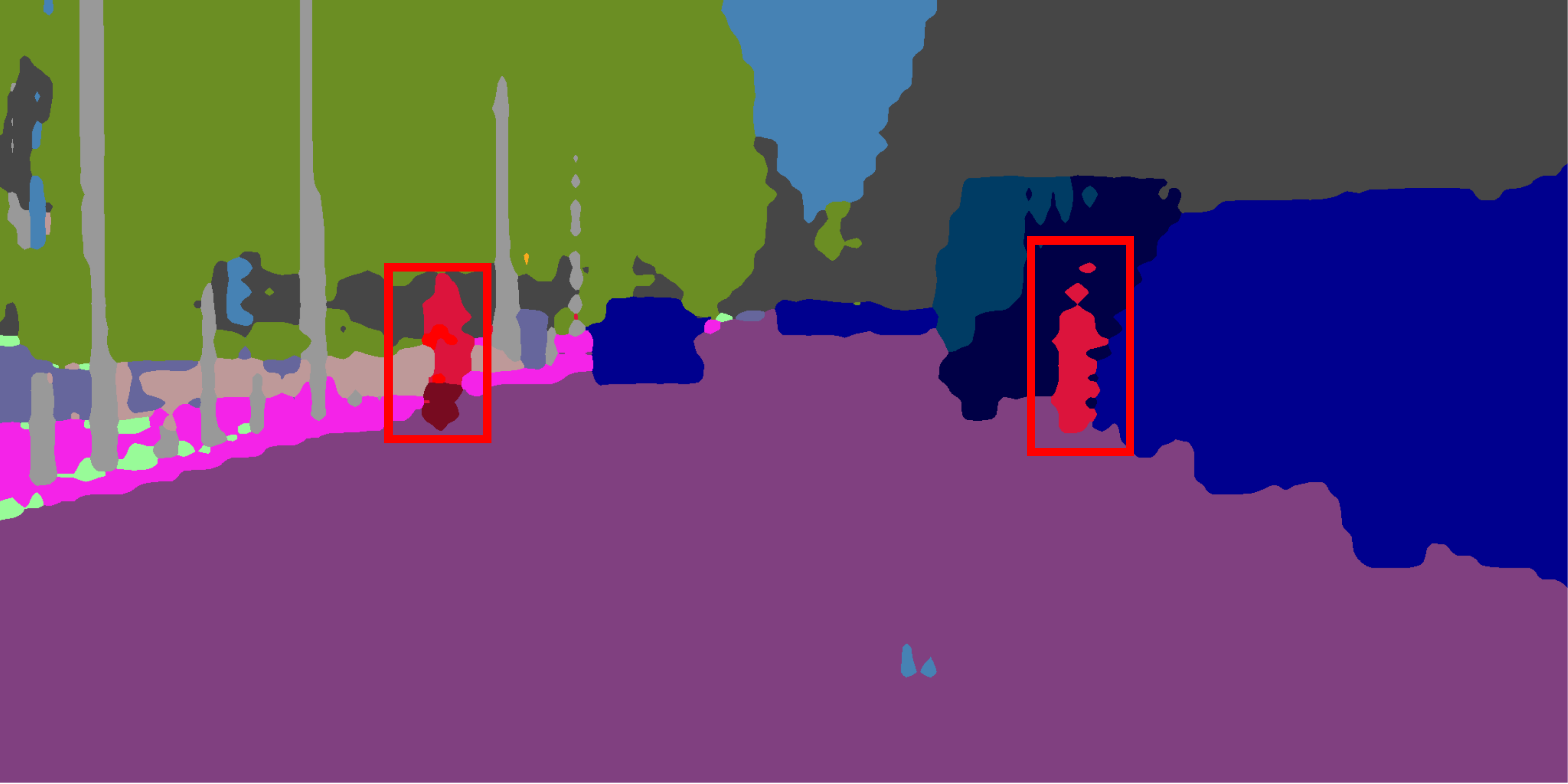}} &\hspace{-4.5mm}
			\subfloat{\includegraphics[width=0.16\linewidth]{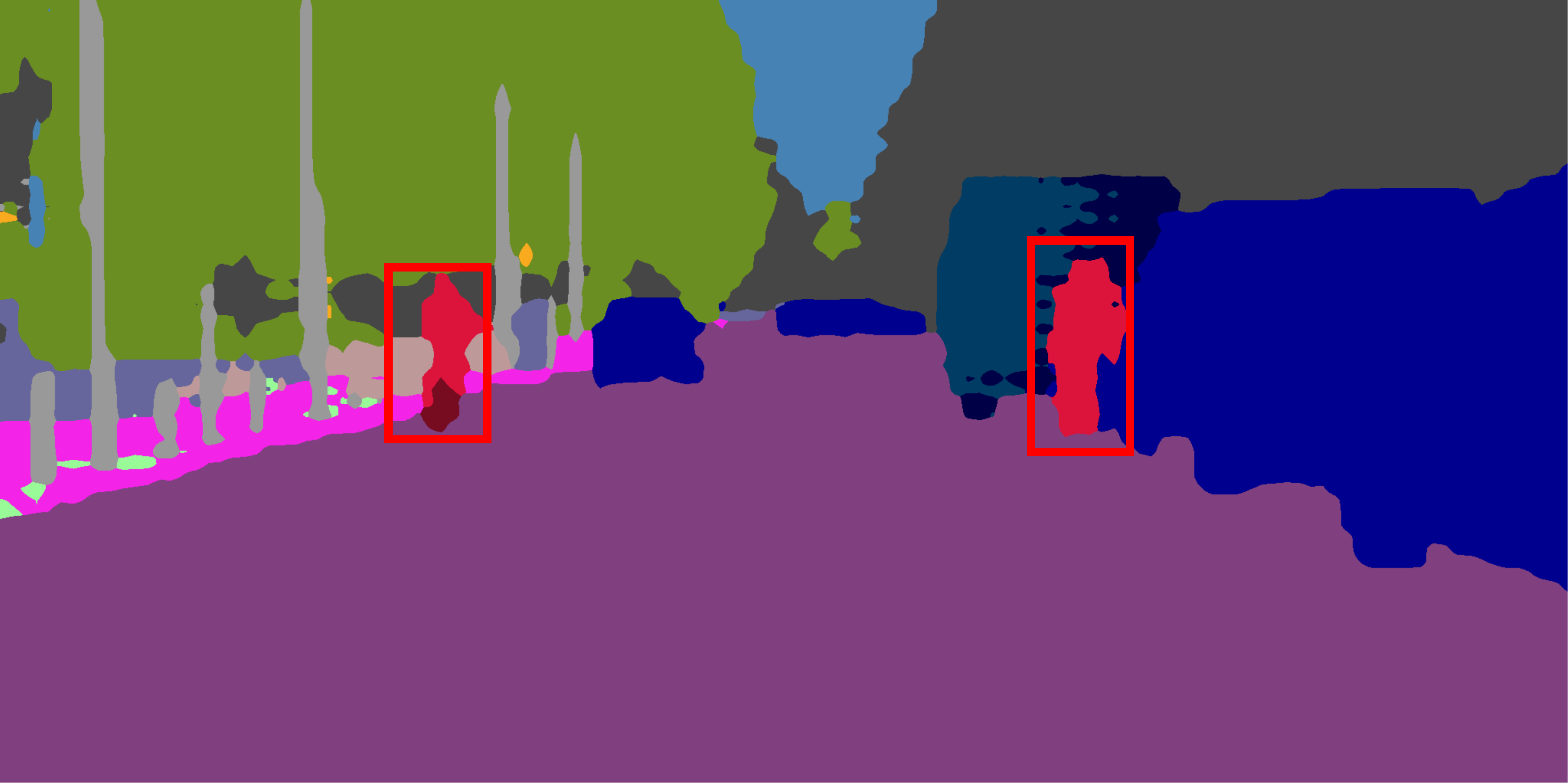}} 
			\\
			\subfloat{\includegraphics[width=0.16\linewidth]{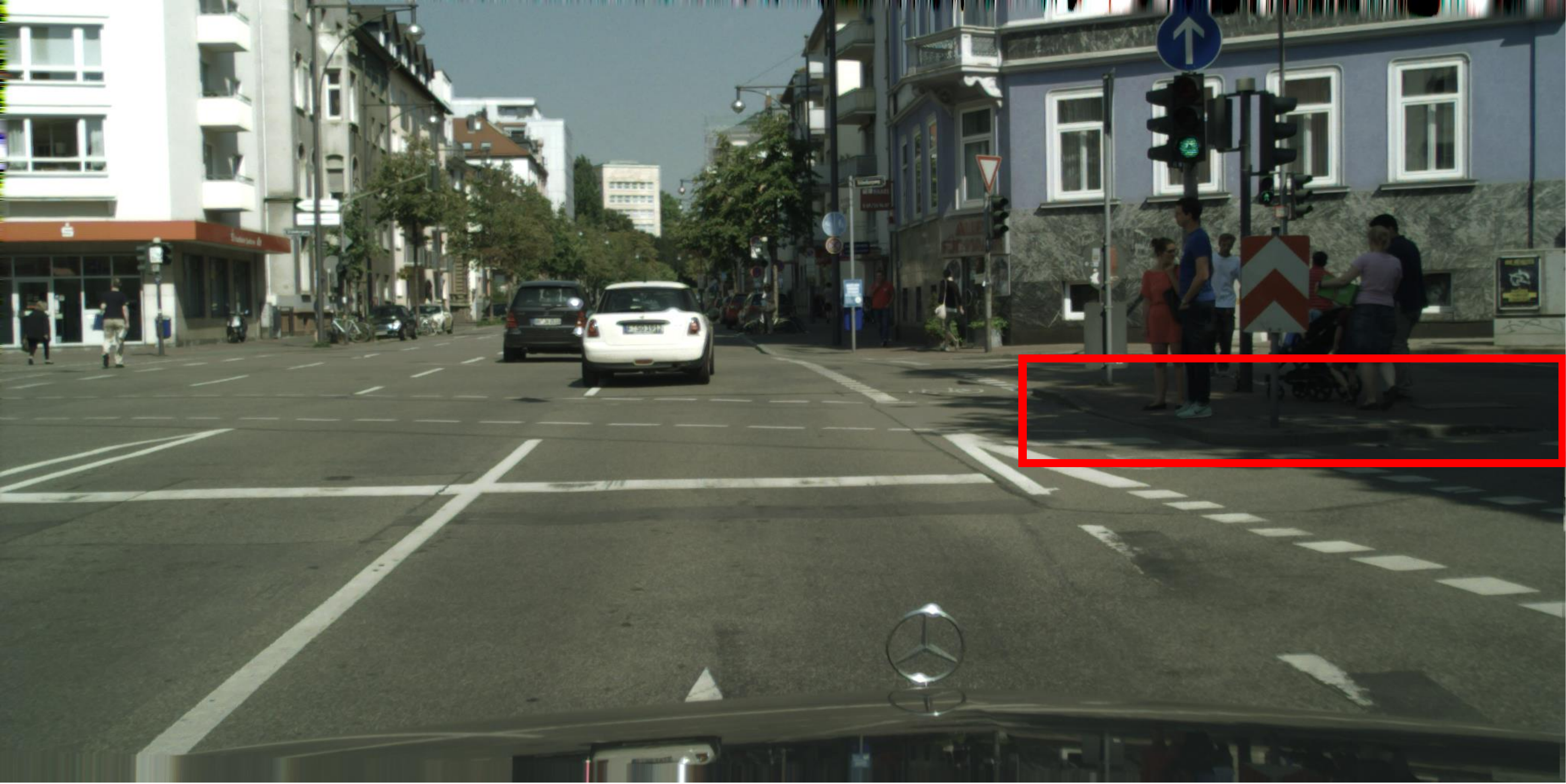}} &\hspace{-4.5mm}
			\subfloat{\includegraphics[width=0.16\linewidth]{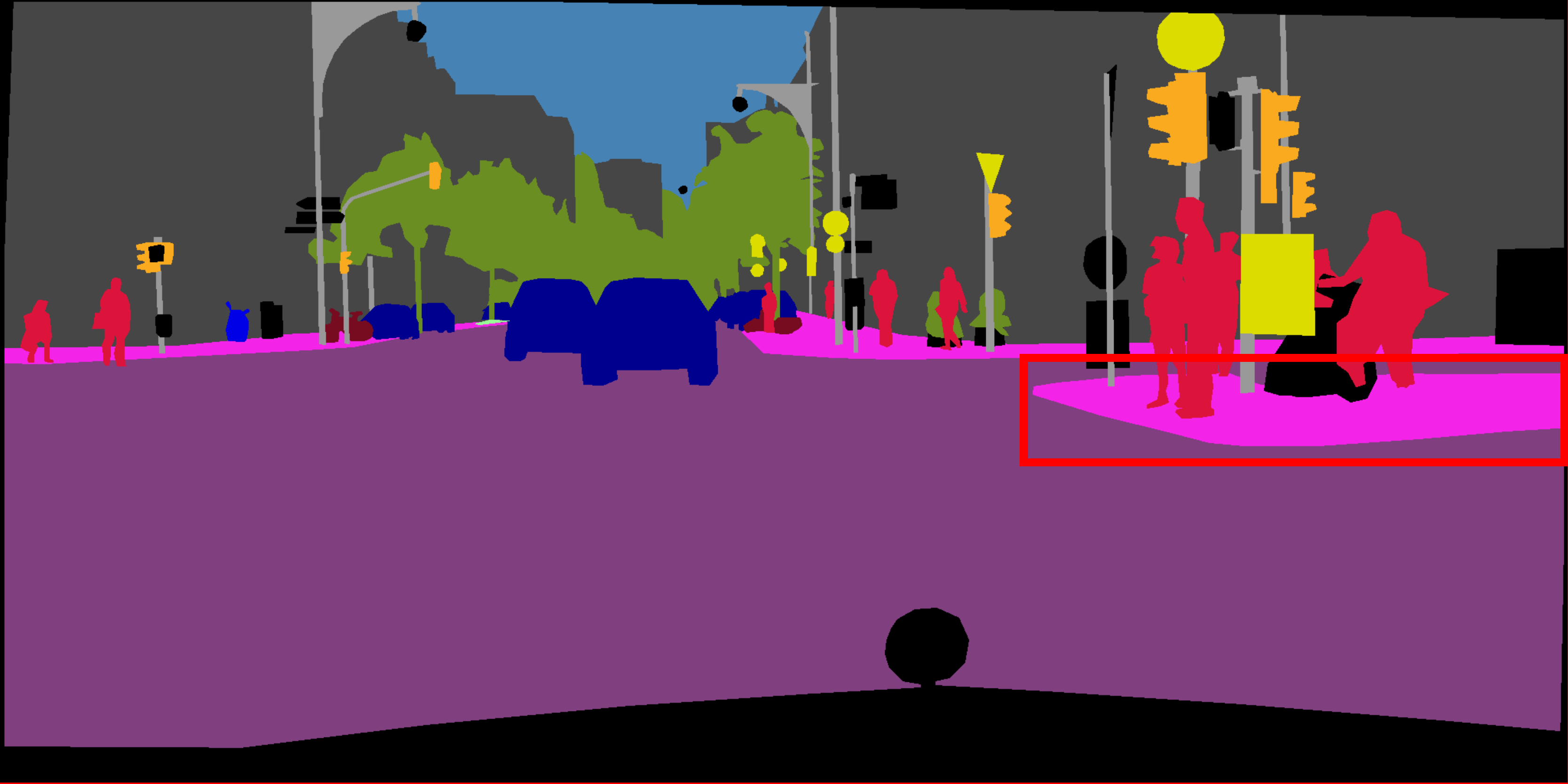}} &\hspace{-4.5mm}
			\subfloat{\includegraphics[width=0.16\linewidth]{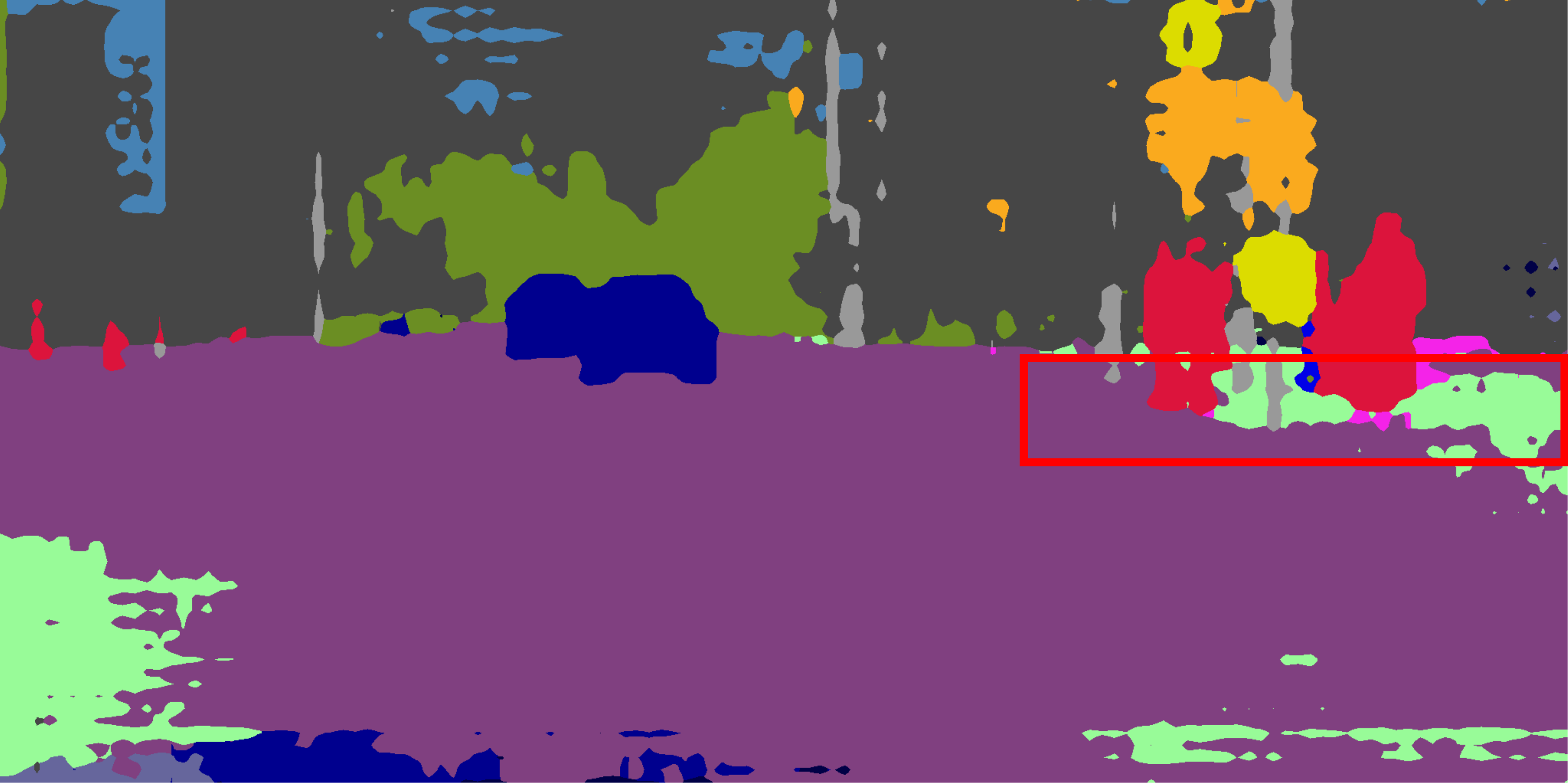}} &\hspace{-4.5mm}
			\subfloat{\includegraphics[width=0.16\linewidth]{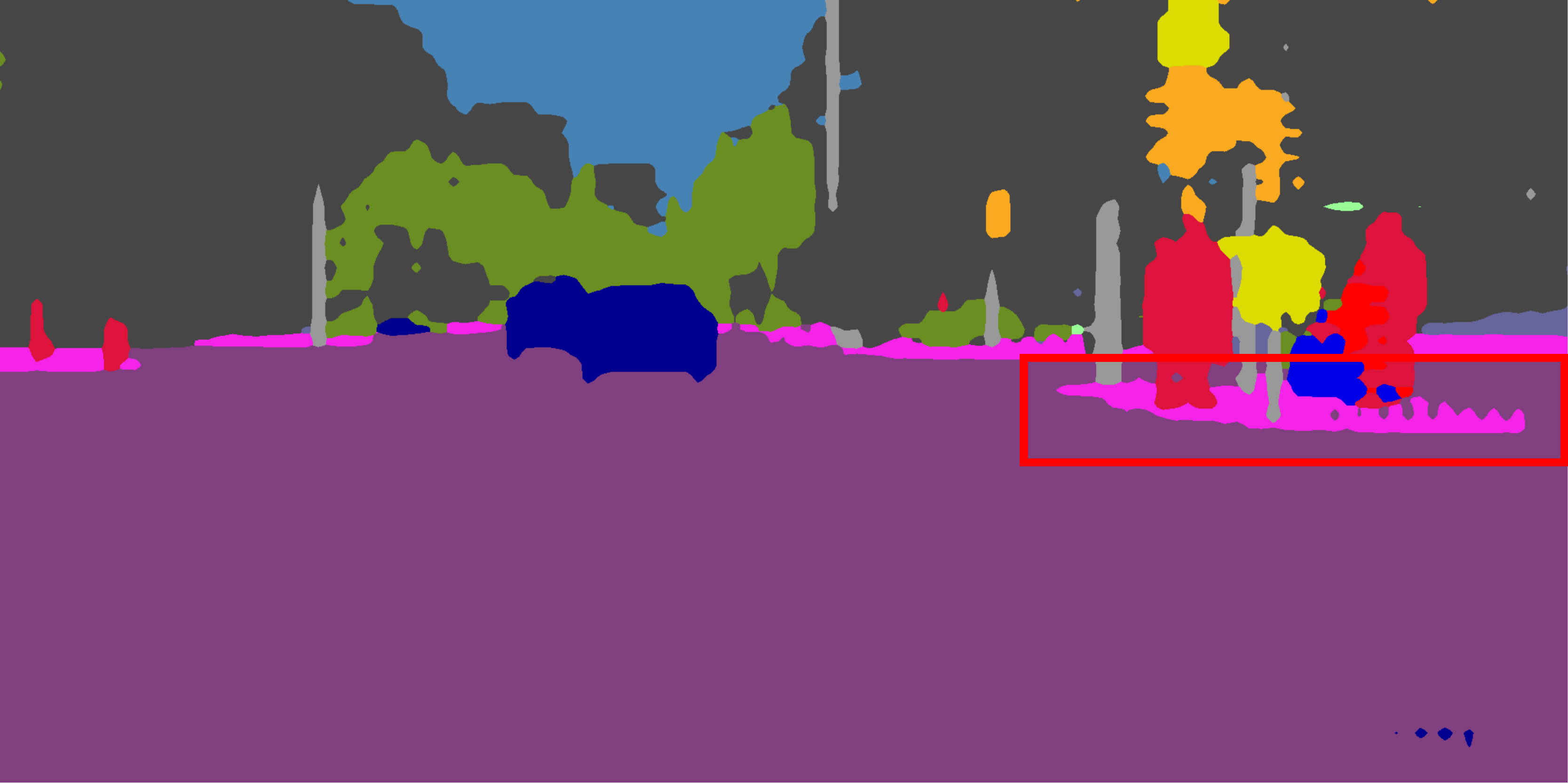}} &\hspace{-4.5mm}
			\subfloat{\includegraphics[width=0.16\linewidth]{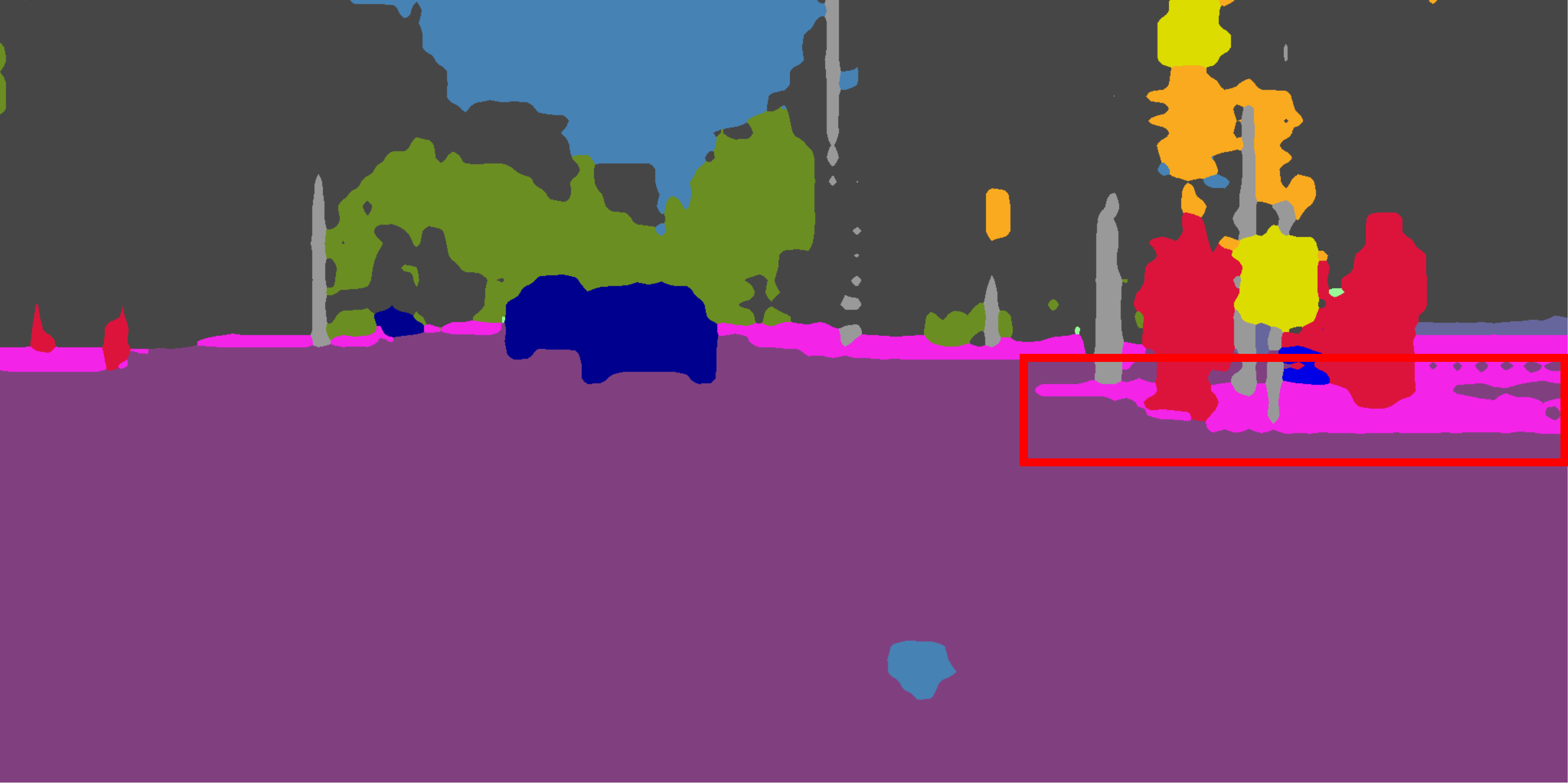}} &\hspace{-4.5mm}
			\subfloat{\includegraphics[width=0.16\linewidth]{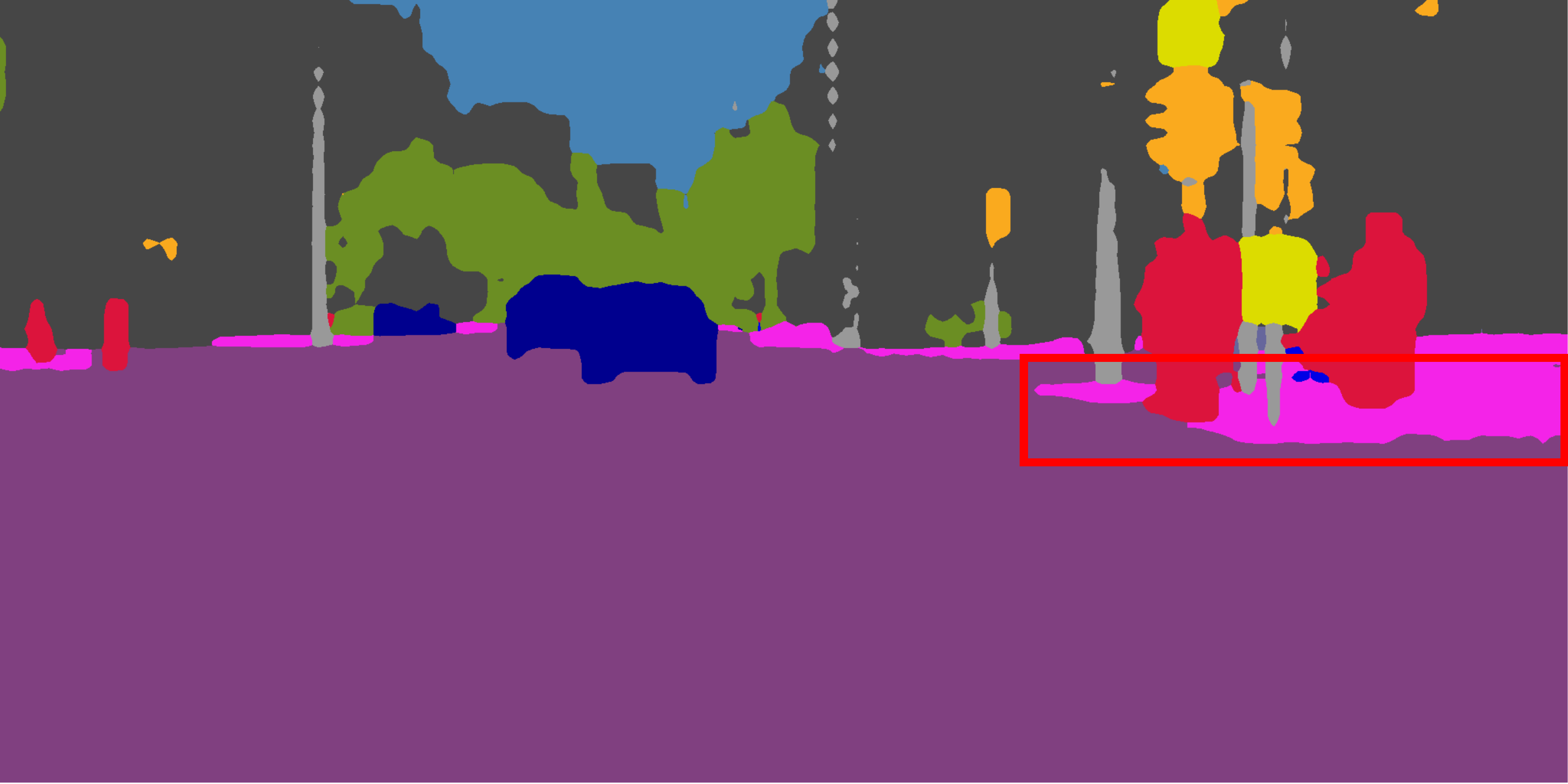}} 
			\\
			\subfloat{\includegraphics[width=0.16\linewidth]{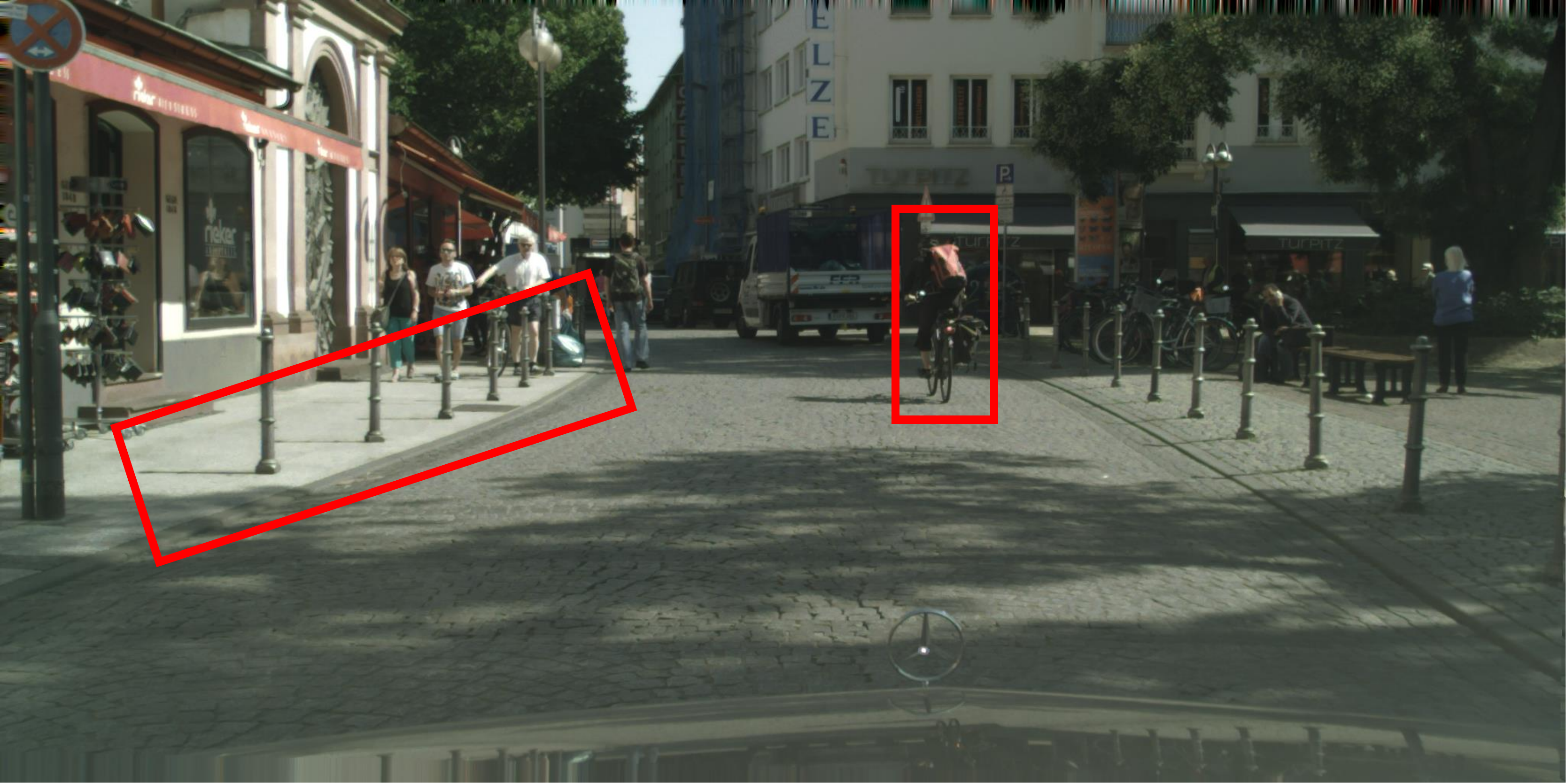}} &\hspace{-4.5mm}
			\subfloat{\includegraphics[width=0.16\linewidth]{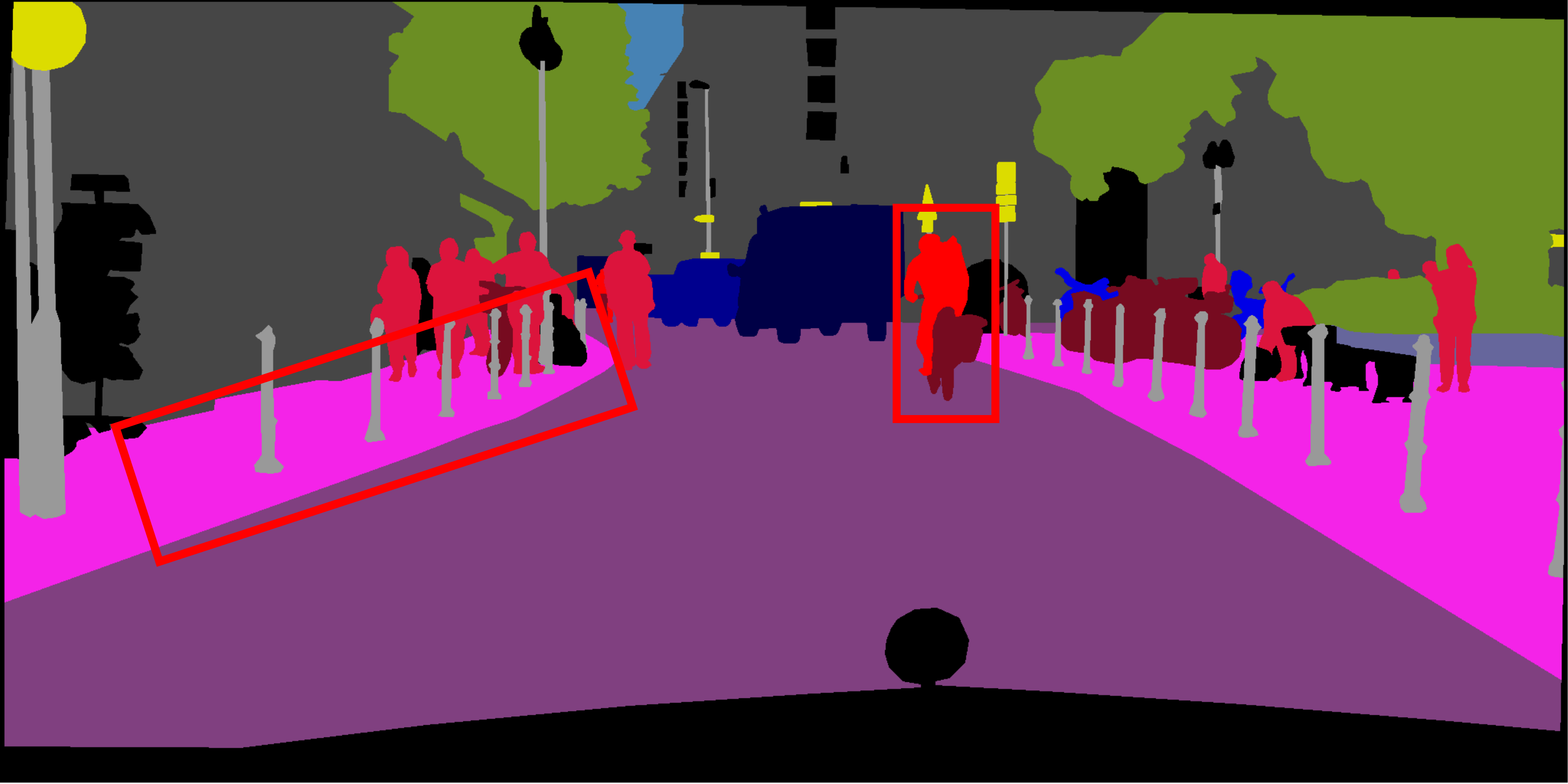}} &\hspace{-4.5mm}
			\subfloat{\includegraphics[width=0.16\linewidth]{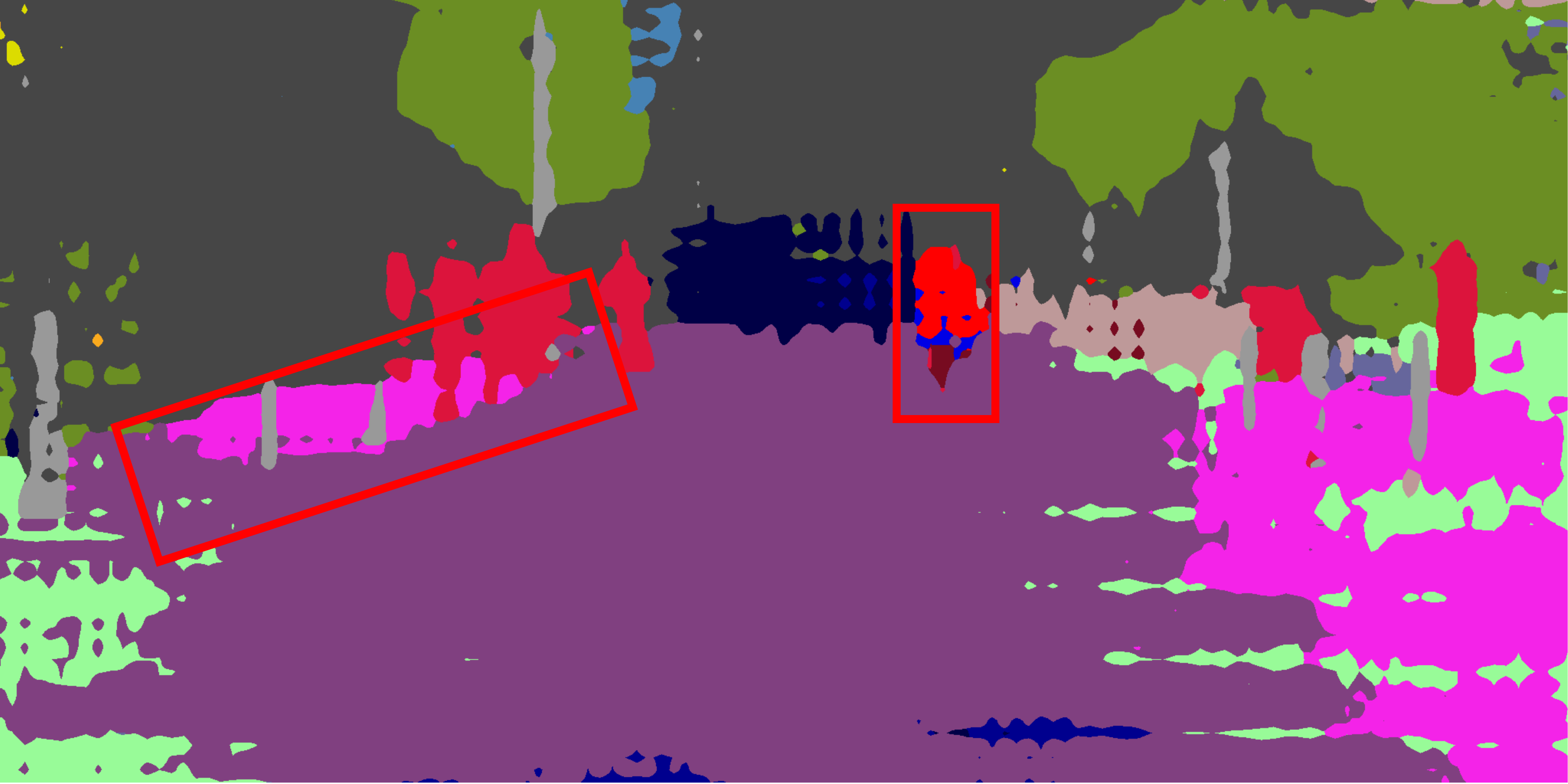}} &\hspace{-4.5mm}
			\subfloat{\includegraphics[width=0.16\linewidth]{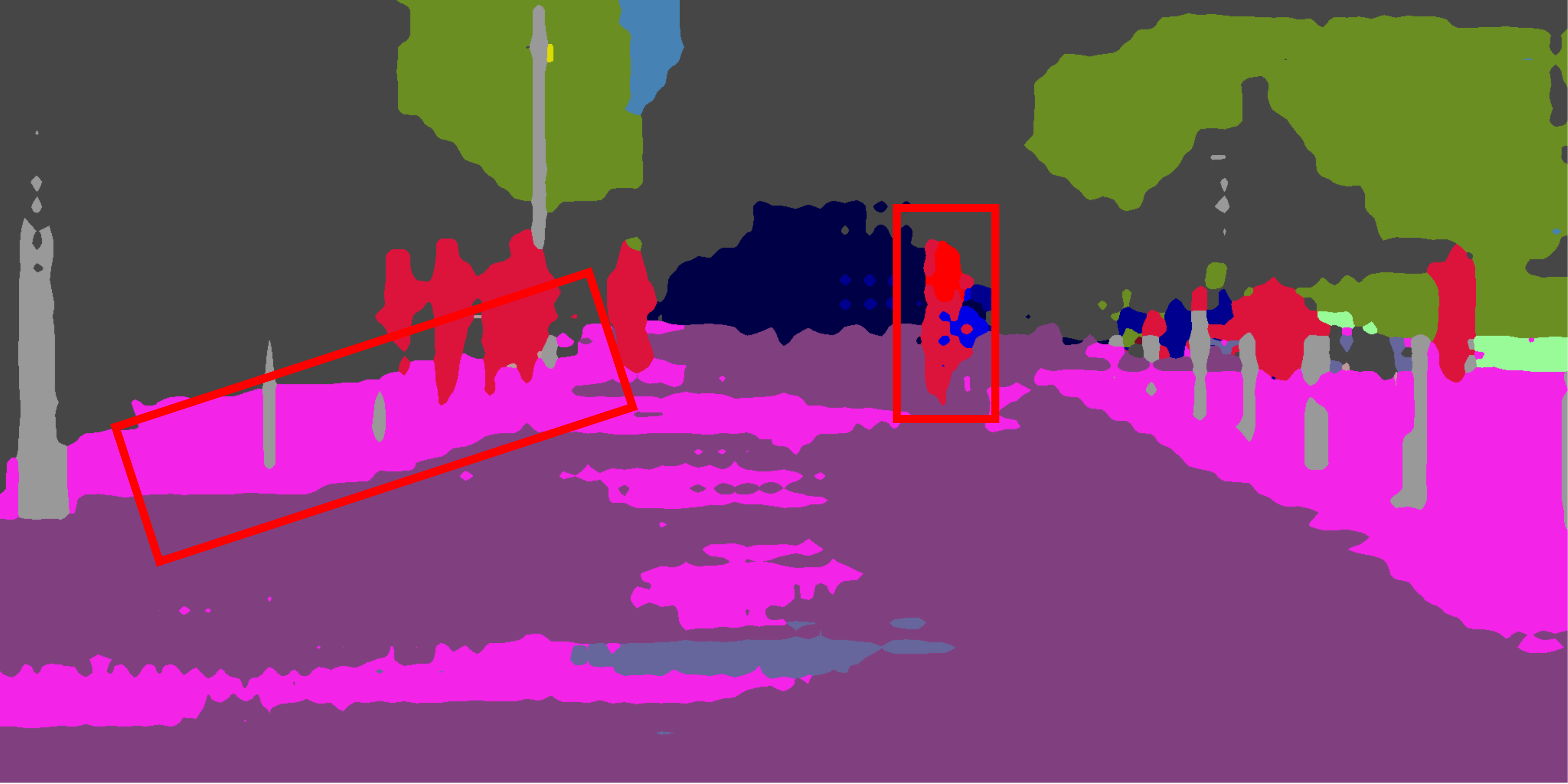}} &\hspace{-4.5mm}
			\subfloat{\includegraphics[width=0.16\linewidth]{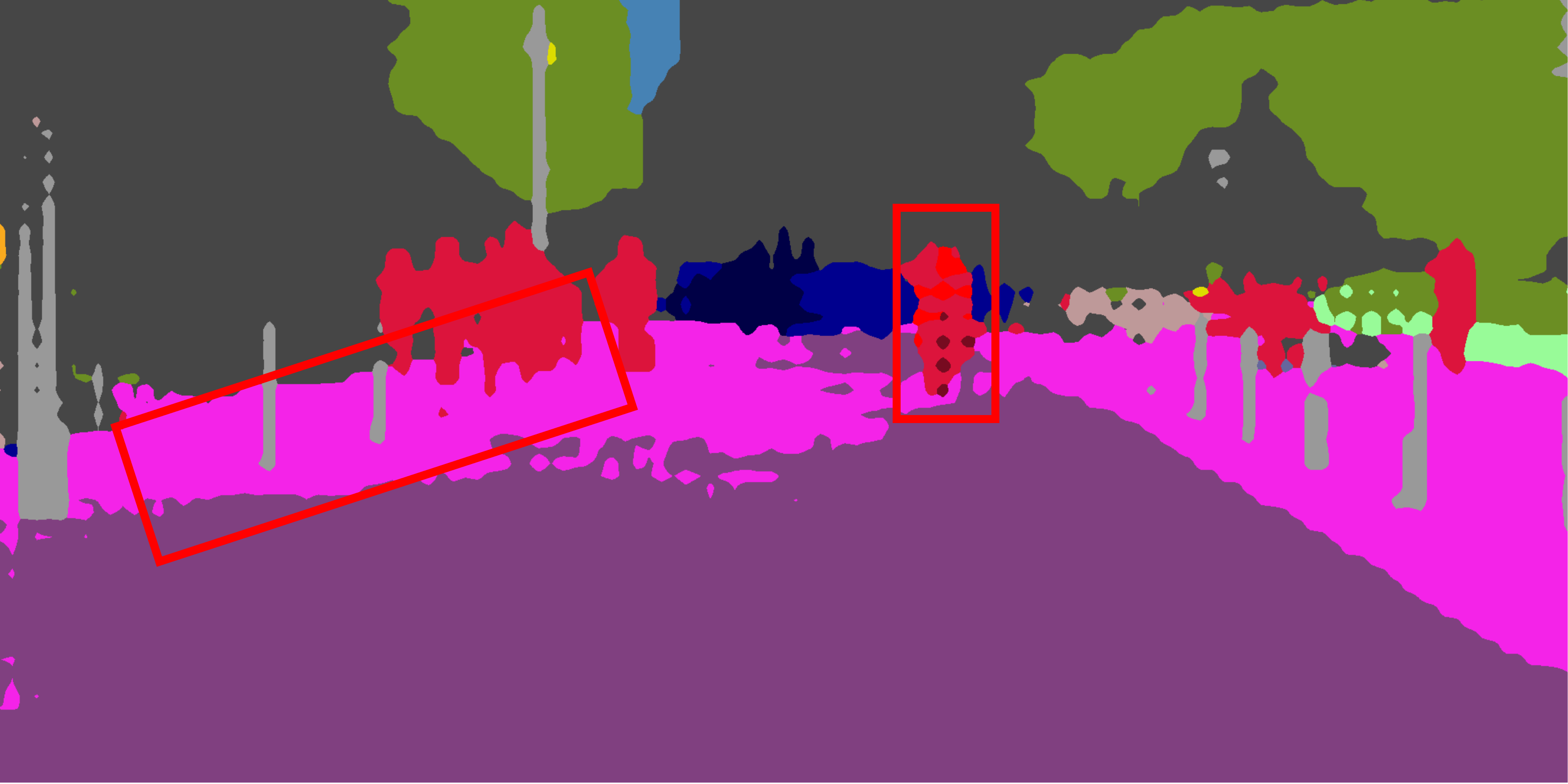}} &\hspace{-4.5mm}
			\subfloat{\includegraphics[width=0.16\linewidth]{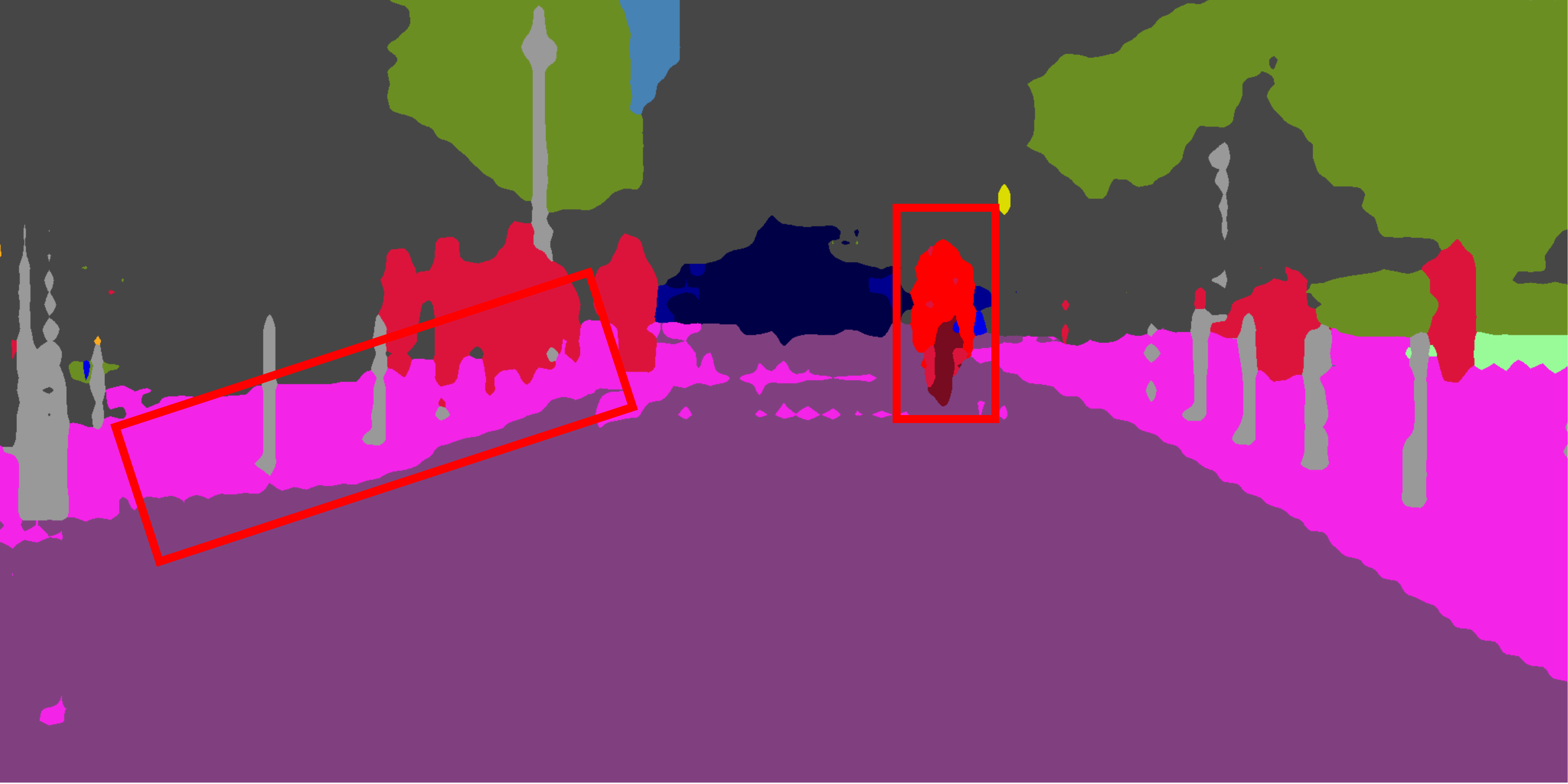}} 
			\\	
			
			\subfloat{\includegraphics[width=0.16\linewidth]{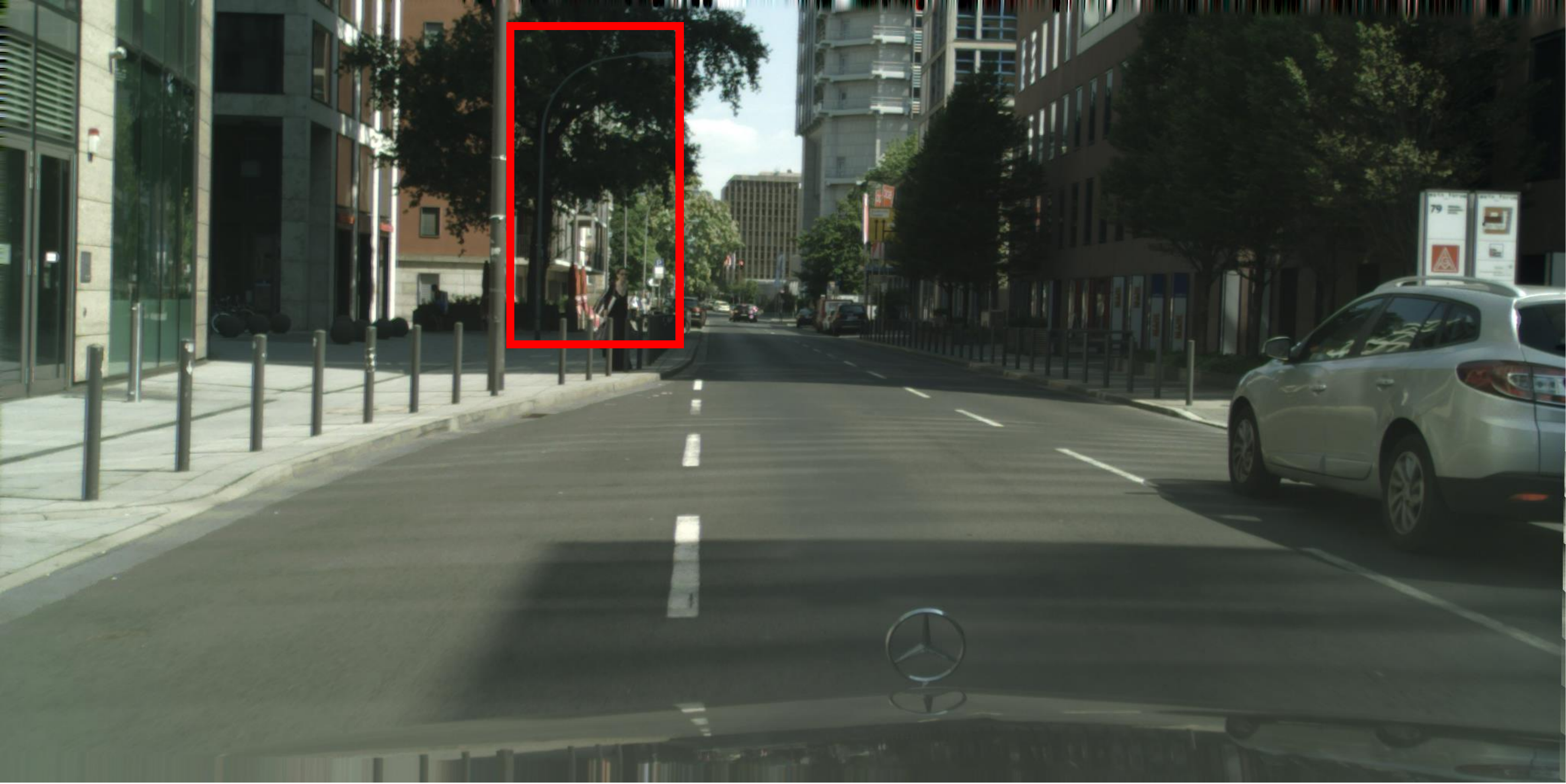}} &\hspace{-4.5mm}			
			\subfloat{\includegraphics[width=0.16\linewidth]{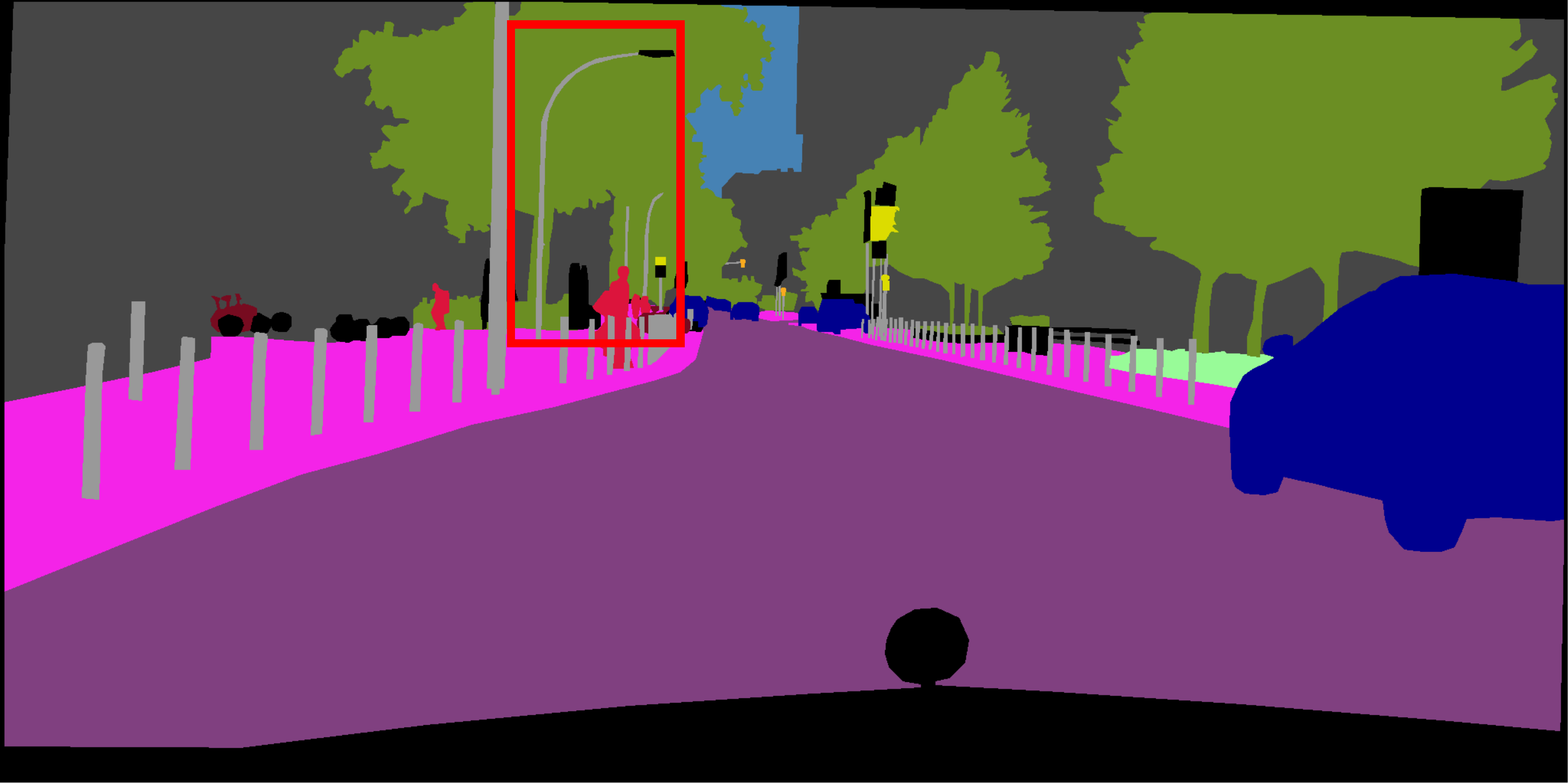}}&\hspace{-4.5mm}
			\subfloat{\includegraphics[width=0.16\linewidth]{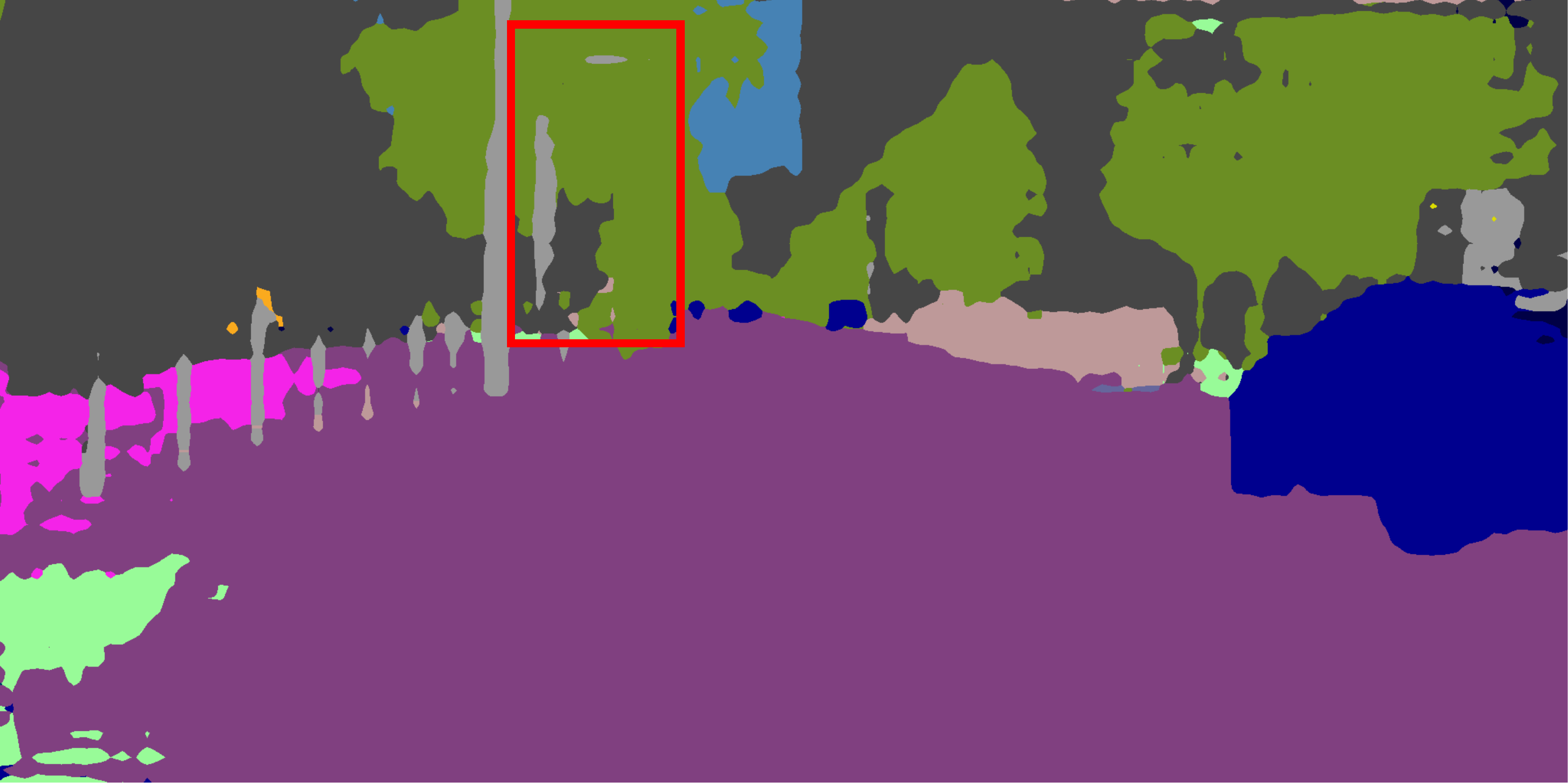}} &\hspace{-4.5mm}
			\subfloat{\includegraphics[width=0.16\linewidth]{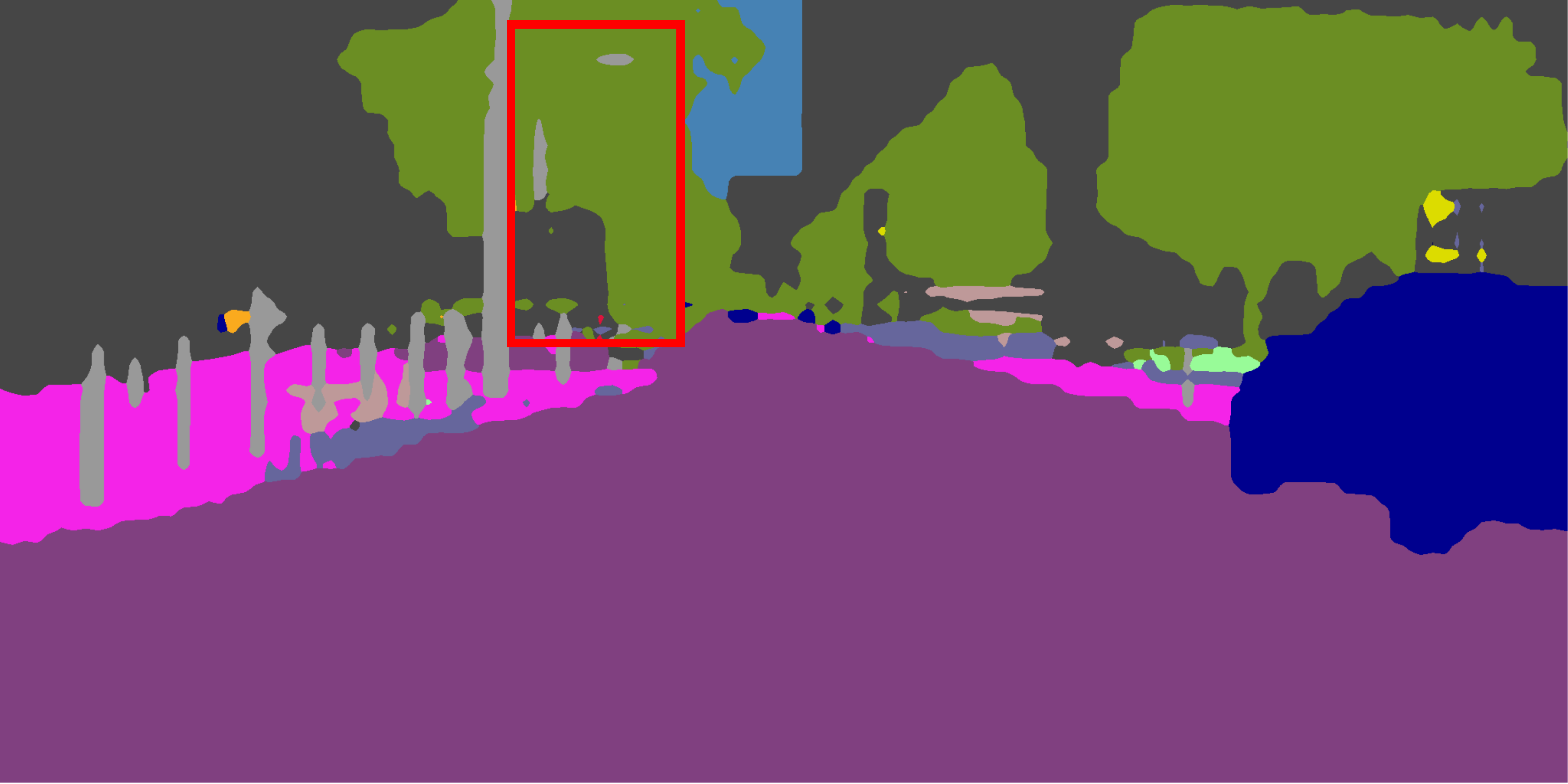}} &\hspace{-4.5mm}
			\subfloat{\includegraphics[width=0.16\linewidth]{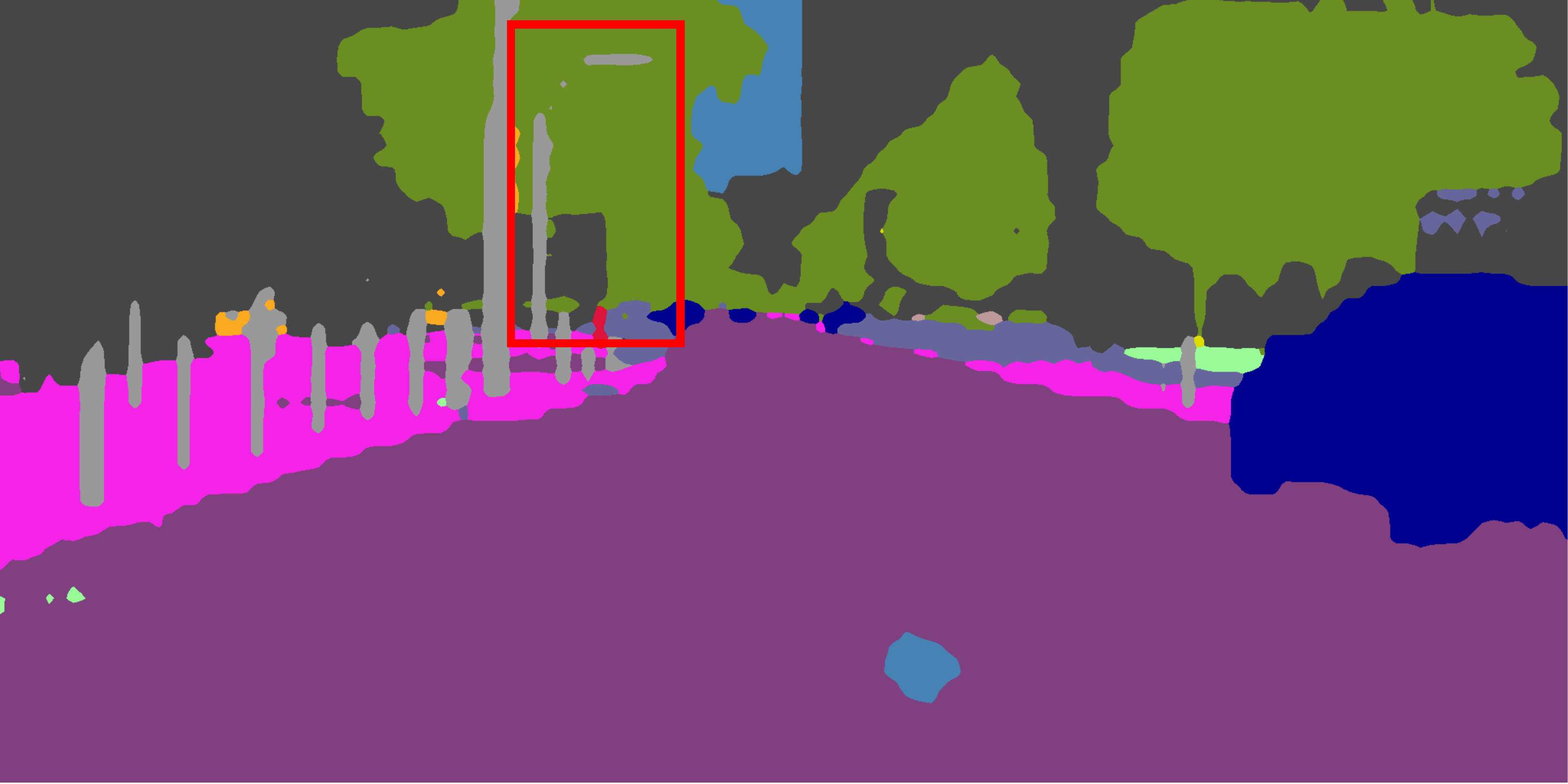}} &\hspace{-4.5mm}
			\subfloat{\includegraphics[width=0.16\linewidth]{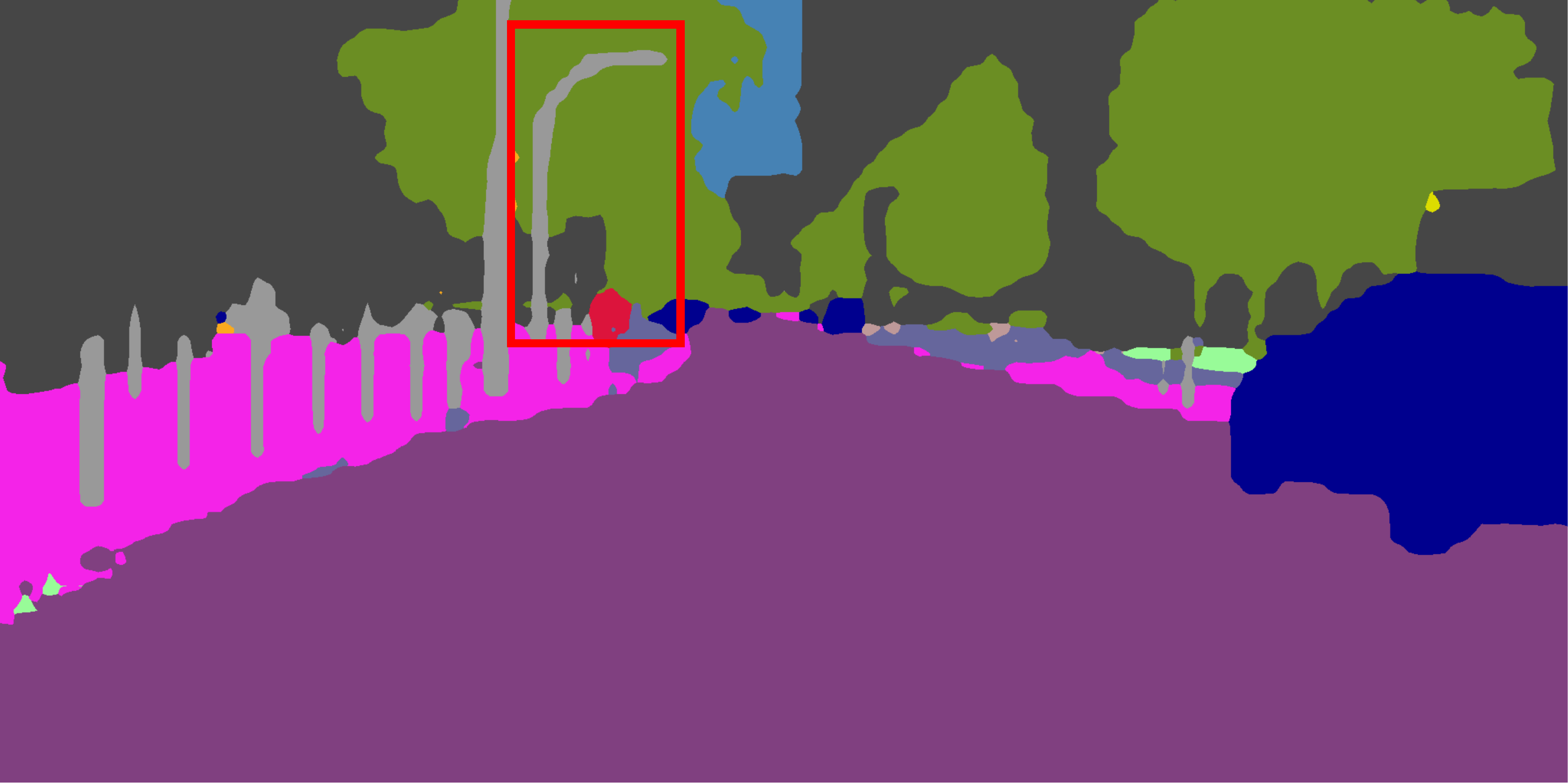}} 
			\\

			\subfloat{\includegraphics[width=0.16\linewidth]{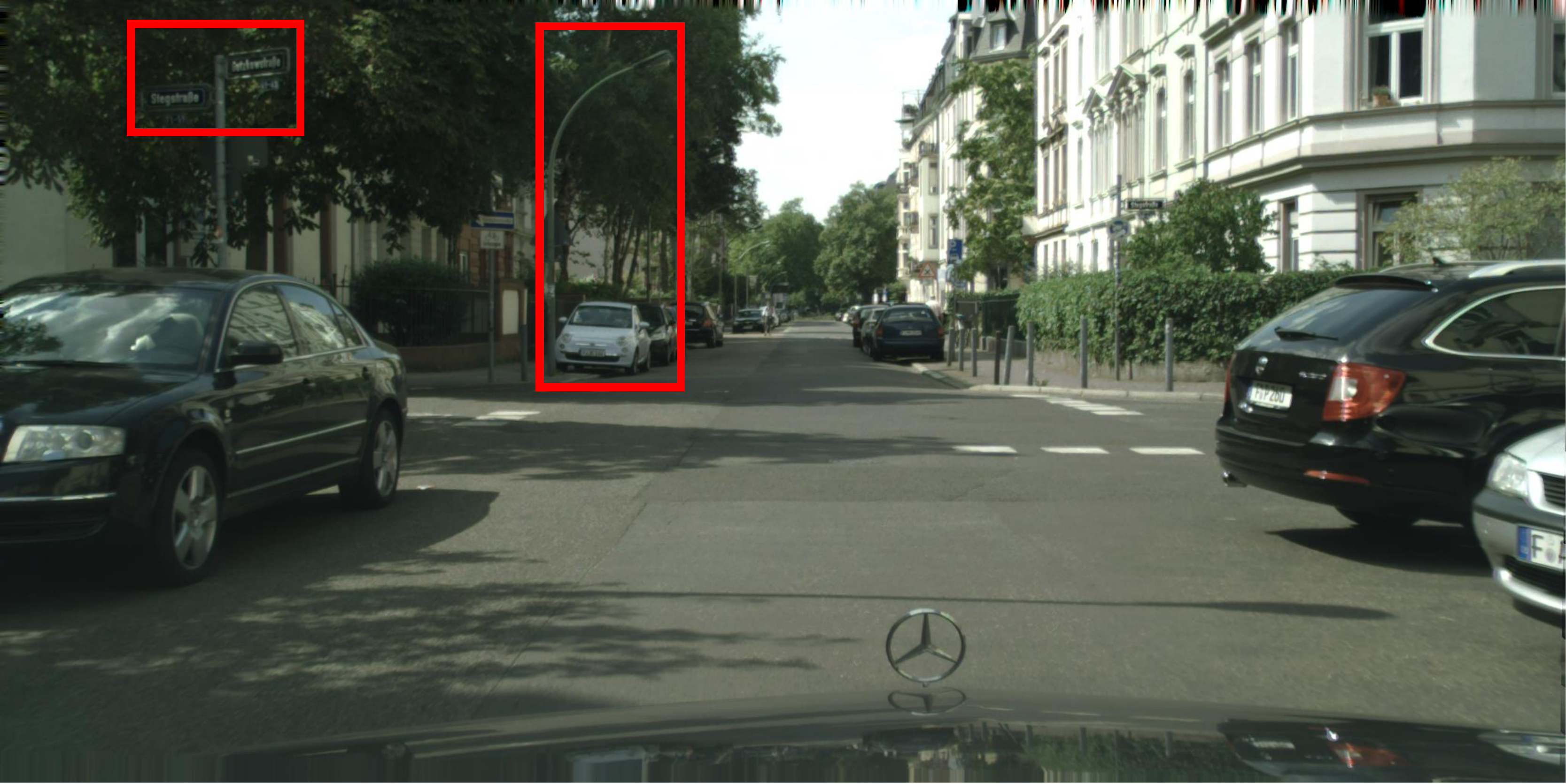}} &\hspace{-4.5mm}
			\subfloat{\includegraphics[width=0.16\linewidth]{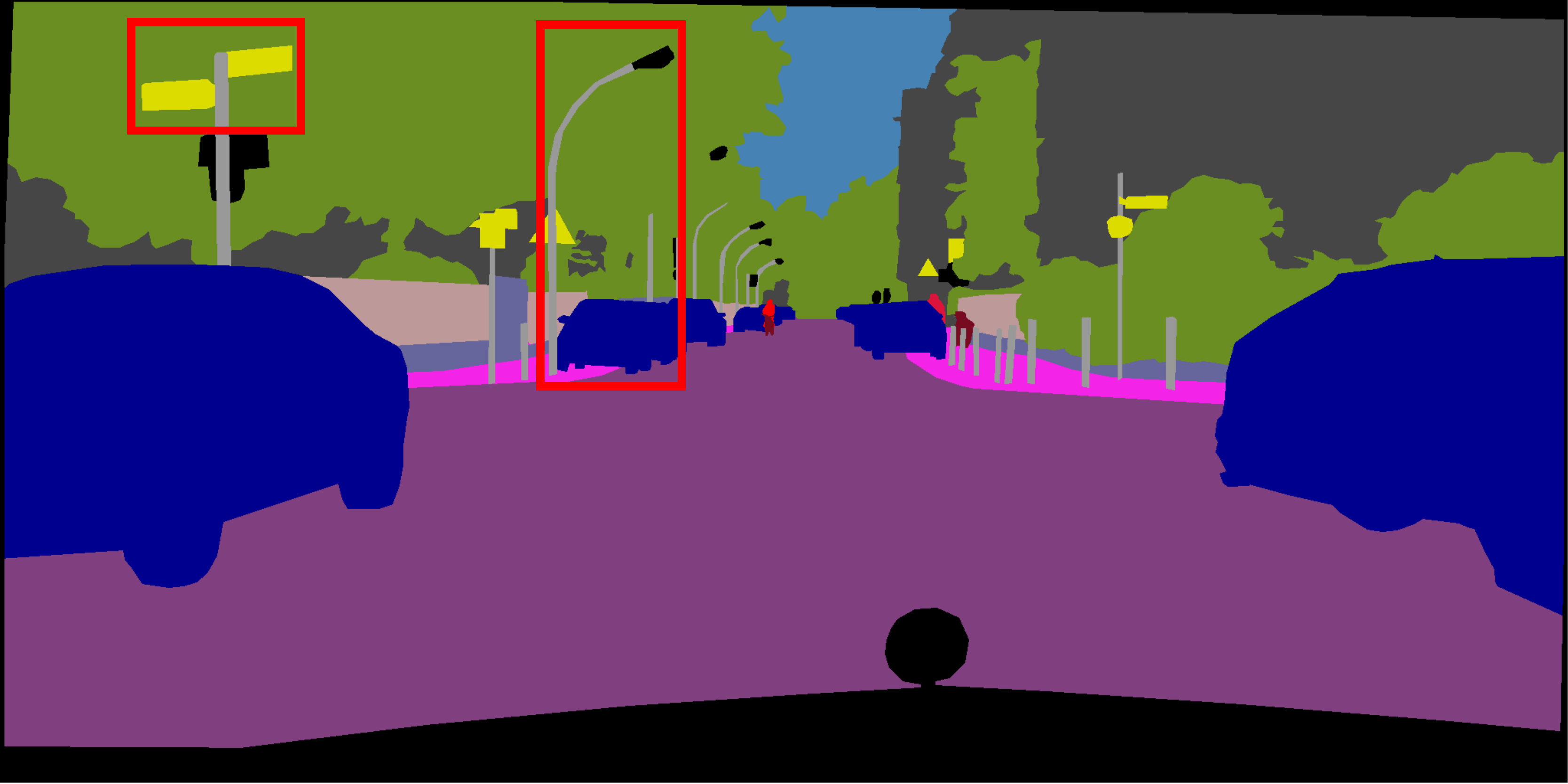}} &\hspace{-4.5mm}
			\subfloat{\includegraphics[width=0.16\linewidth]{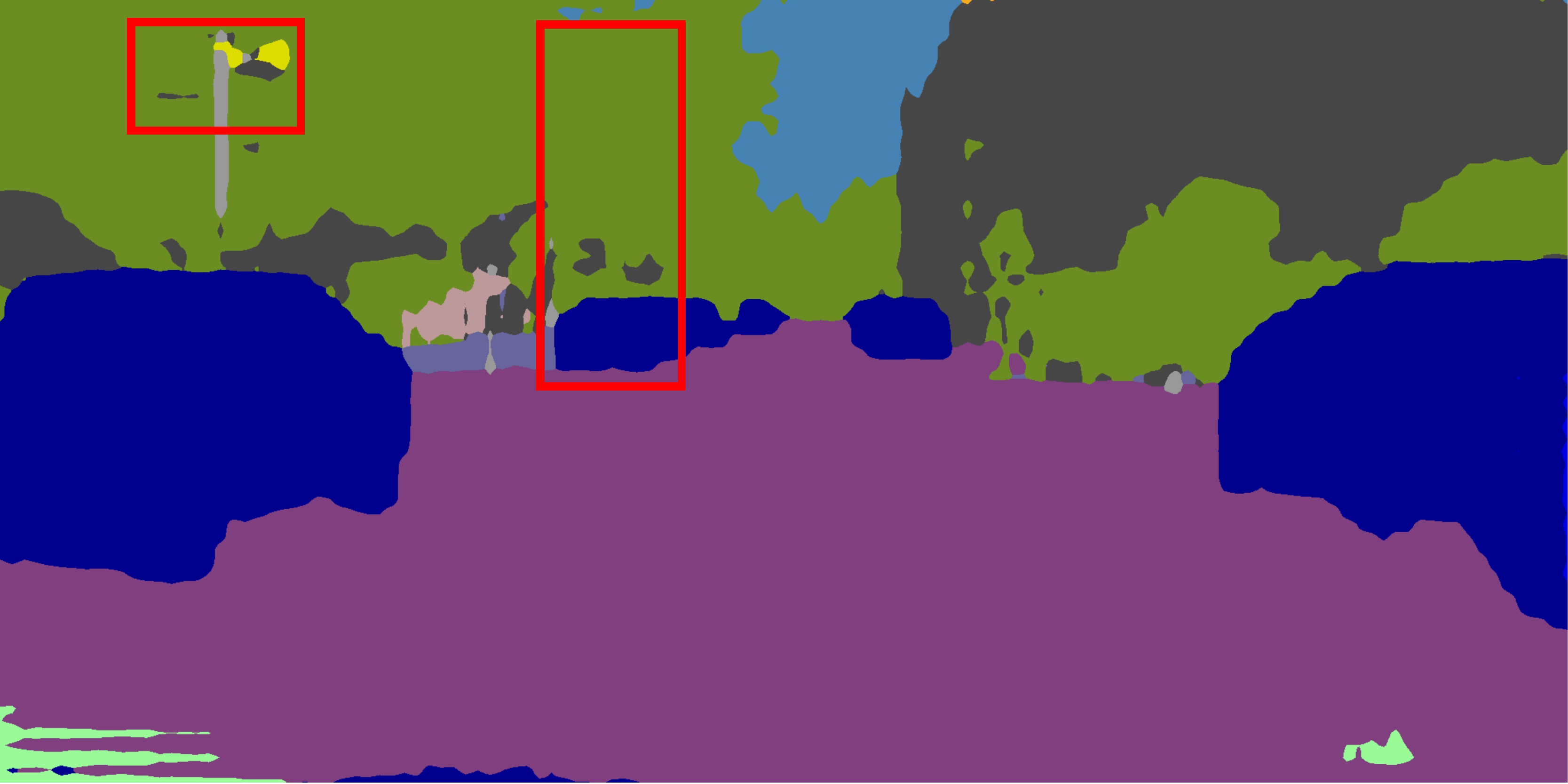}} &\hspace{-4.5mm}
			\subfloat{\includegraphics[width=0.16\linewidth]{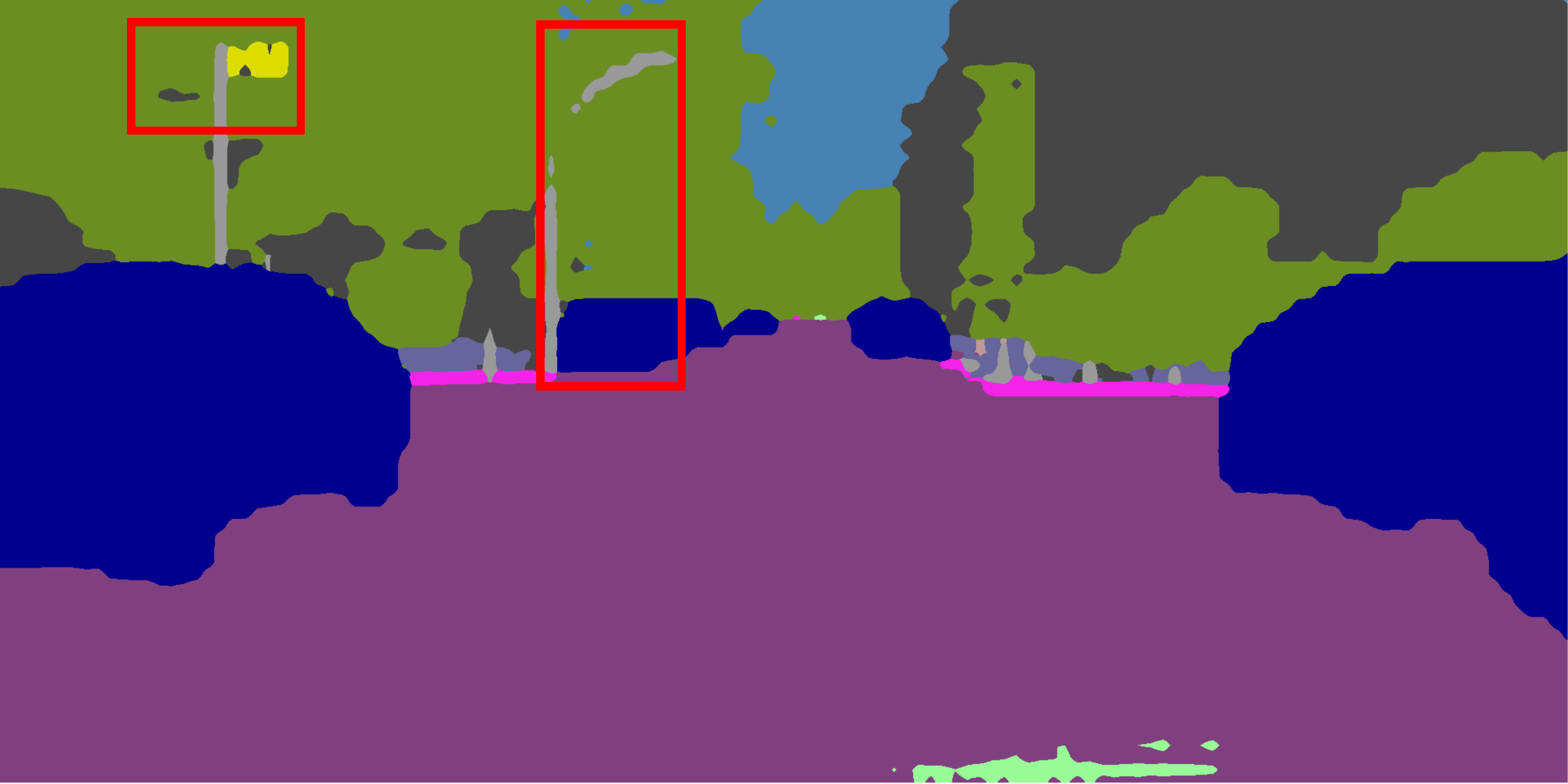}} &\hspace{-4.5mm}
			\subfloat{\includegraphics[width=0.16\linewidth]{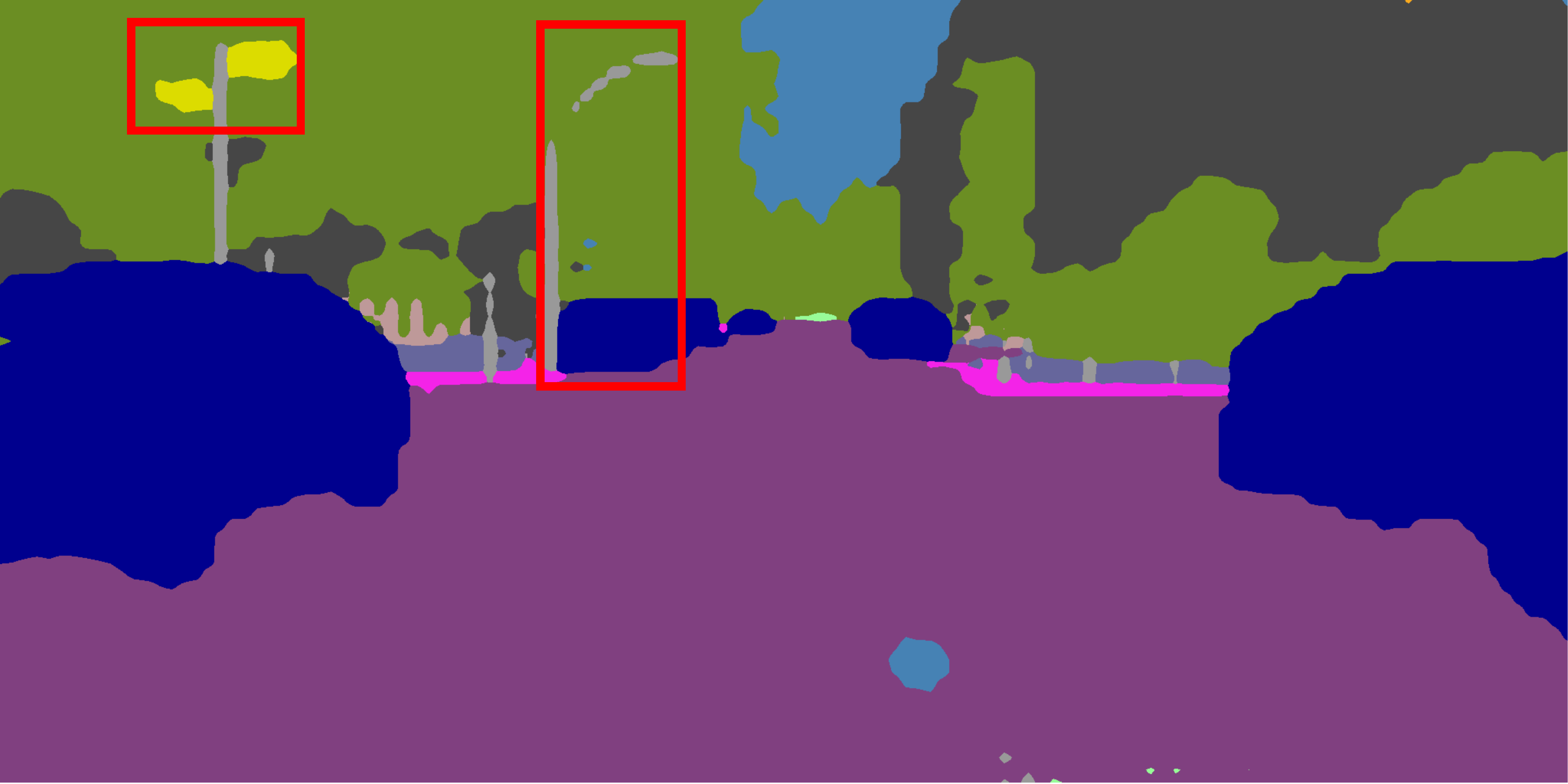}} &\hspace{-4.5mm}
			\subfloat{\includegraphics[width=0.16\linewidth]{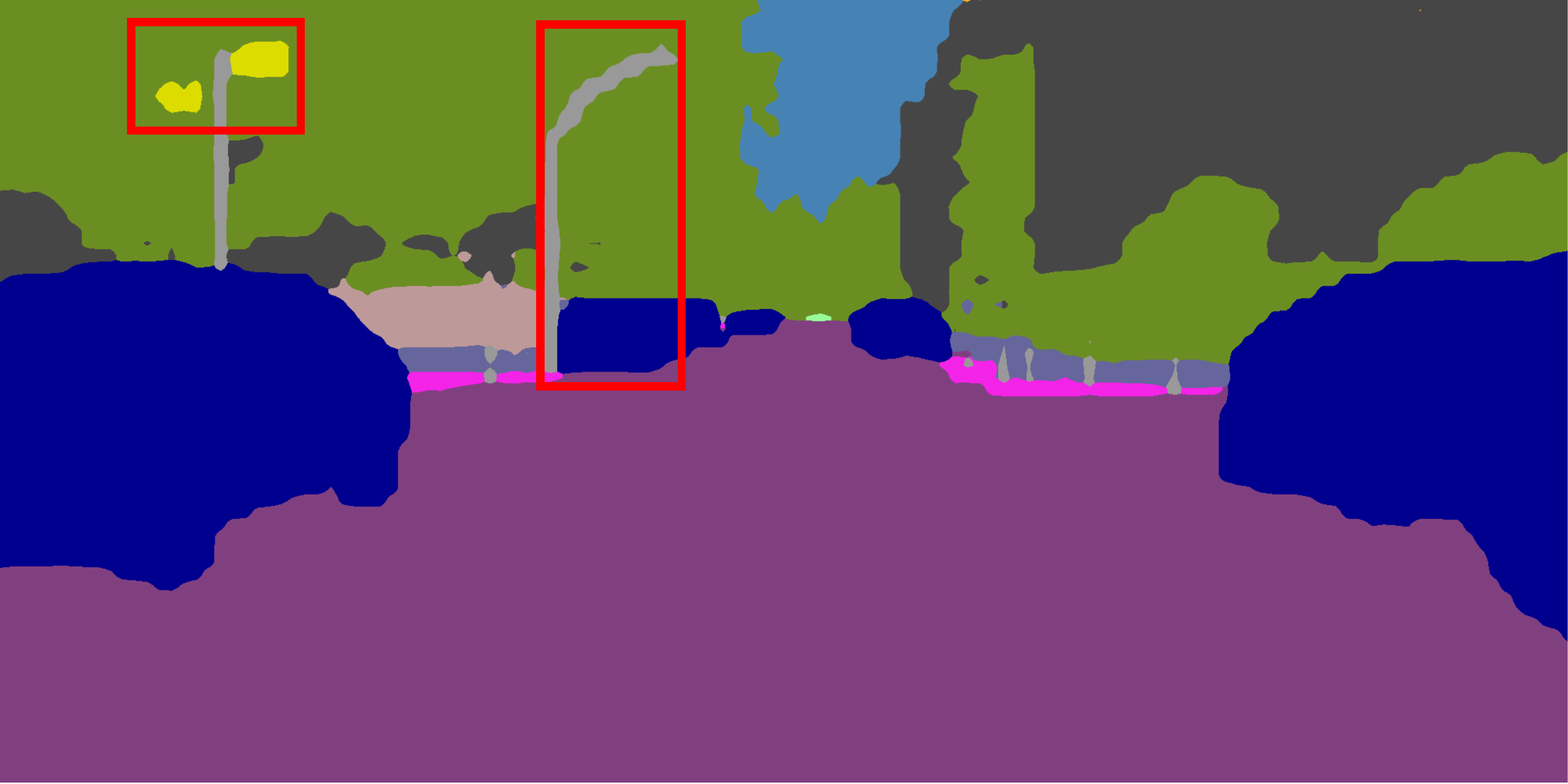}} 
			\\

			(a) Raw images& (b) Ground-truth & (c) $M_\mathcal{S}^{(0)}$ & (d) $M_\mathcal{T}^{(0)}$ & (e) DPL & (f) DPL-Dual
	\end{tabular}}
	\caption{Visualization of segmentation results (GTA5$\rightarrow$Cityscapes). (a) raw images from Cityscapes dataset; (b) ground-truth; (c) segmentation prediction of $M_\mathcal{S}^{(0)}$; (d) segmentation prediction of $M_\mathcal{T}^{(0)}$; (e) segmentation results of DPL; (f) segmentation results of DPL-Dual. Red rectangles highlight the differences.}
	\label{fig:seg_vis}
\end{figure*}

{\noindent \textbf{Visualization of Segmentation Results of DPL.}}\hspace{3pt}
In Figure \ref{fig:seg_vis}, we visualize the segmentation results of different stages of DPL framework. $M_\mathcal{S}^{(0)}$ and $M_\mathcal{T}^{(0)}$ are segmentation models initialized by the proposed warm-up strategy. Step by step, segmentation results appear smoother with clearer boundaries among different classes. DPL and DPL-Dual further improve the performance on maintaining whole structures of small objectives (e.g., person and rider in row 1,2,3,4,6, sign and pole in row 7,8) and distinguishing similar categories(e.g., sidewalks and road in row 5,6). The stage-wise improvements demonstrate the effectiveness of the proposed label correction strategy, dual path image translation and dual path adaptive segmentation.
\renewcommand{\arraystretch}{1}
\clearpage

\begin{figure*}[htbp]
	\vspace{1cm}
	\scalebox{1}{
		\begin{tabular}{ccccc}

			\subfloat{\includegraphics[width=0.2\linewidth]{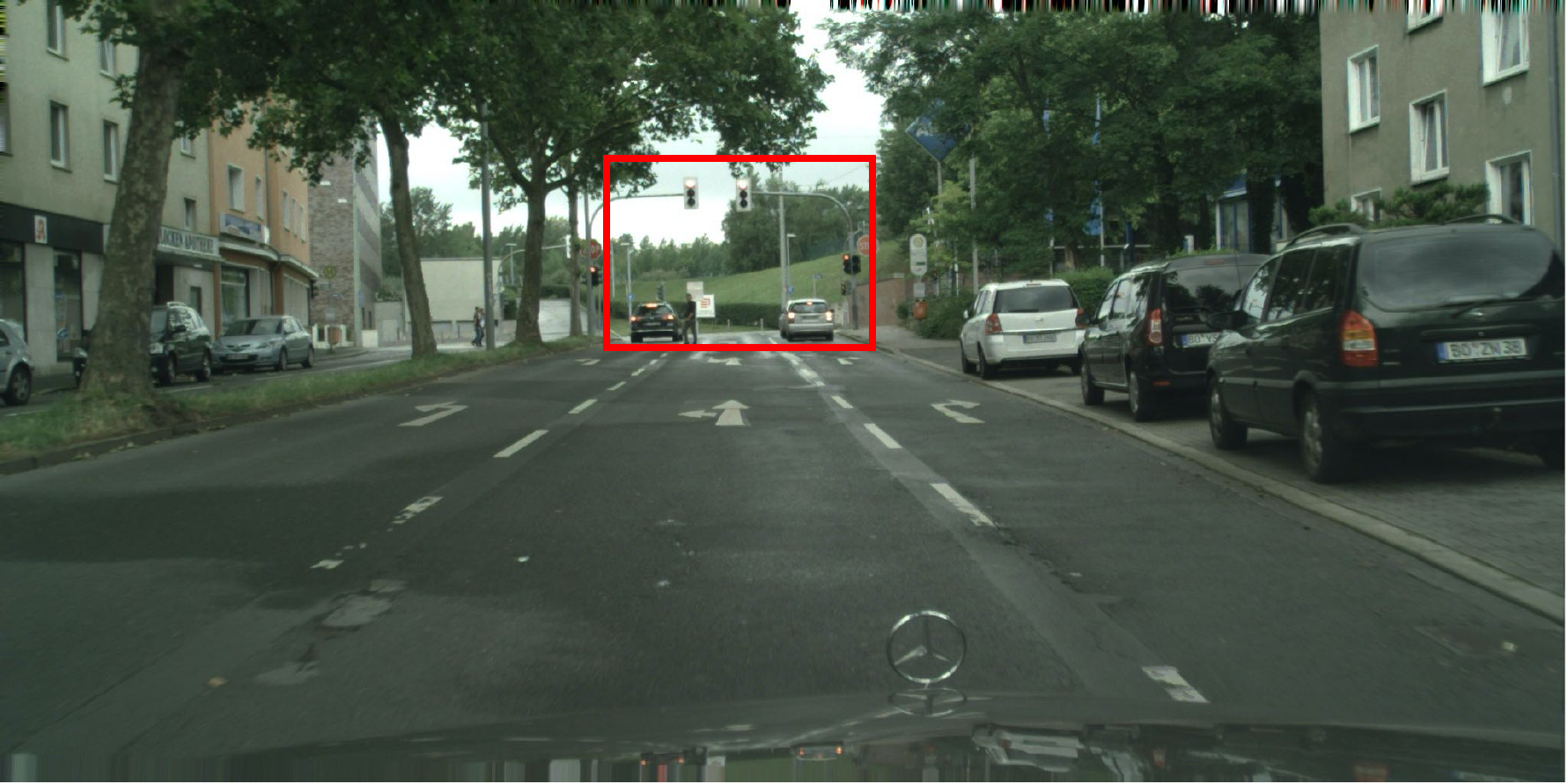}} & \hspace{-4.5mm}		
			\subfloat{\includegraphics[width=0.2\linewidth]{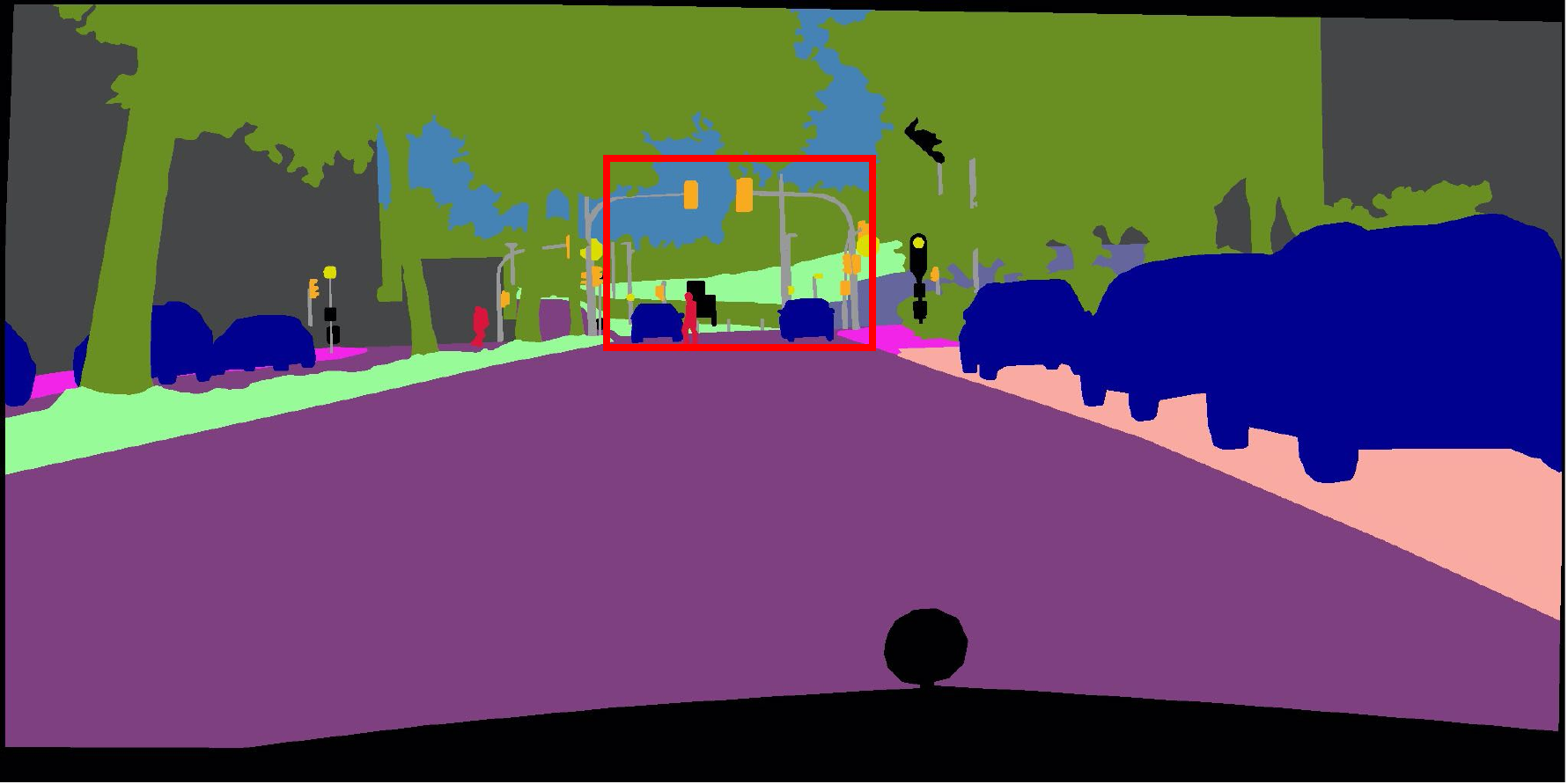}} & \hspace{-4.5mm}
			\subfloat{\includegraphics[width=0.2\linewidth]{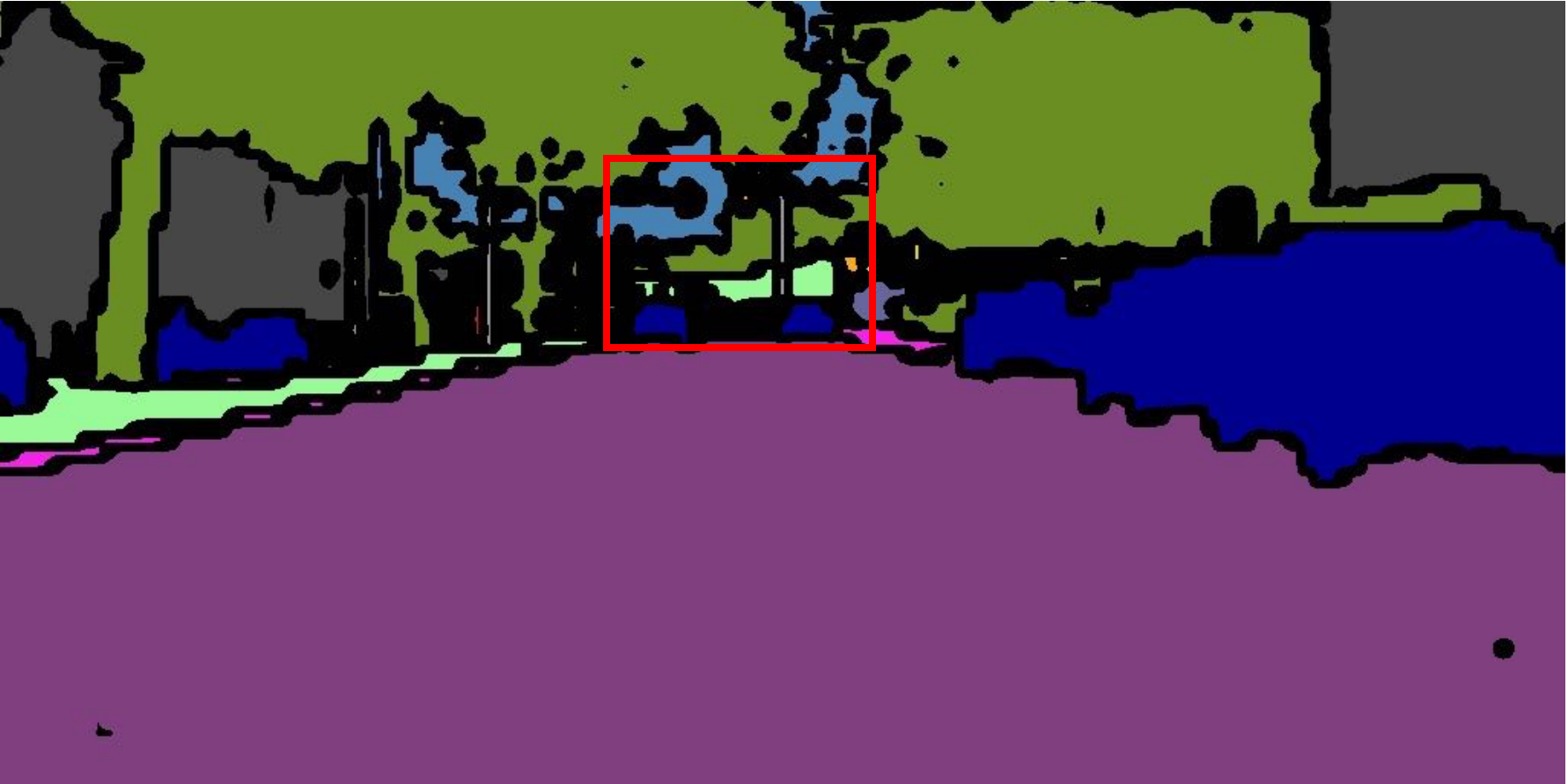}} & \hspace{-4.5mm}
			\subfloat{\includegraphics[width=0.2\linewidth]{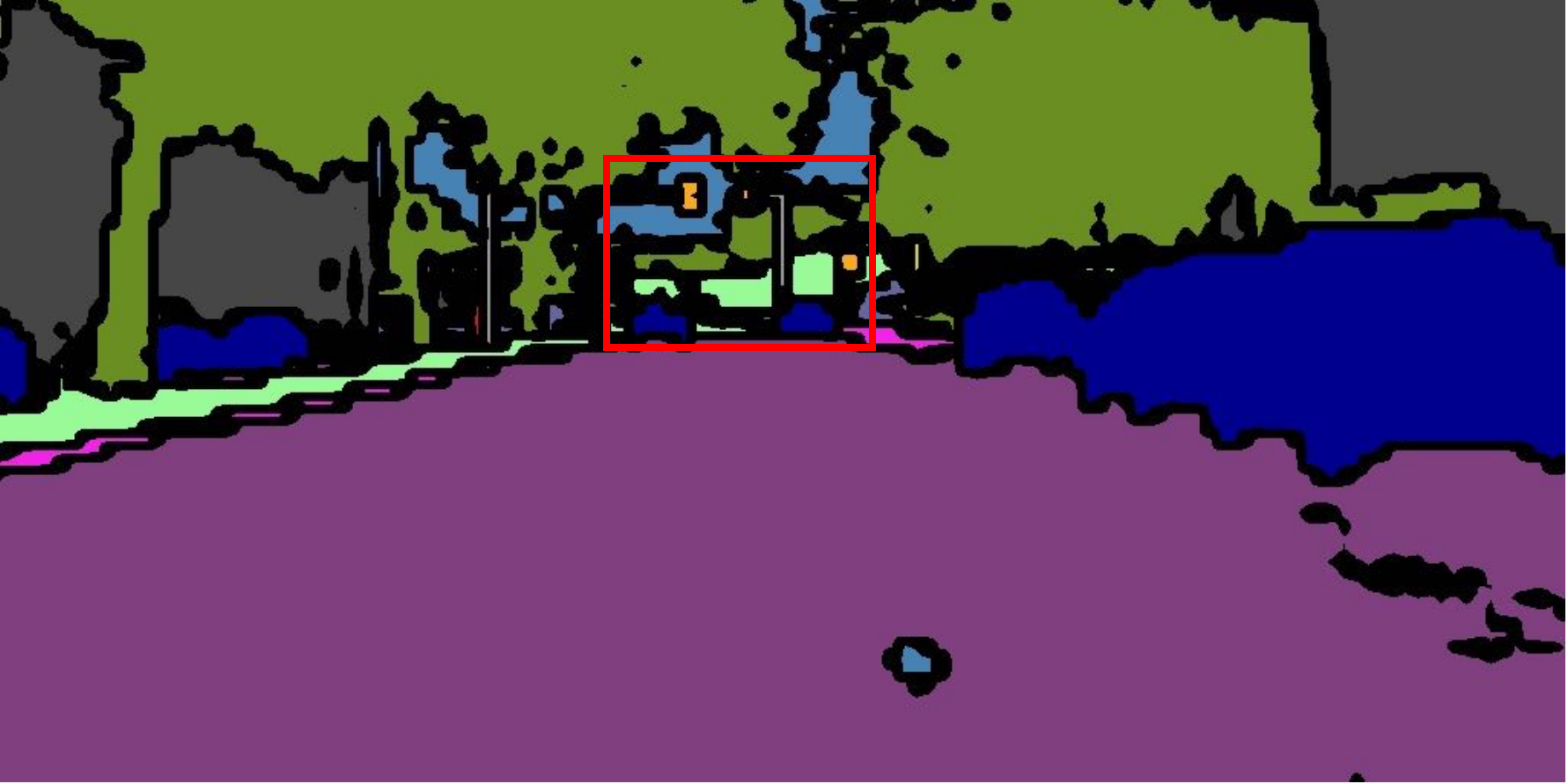}} & \hspace{-4.5mm}
			\subfloat{\includegraphics[width=0.2\linewidth]{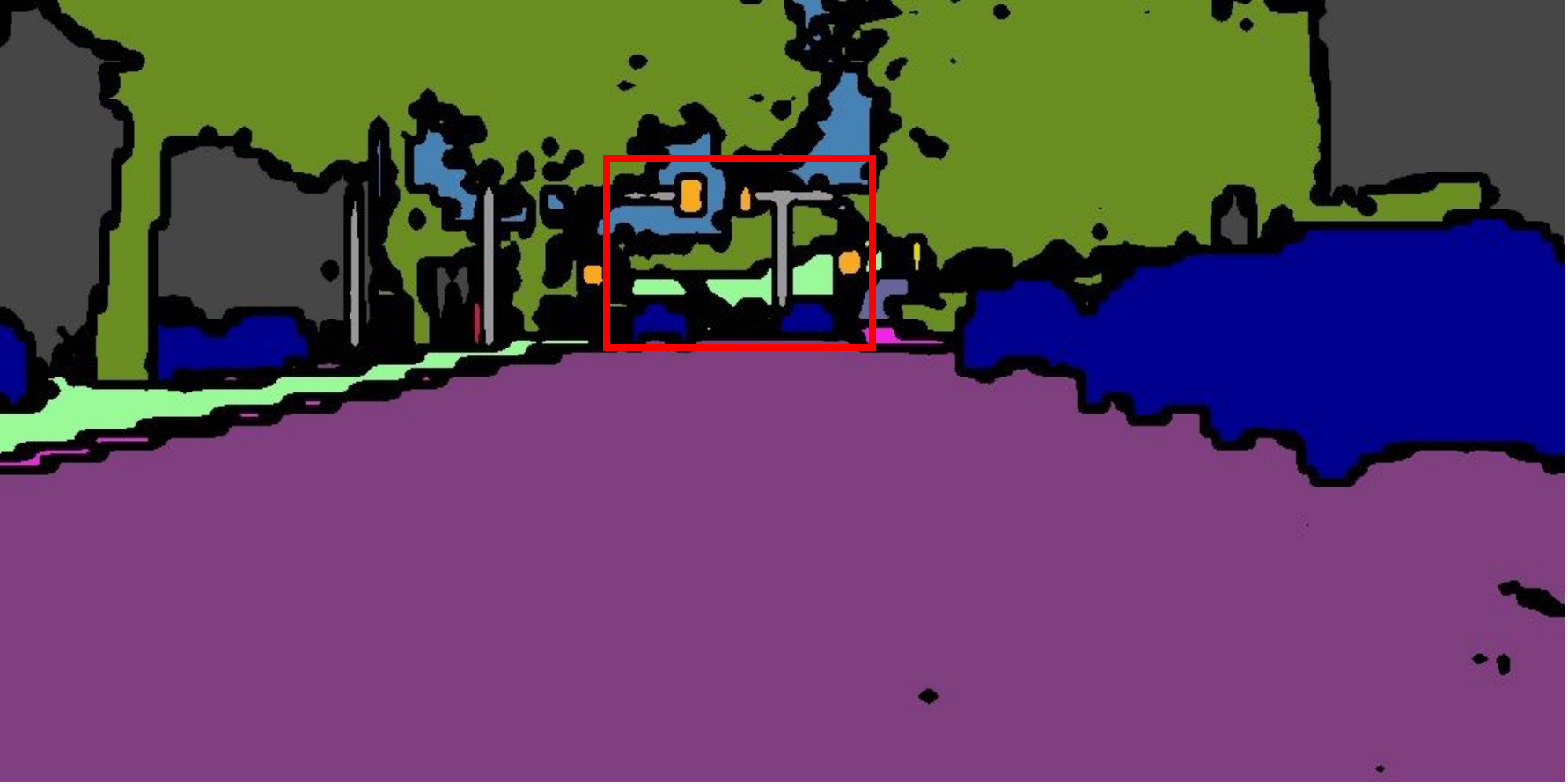}}
			\\
		
			\subfloat{\includegraphics[width=0.2\linewidth]{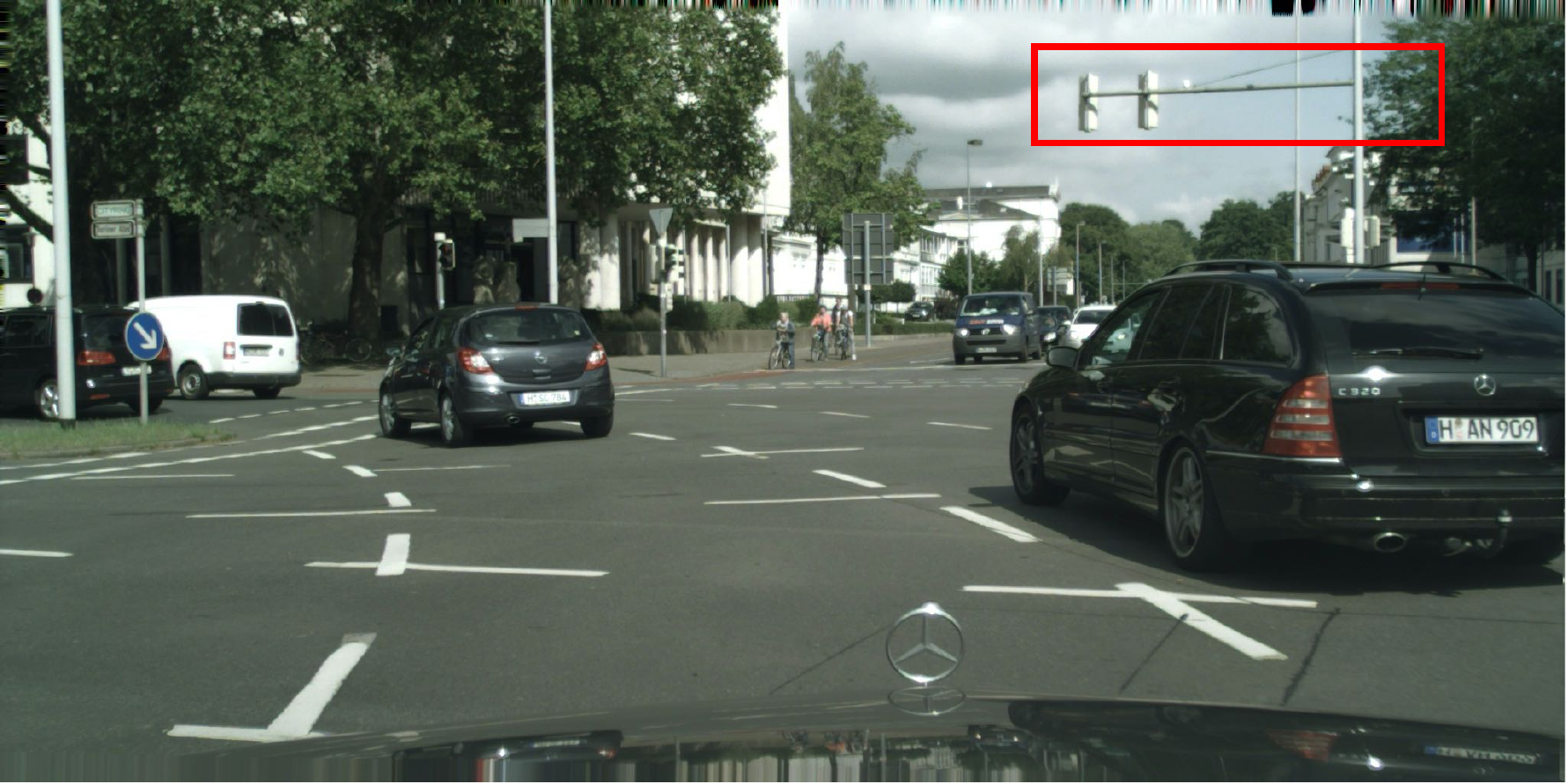}} & \hspace{-4.5mm}		
			\subfloat{\includegraphics[width=0.2\linewidth]{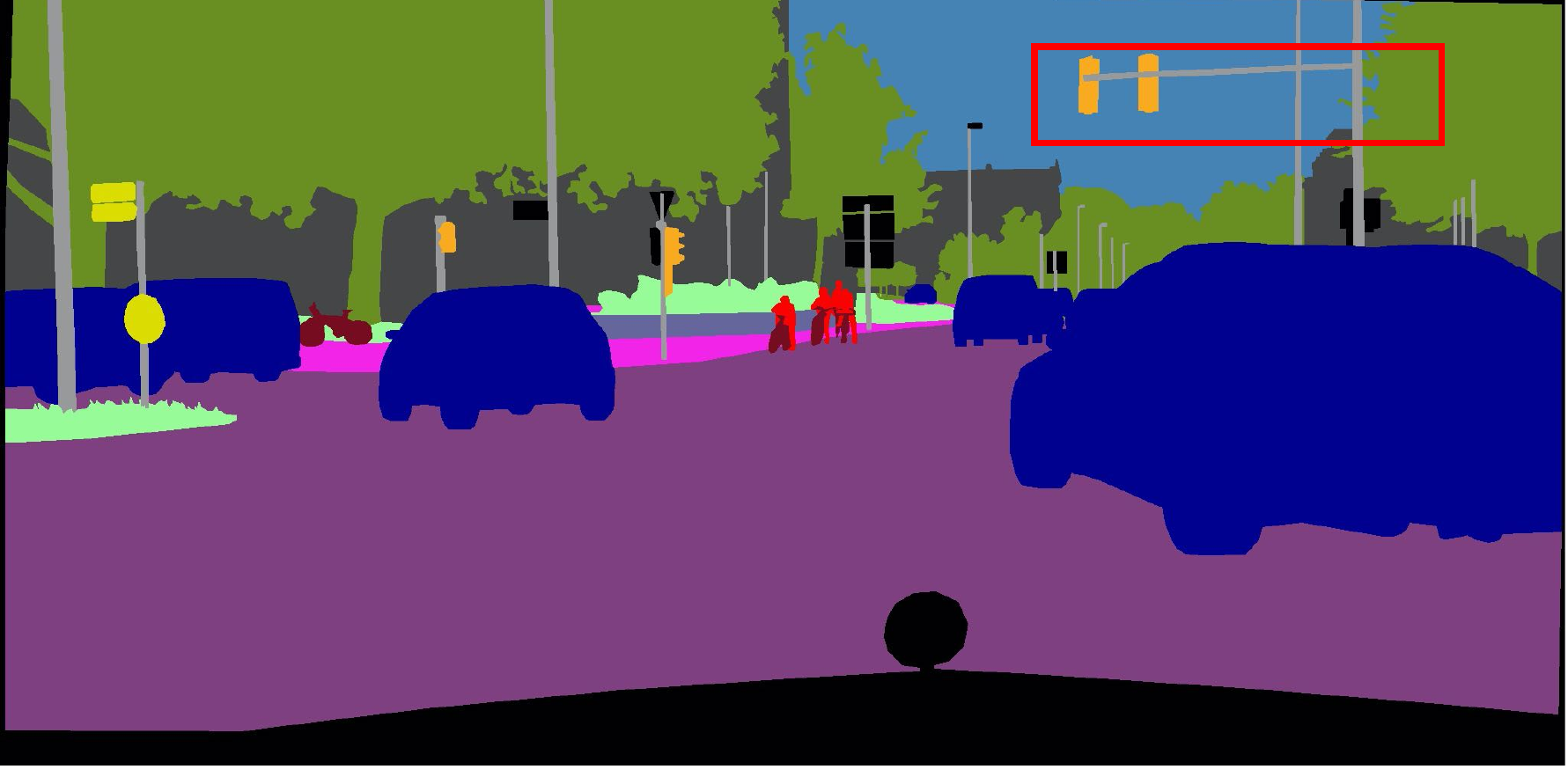}} & \hspace{-4.5mm}
			\subfloat{\includegraphics[width=0.2\linewidth]{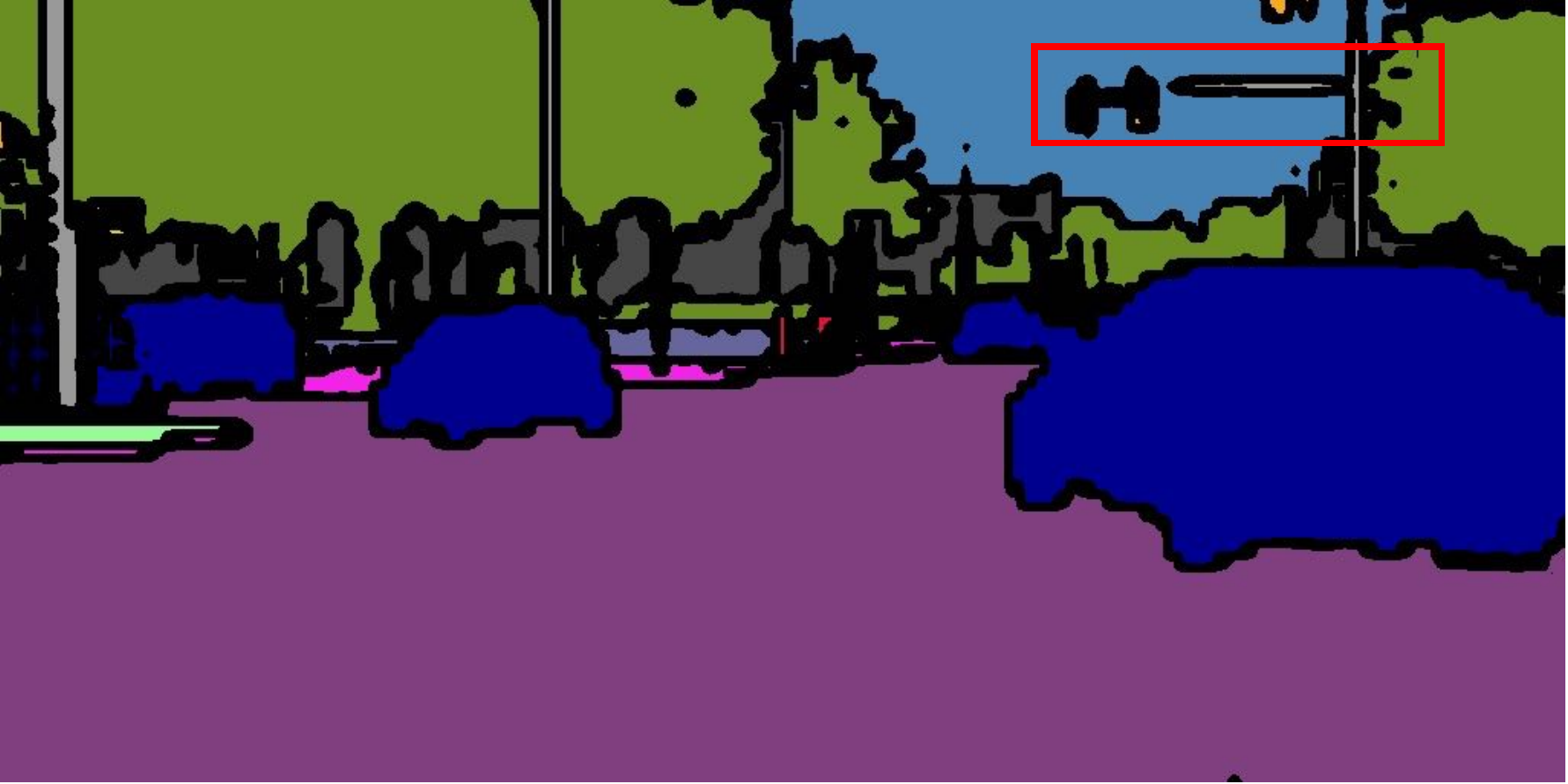}} & \hspace{-4.5mm}
			\subfloat{\includegraphics[width=0.2\linewidth]{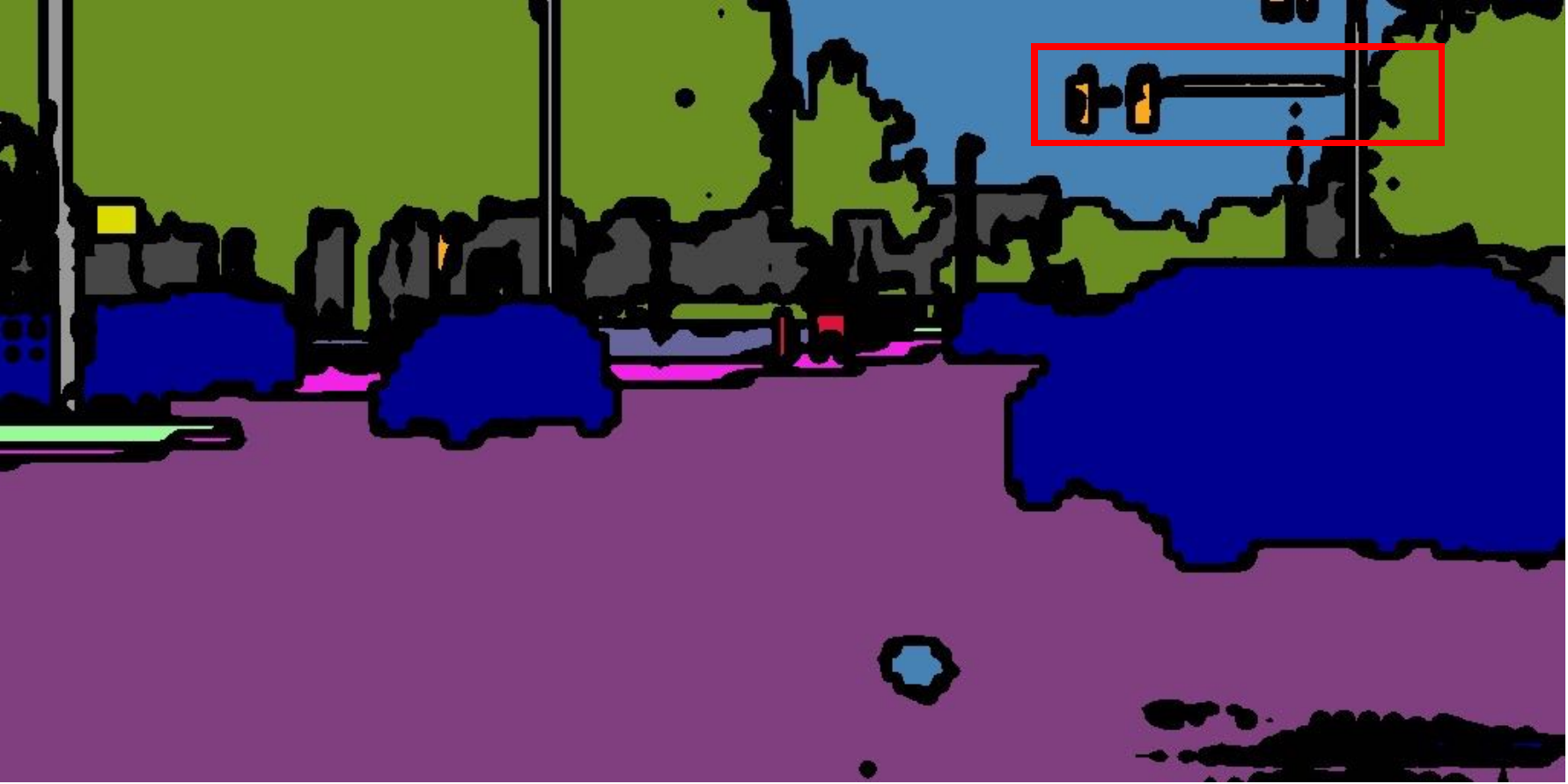}} & \hspace{-4.5mm}
			\subfloat{\includegraphics[width=0.2\linewidth]{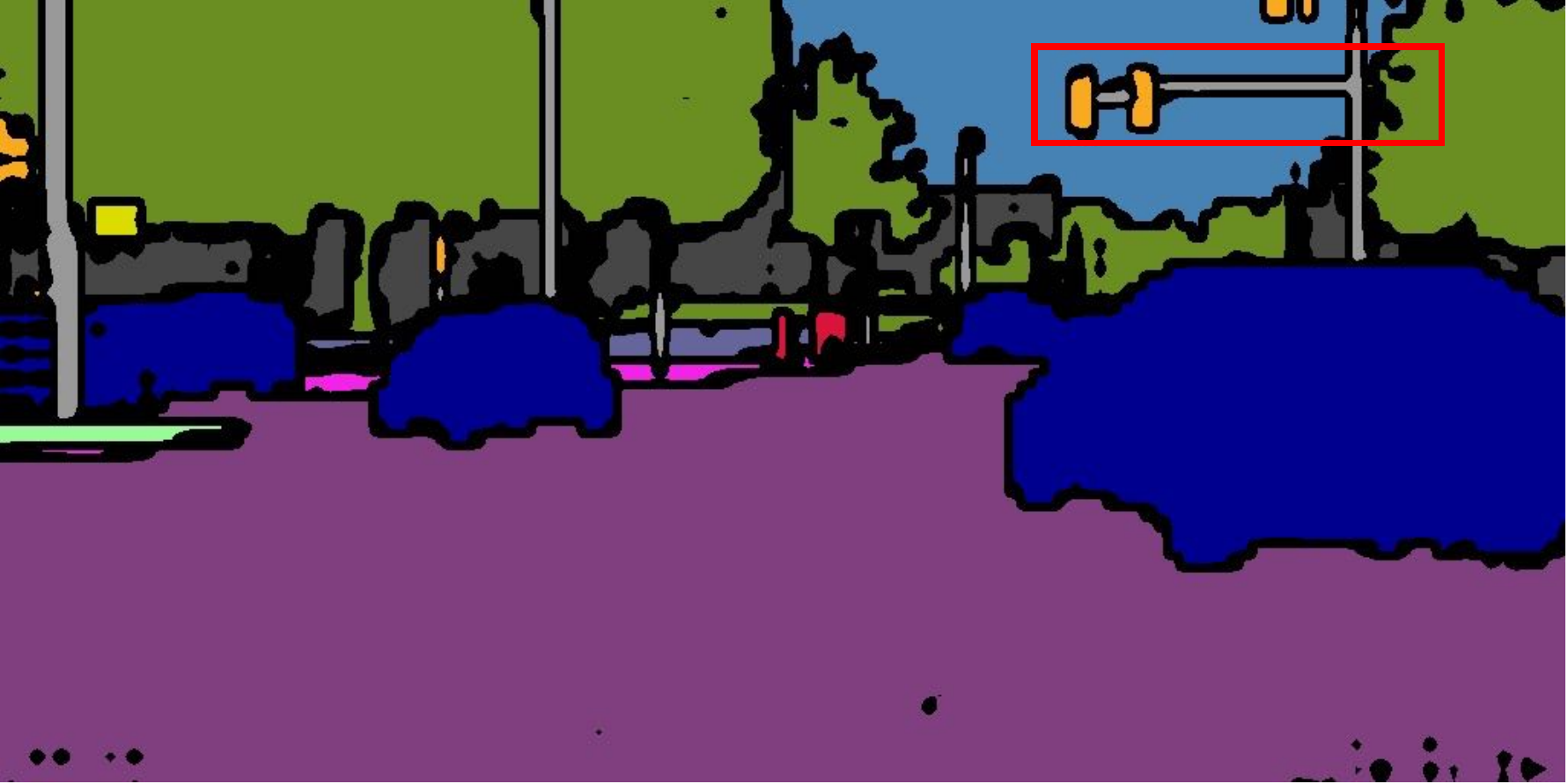}}
			\\
			\subfloat{\includegraphics[width=0.2\linewidth]{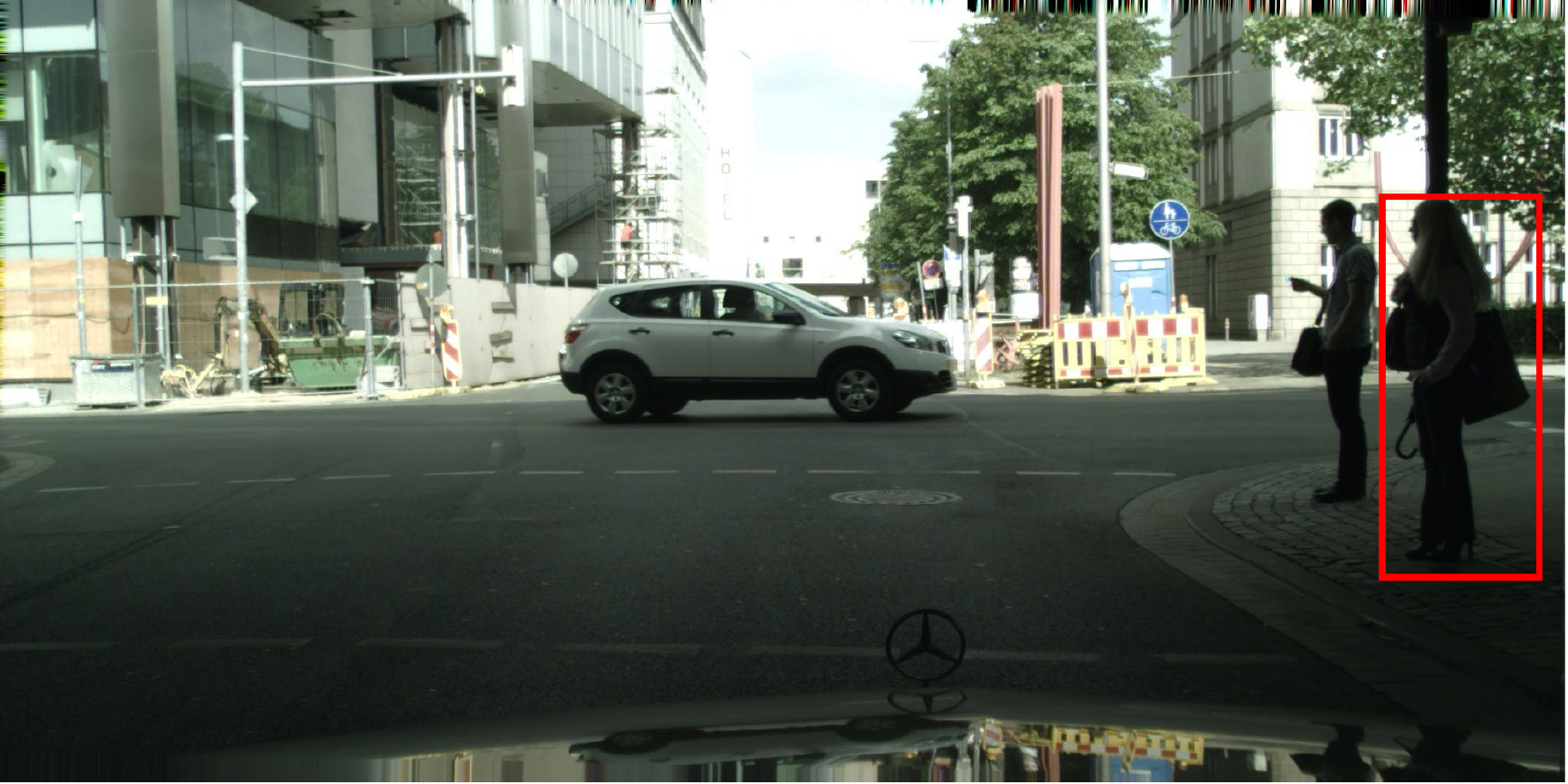}} & \hspace{-4.5mm}		
			\subfloat{\includegraphics[width=0.2\linewidth]{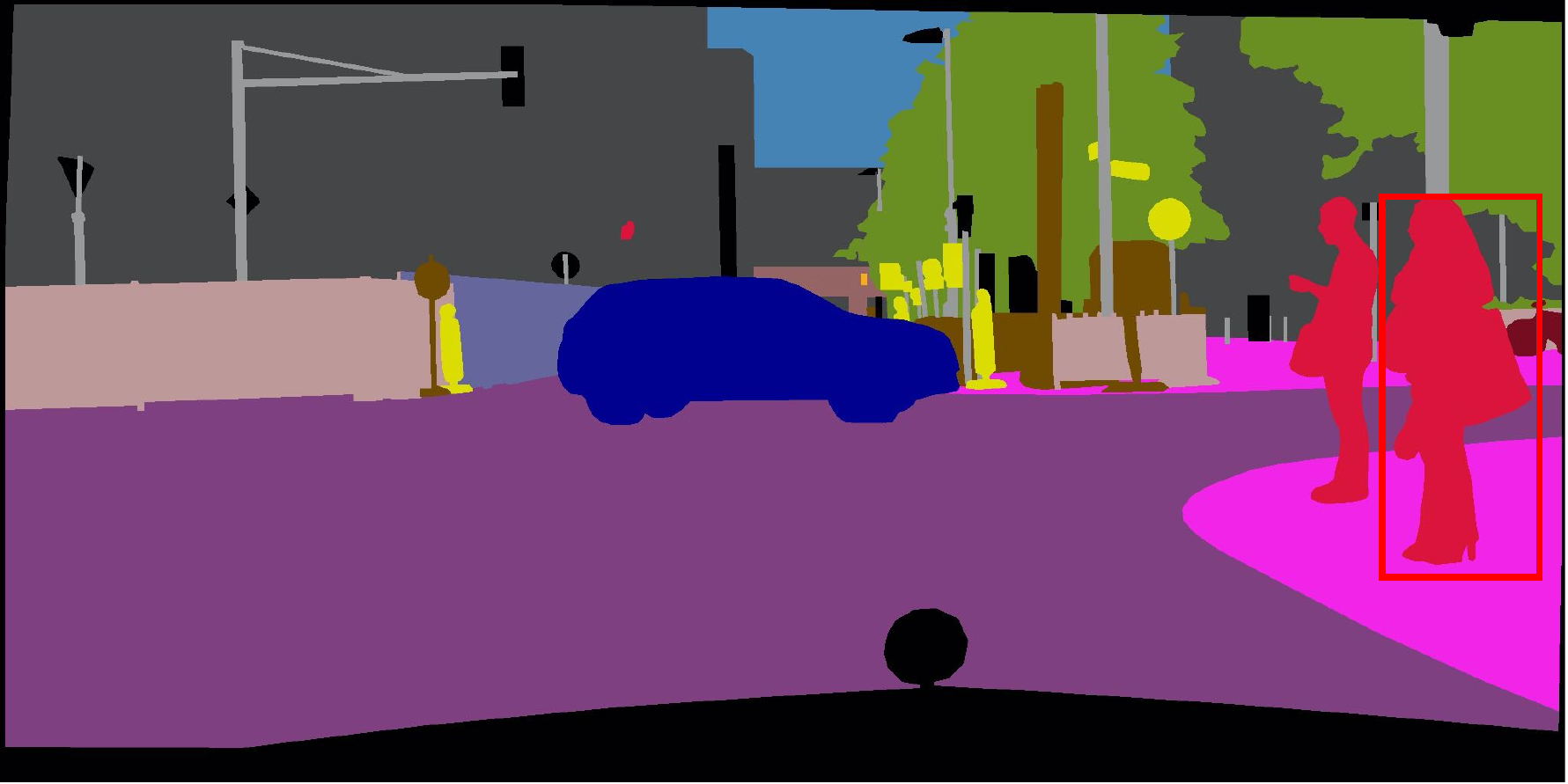}} & \hspace{-4.5mm}
			\subfloat{\includegraphics[width=0.2\linewidth]{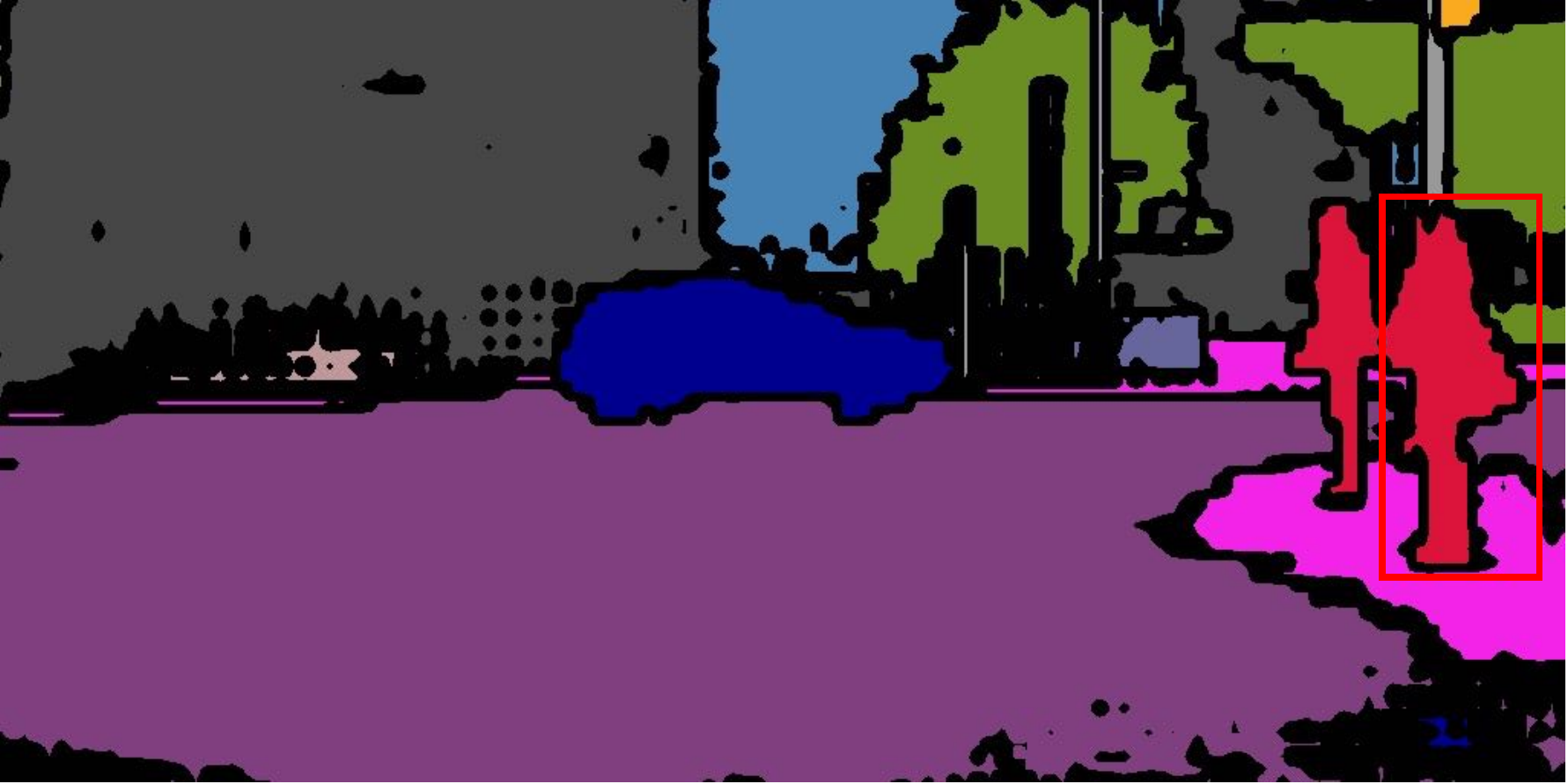}} & \hspace{-4.5mm}
			\subfloat{\includegraphics[width=0.2\linewidth]{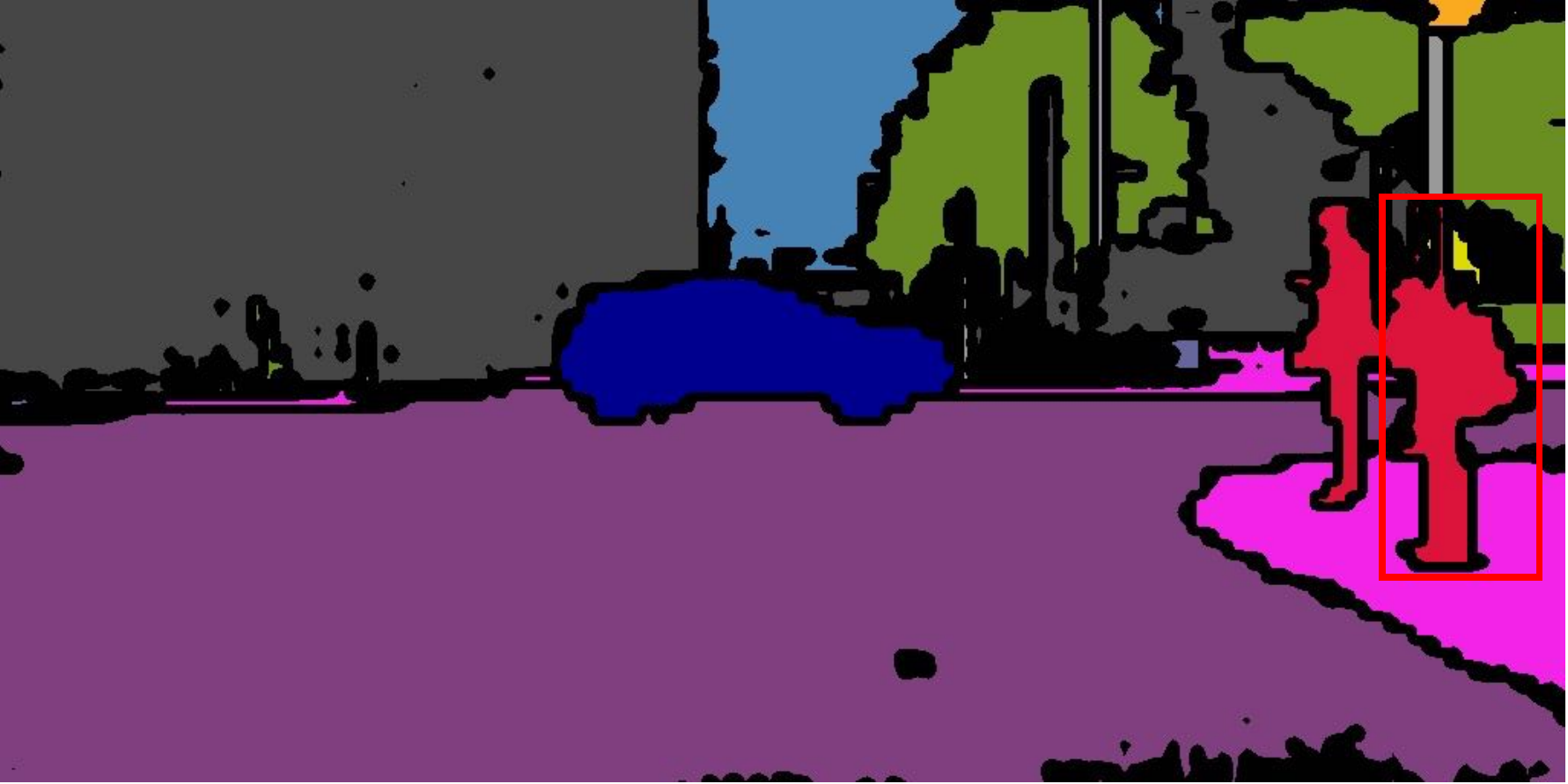}} & \hspace{-4.5mm}
			\subfloat{\includegraphics[width=0.2\linewidth]{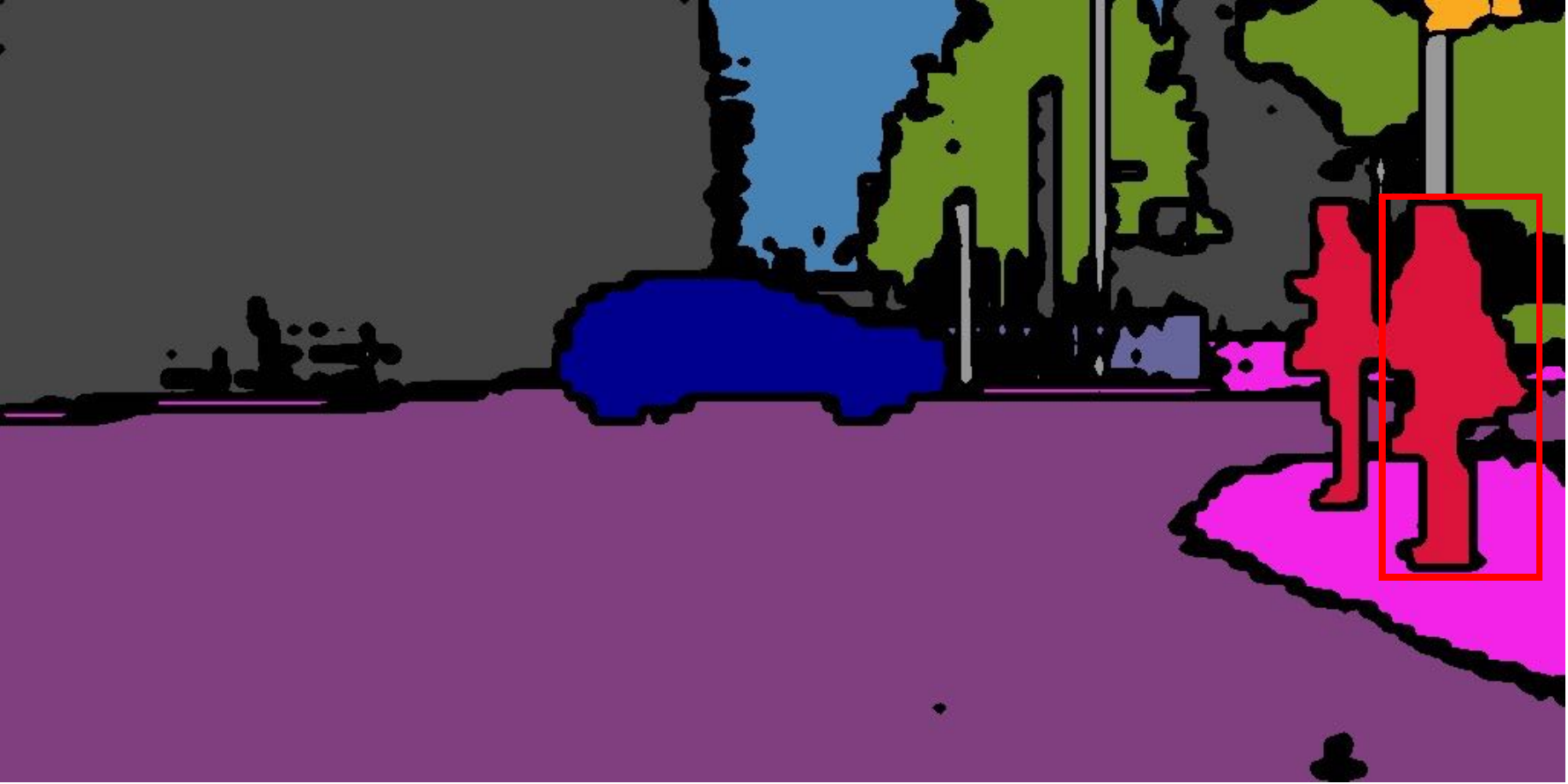}}
			\\
		
			\subfloat{\includegraphics[width=0.2\linewidth]{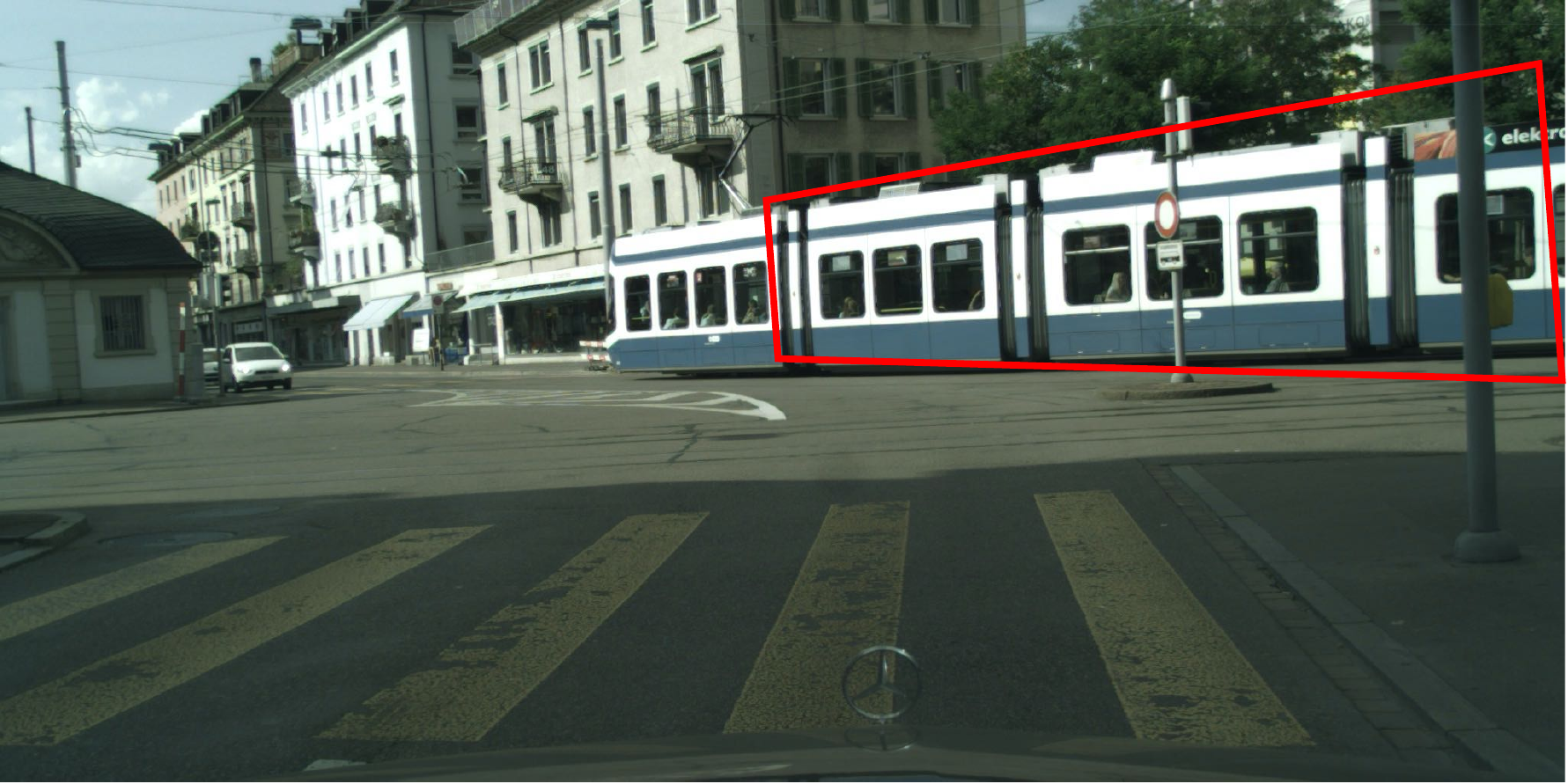}} & \hspace{-4.5mm}		
			\subfloat{\includegraphics[width=0.2\linewidth]{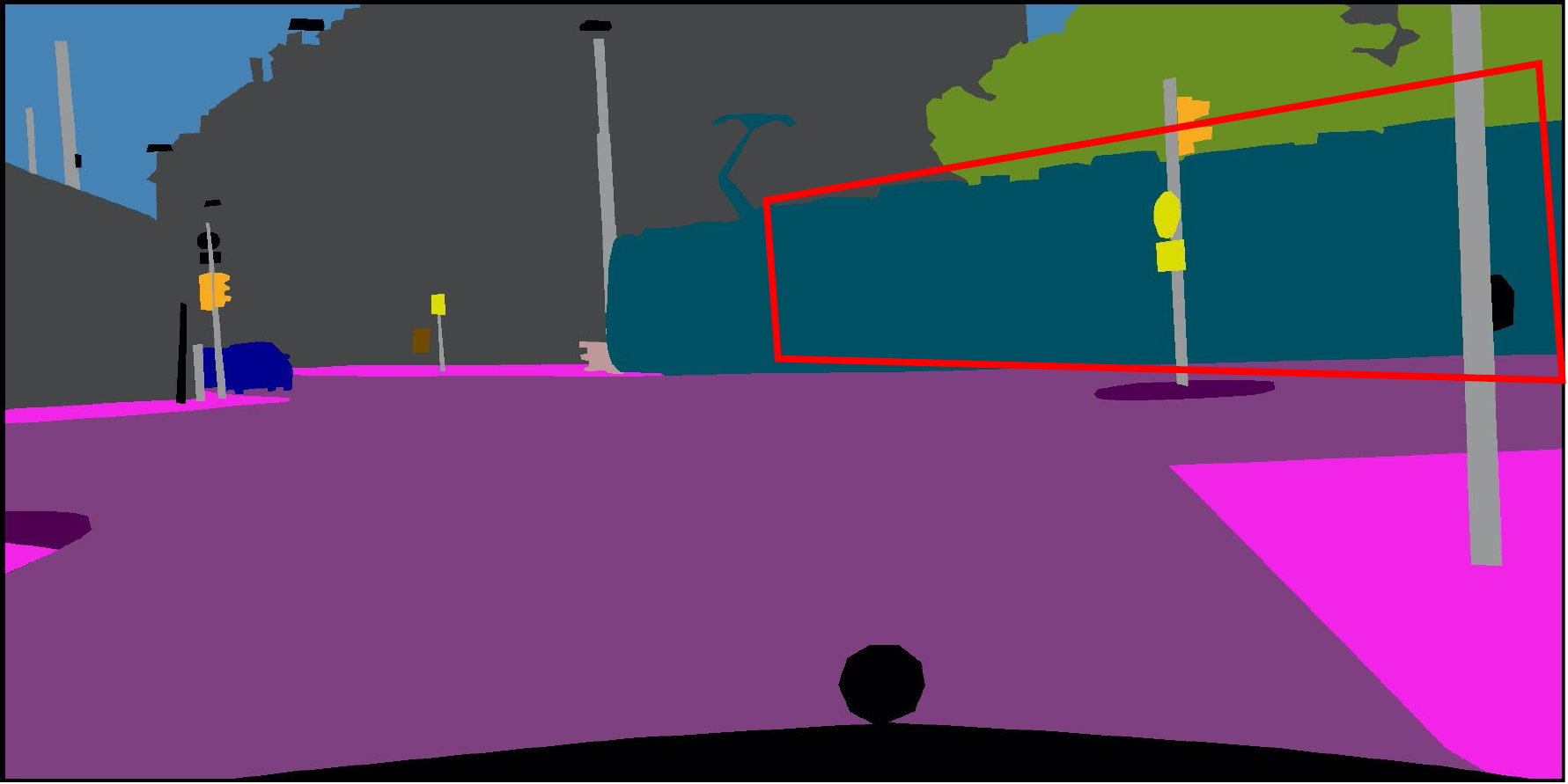}} & \hspace{-4.5mm}
			\subfloat{\includegraphics[width=0.2\linewidth]{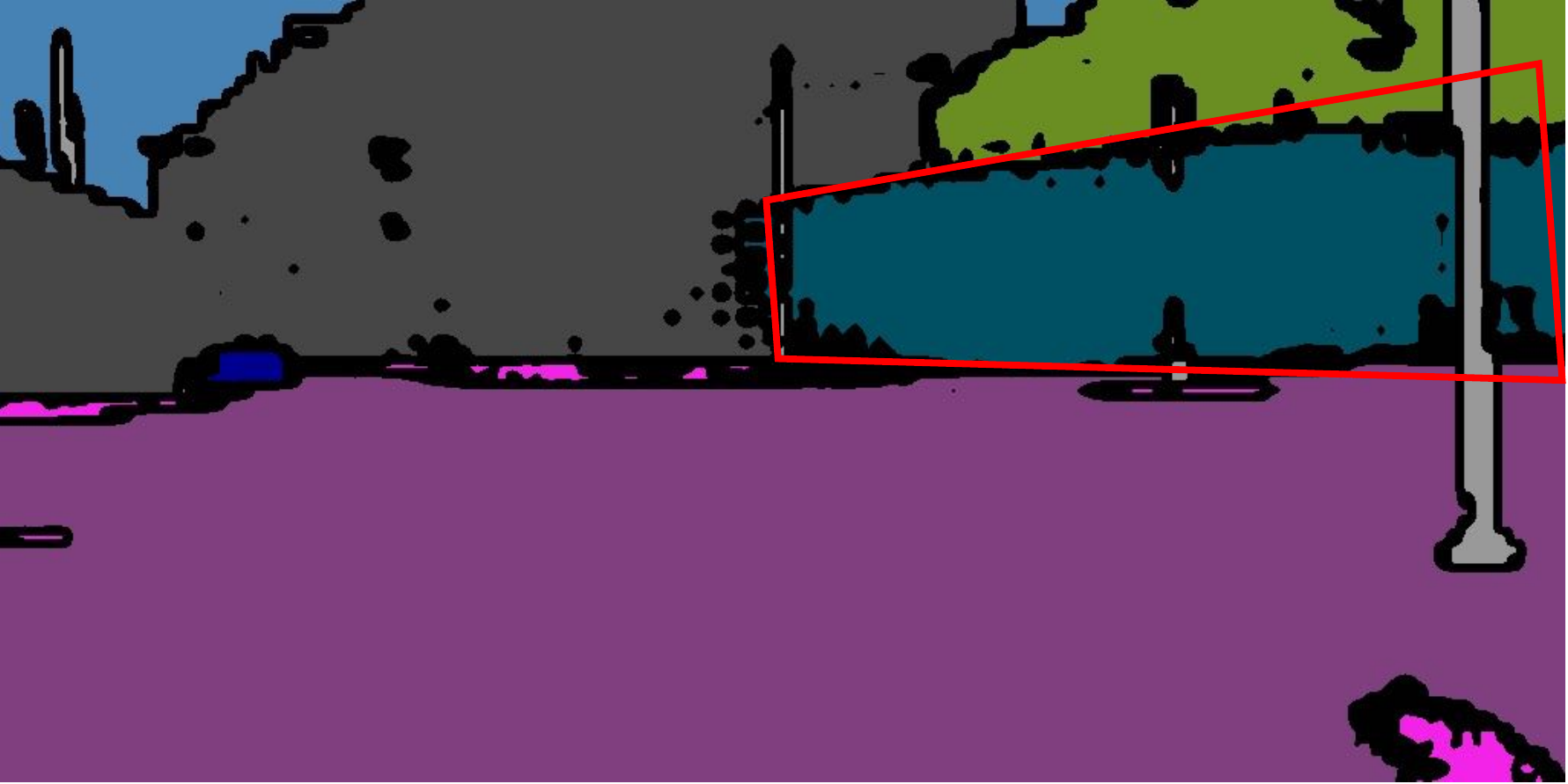}} & \hspace{-4.5mm}
			\subfloat{\includegraphics[width=0.2\linewidth]{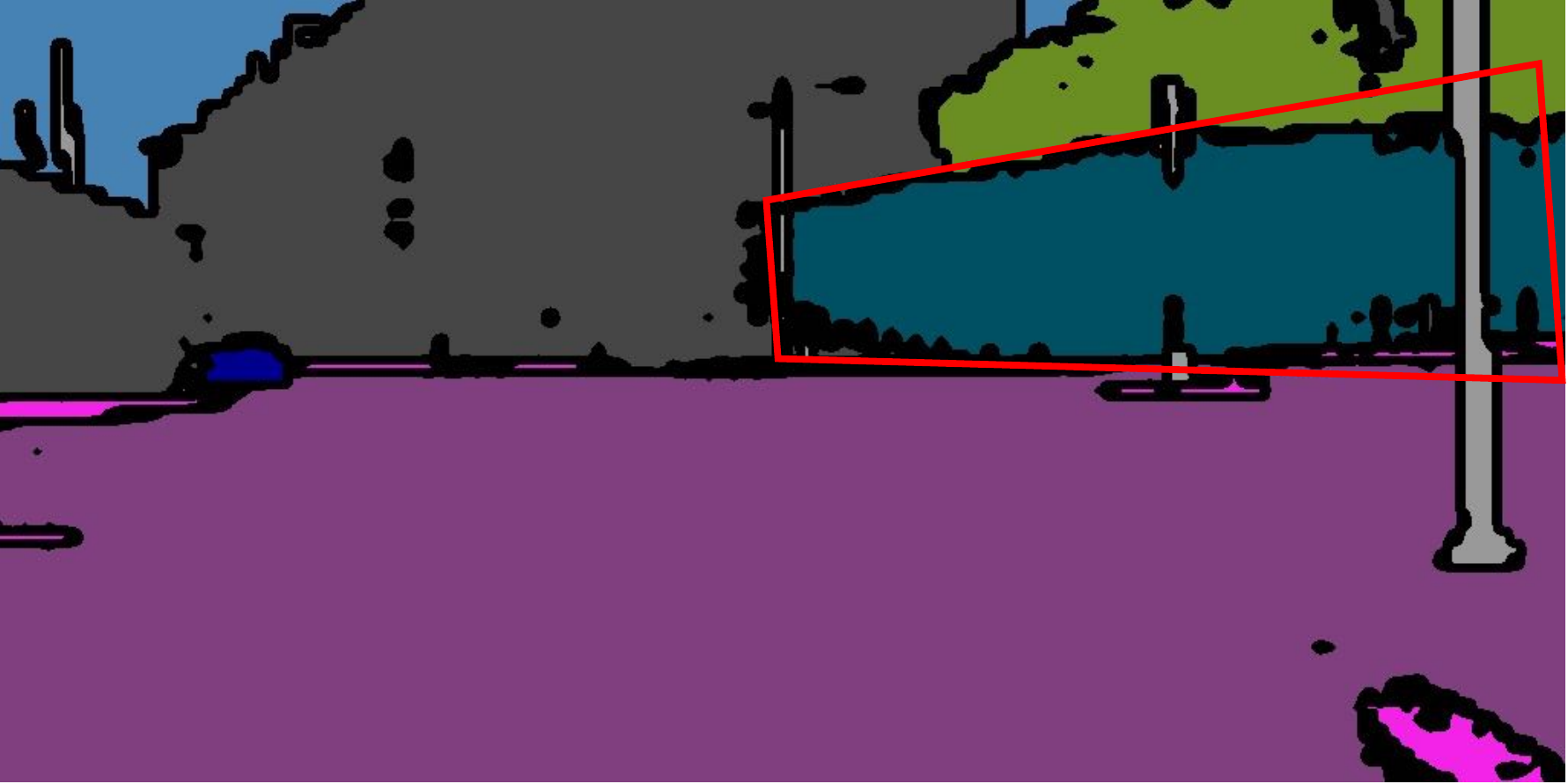}} & \hspace{-4.5mm}
			\subfloat{\includegraphics[width=0.2\linewidth]{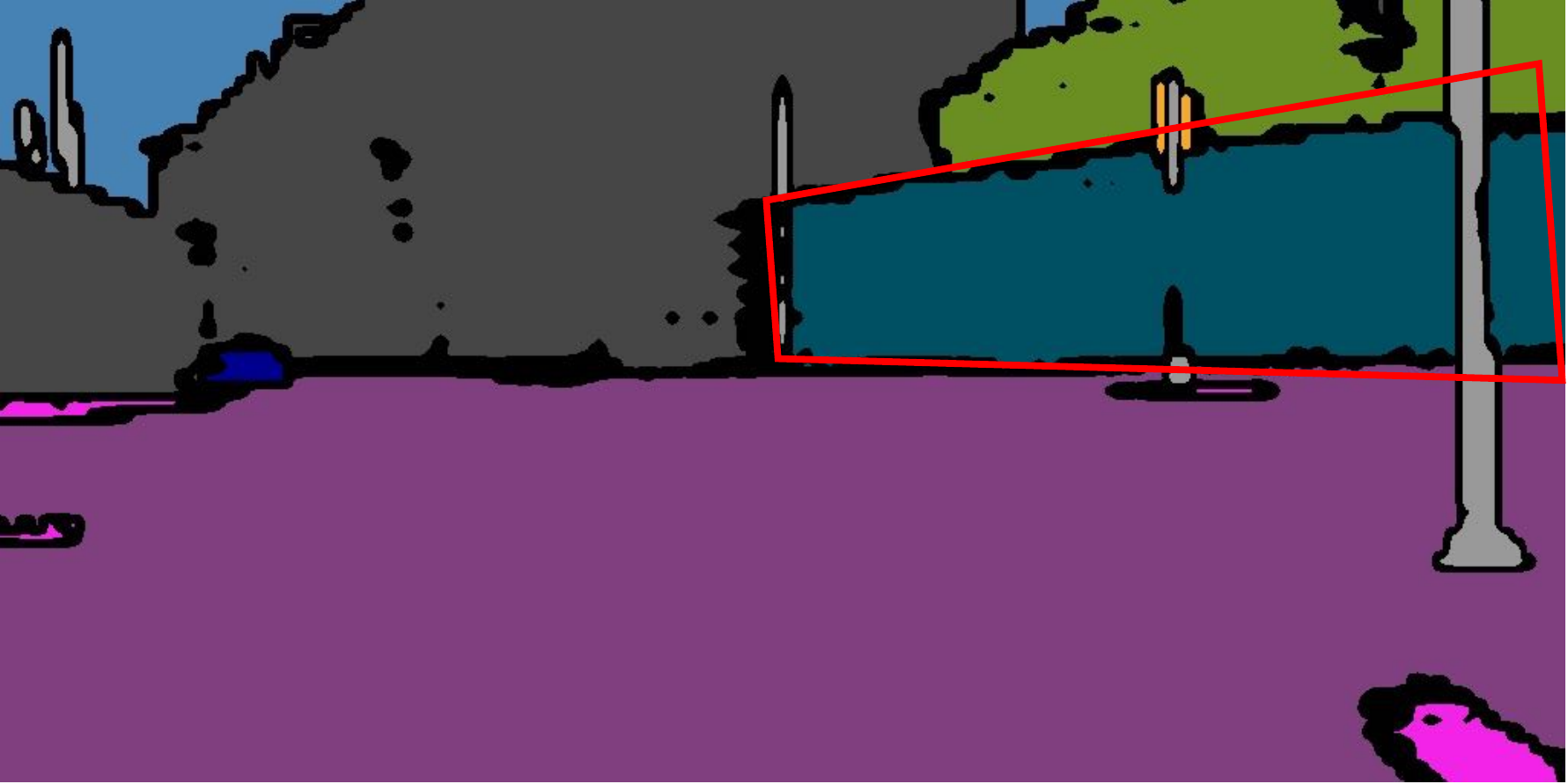}}
			\\
			\subfloat{\includegraphics[width=0.2\linewidth]{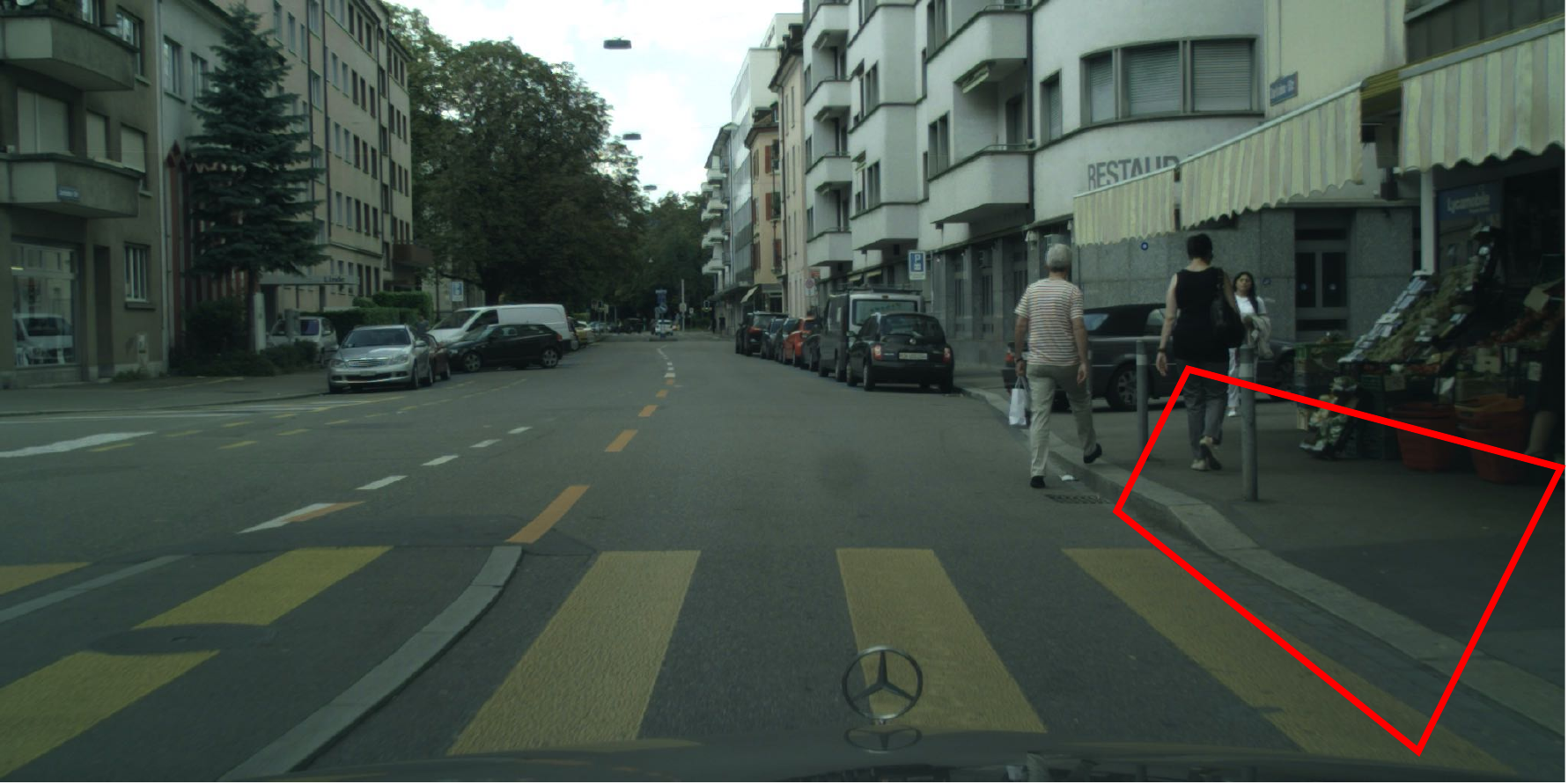}} & \hspace{-4.5mm}		
			\subfloat{\includegraphics[width=0.2\linewidth]{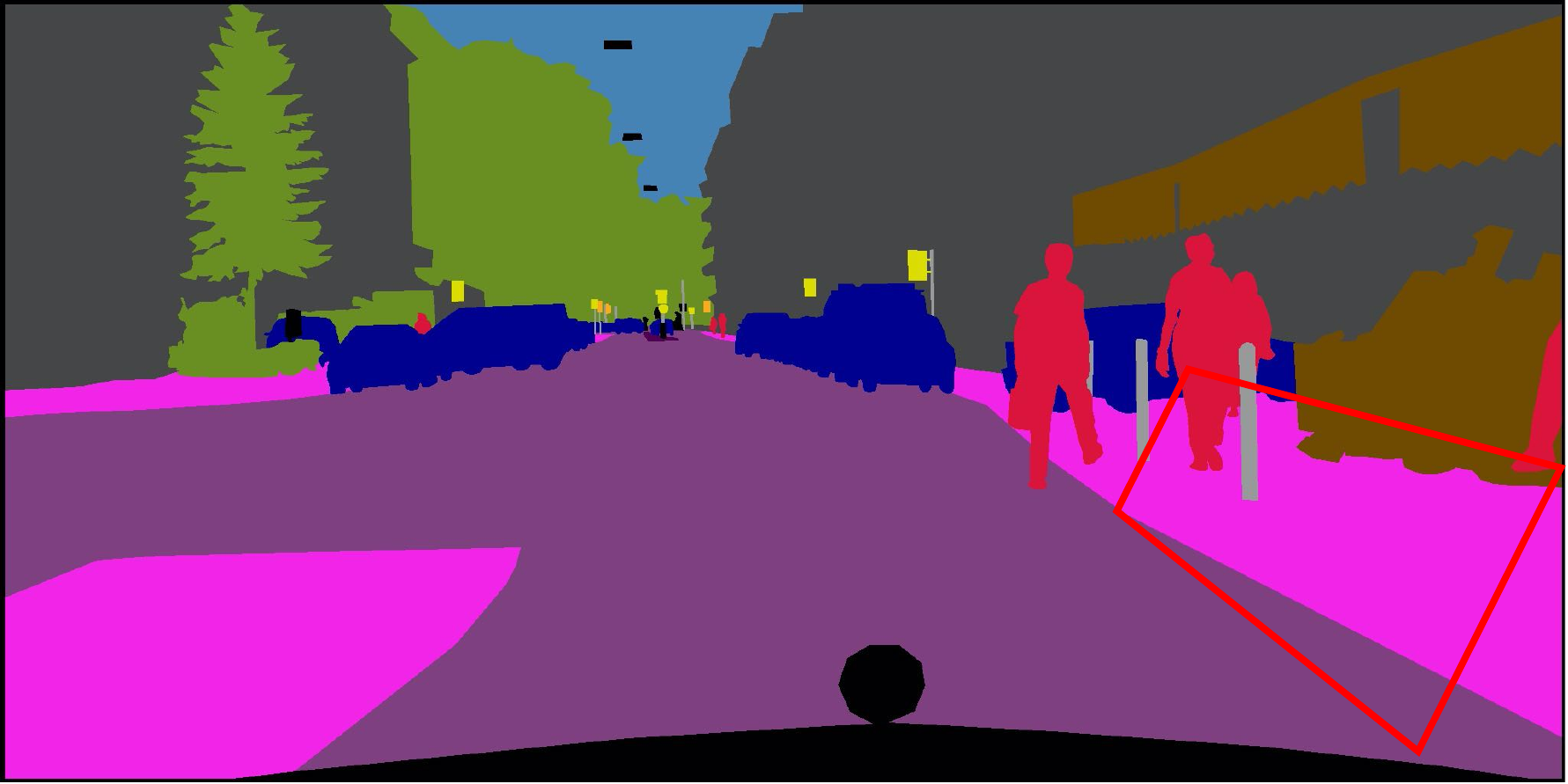}} & \hspace{-4.5mm}
			\subfloat{\includegraphics[width=0.2\linewidth]{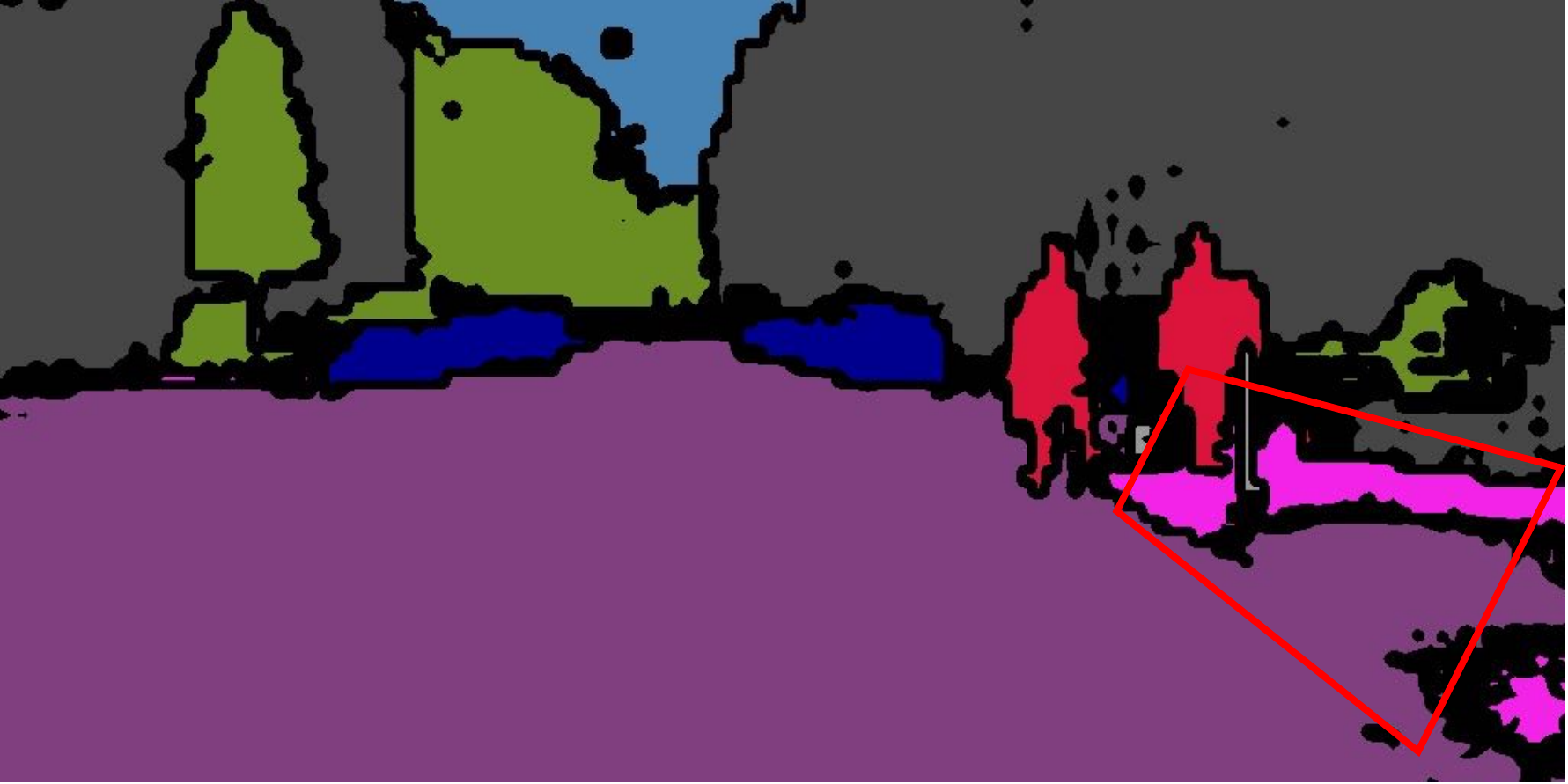}} & \hspace{-4.5mm}
			\subfloat{\includegraphics[width=0.2\linewidth]{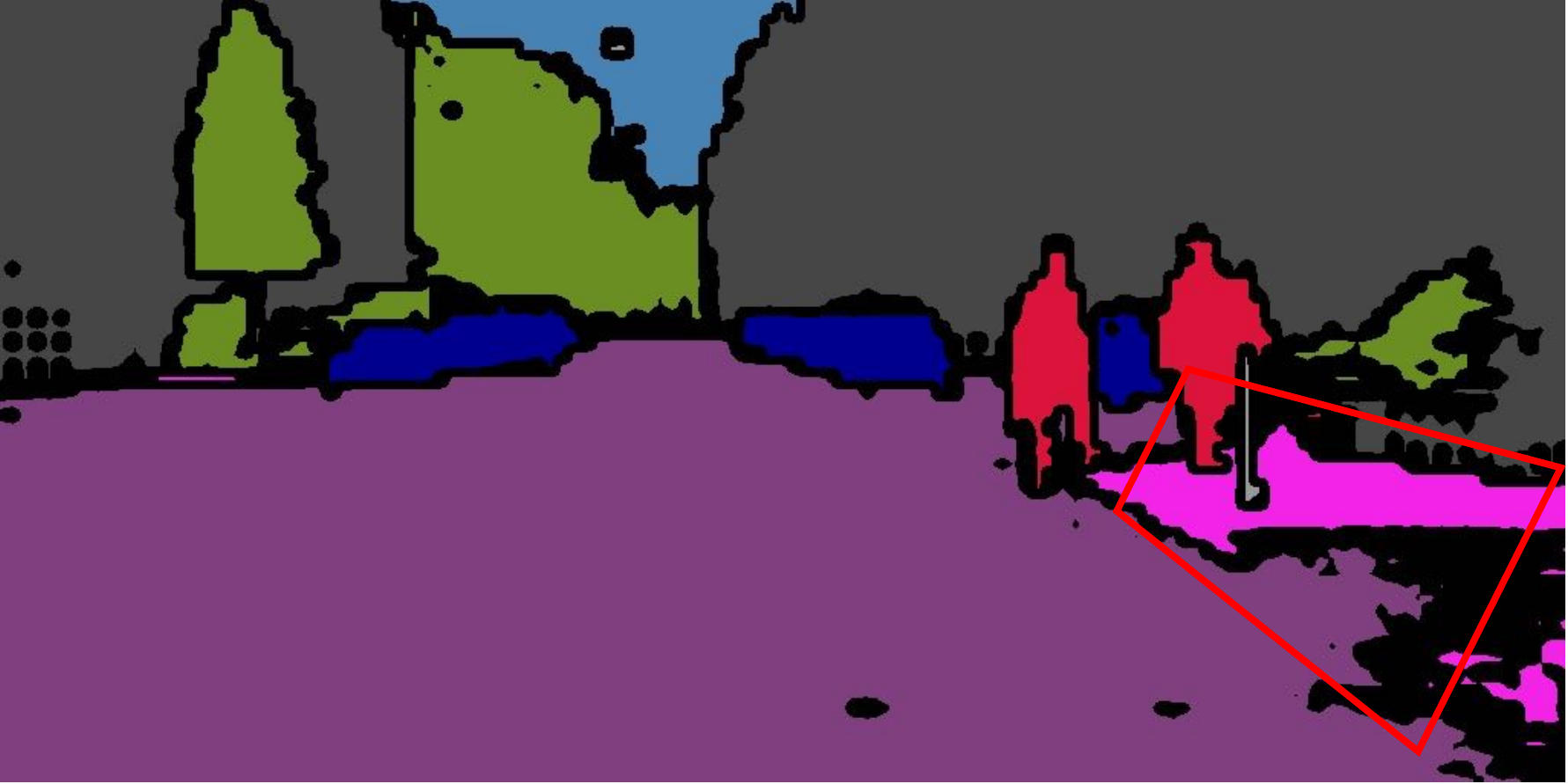}} & \hspace{-4.5mm}
			\subfloat{\includegraphics[width=0.2\linewidth]{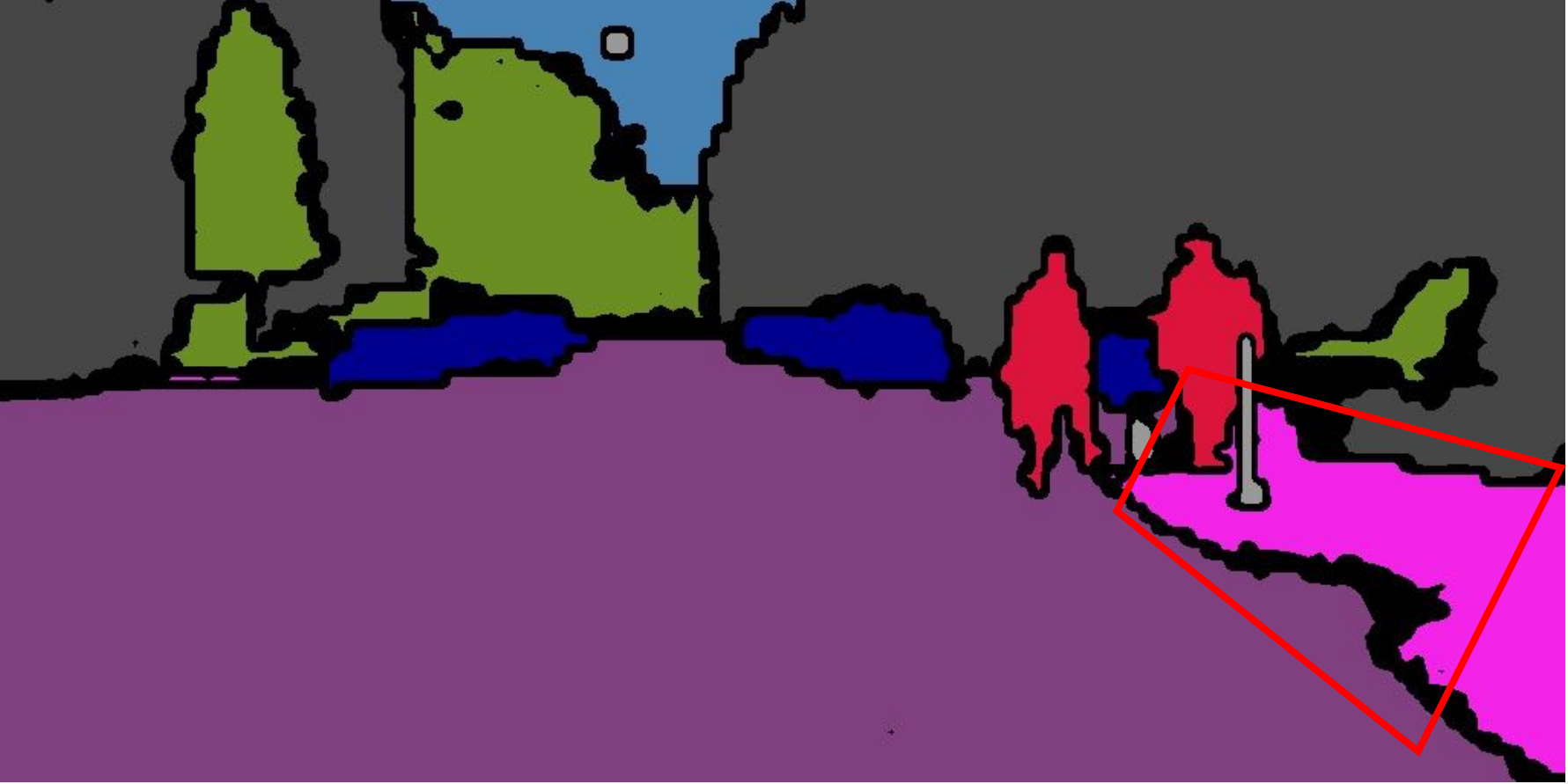}}
			\\
		
			\subfloat{\includegraphics[width=0.2\linewidth]{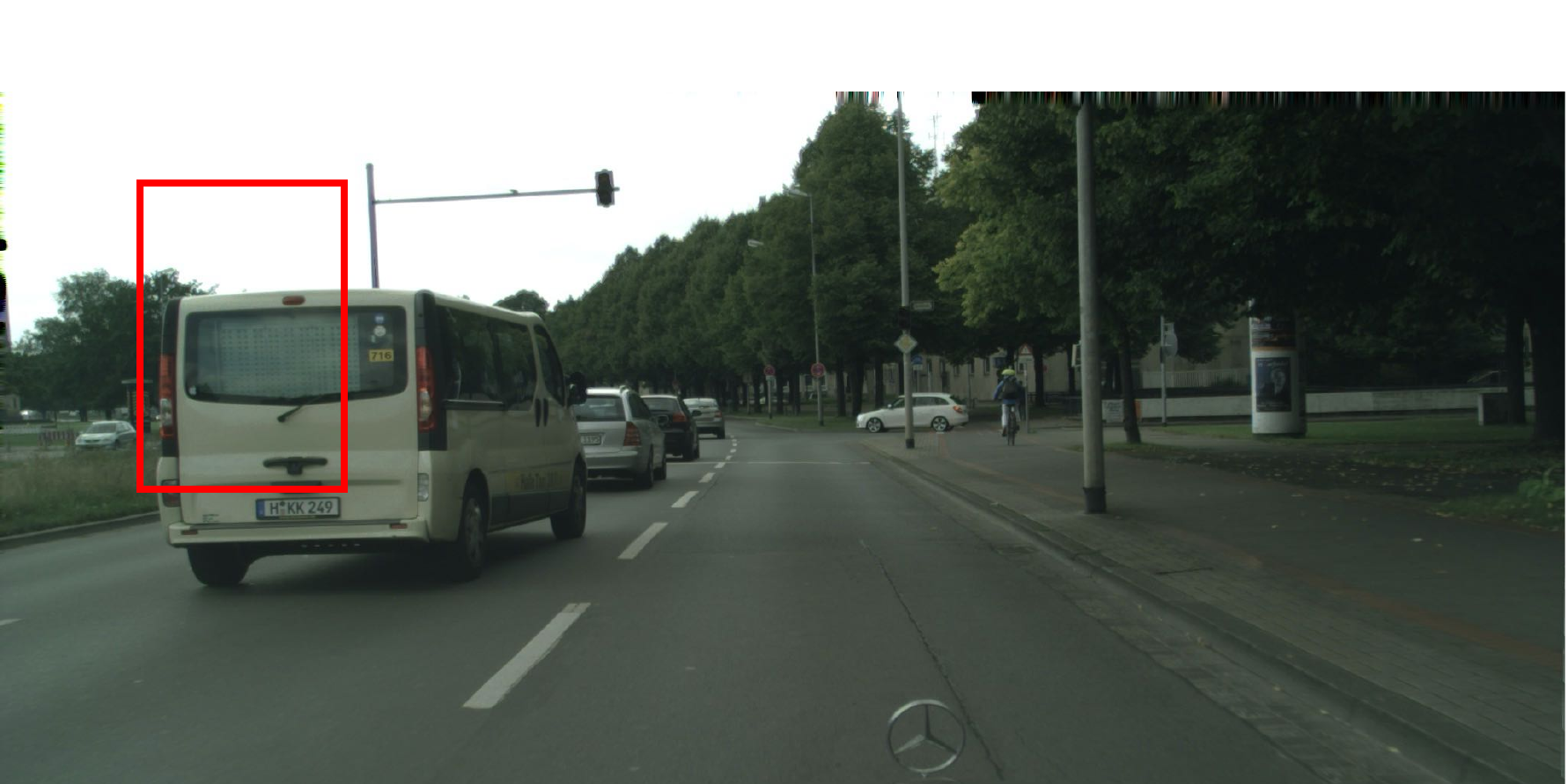}} & \hspace{-4.5mm}		
			\subfloat{\includegraphics[width=0.2\linewidth]{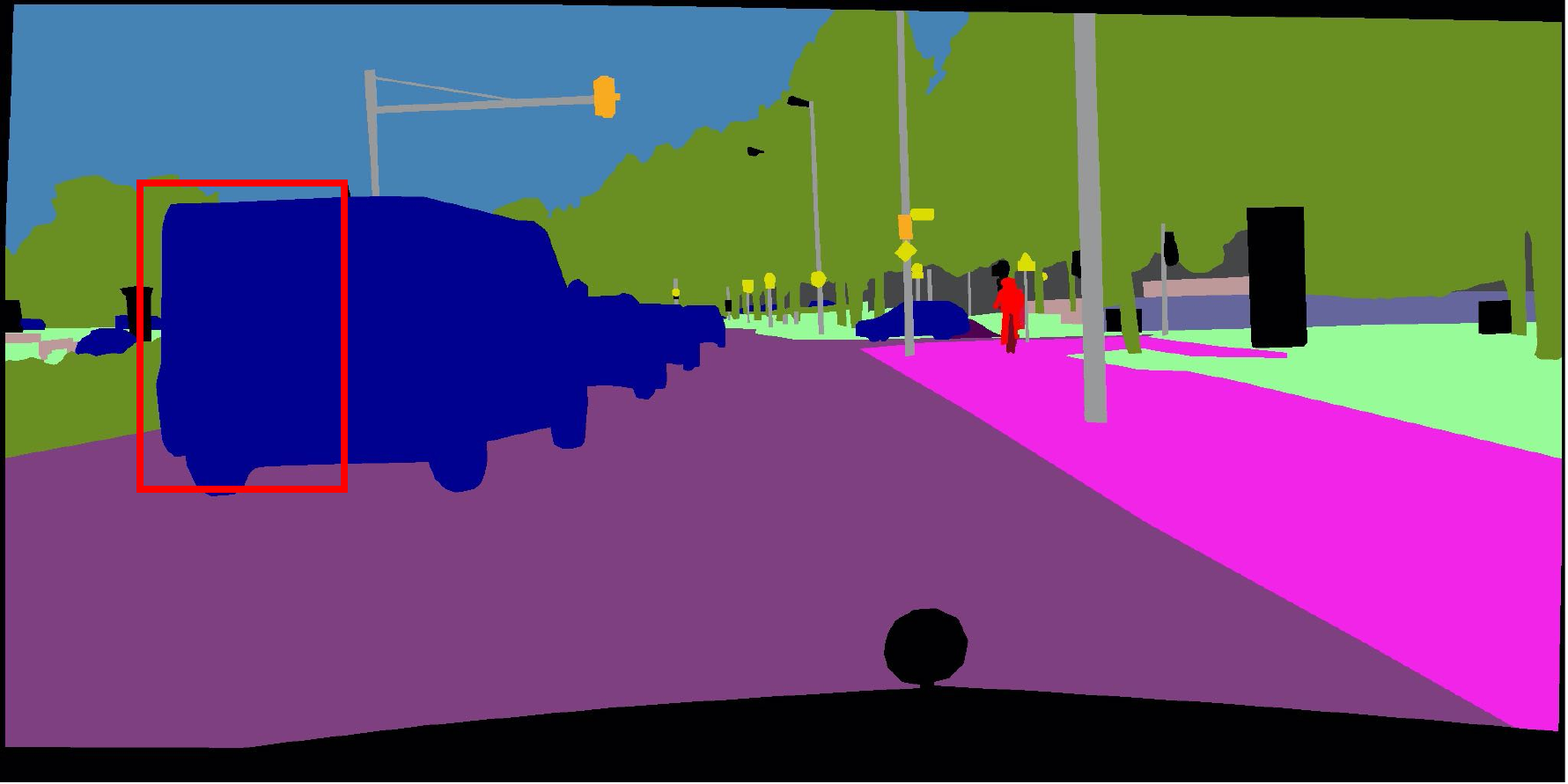}} & \hspace{-4.5mm}
			\subfloat{\includegraphics[width=0.2\linewidth]{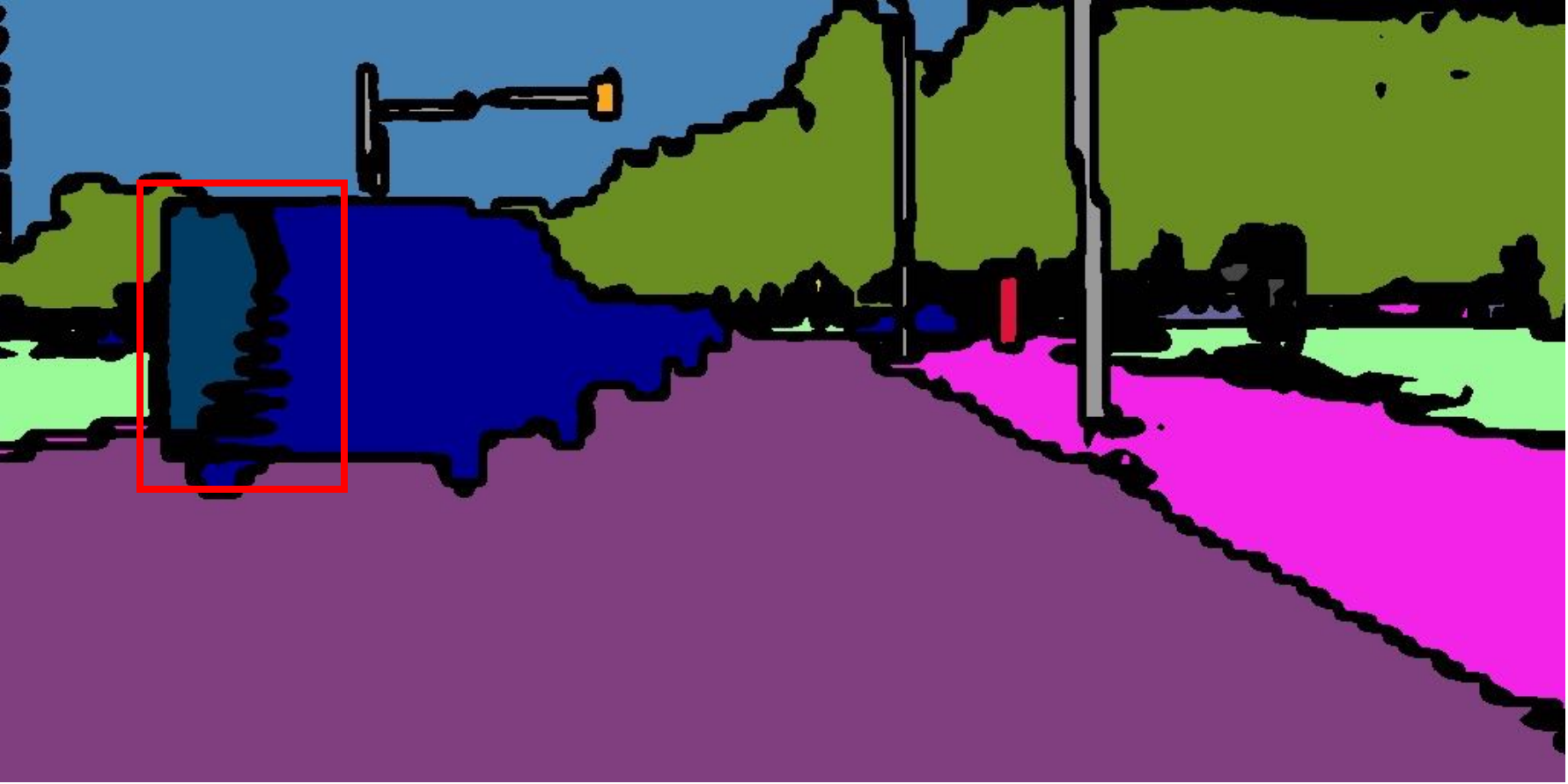}} & \hspace{-4.5mm}
			\subfloat{\includegraphics[width=0.2\linewidth]{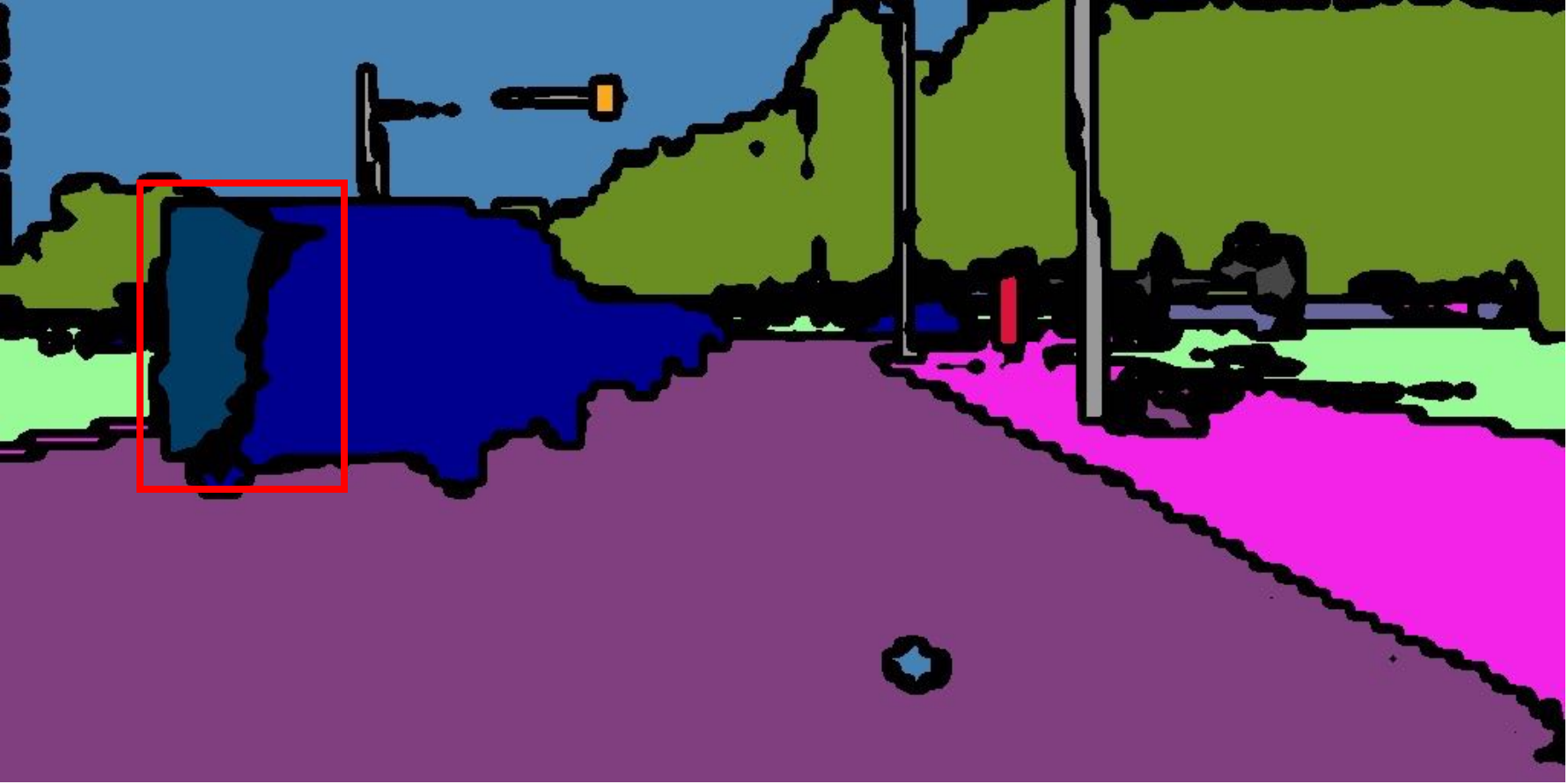}} & \hspace{-4.5mm}
			\subfloat{\includegraphics[width=0.2\linewidth]{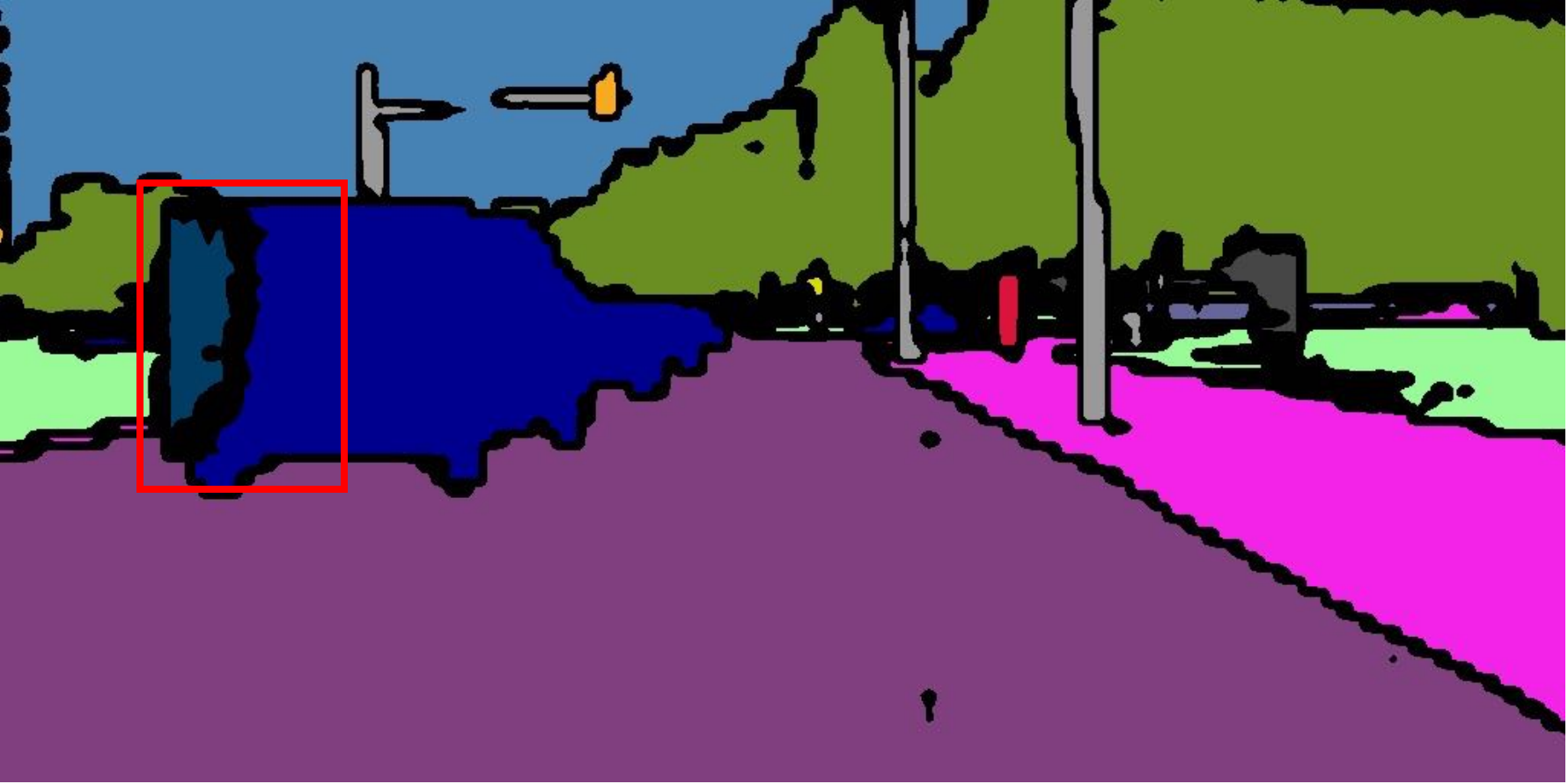}}
			\\
			(a) Raw images& (b) Ground-truth & (c) SPPLG (path-$\mathcal{S}$) & (d) SPPLG (path-$\mathcal{T}$) & (e) DPPLG 
	\end{tabular}}
	\caption{Pseudo label visualization of different pseudo label generation strategies (GTA5$\rightarrow$Cityscapes). (a) raw images from Cityscapes dataset; (b) ground-truth; (c) pseudo label generated by SPPLG (path-$\mathcal{S}$); (d) pseudo label generated by SPPLG (path-$\mathcal{T}$); (e) pseudo label generated by DPPLG strategy. Red rectangles highlight the differences.}
	\label{fig:pseudo_vis}
\end{figure*}

{\noindent \textbf{Visualization of Pseudo Labels.}}\hspace{3pt}
In Figure \ref{fig:pseudo_vis}, we visualize pseudo labels generated by different pseudo label generation strategies. Recall that path-$\mathcal{S}$ and path-$\mathcal{T}$ can generate respective pseudo label separately, which is named as single path pseudo label generation (SPPLG) strategy. We visualize pseudo label generated by SPPLG (in path-$\mathcal{S}$ and path-$\mathcal{T}$ respectively) and DPPLG for comparison. Obviously, DPPLG generates more accurate pseudo labels in both small objects (e.g., pole and t-light in row 1,2 and person in row 3) and indistinguishable categories (e.g., train in row 4, sidewalk in row 5 and car in row 6). The visualization further demonstrates the effectiveness of our DPPLG strategy. 

\clearpage
 
\begin{figure*}[h]
	\centering
	\scalebox{0.8}{
		\begin{tabular}{cccc}
			\subfloat{\includegraphics[width=0.24\linewidth]{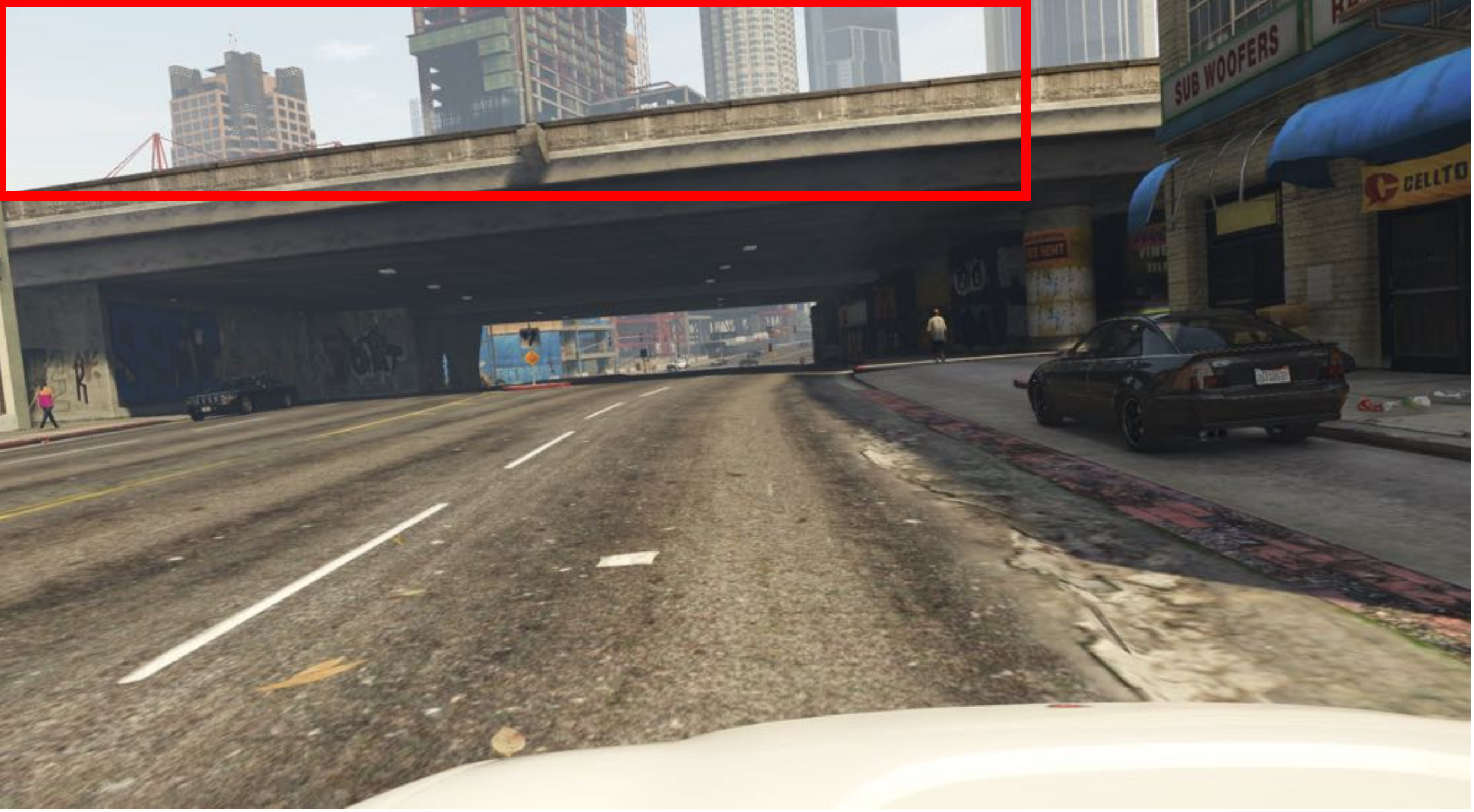}} &
			\subfloat{\includegraphics[width=0.24\linewidth]{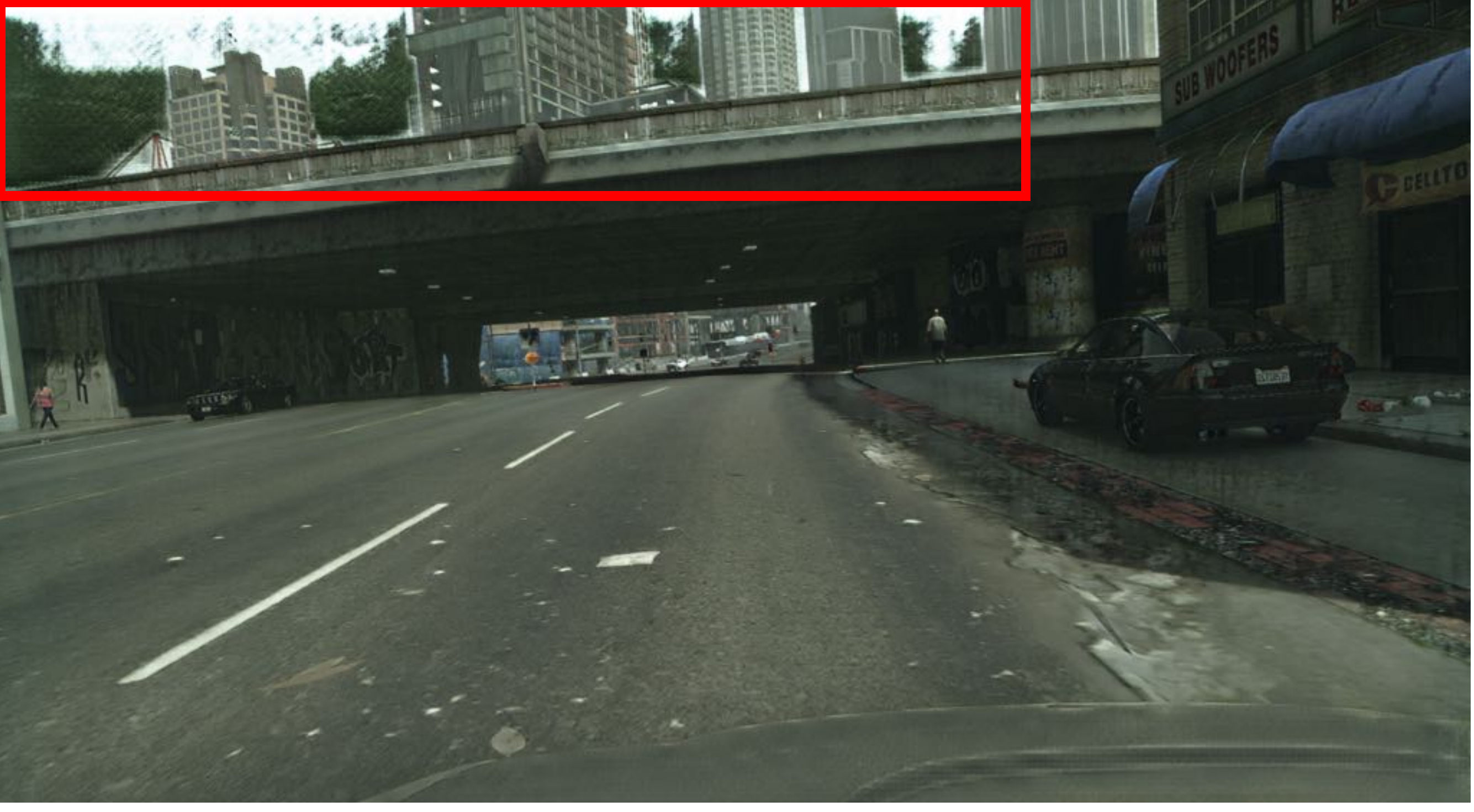}} &
			\subfloat{\includegraphics[width=0.24\linewidth]{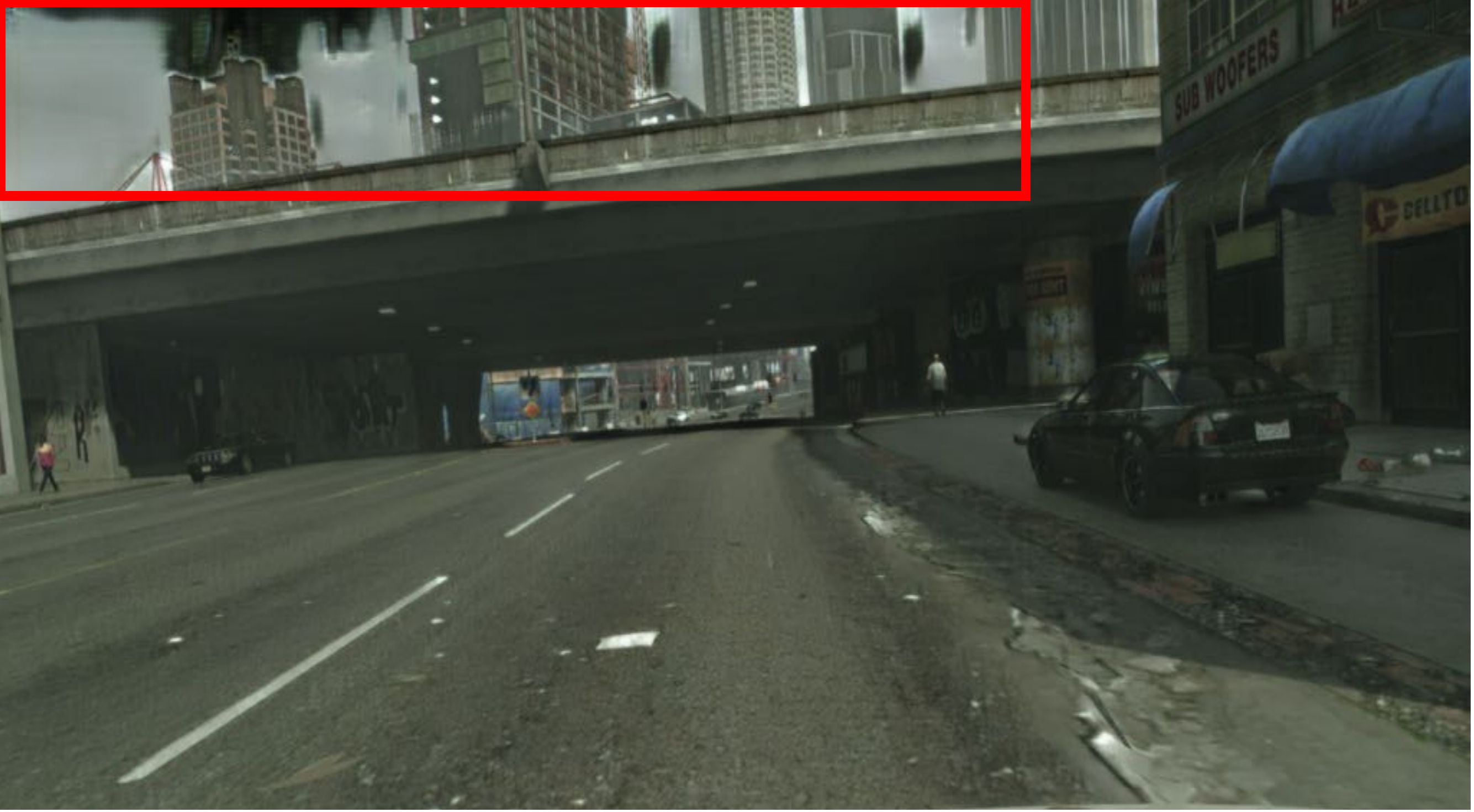}} &
			\subfloat{\includegraphics[width=0.24\linewidth]{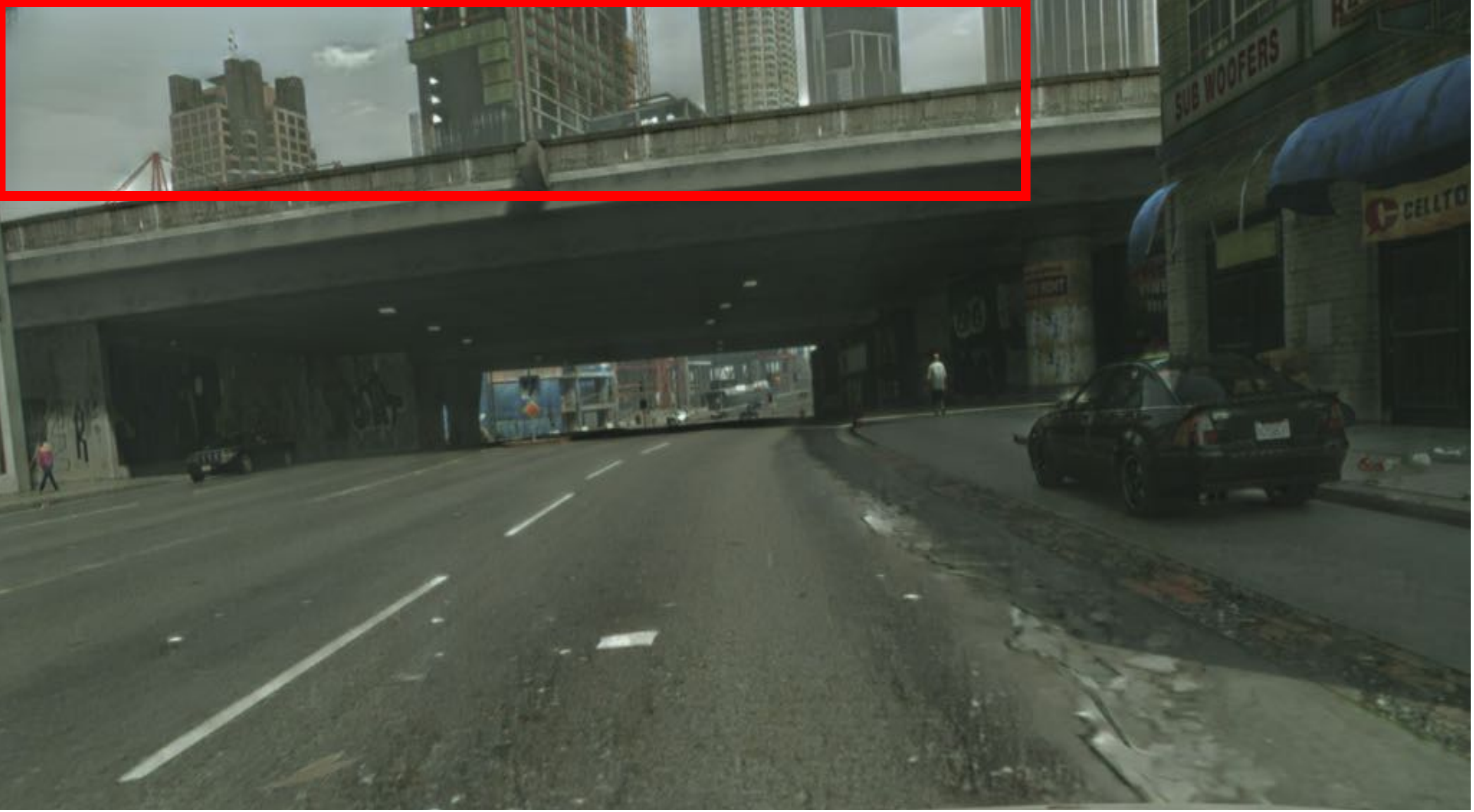}} 
			\\
			\subfloat{\includegraphics[width=0.24\linewidth]{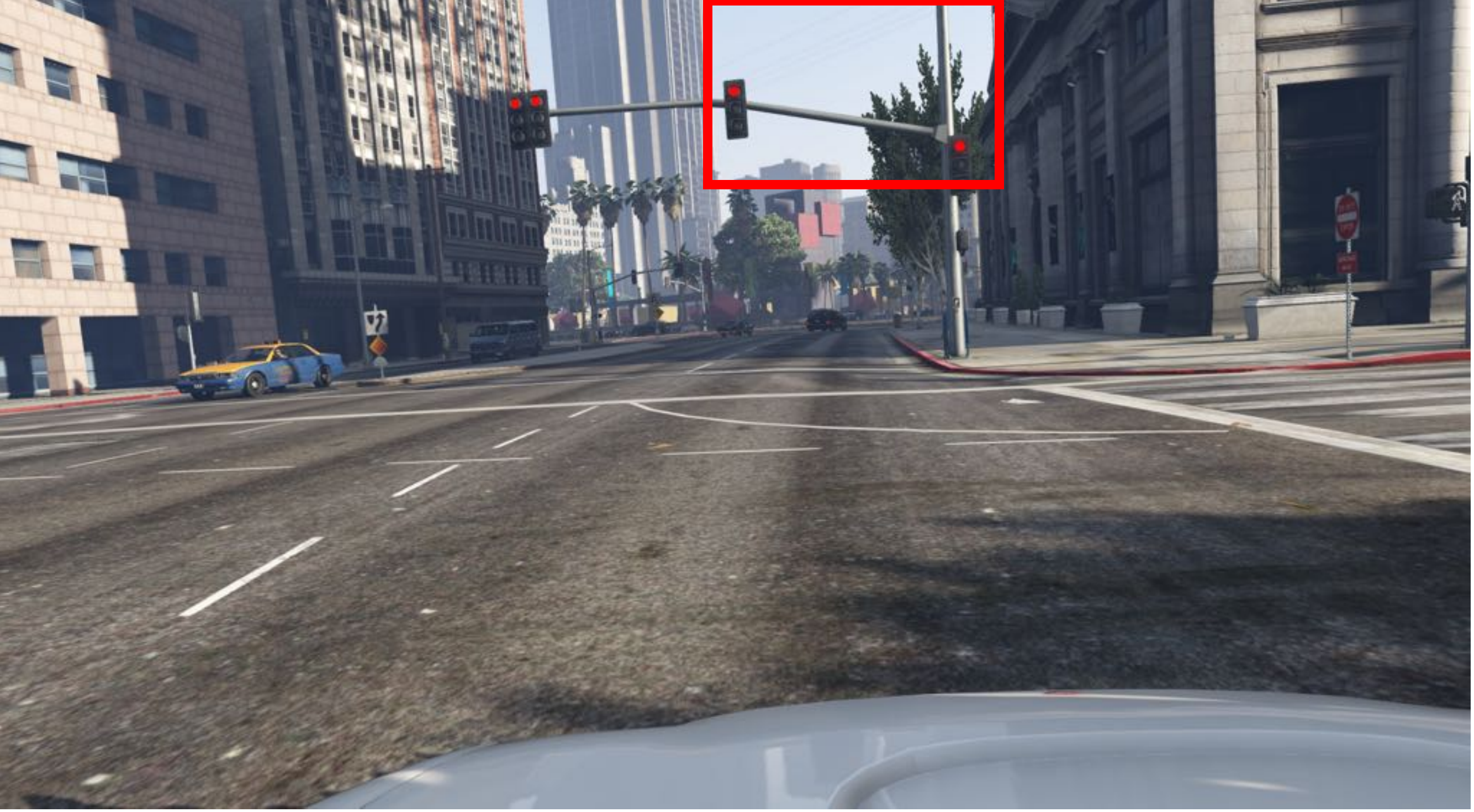}} &
			\subfloat{\includegraphics[width=0.24\linewidth]{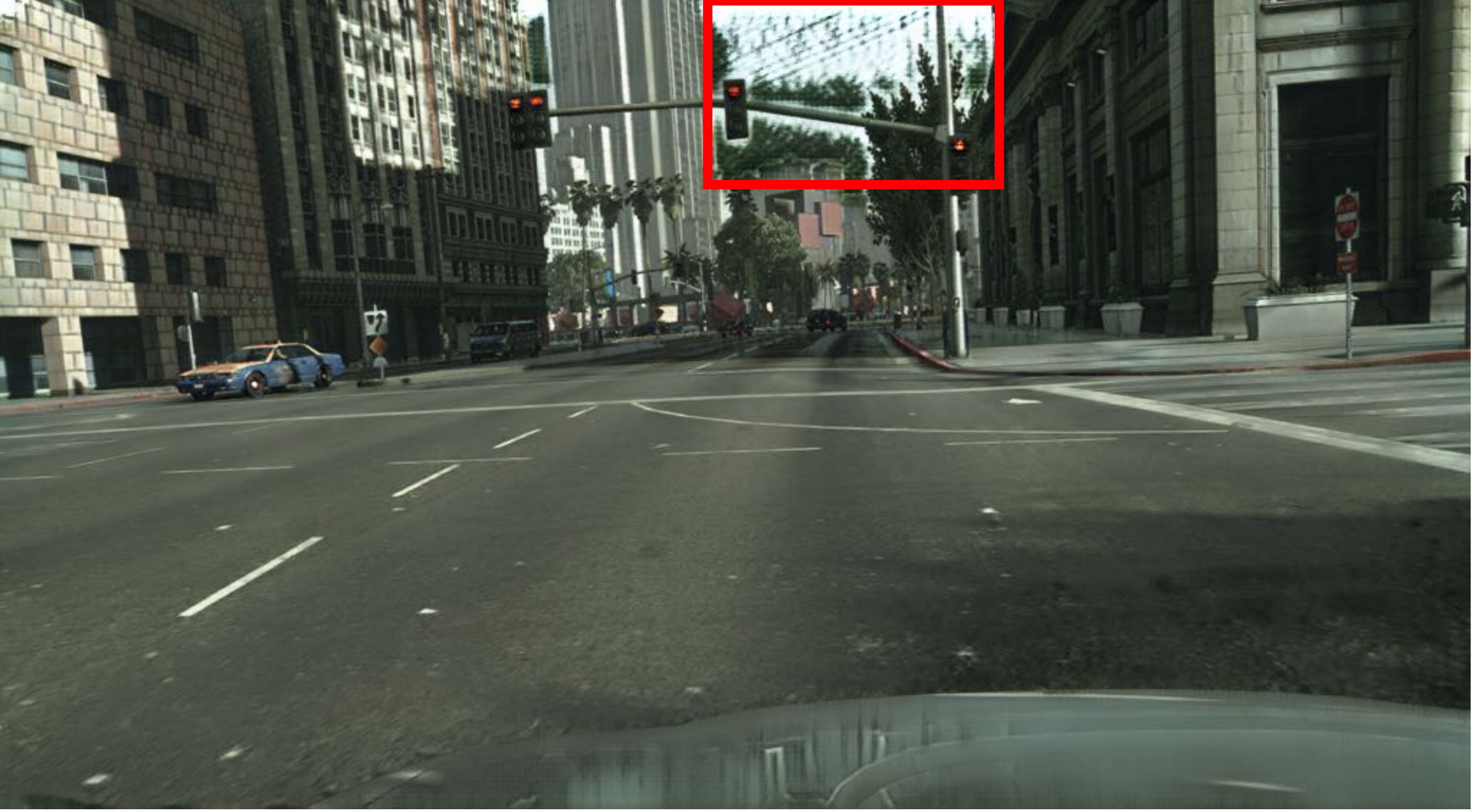}} &
			\subfloat{\includegraphics[width=0.24\linewidth]{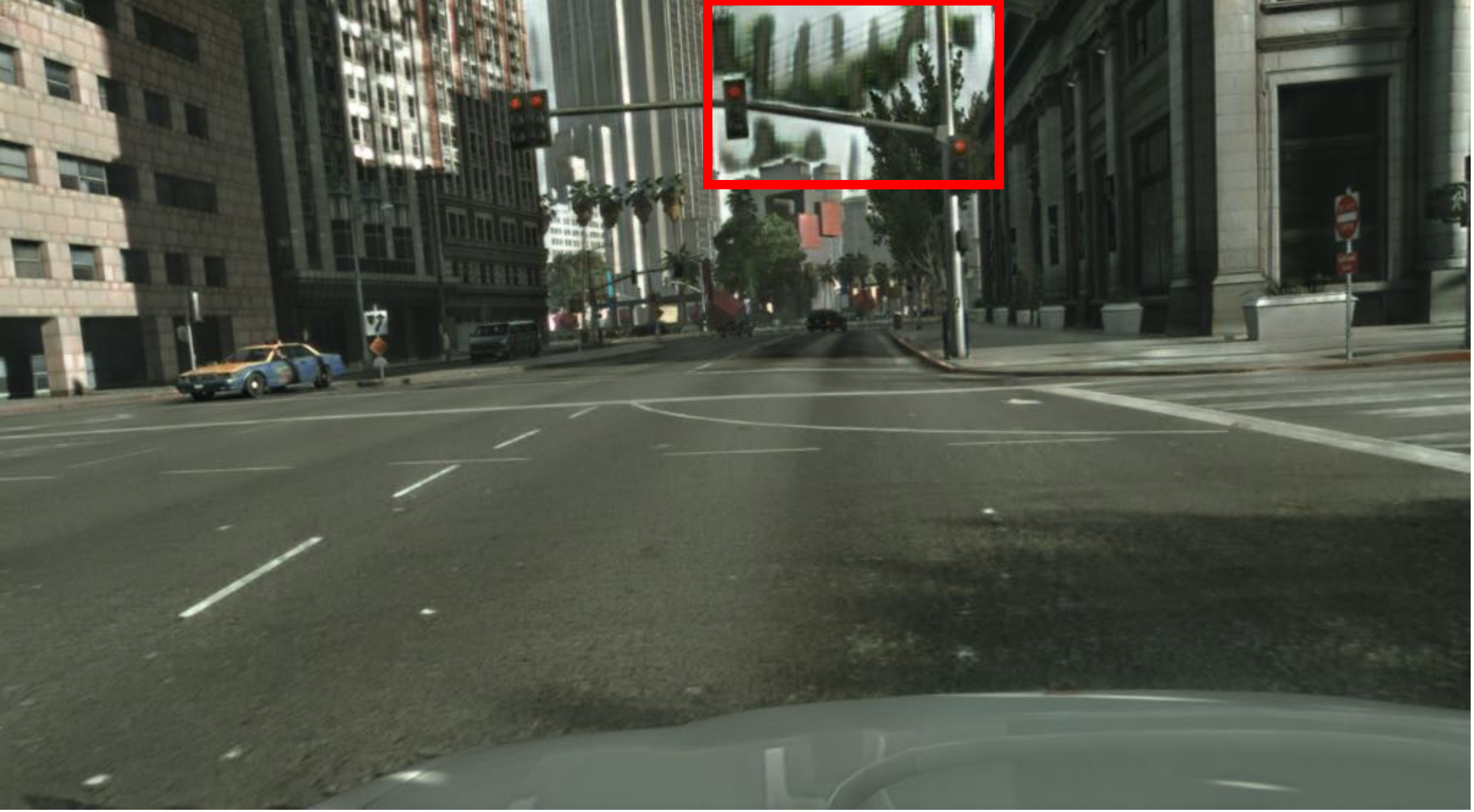}} &
			\subfloat{\includegraphics[width=0.24\linewidth]{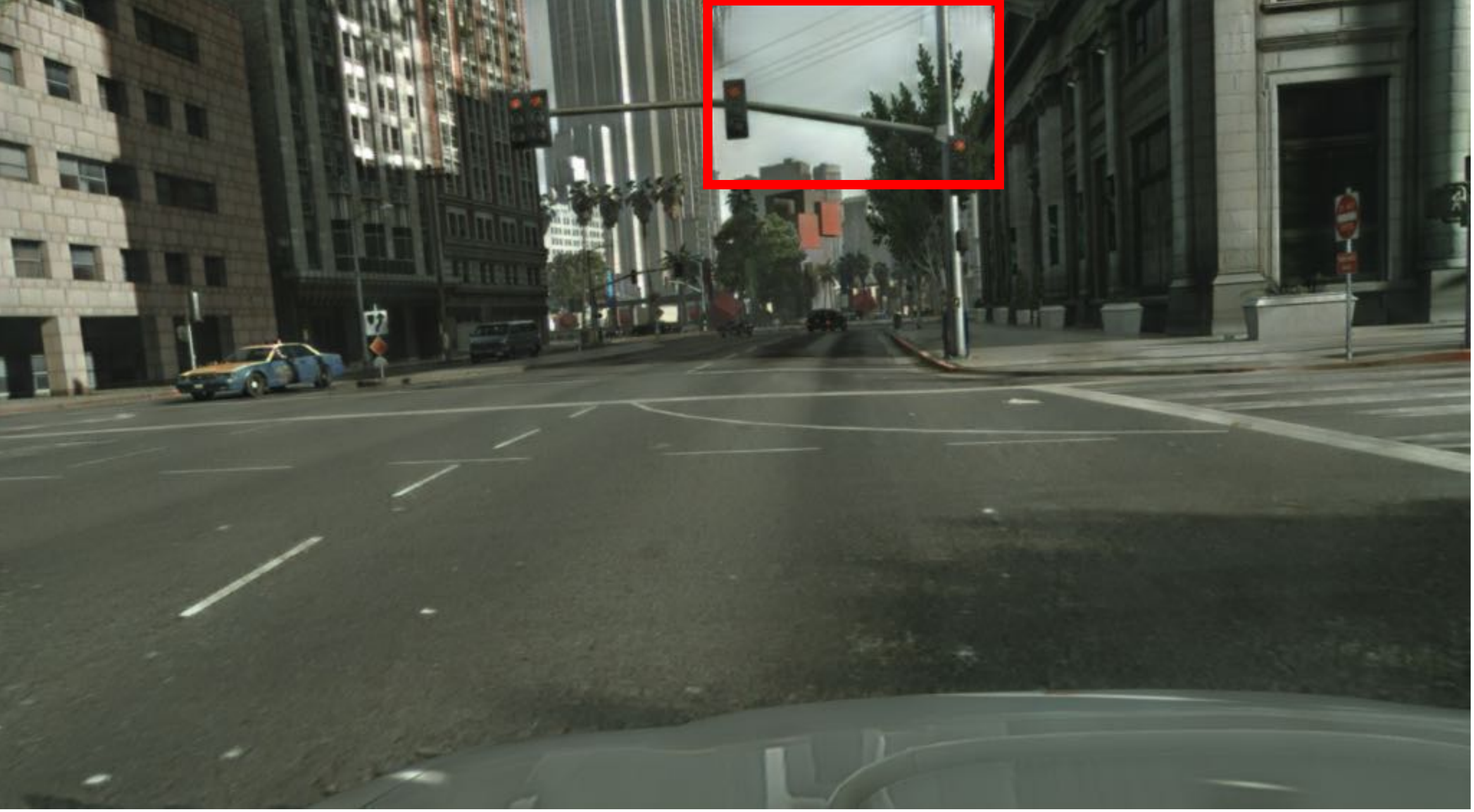}} 
			\\
			
			\subfloat{\includegraphics[width=0.24\linewidth]{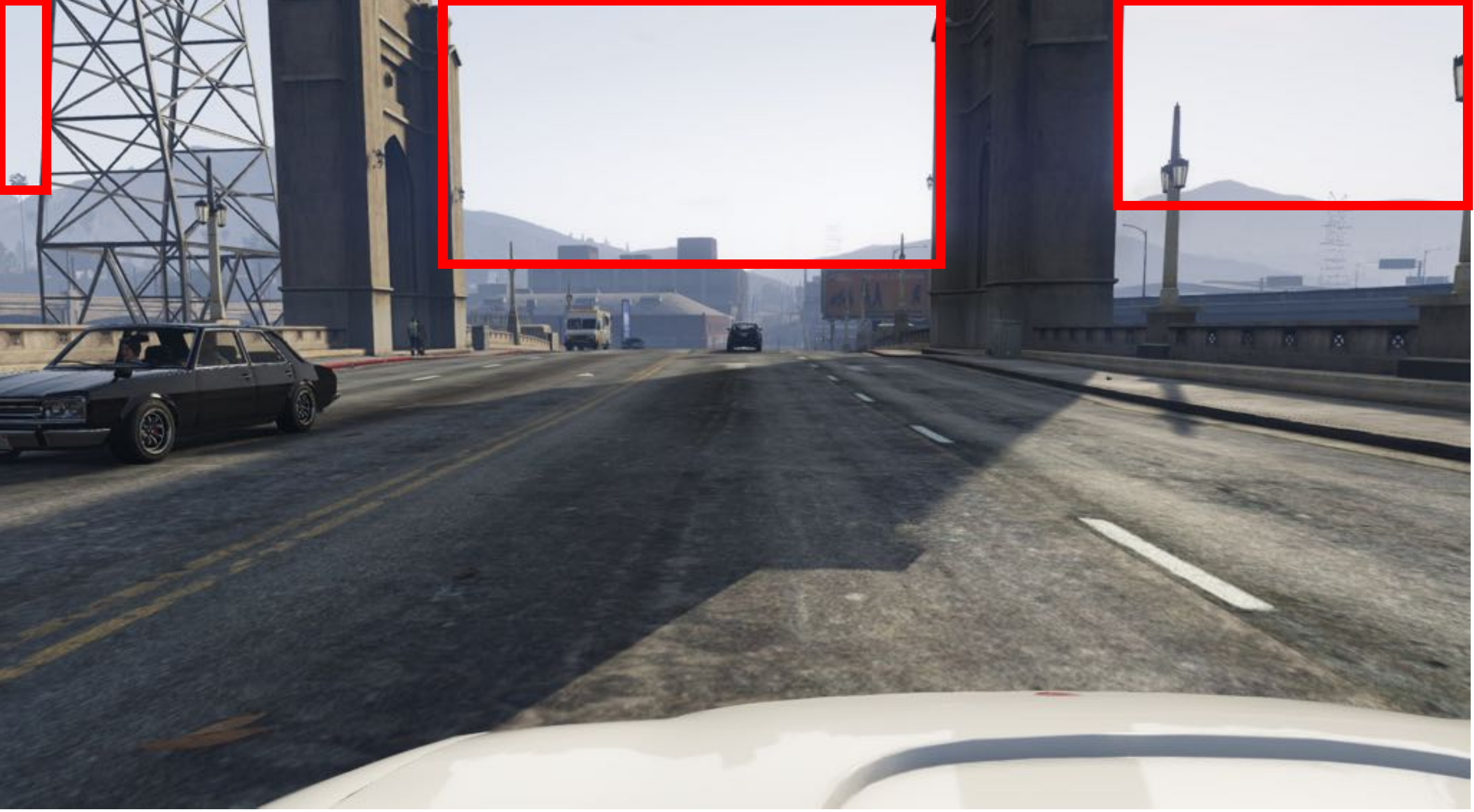}} &
			\subfloat{\includegraphics[width=0.24\linewidth]{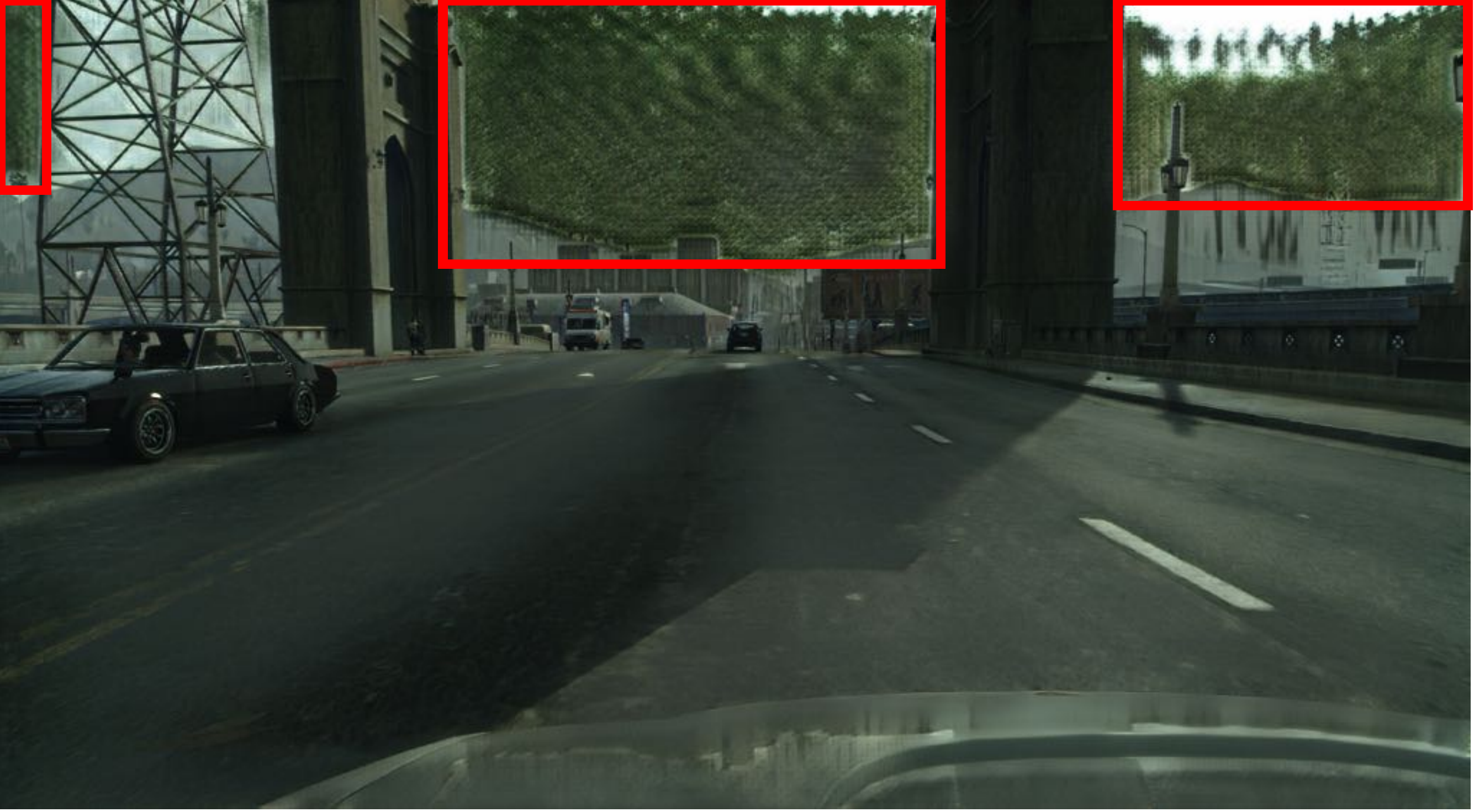}} &
			\subfloat{\includegraphics[width=0.24\linewidth]{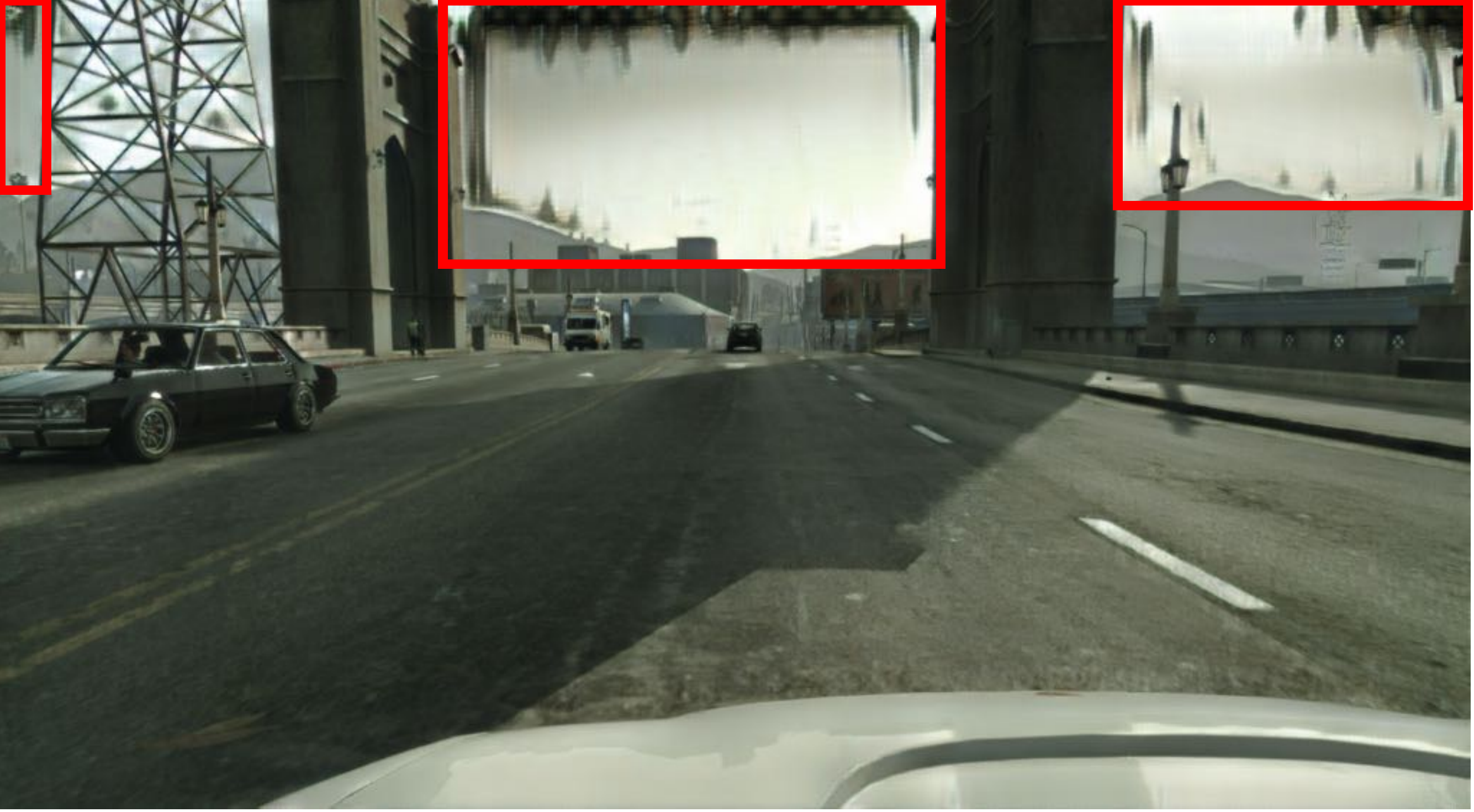}} &
			\subfloat{\includegraphics[width=0.24\linewidth]{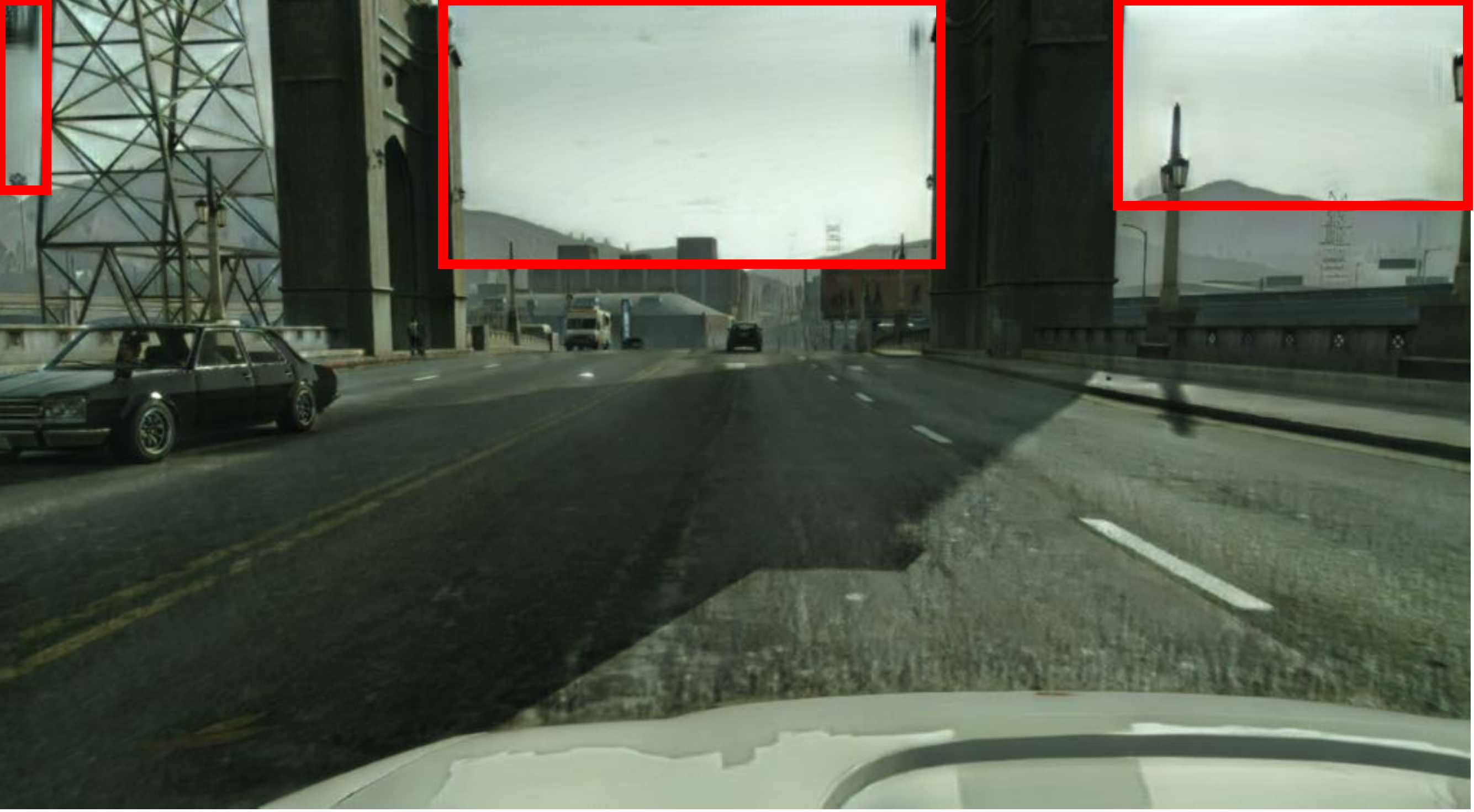}} 
			\\
			
			\subfloat{\includegraphics[width=0.24\linewidth]{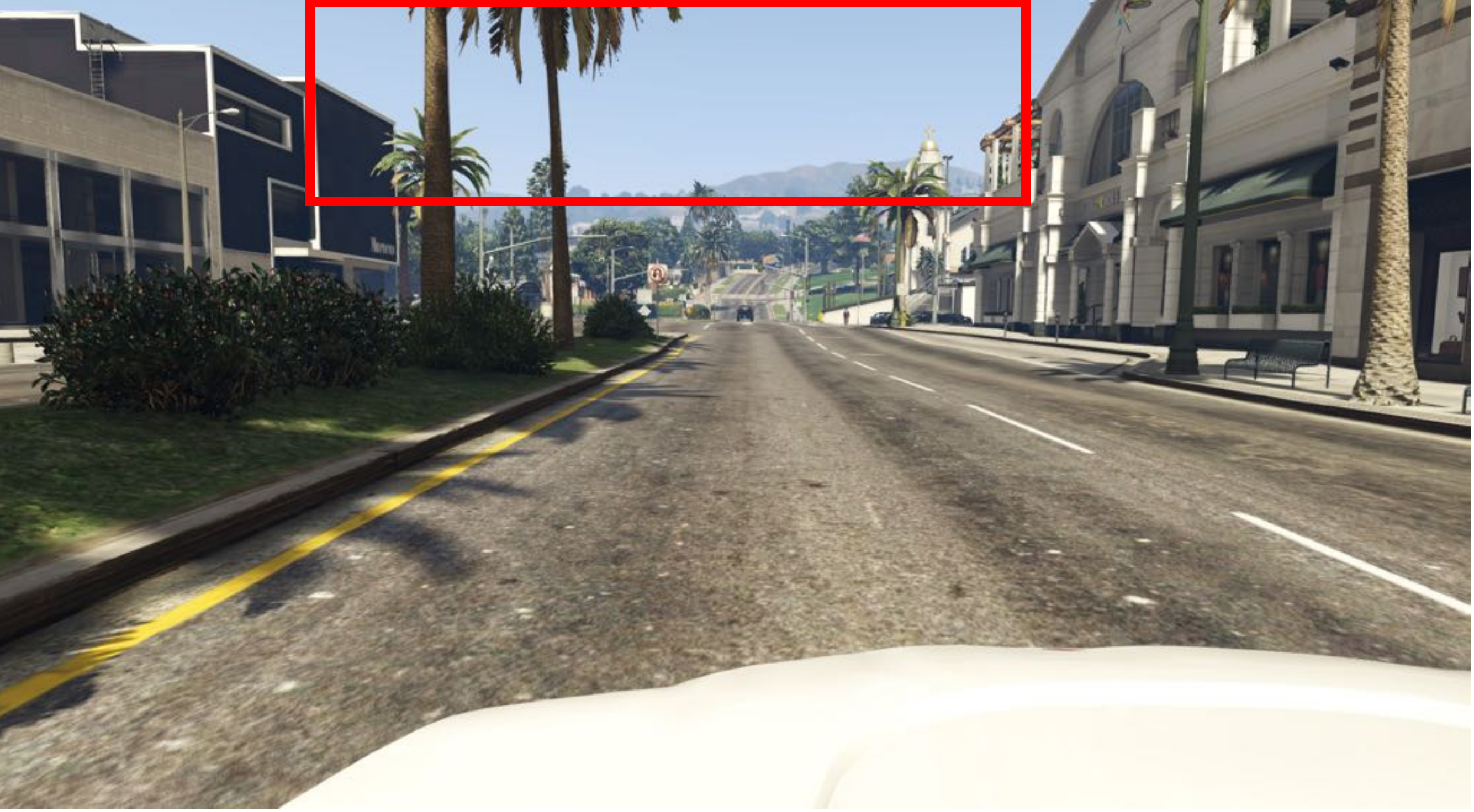}} &
			\subfloat{\includegraphics[width=0.24\linewidth]{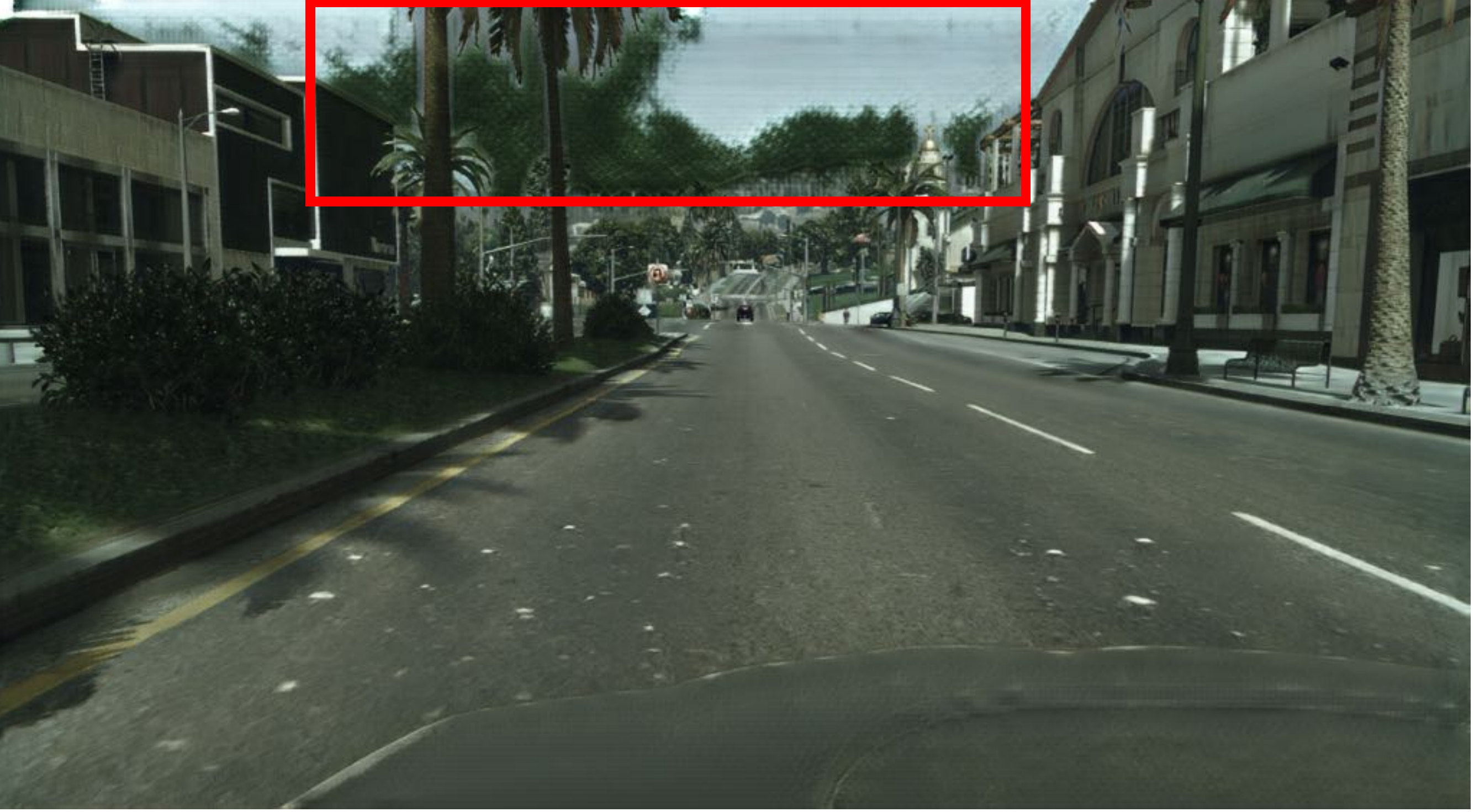}} &
			\subfloat{\includegraphics[width=0.24\linewidth]{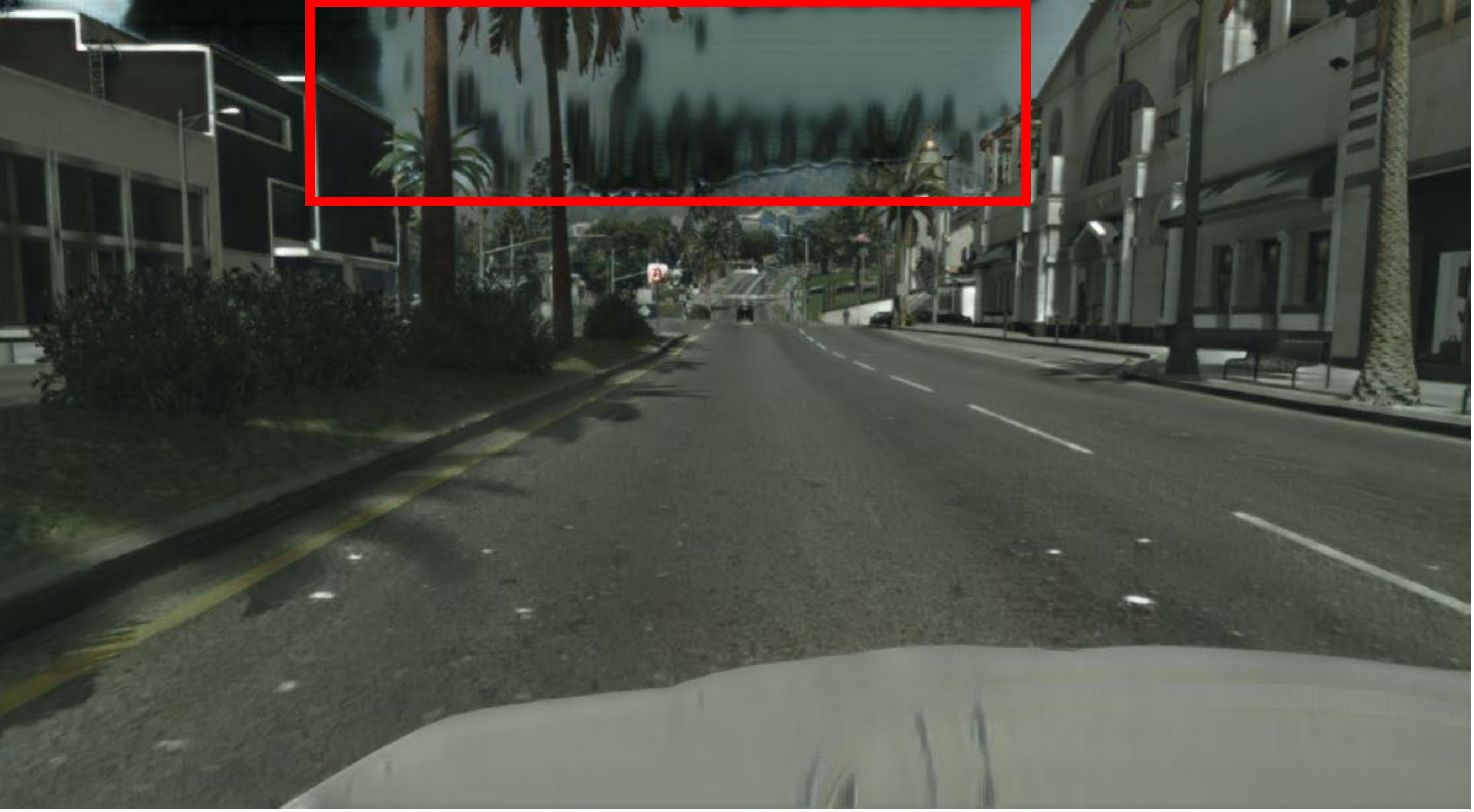}} &
			\subfloat{\includegraphics[width=0.24\linewidth]{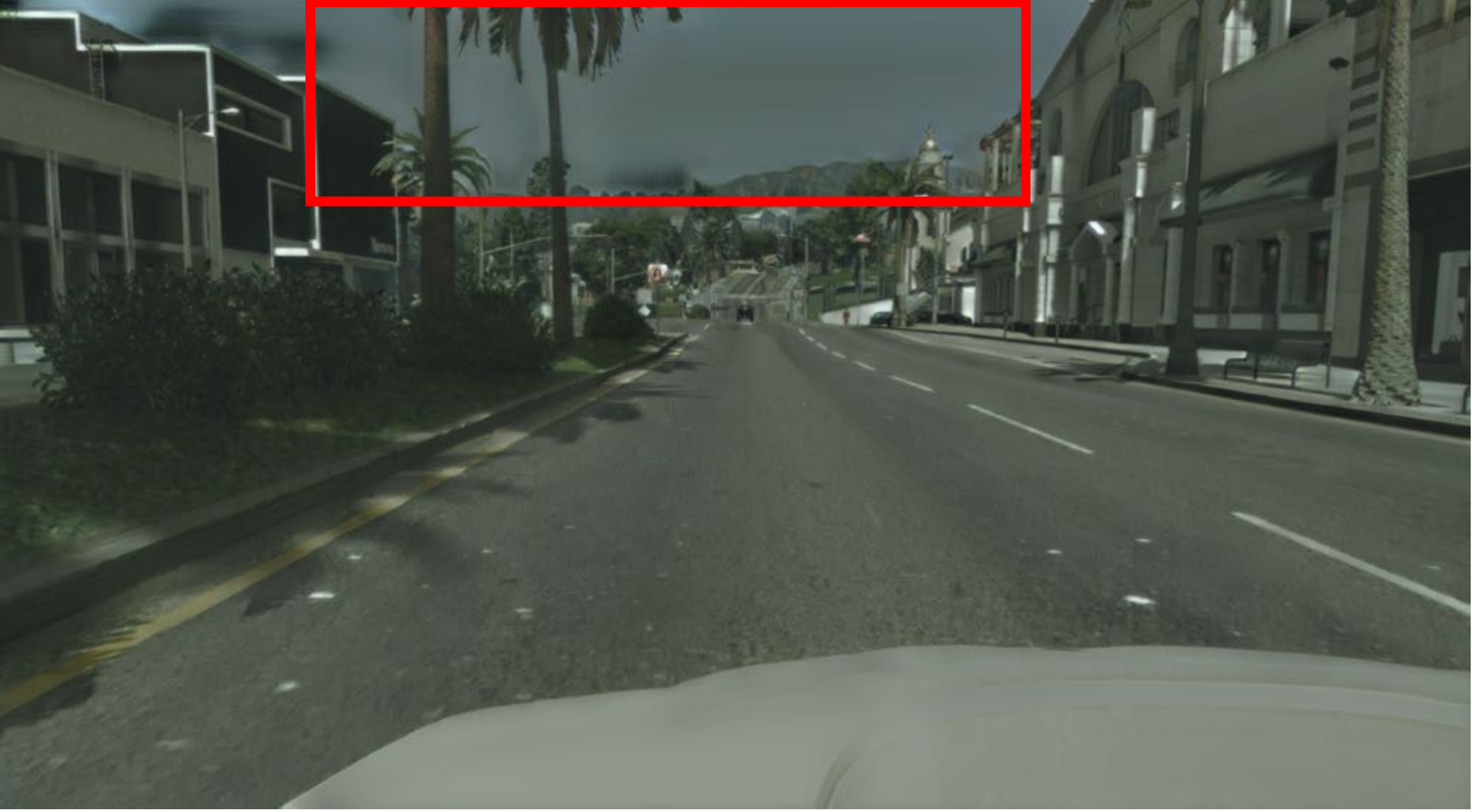}} 
			\\
			
			\subfloat{\includegraphics[width=0.24\linewidth]{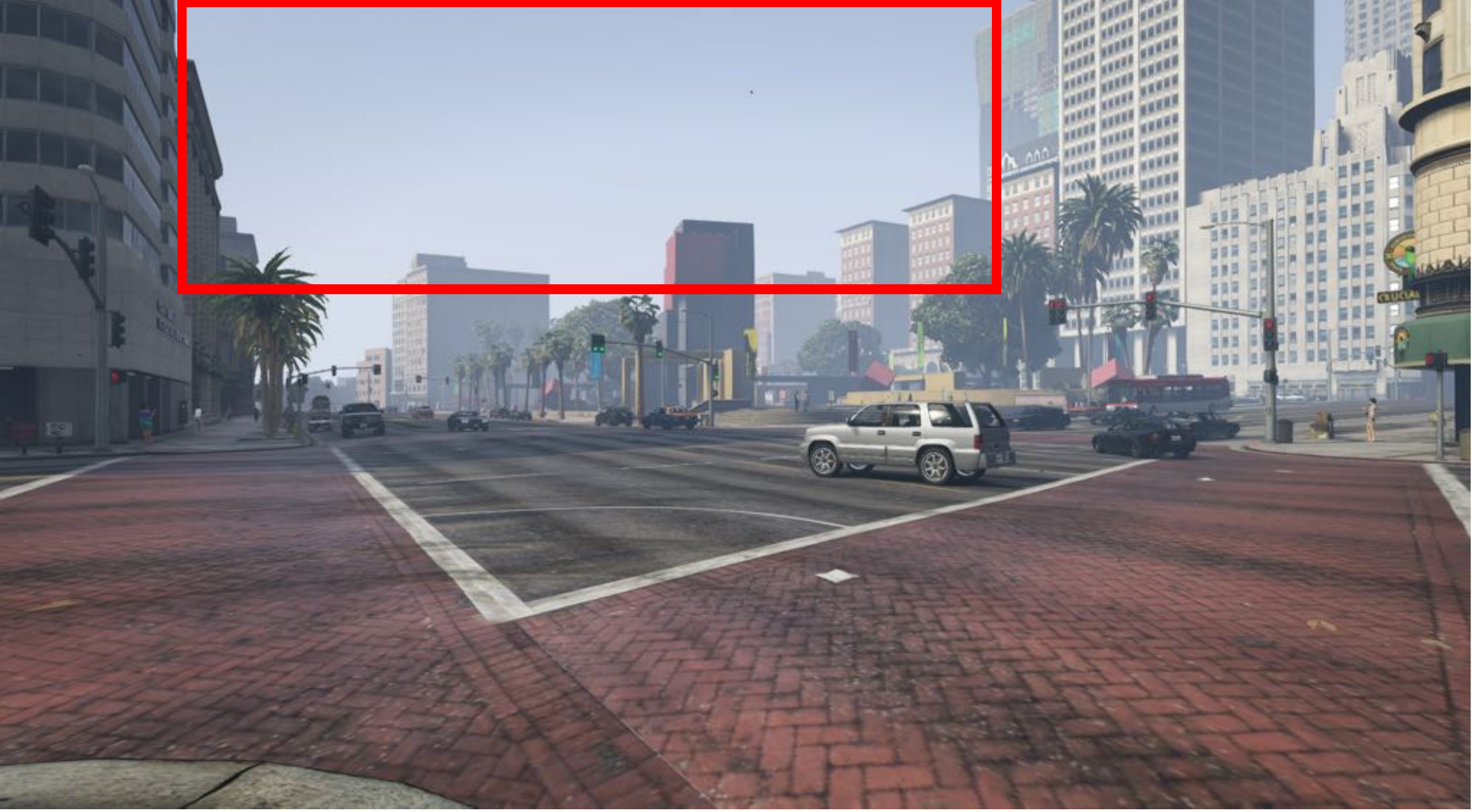}} &
			\subfloat{\includegraphics[width=0.24\linewidth]{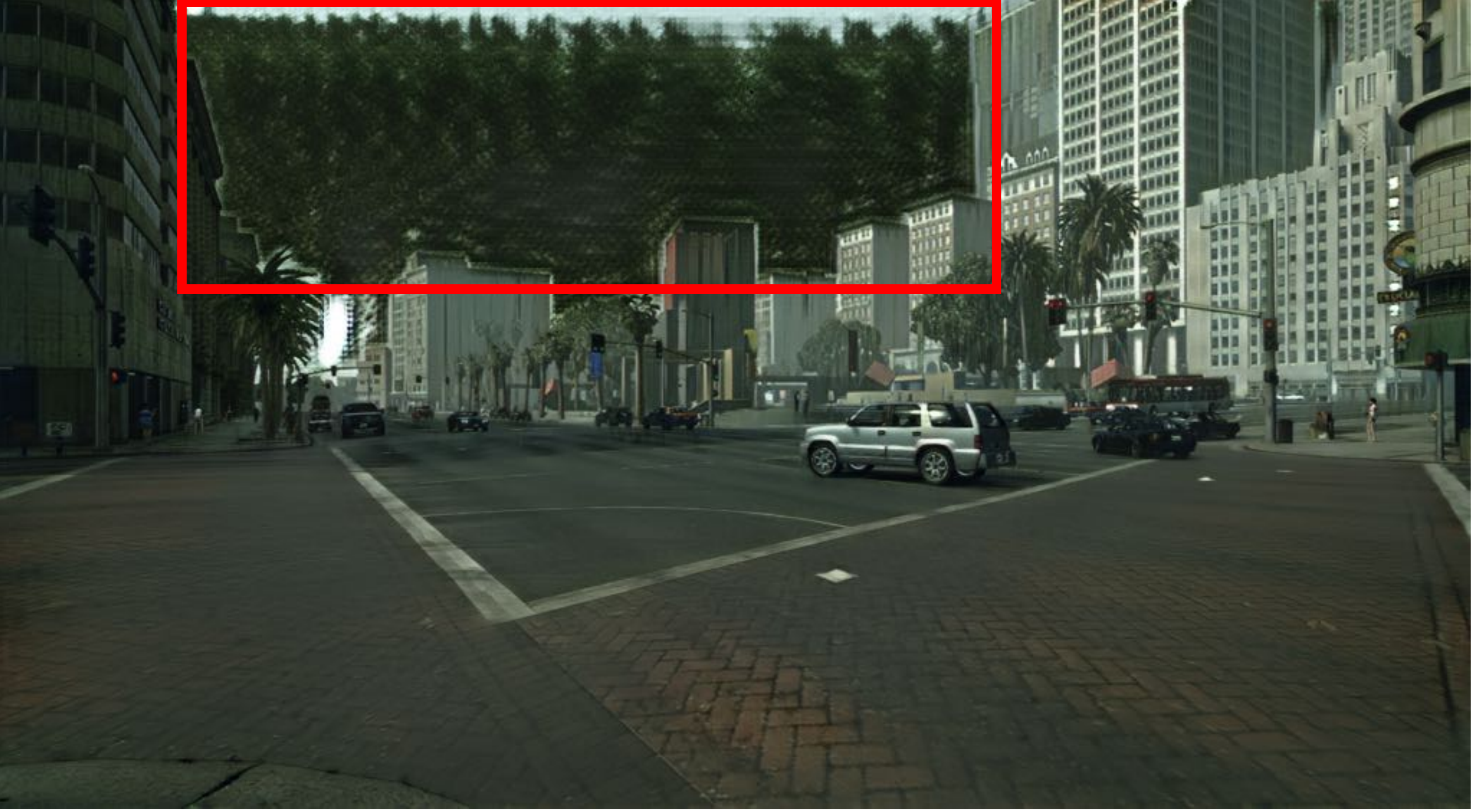}} &
			\subfloat{\includegraphics[width=0.24\linewidth]{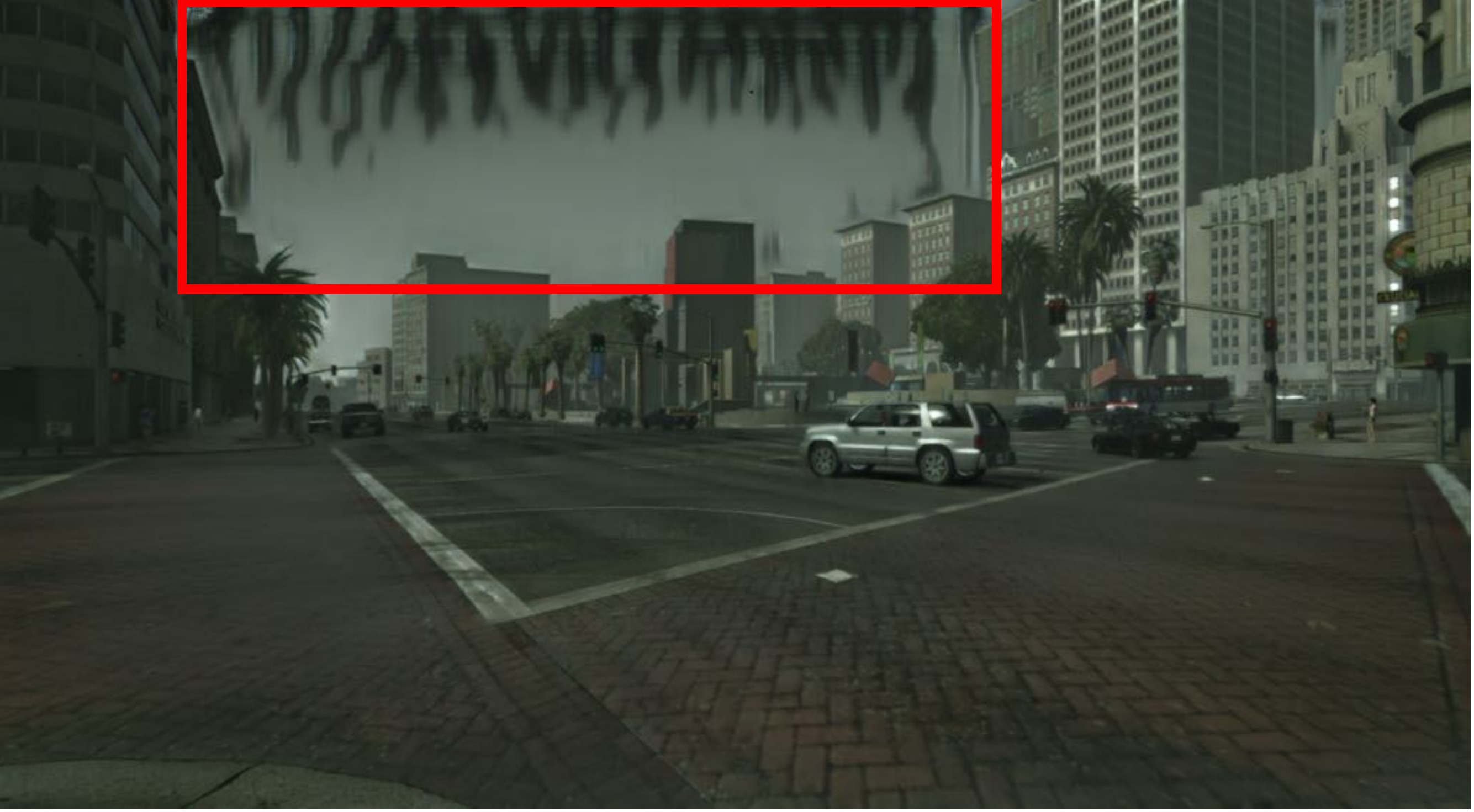}} &
			\subfloat{\includegraphics[width=0.24\linewidth]{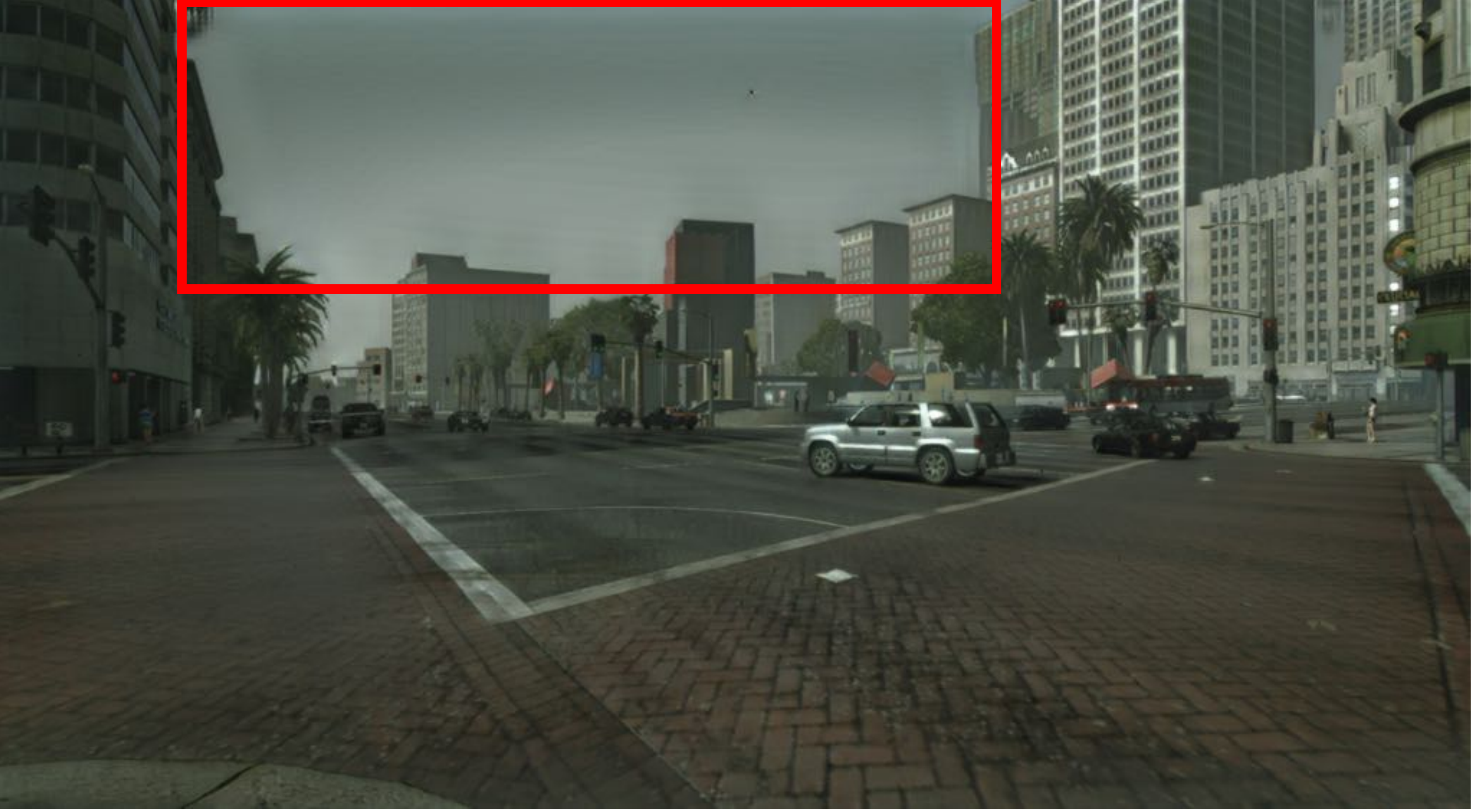}} 
			\\		
			
			\subfloat{\includegraphics[width=0.24\linewidth]{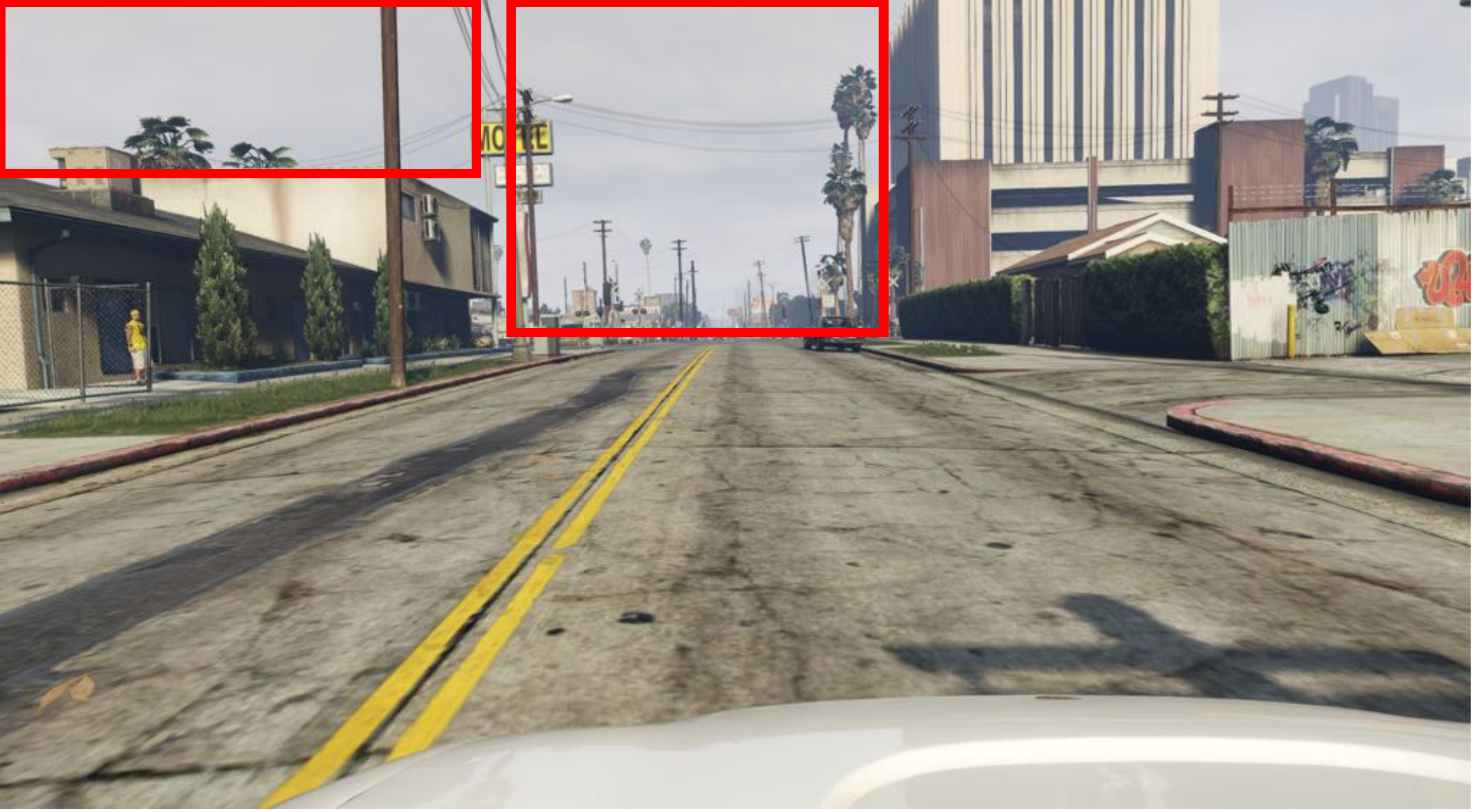}} &
			\subfloat{\includegraphics[width=0.24\linewidth]{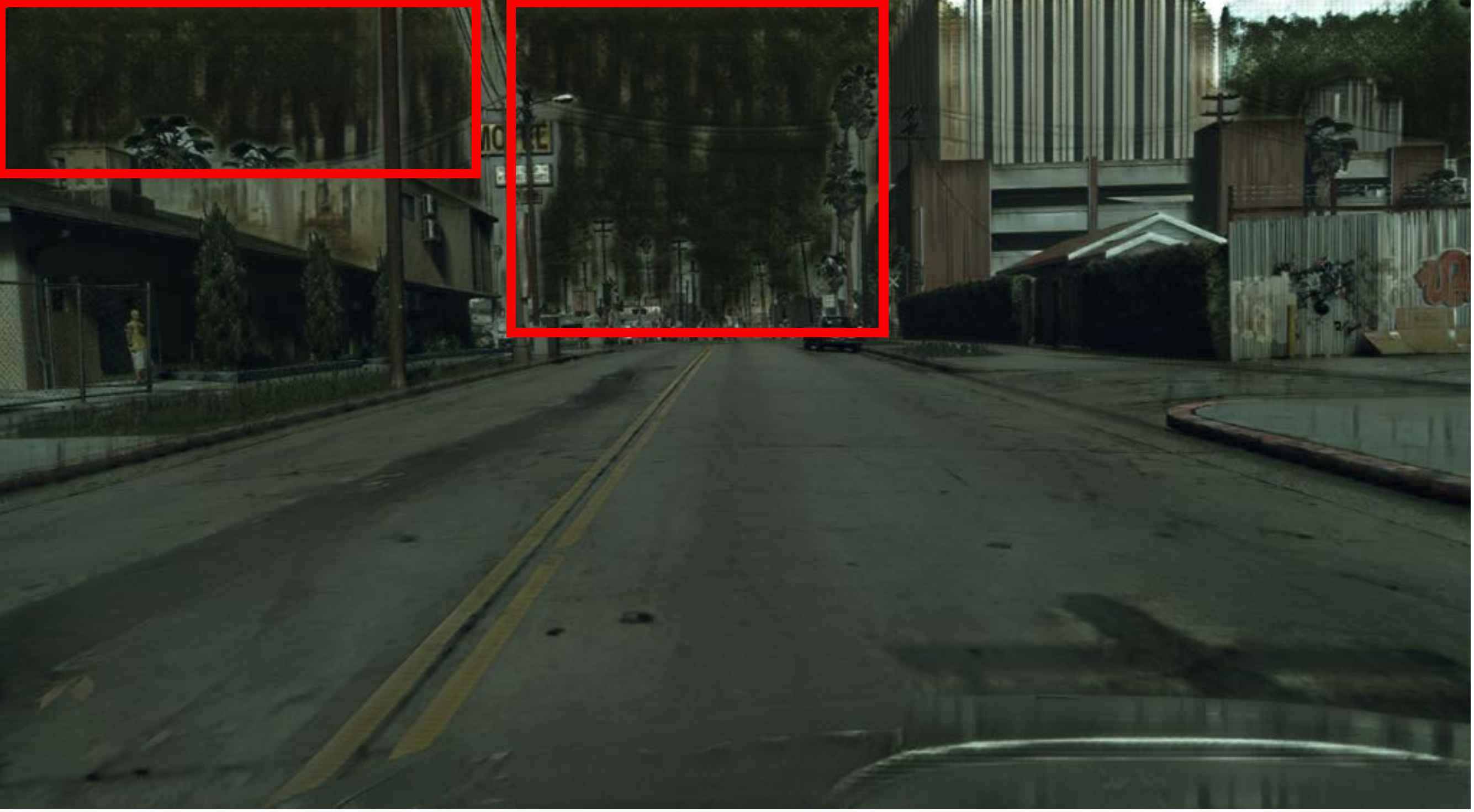}} &
			\subfloat{\includegraphics[width=0.24\linewidth]{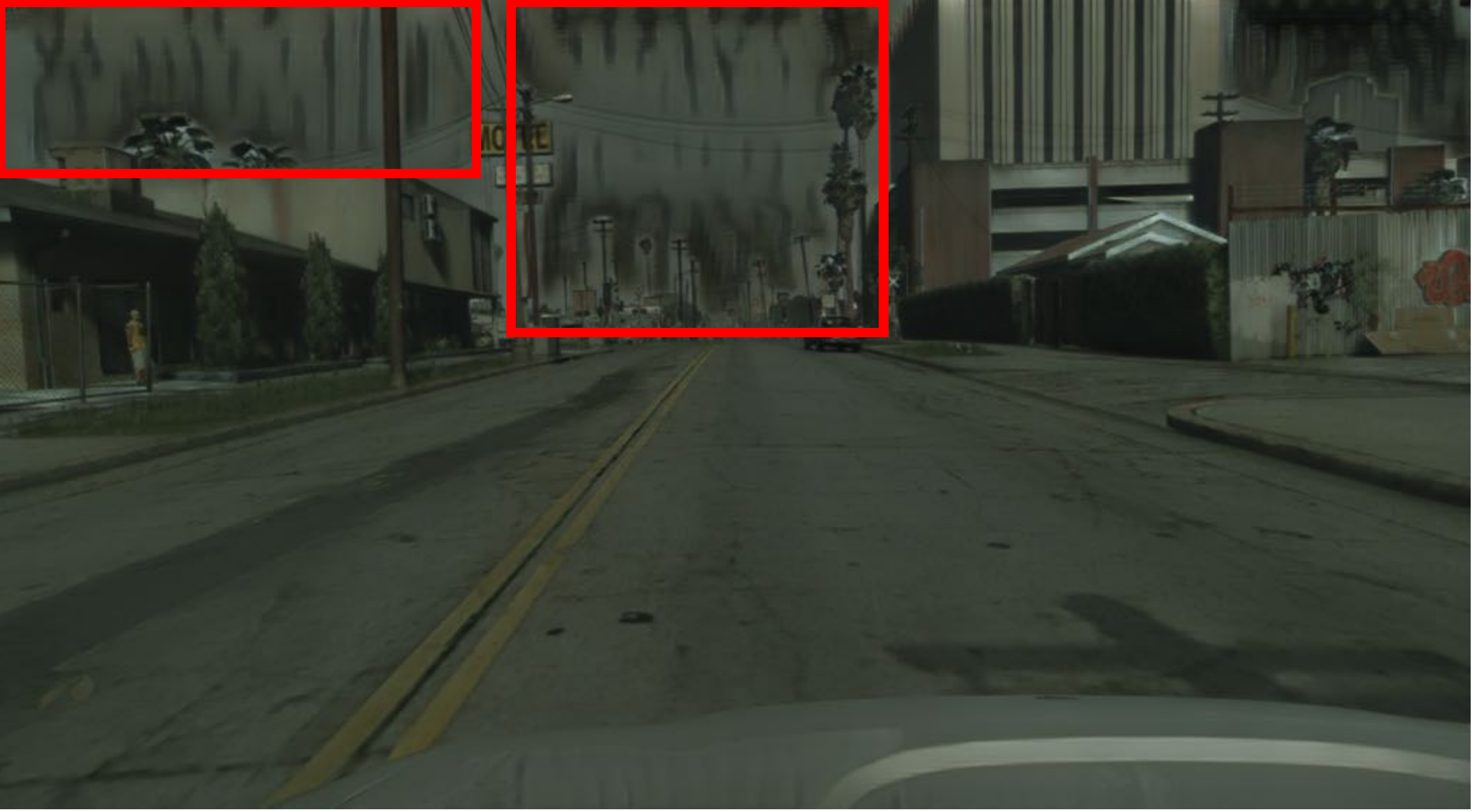}} &
			\subfloat{\includegraphics[width=0.24\linewidth]{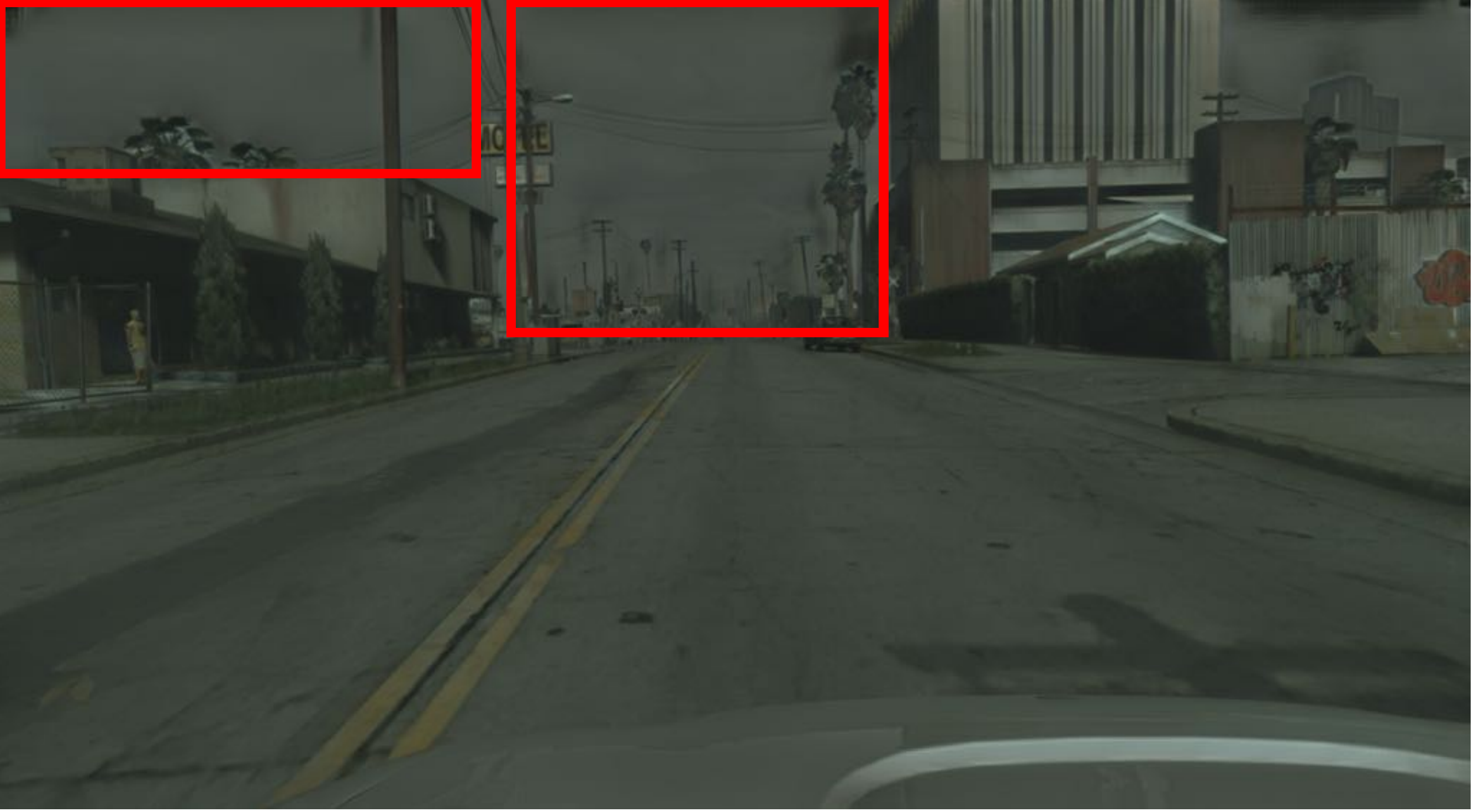}} 
			\\				
			
			\subfloat{\includegraphics[width=0.24\linewidth]{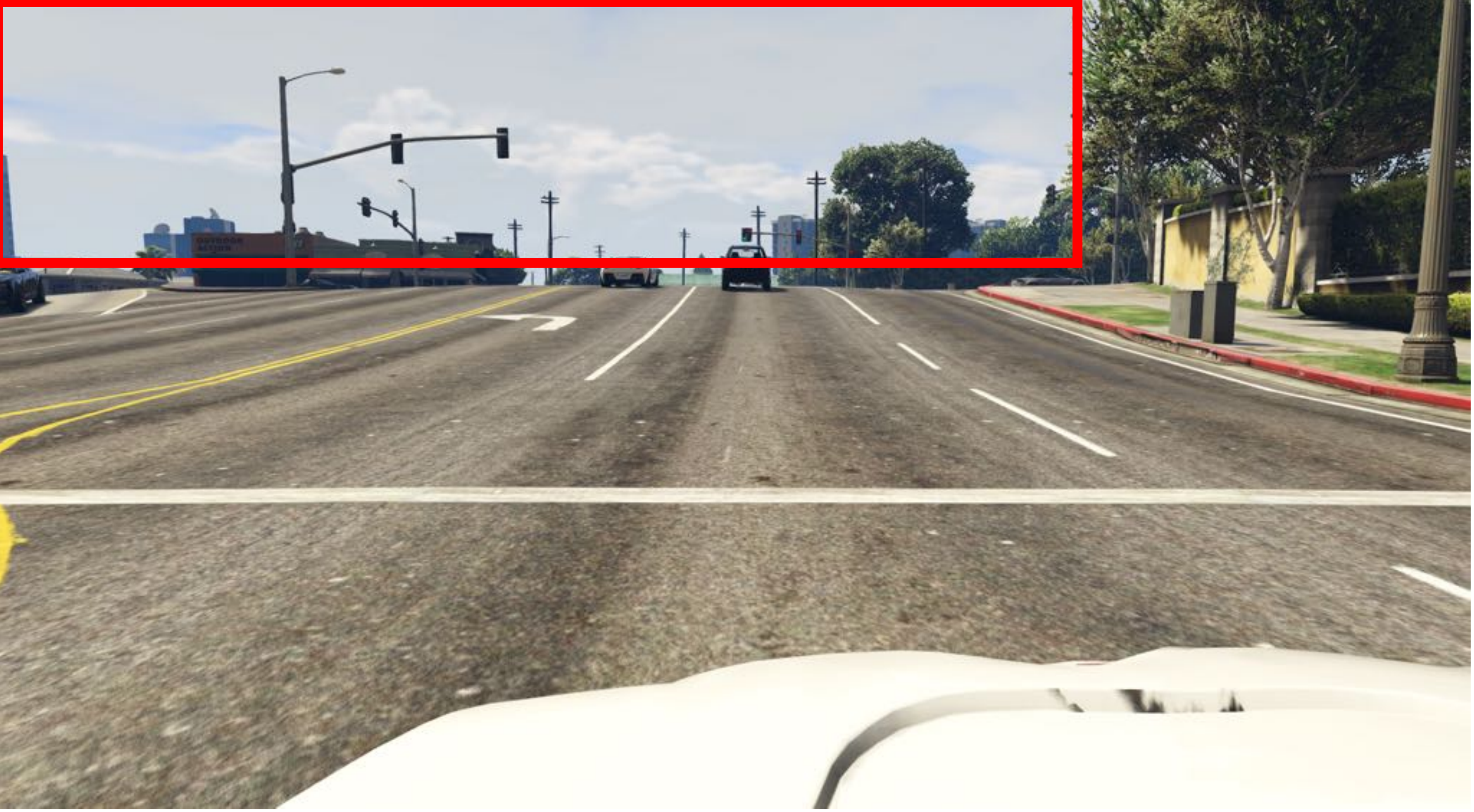}} &
			\subfloat{\includegraphics[width=0.24\linewidth]{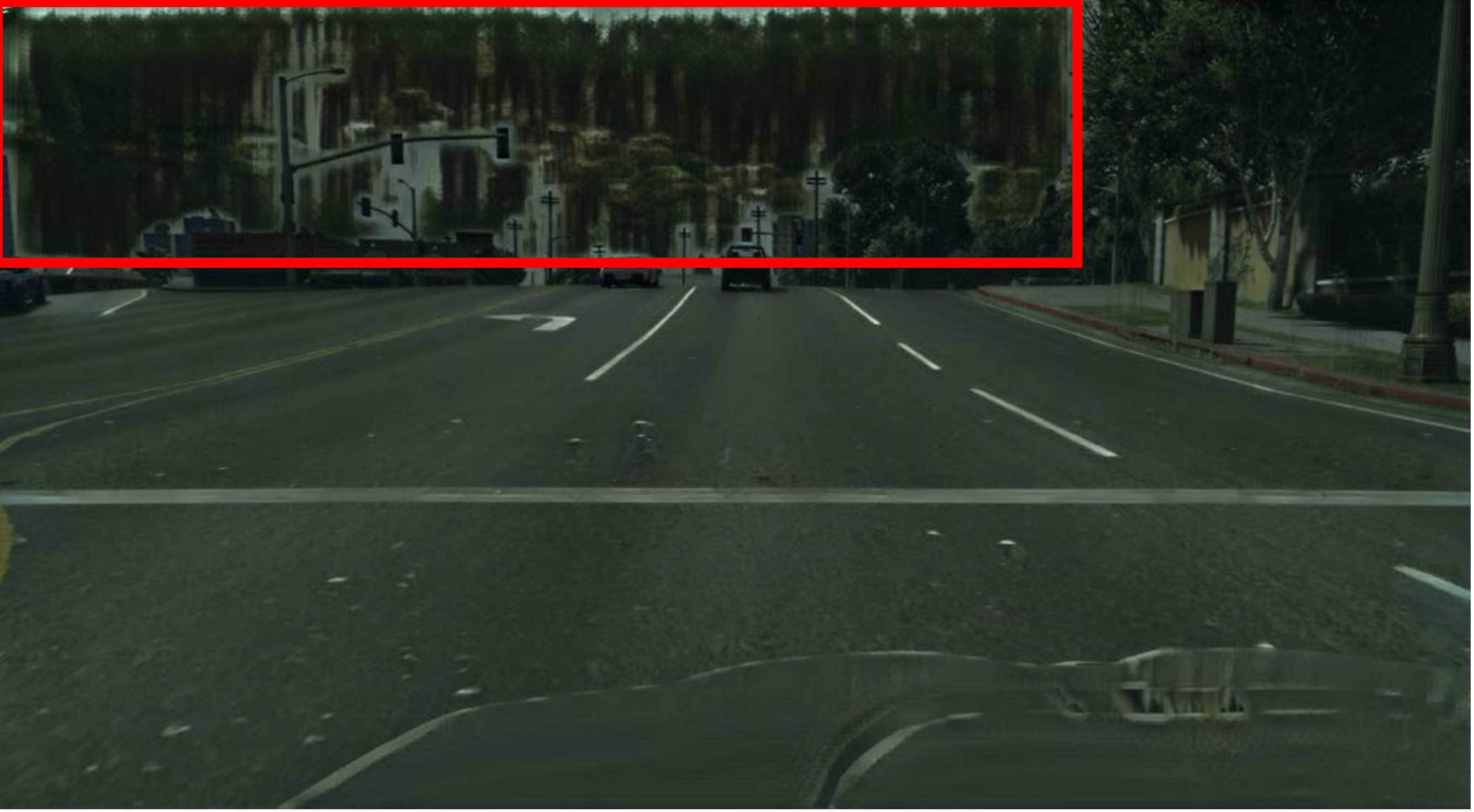}} &
			\subfloat{\includegraphics[width=0.24\linewidth]{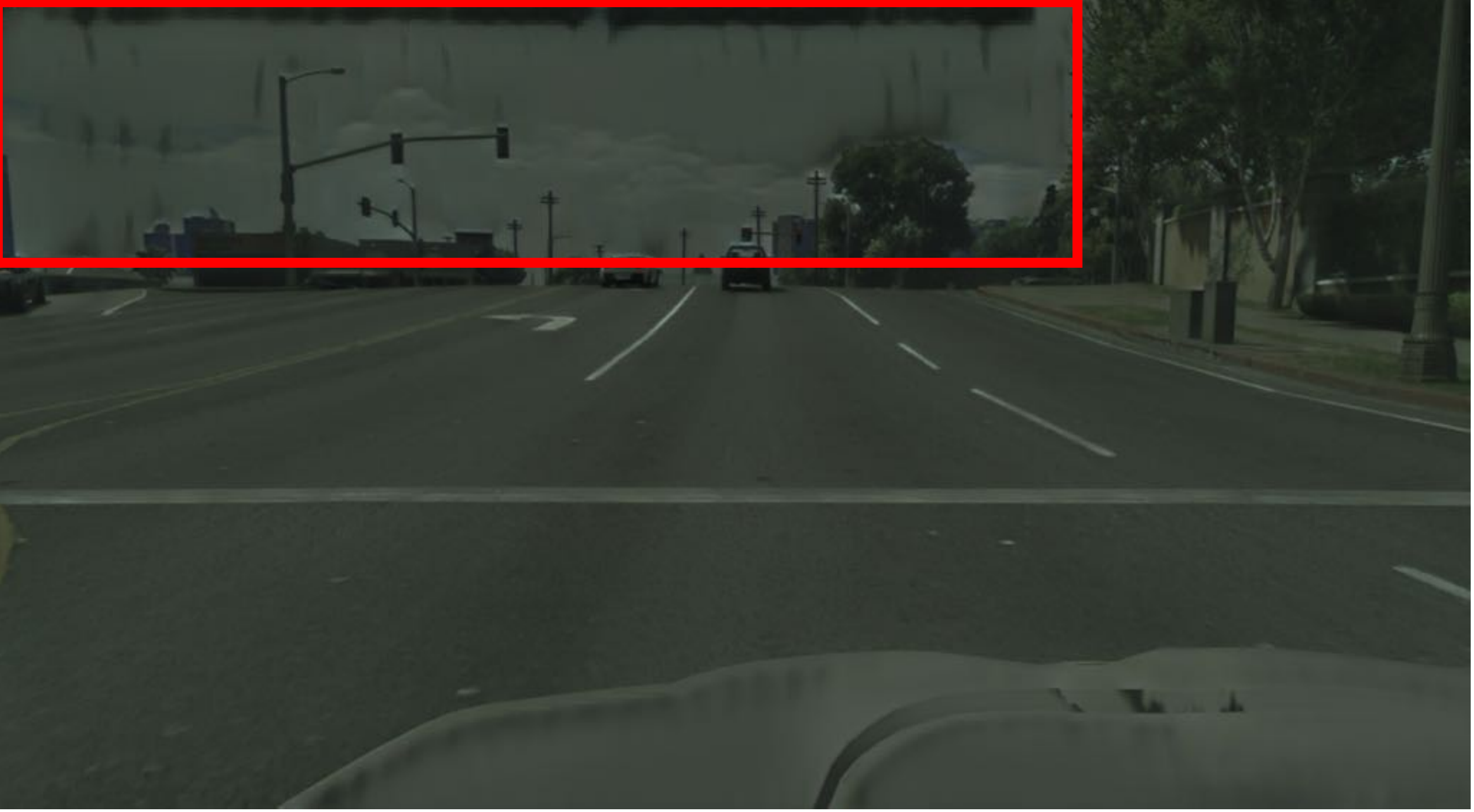}} &
			\subfloat{\includegraphics[width=0.24\linewidth]{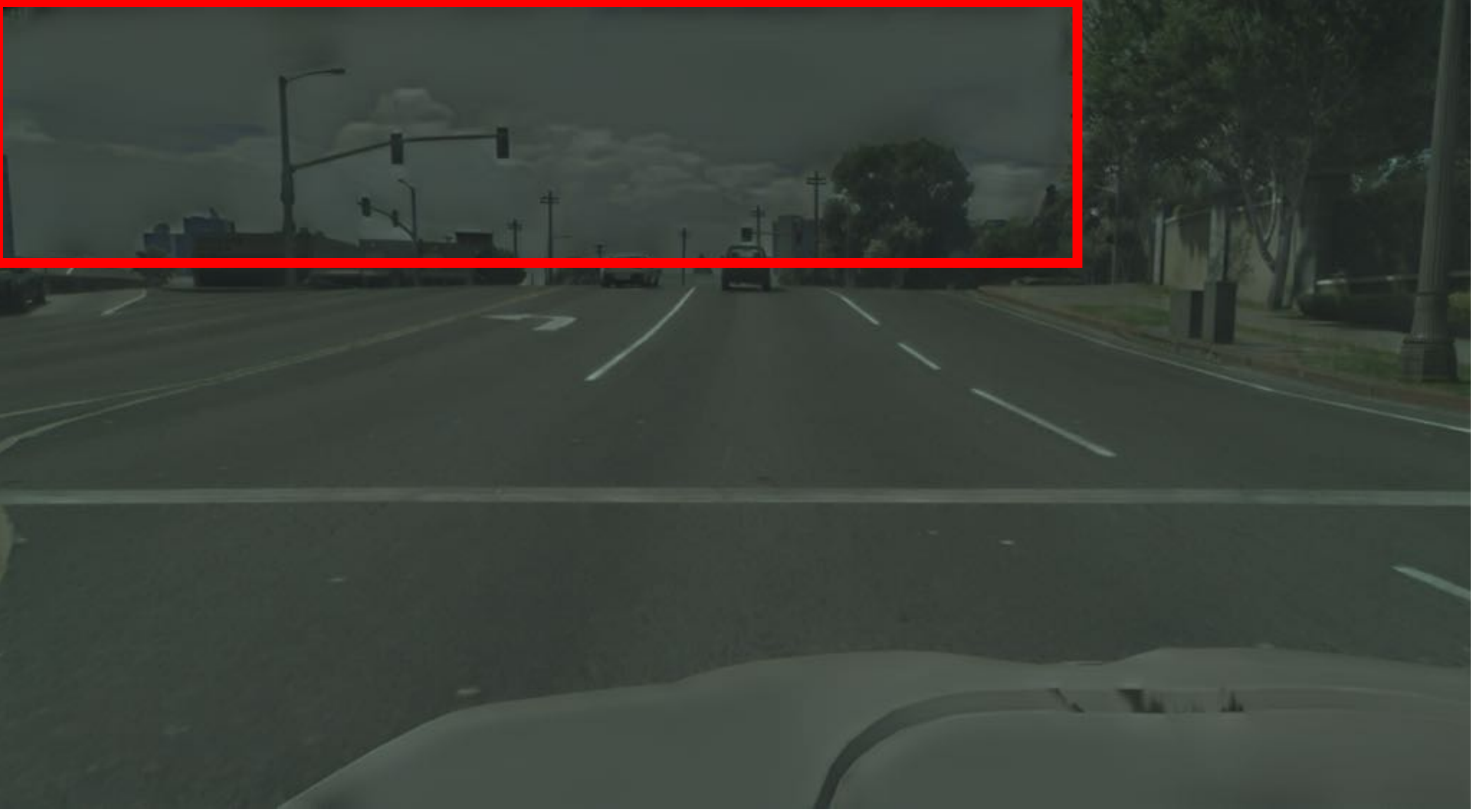}} 
			\\	
			
			\subfloat{\includegraphics[width=0.24\linewidth]{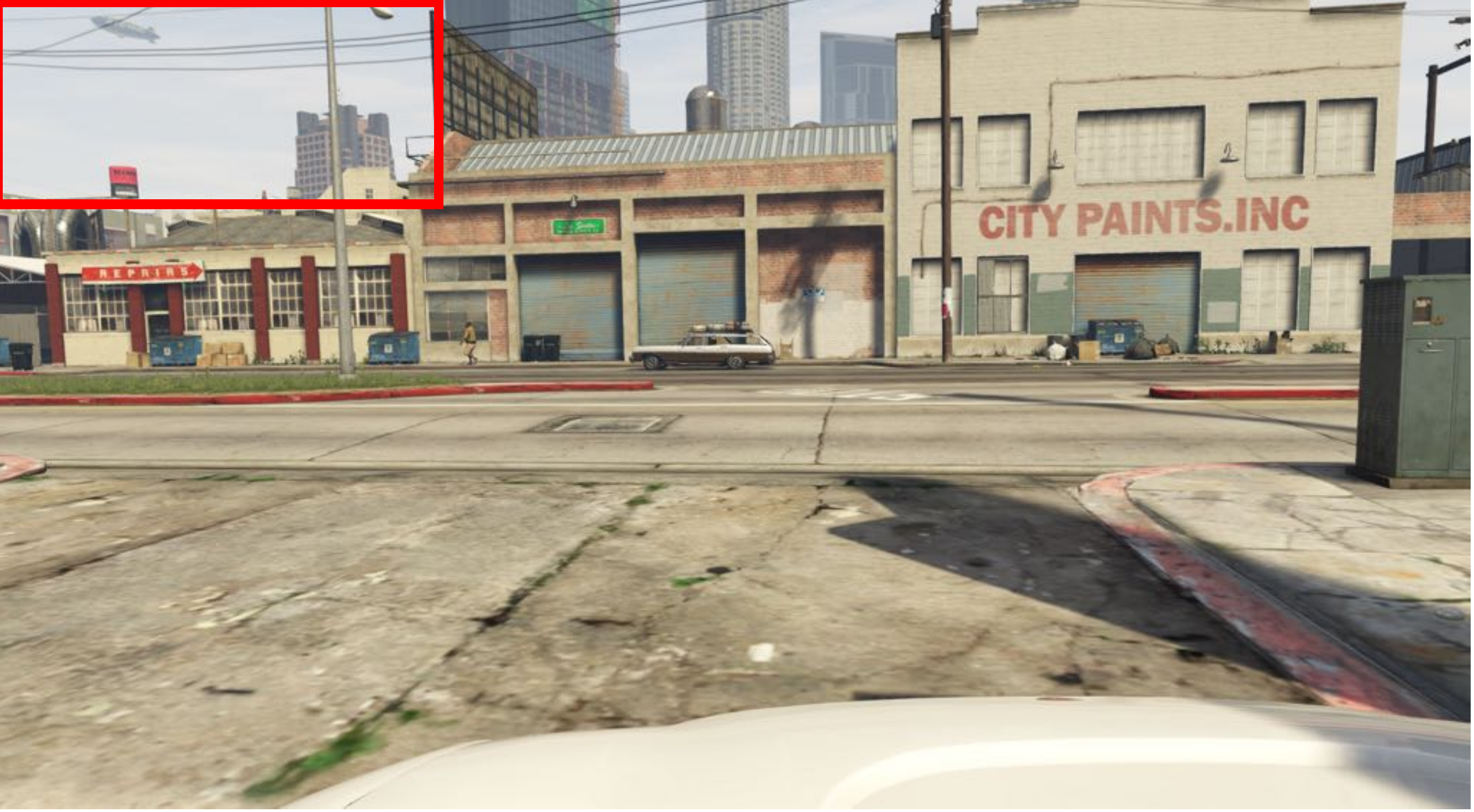}} &
			\subfloat{\includegraphics[width=0.24\linewidth]{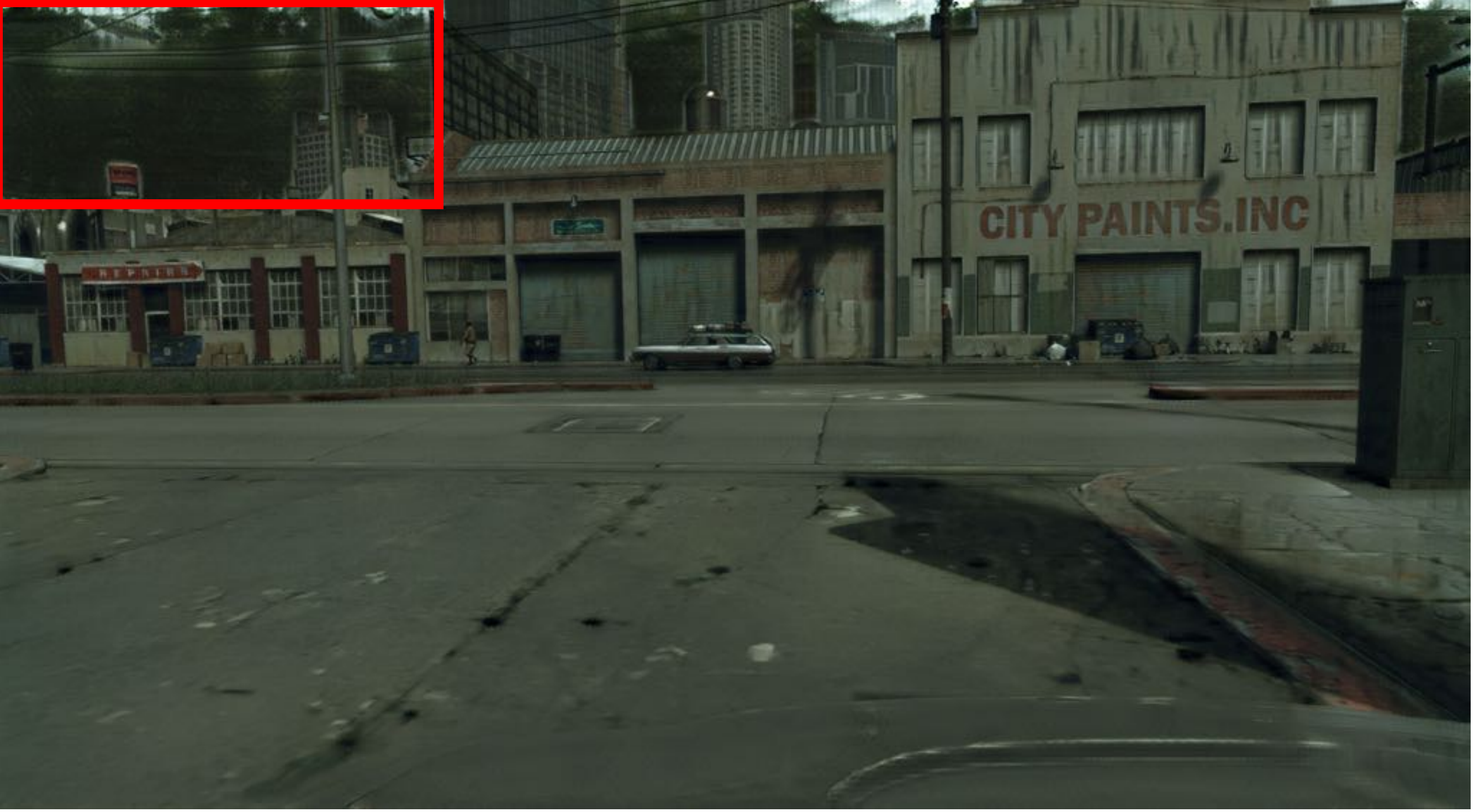}} &
			\subfloat{\includegraphics[width=0.24\linewidth]{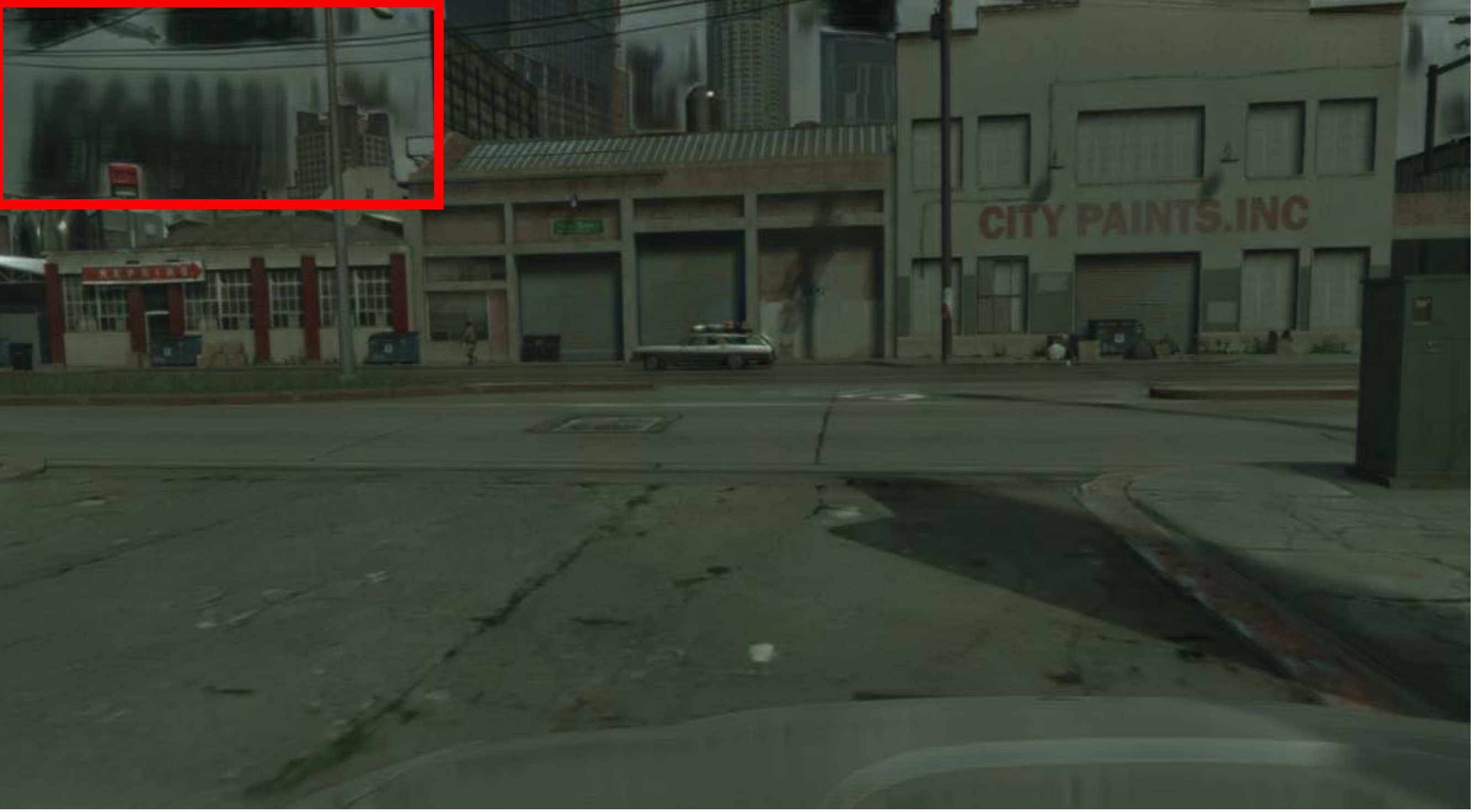}} &
			\subfloat{\includegraphics[width=0.24\linewidth]{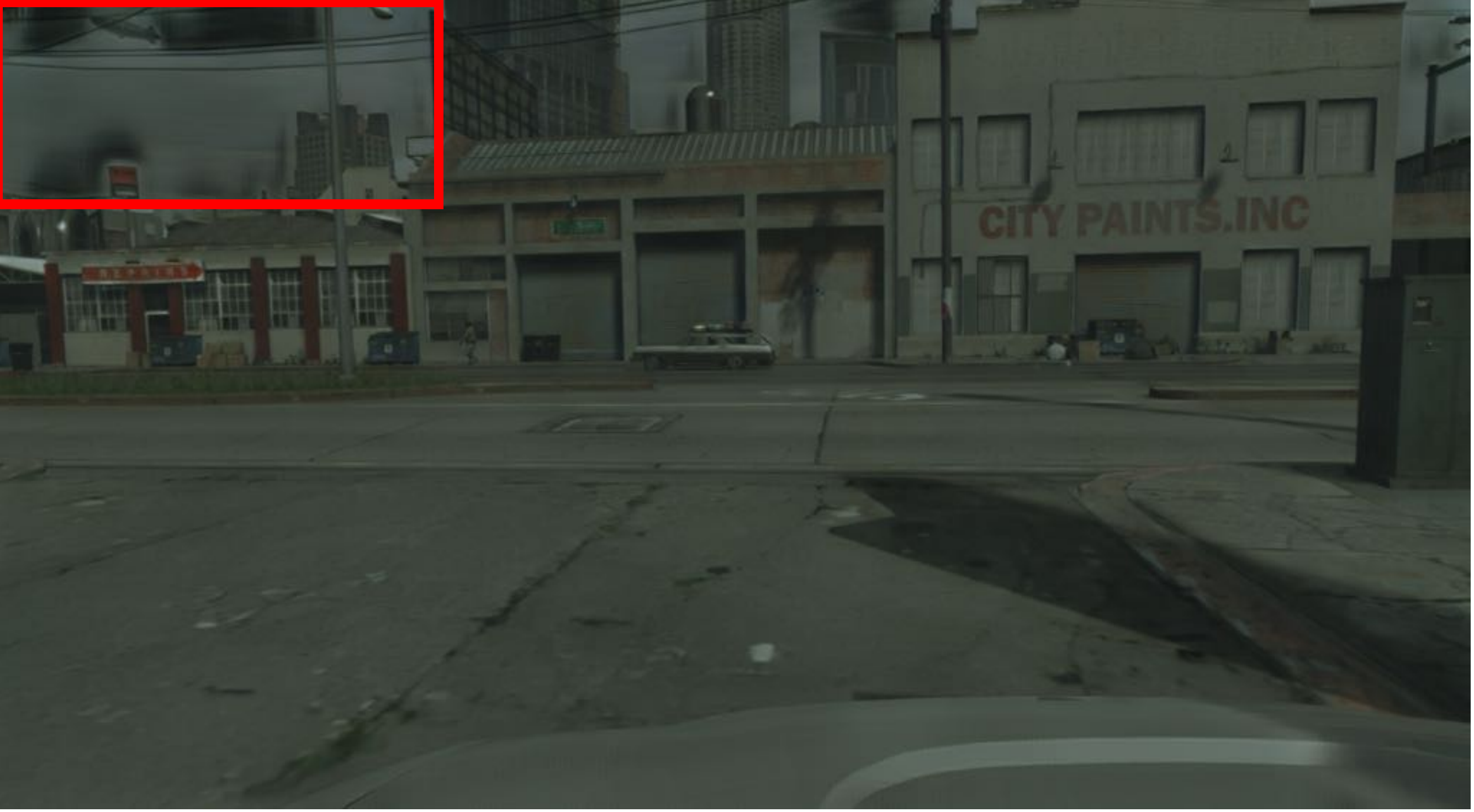}} 		
			\\
			(a) Raw images & (b) CycleGAN & (c) SPIT & (d) DPIT
		\end{tabular}
		}
	\caption{Qualitative results of different image translation methods on GTA5$\rightarrow$Cityscapes scenario. (a) raw images from GTA5 dataset; (b) translated images by CycleGAN; (c) translated images by single path image translation (SPIT); (d) translated images by our dual path image translation (DPIT). Red rectangles highlight the differences.}
 \label{fig:image_tranlaton_contrast}
\end{figure*}

{\noindent \textbf{Visualization of Image Translation Results.}}\hspace{3pt}
In Figure~\ref{fig:image_tranlaton_contrast}, we show the qualitative results of different image translation methods, i.e., CycleGAN, single path image translation (SPIT) used in BDL~\cite{li2019bidirectional} and dual path image translation (DPIT) used in our DPL, on GTA5$\rightarrow$Cityscapes scenario. As highlighted by red rectangles in Figure~\ref{fig:image_tranlaton_contrast}, CycleGAN and SPIT tend to generate vegetation-like artifacts in the sky to fool the discriminator, which can cause image distortion and visual inconsistency in the following adaptive learning. Benefit from the cross-domain perceptual loss which better maintains visual consistency, our DPIT achieves superior performance in terms of content preserving and style translation.

\clearpage

\begin{figure*}[htbp]
	\centering
		\begin{tabular}{ccc}

			\subfloat[Raw image]{\includegraphics[width=0.28\linewidth]{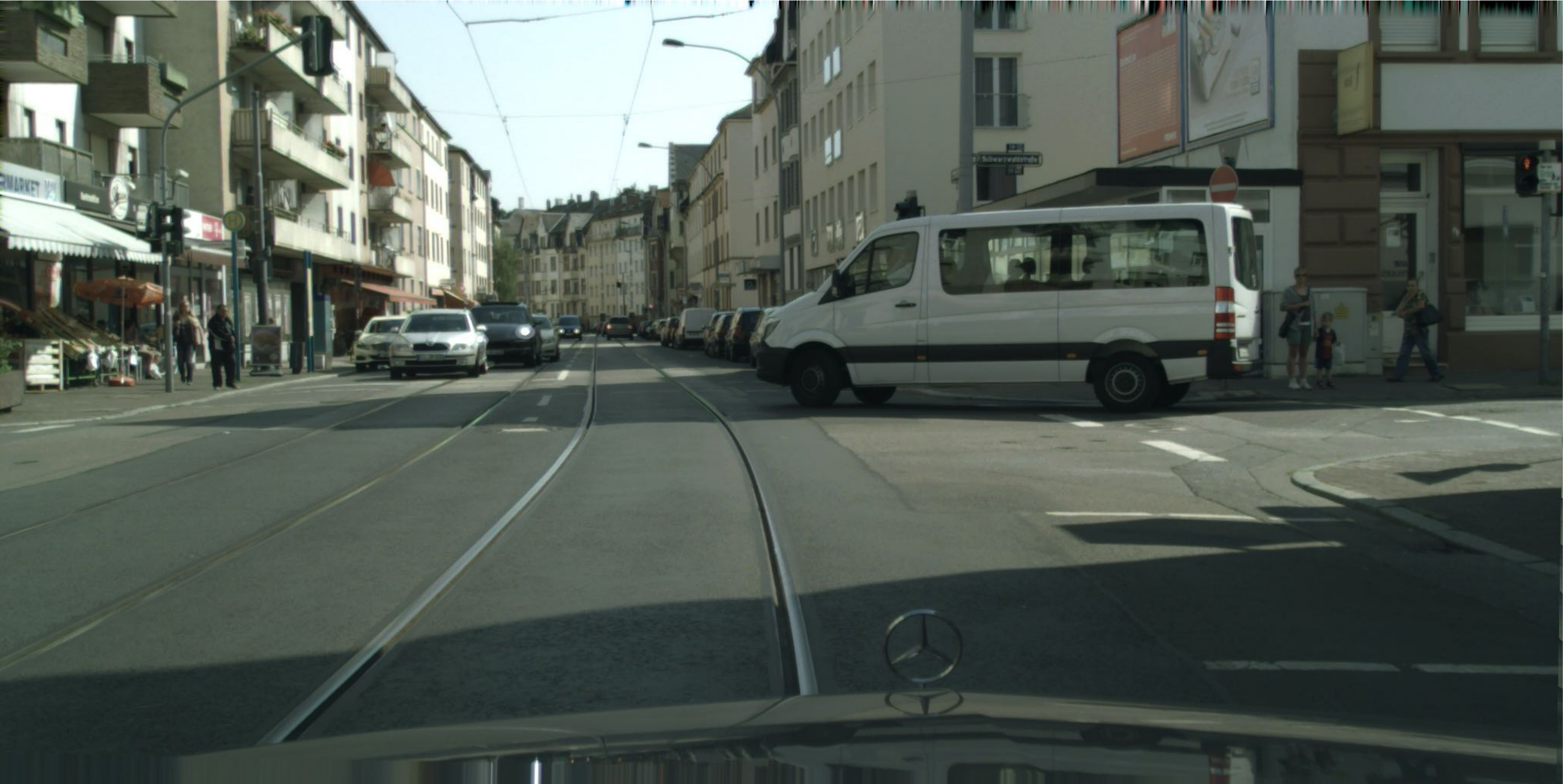}} & 
			\subfloat[Ground-truth]{\includegraphics[width=0.28\linewidth]{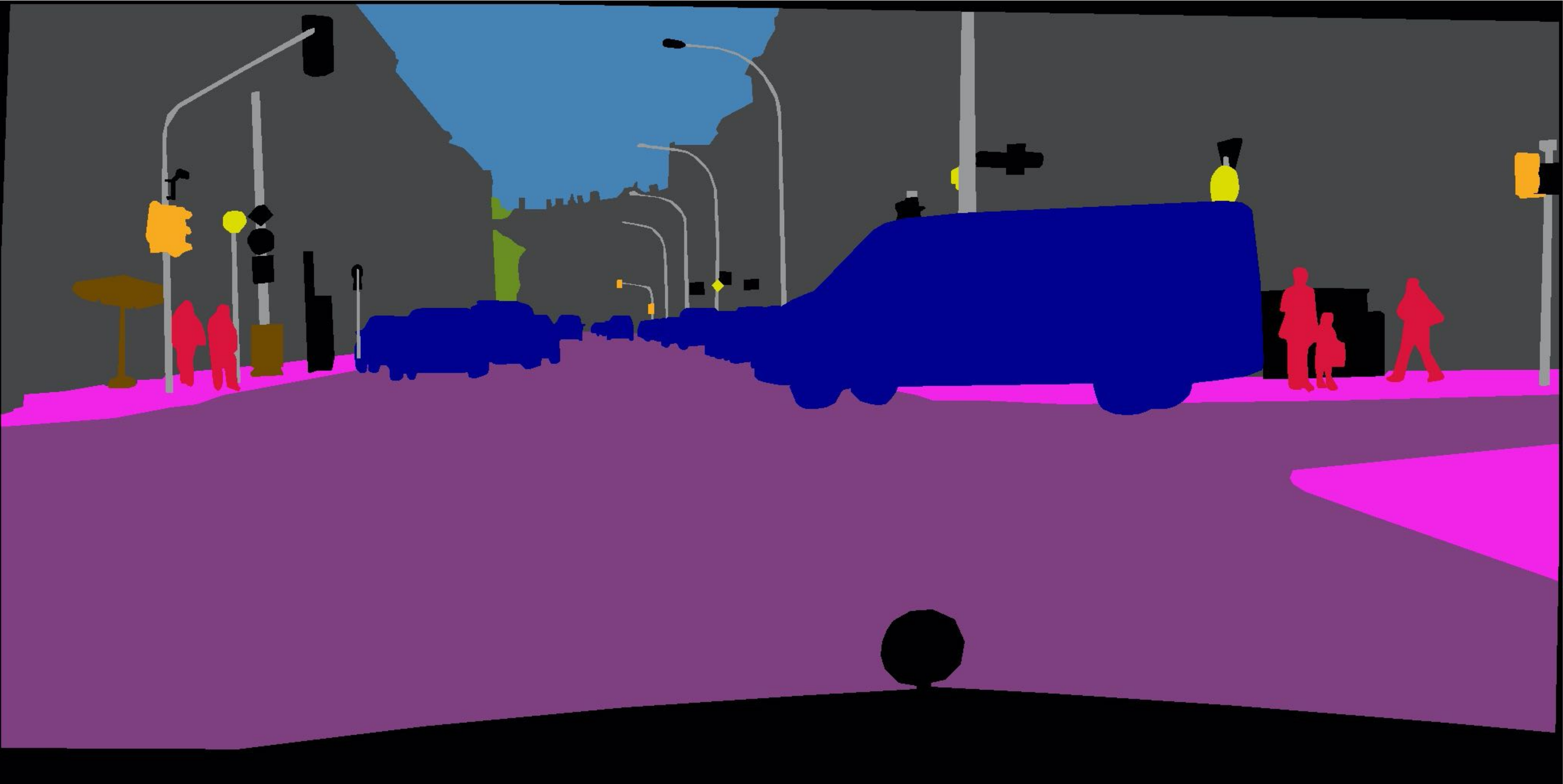}} & 
			\subfloat[Prediction without adaptation]{\includegraphics[width=0.28\linewidth]{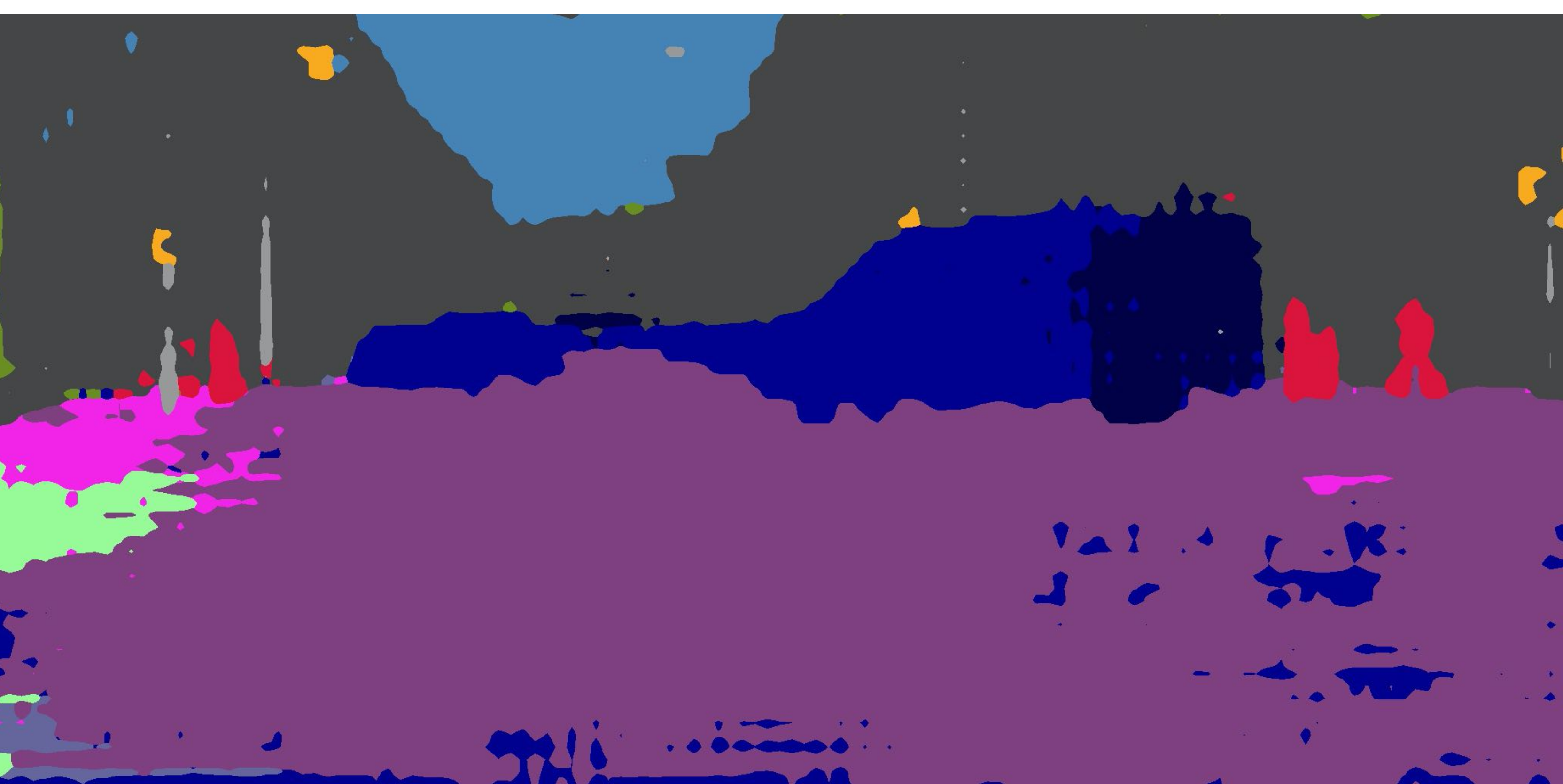}} 
			\\

			\subfloat[ADVENT~\cite{vu2018advent}]{\includegraphics[width=0.28\linewidth]{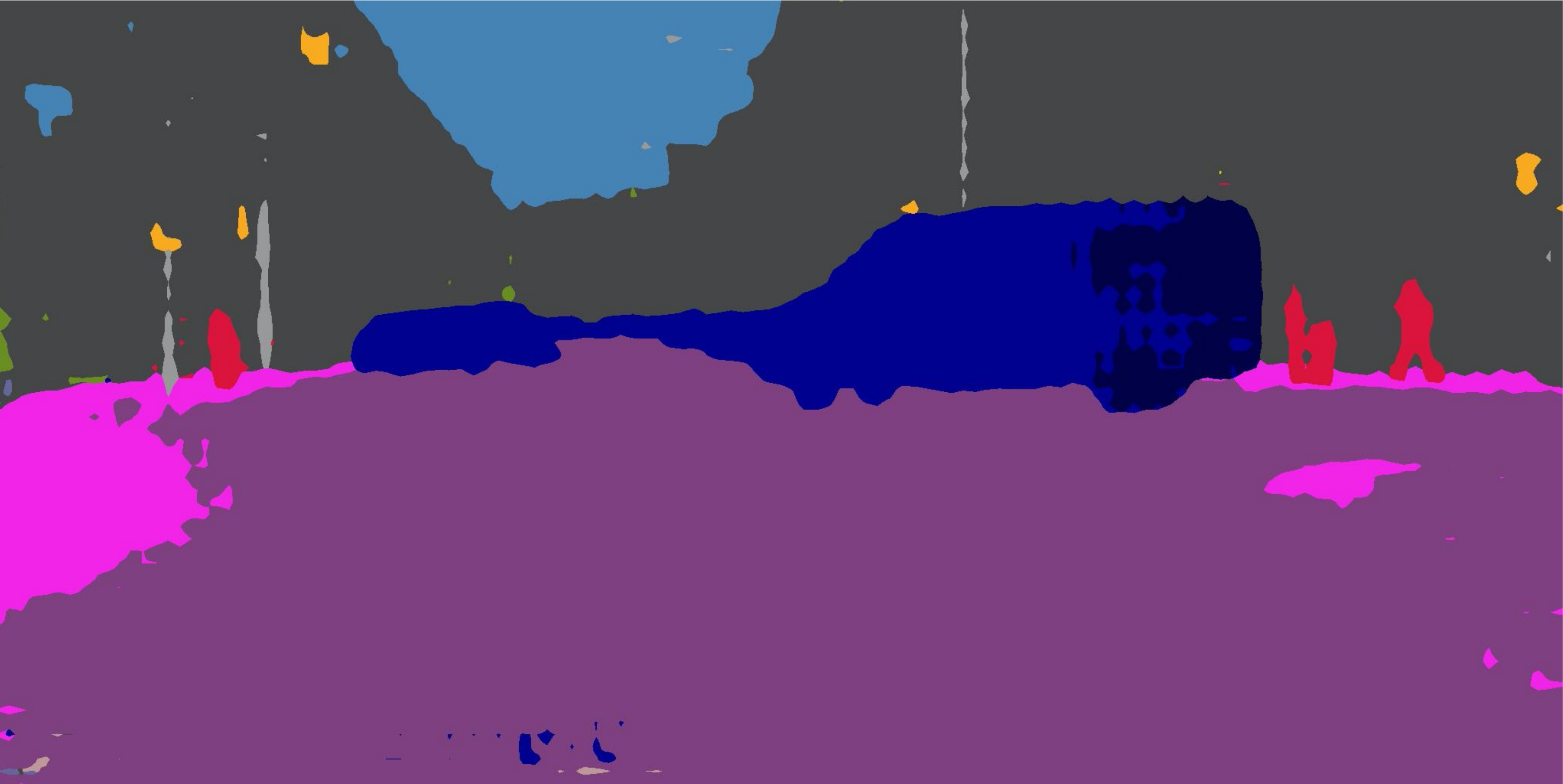}} & 
			\subfloat[BDL~\cite{li2019bidirectional}]{\includegraphics[width=0.28\linewidth]{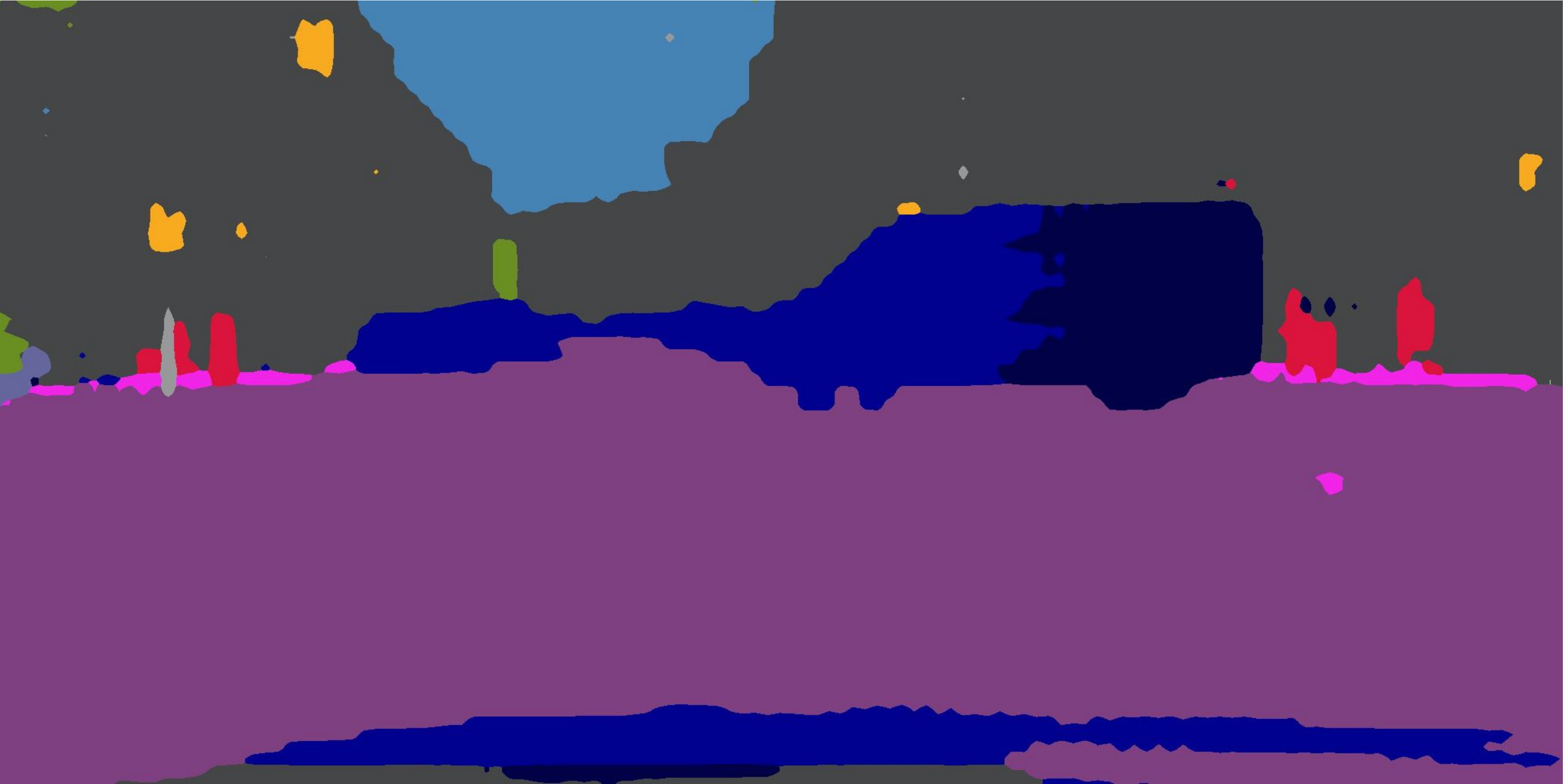}} & 
			\subfloat[CLAN~\cite{luo2019taking}]{\includegraphics[width=0.28\linewidth]{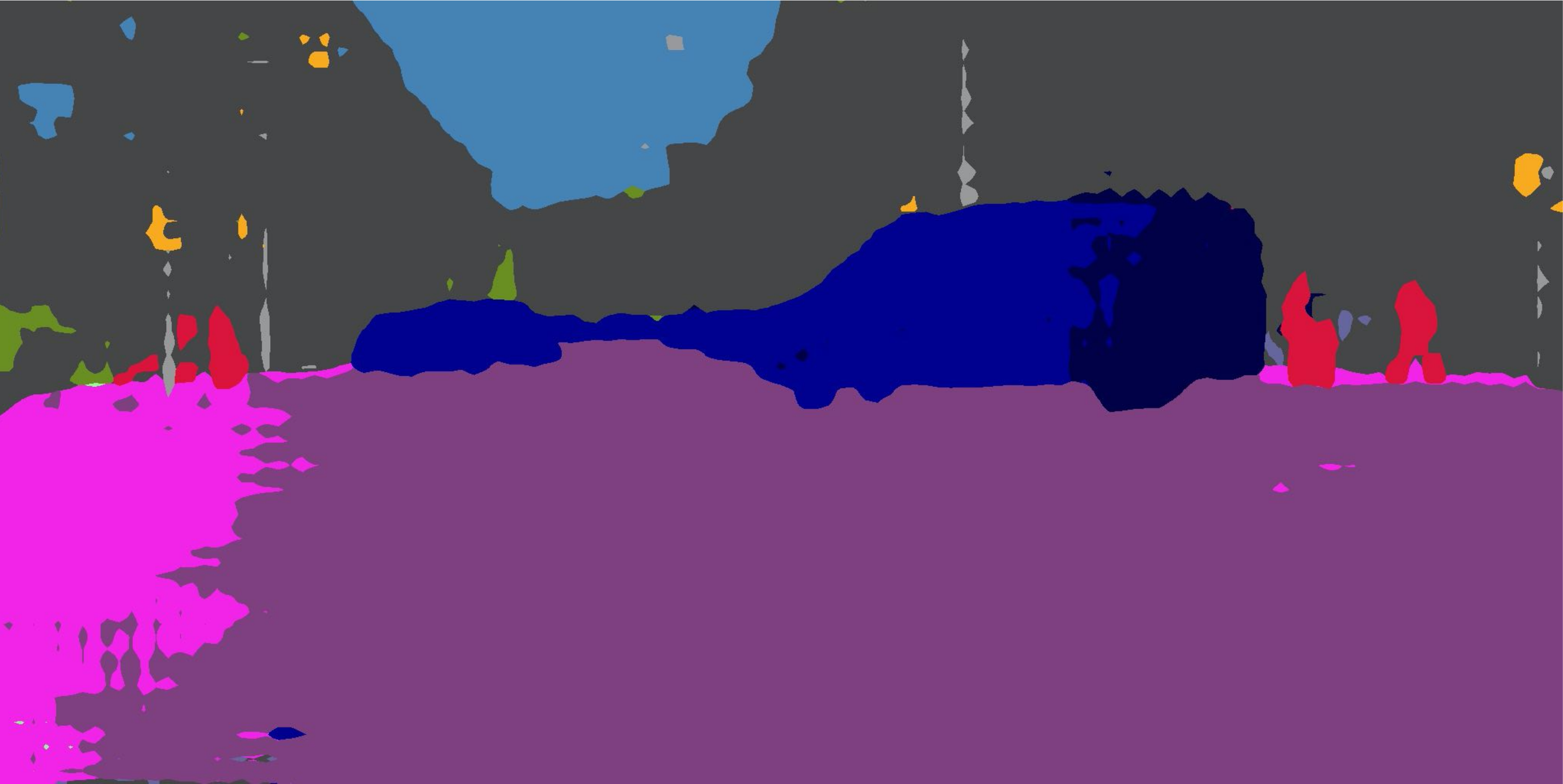}} 
			\\

			\subfloat[SIM~\cite{wang2020differential}]{\includegraphics[width=0.28\linewidth]{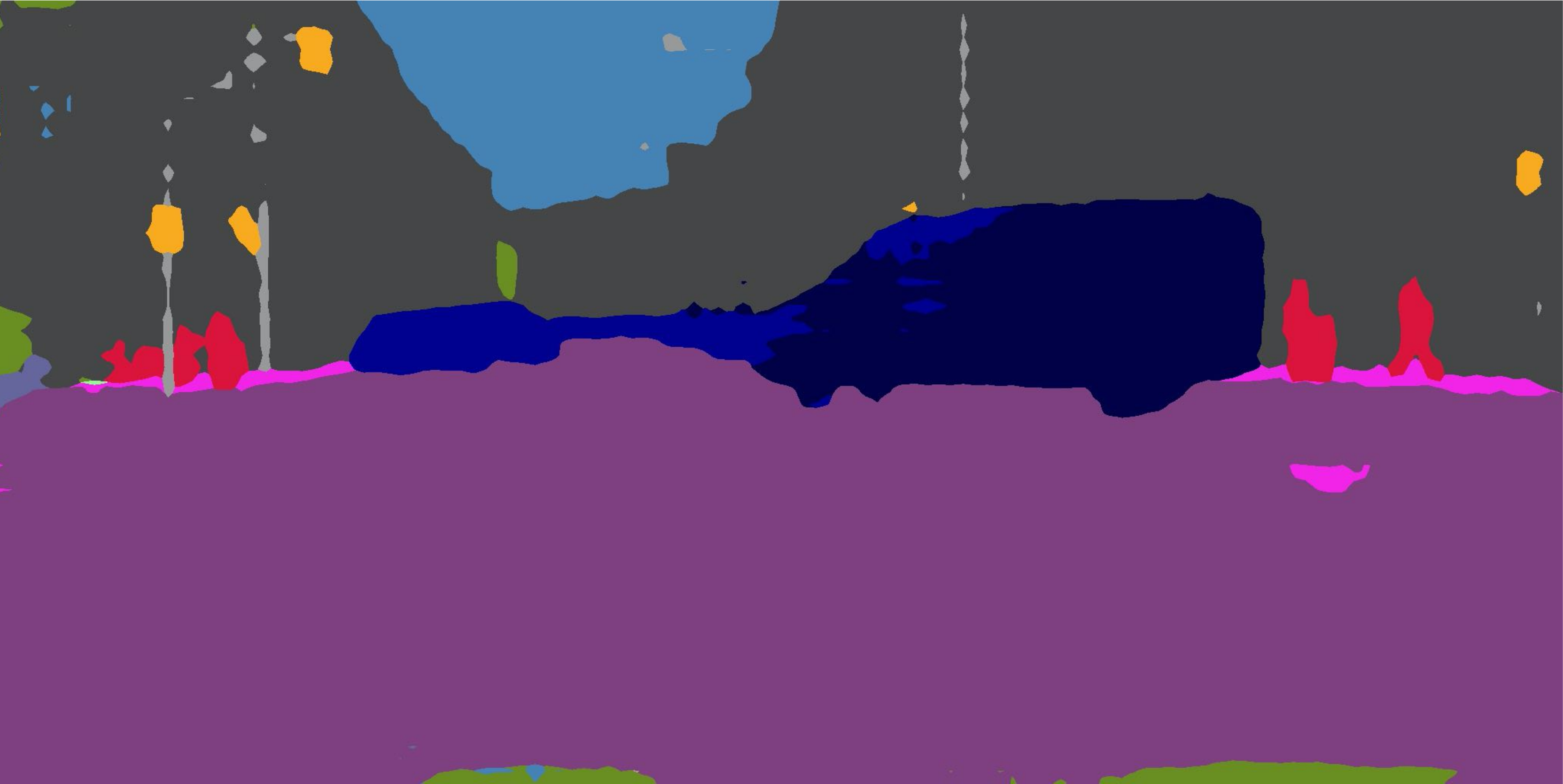}} & 
			\subfloat[FADA~\cite{wang2020classes}]{\includegraphics[width=0.28\linewidth]{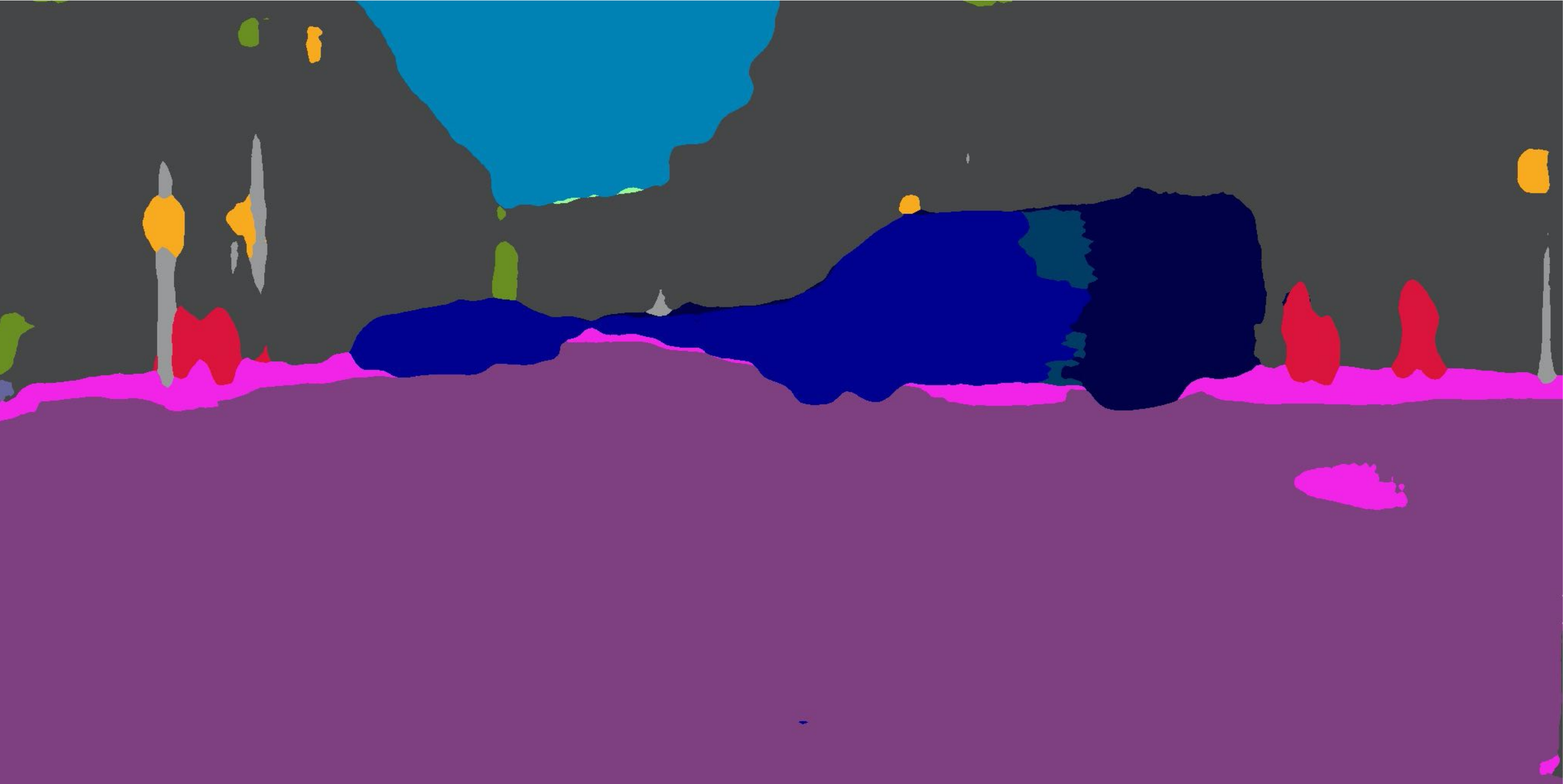}} & 
			\subfloat[FDA-MBT~\cite{yang2020fda}]{\includegraphics[width=0.28\linewidth]{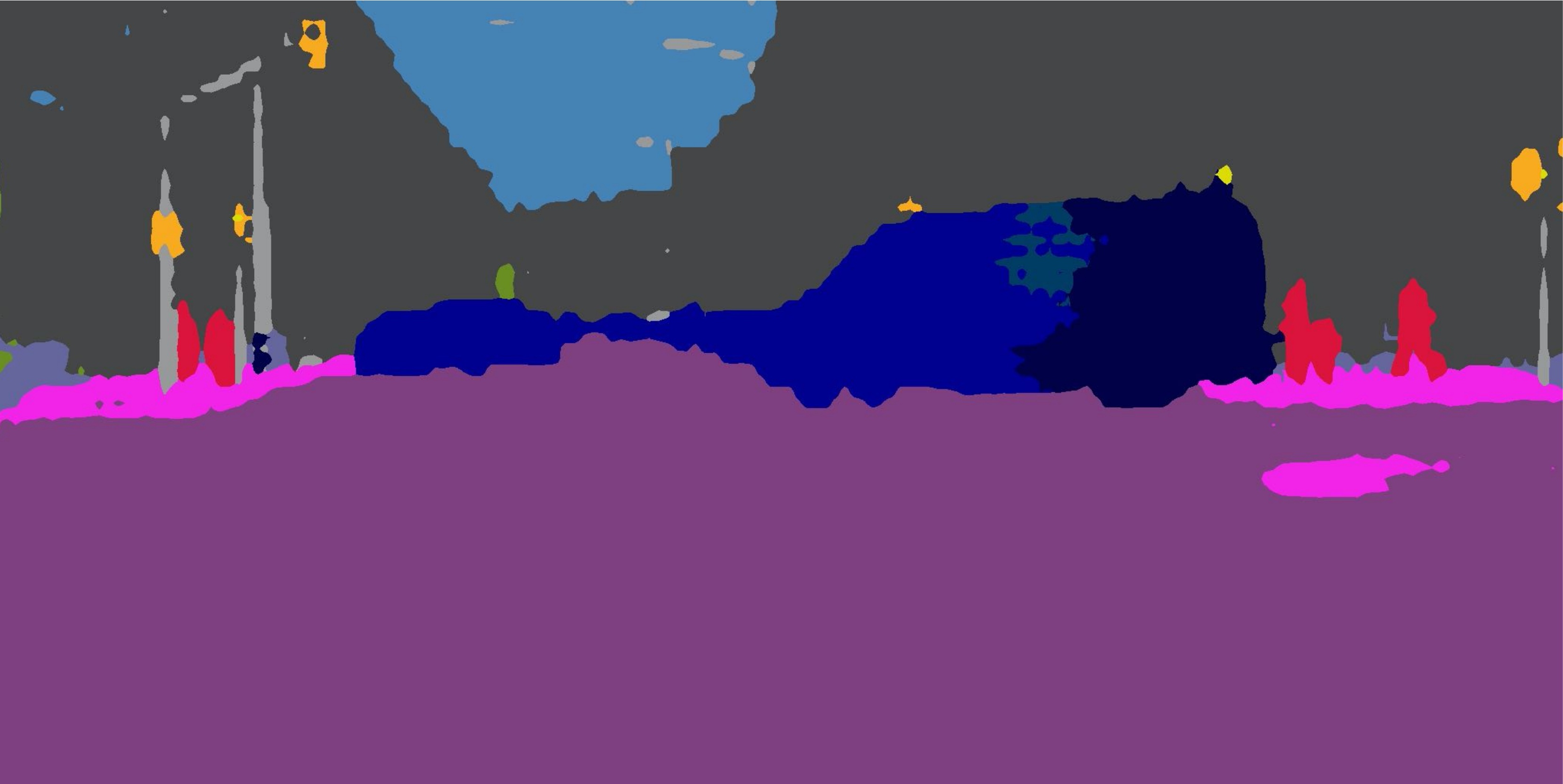}}  
			\\
 
			\subfloat[Kim et al.~\cite{kim2020learning}]{\includegraphics[width=0.28\linewidth]{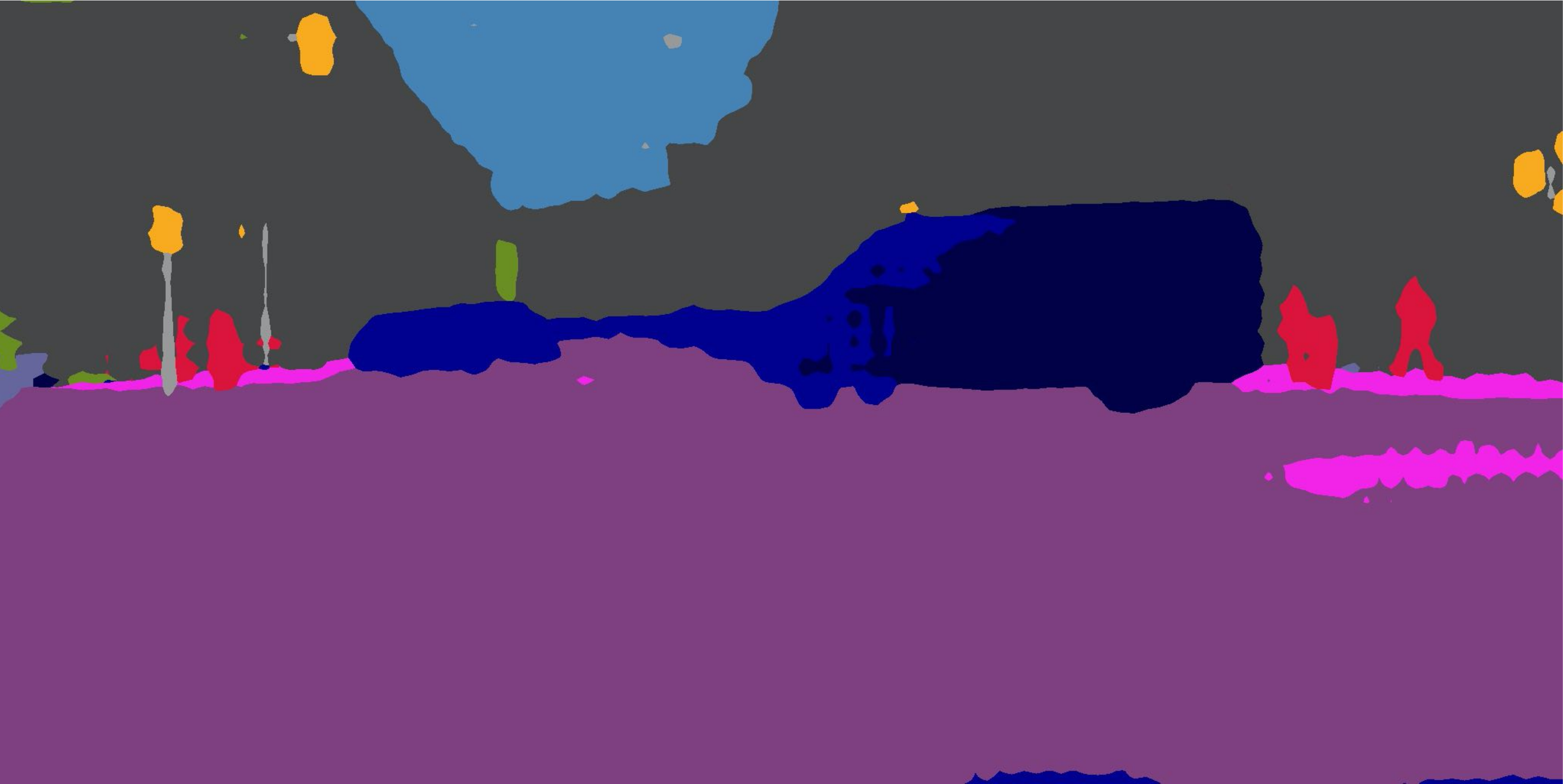}} & 
			\subfloat[DPL]{\includegraphics[width=0.28\linewidth]{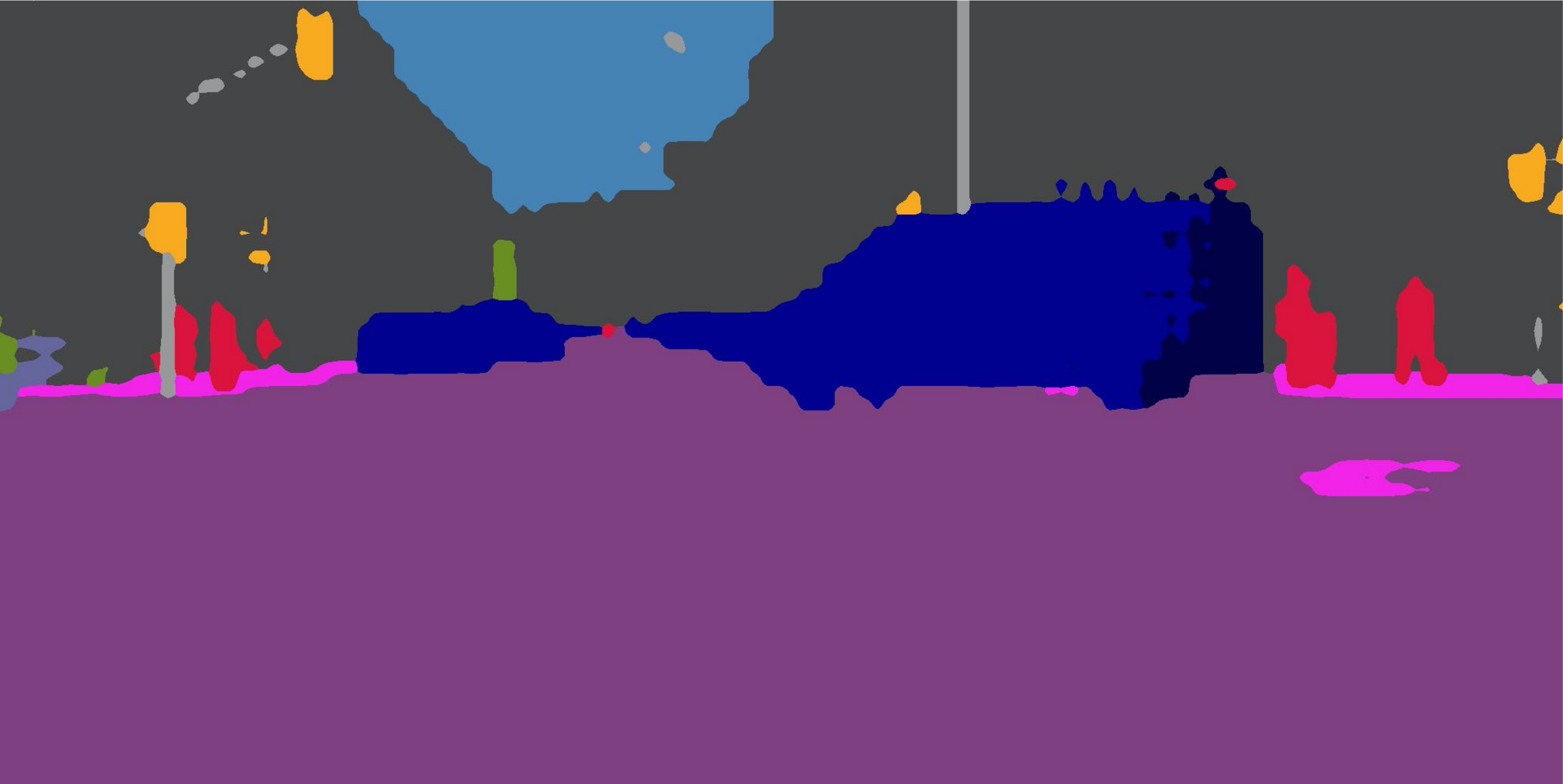}} & 
			\subfloat[DPL-Dual]{\includegraphics[width=0.28\linewidth]{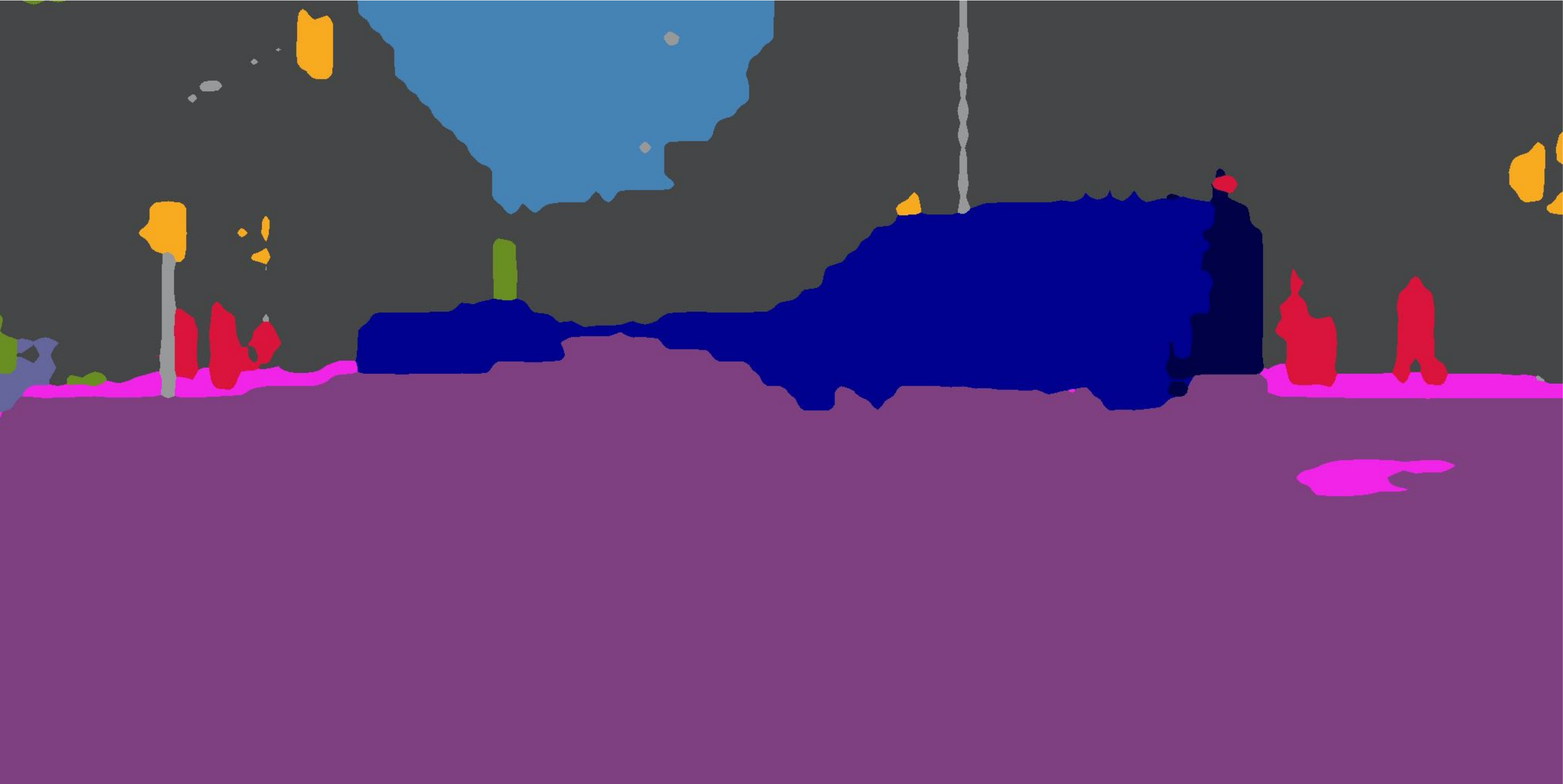}} \\
            \vspace{0.5cm}
	\end{tabular}
	\caption{Qualitative comparison of different methods on GTA5$\rightarrow$Cityscapes scenario.}
	\label{fig:comp1}
 \vspace{3cm}
\end{figure*}

\begin{figure*}[htbp]
	\centering
		\begin{tabular}{ccc}
			\subfloat[Raw image]{\includegraphics[width=0.28\linewidth]{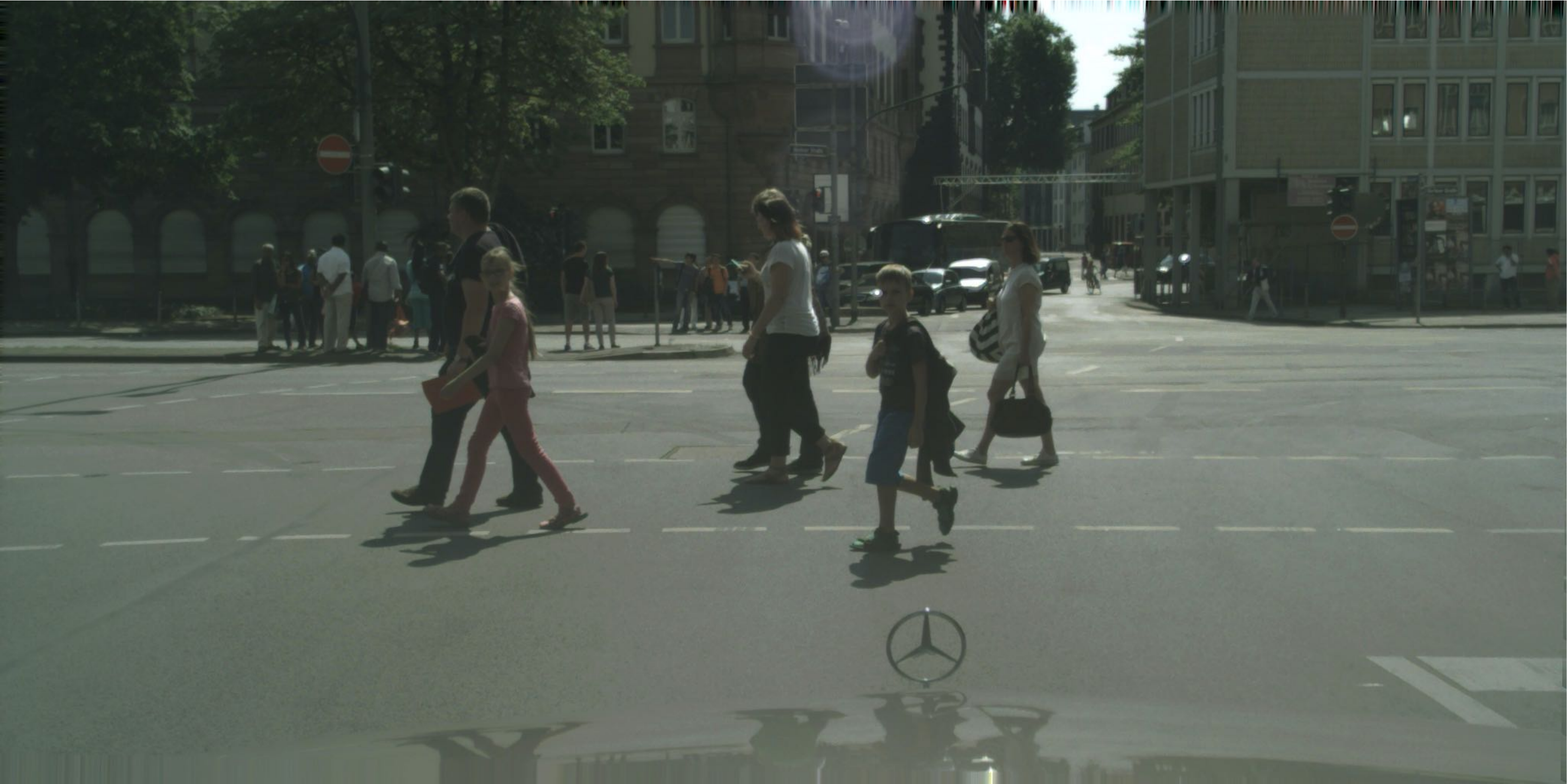}} & 
			\subfloat[Ground-truth]{\includegraphics[width=0.28\linewidth]{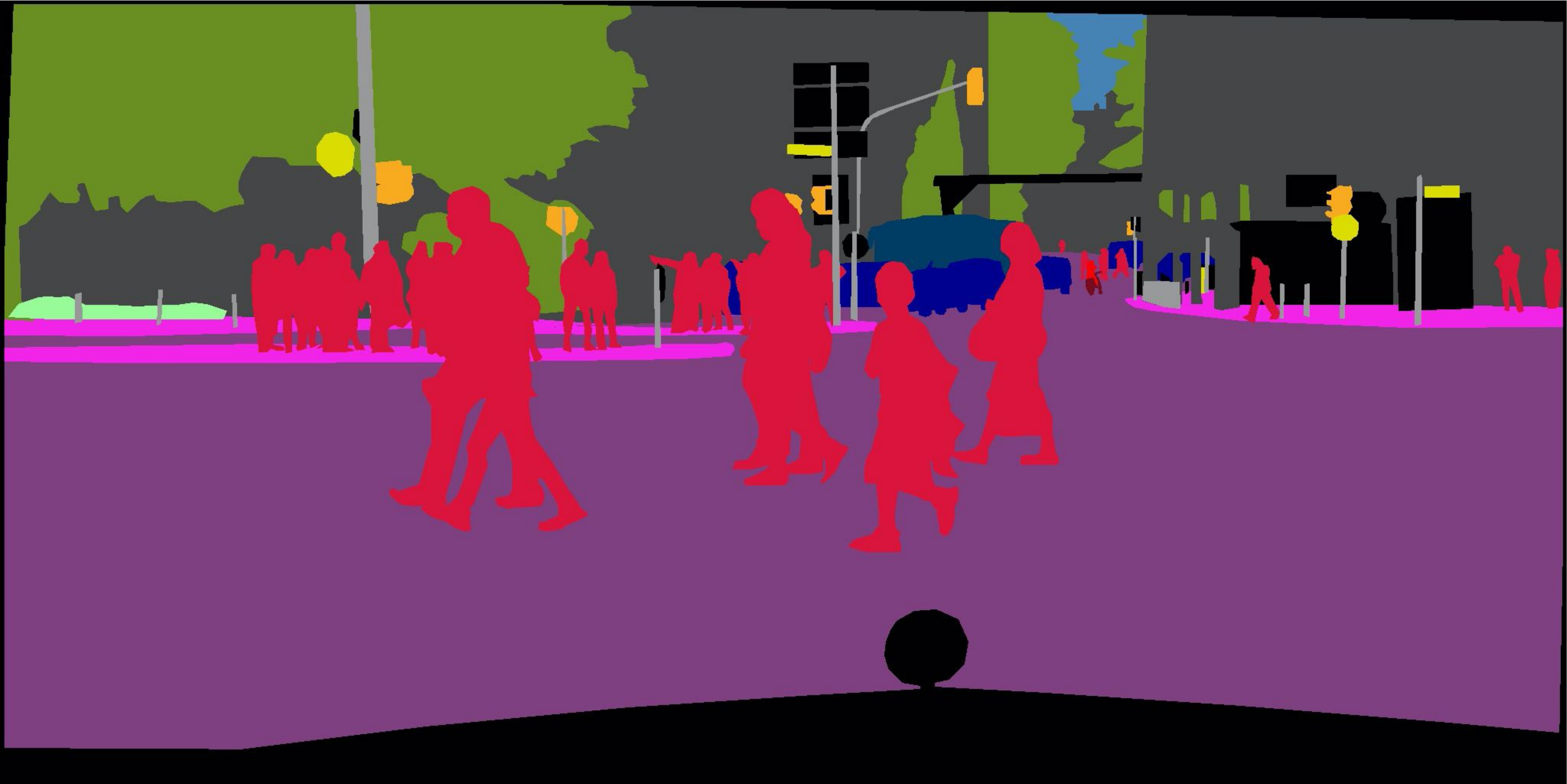}} & 
			\subfloat[Prediction without adaptation]{\includegraphics[width=0.28\linewidth]{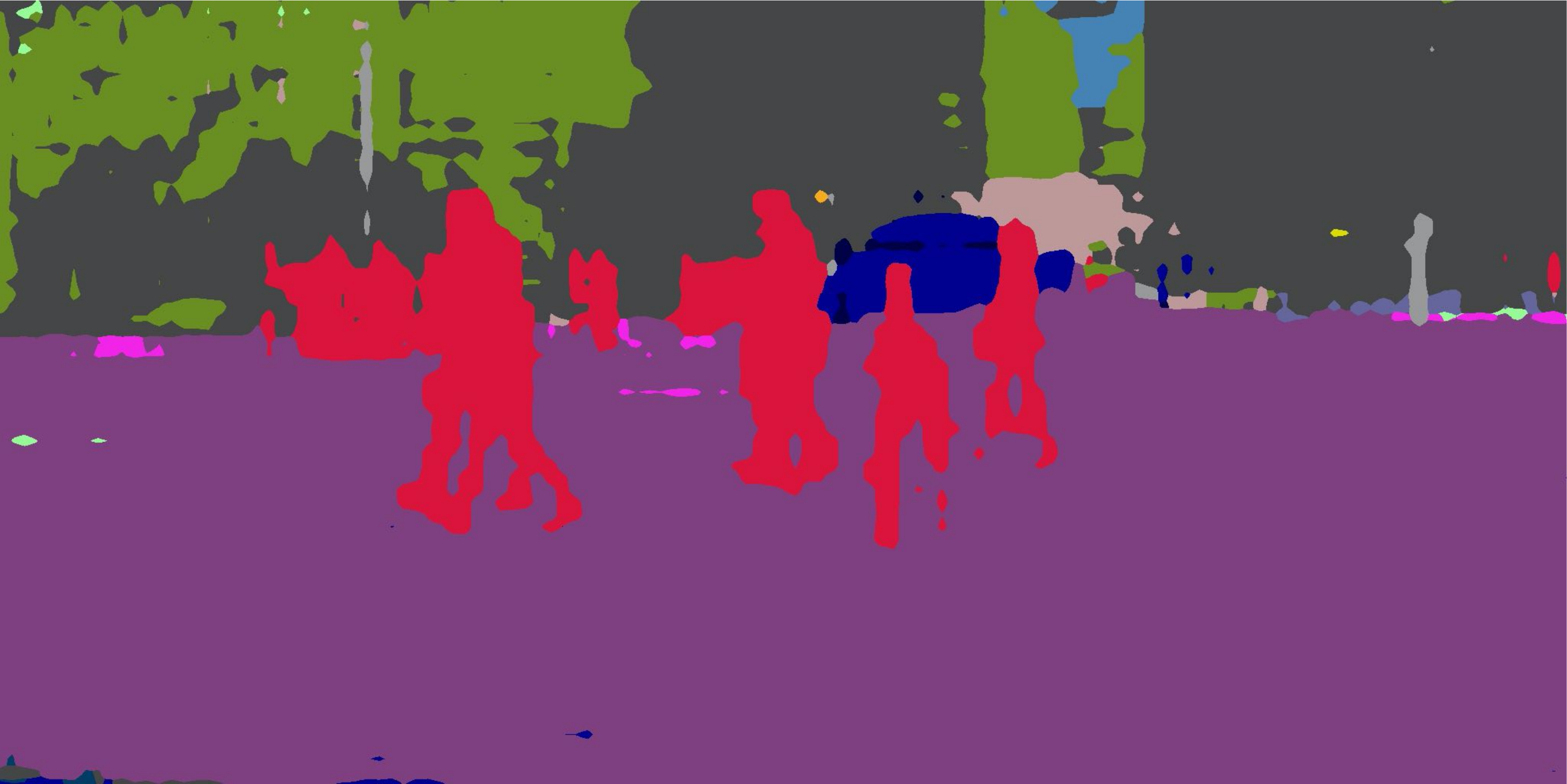}} 
			\\

			\subfloat[ADVENT~\cite{vu2018advent}]{\includegraphics[width=0.28\linewidth]{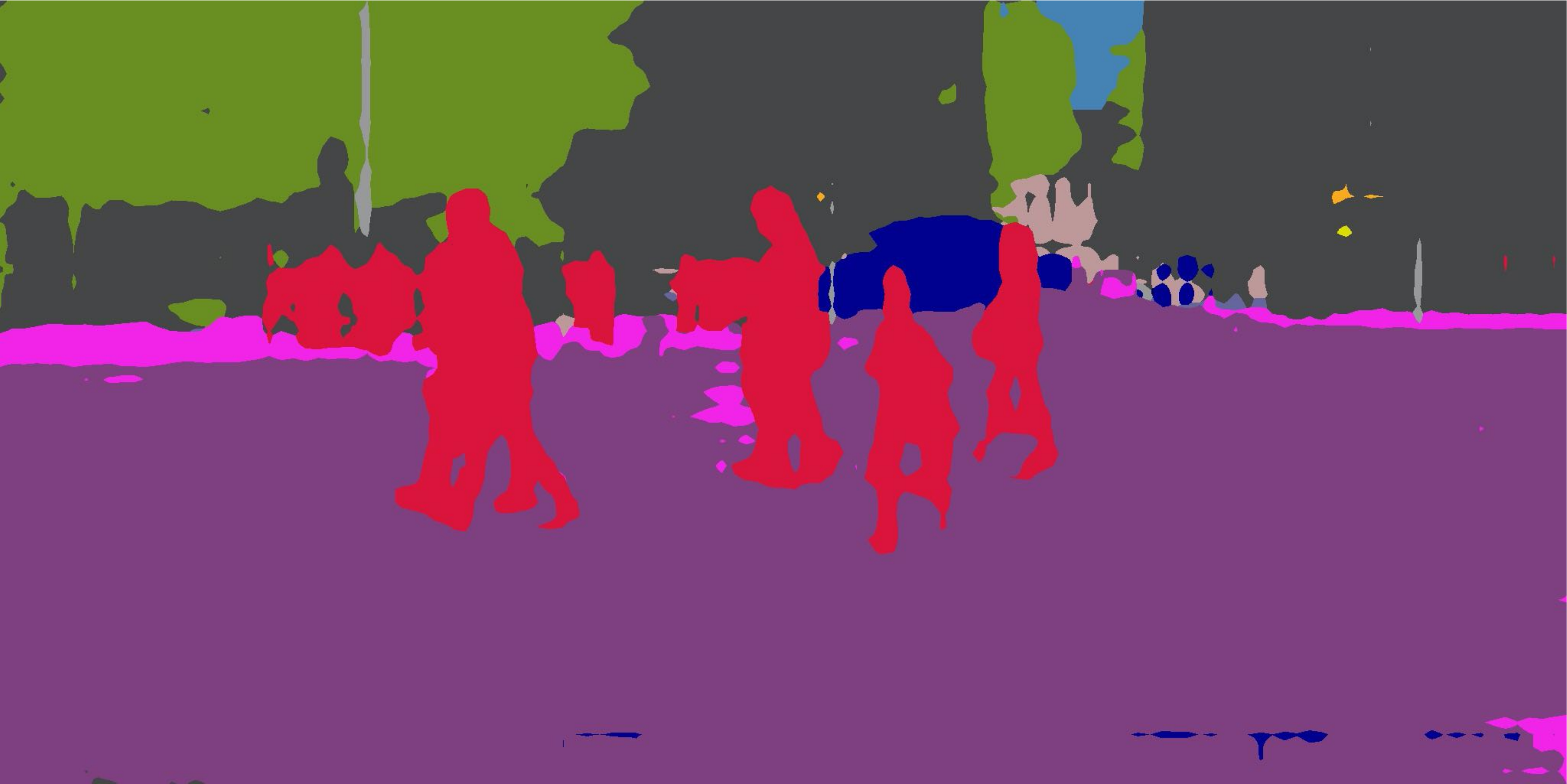}} & 
			\subfloat[BDL~\cite{li2019bidirectional}]{\includegraphics[width=0.28\linewidth]{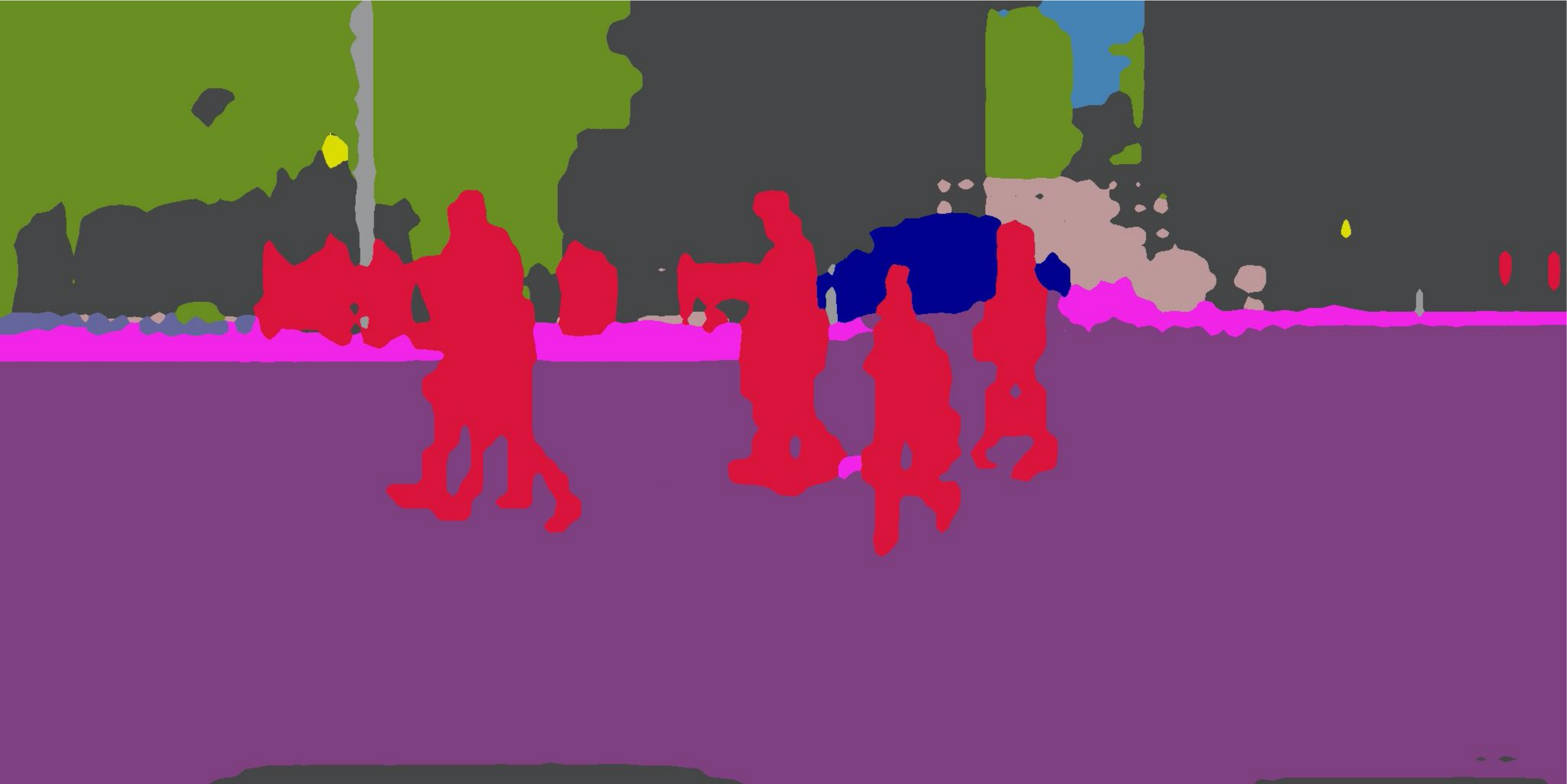}} & 
			\subfloat[CLAN~\cite{luo2019taking}]{\includegraphics[width=0.28\linewidth]{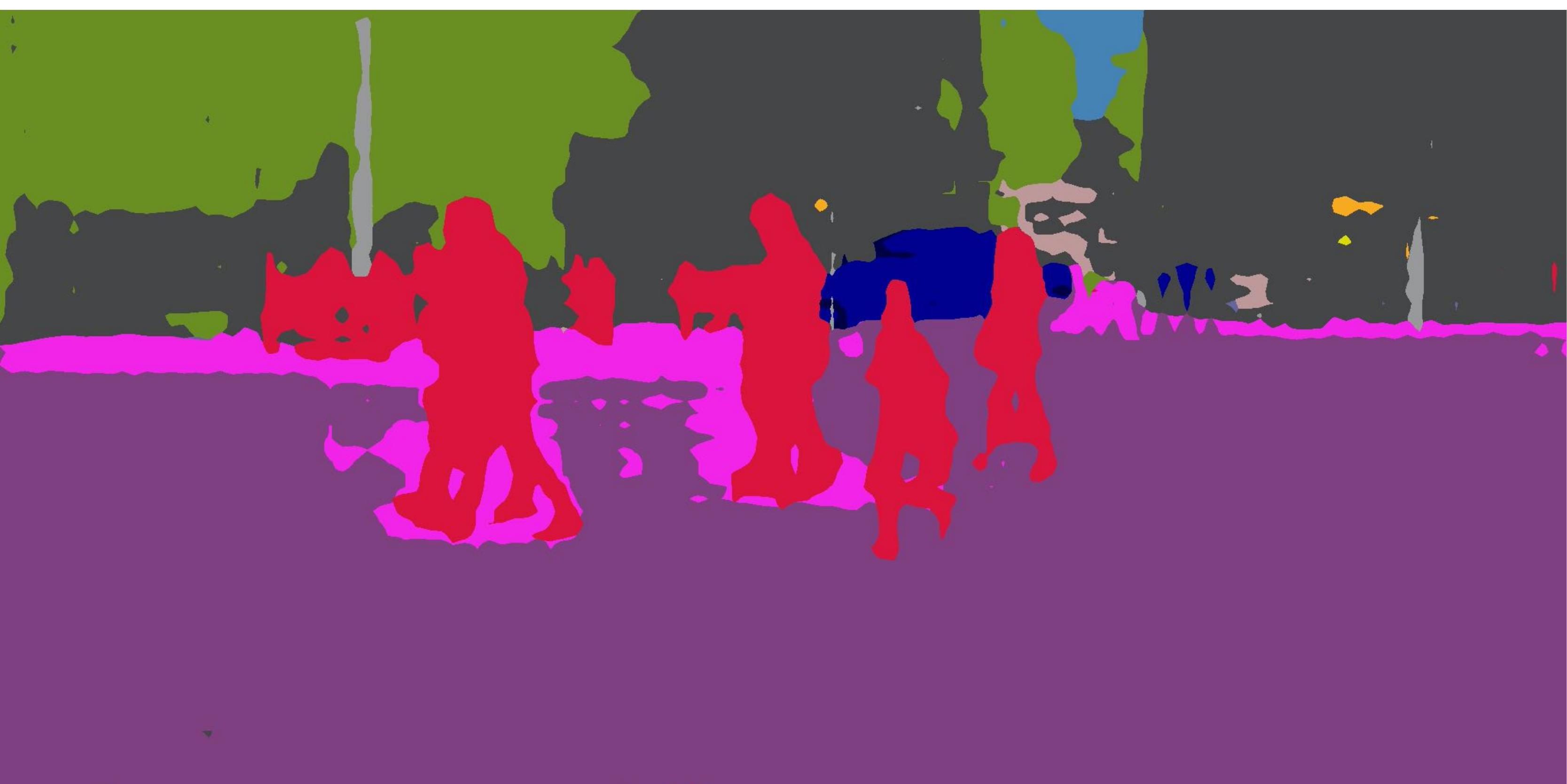}} 
			\\

			\subfloat[SIM~\cite{wang2020differential}]{\includegraphics[width=0.28\linewidth]{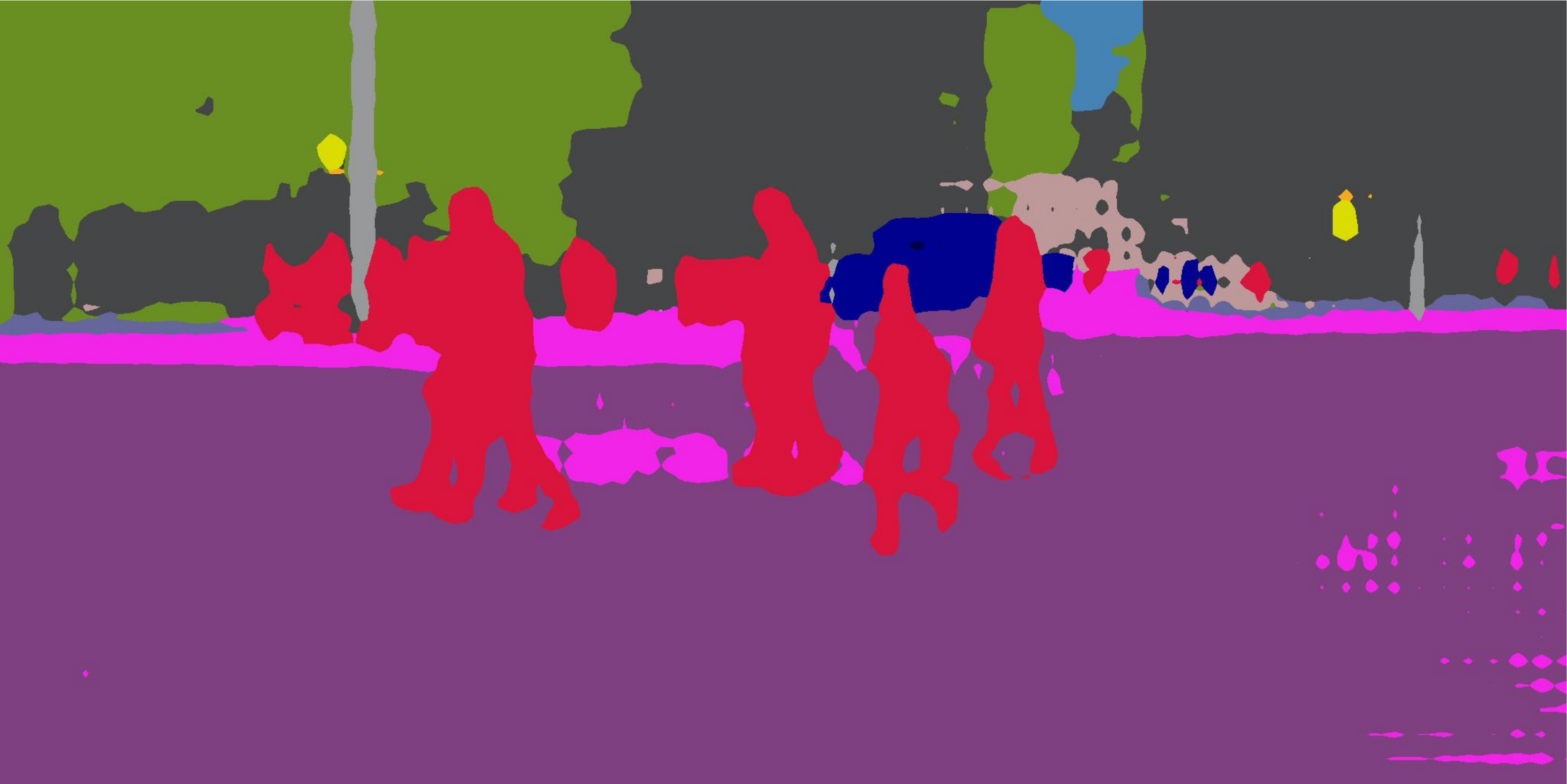}} & 
			\subfloat[FADA~\cite{wang2020classes}]{\includegraphics[width=0.28\linewidth]{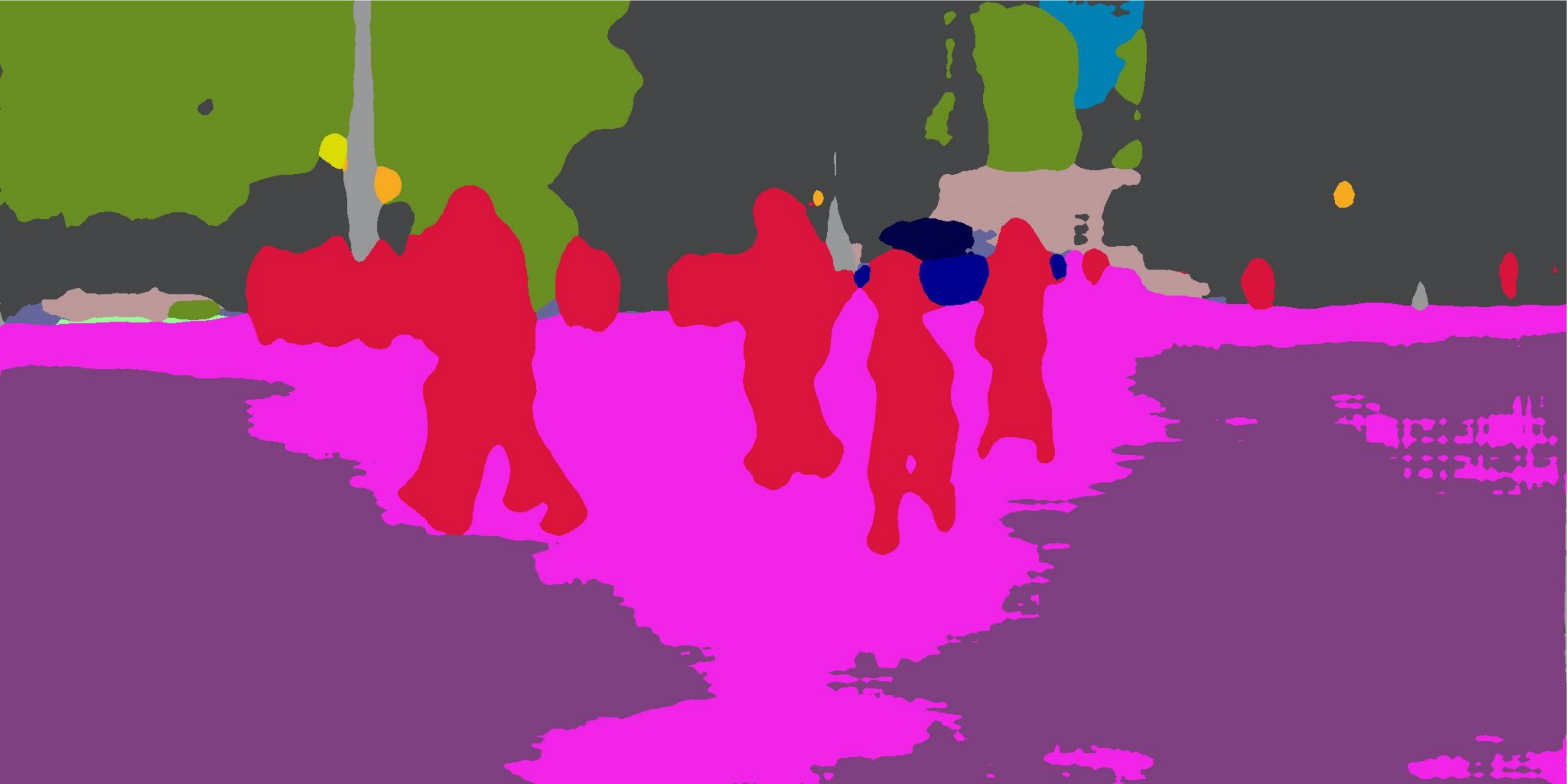}} & 
			\subfloat[FDA-MBT~\cite{yang2020fda}]{\includegraphics[width=0.28\linewidth]{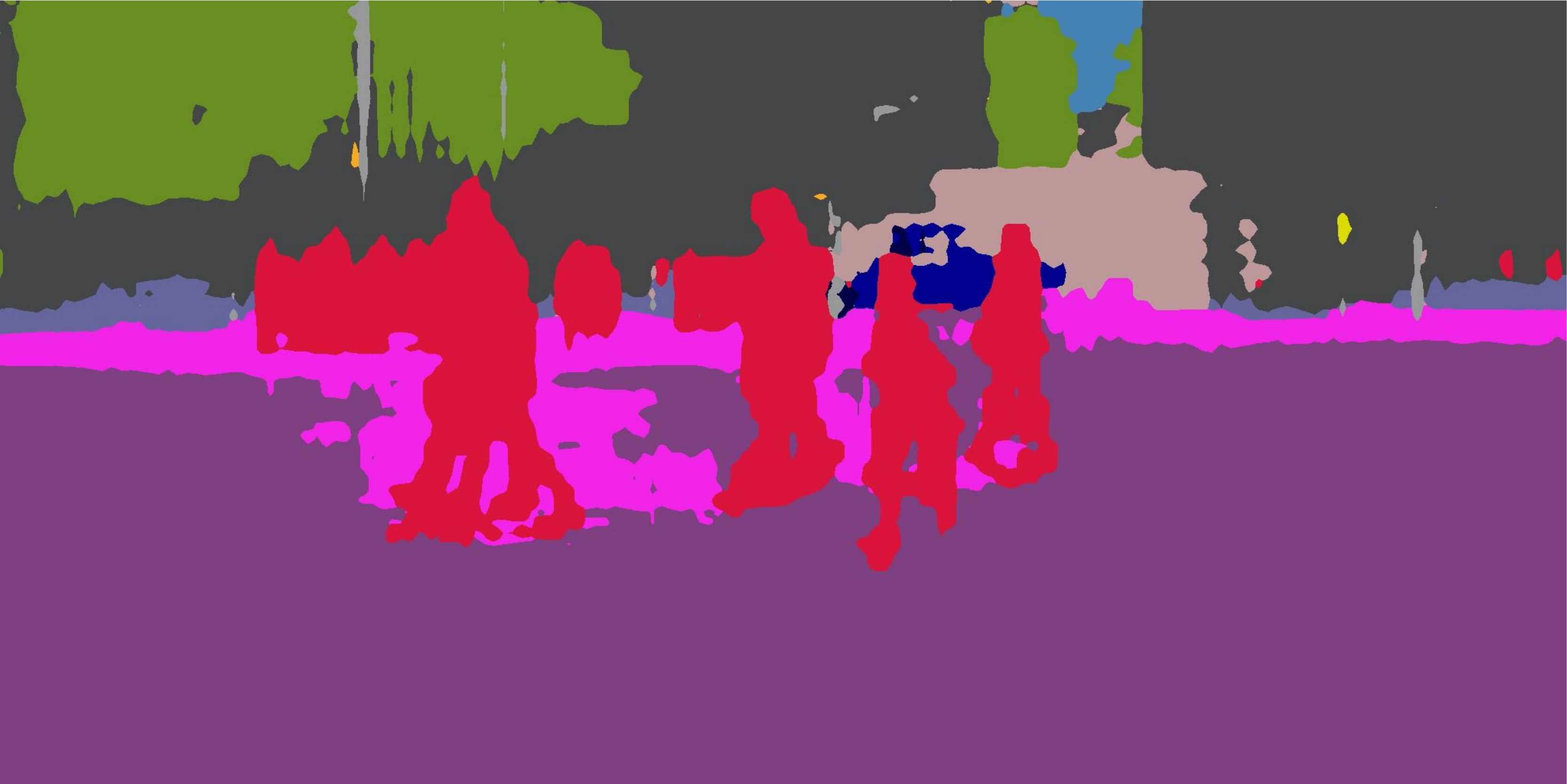}}  
			\\
 
			\subfloat[Kim et al.~\cite{kim2020learning}]{\includegraphics[width=0.28\linewidth]{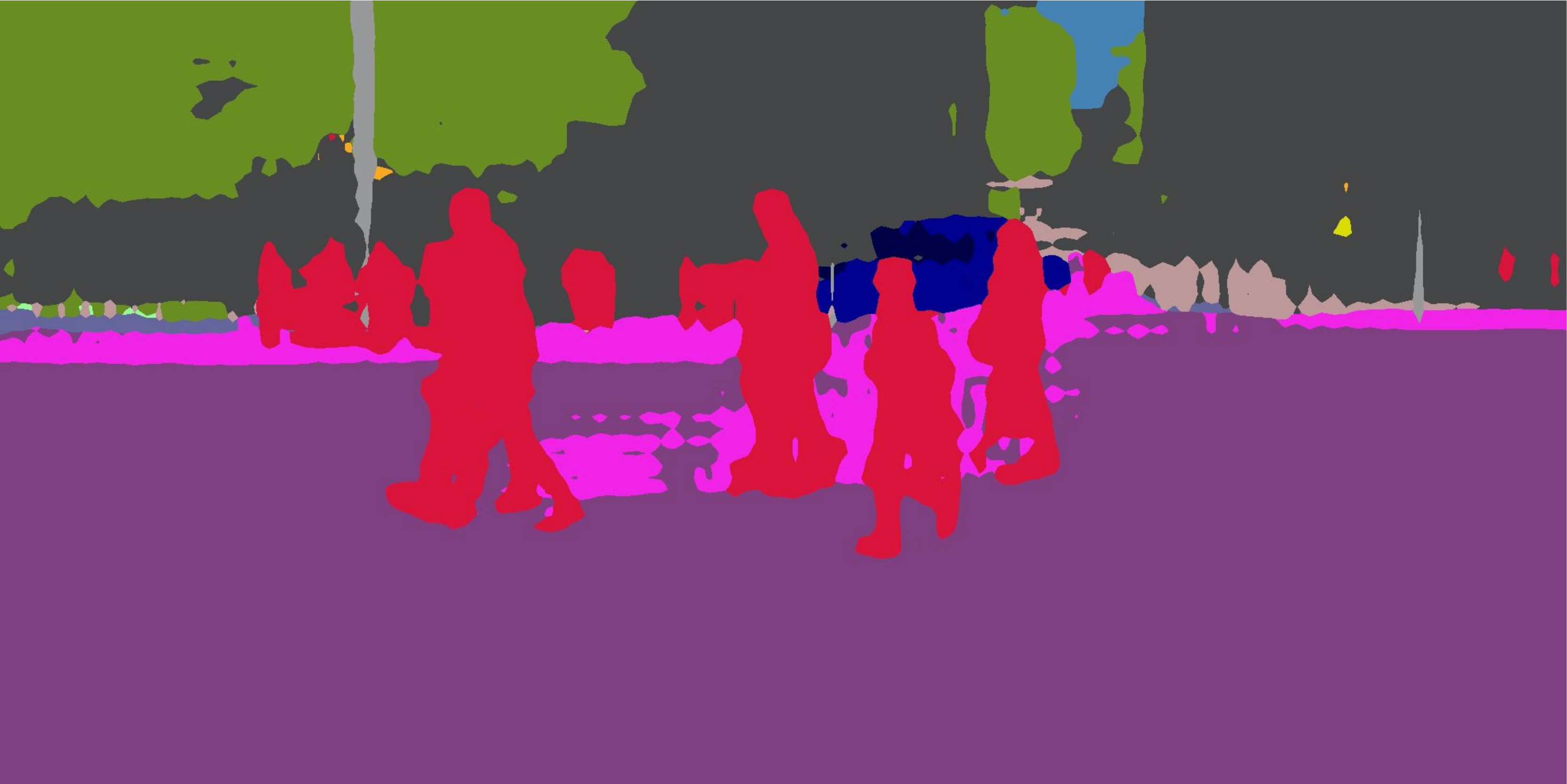}} & 
			\subfloat[DPL]{\includegraphics[width=0.28\linewidth]{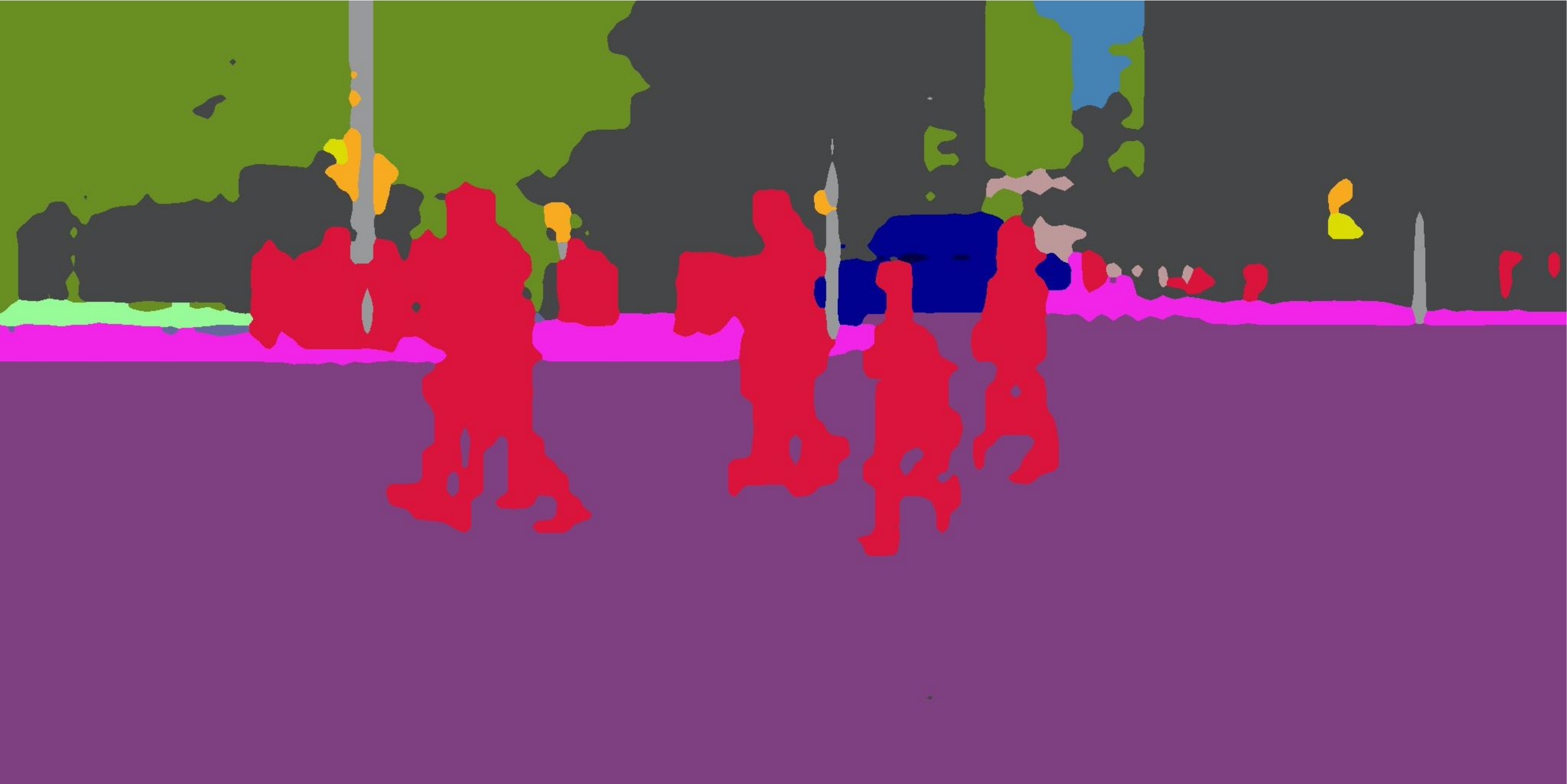}} & 
			\subfloat[DPL-Dual]{\includegraphics[width=0.28\linewidth]{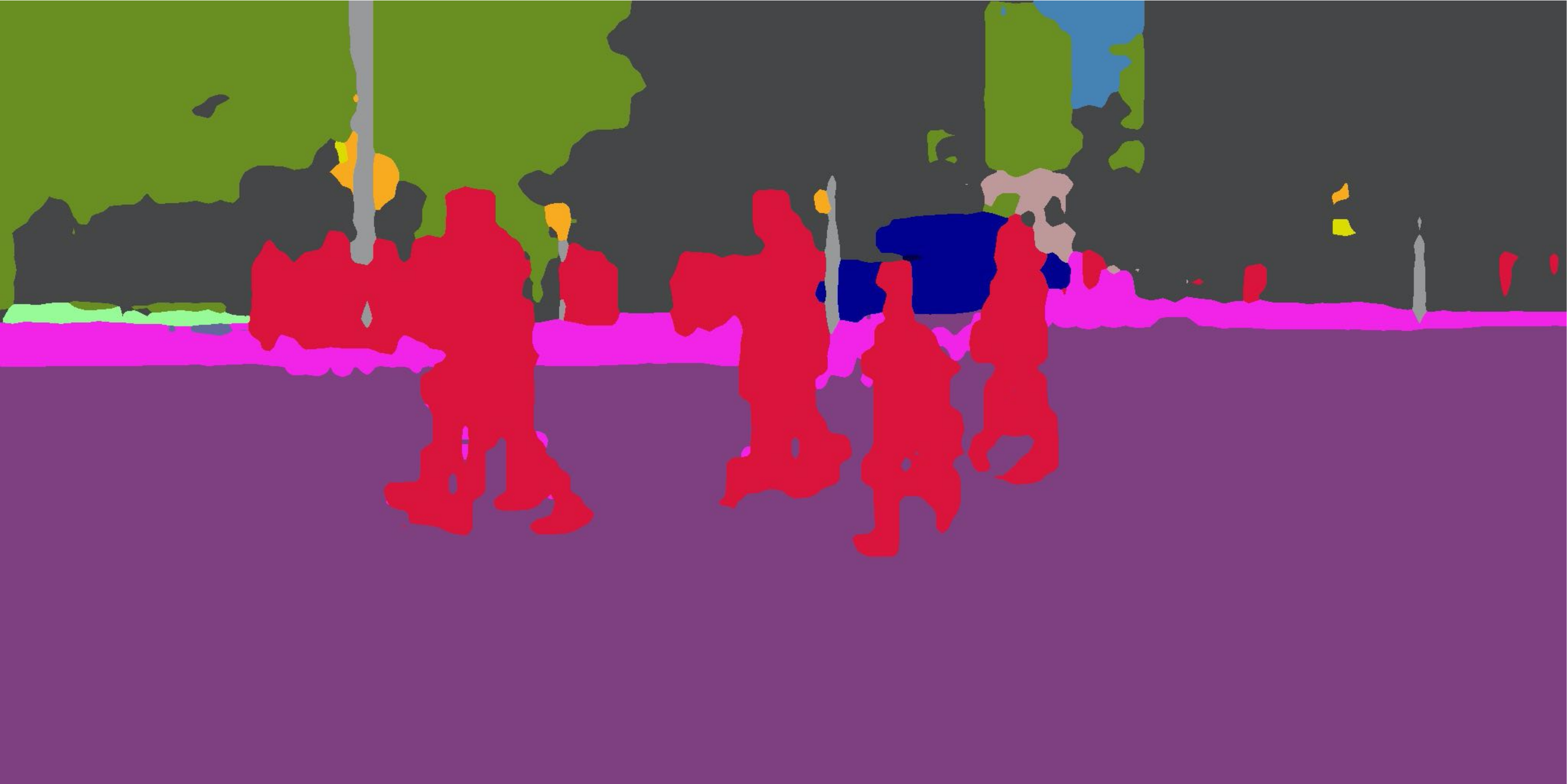}} \\
            \vspace{0.5cm}
	\end{tabular}
	\caption{Qualitative comparison of different methods on GTA5$\rightarrow$Cityscapes scenario.}
	\label{fig:comp2}
 \vspace{3cm}
\end{figure*}

{\noindent \textbf{Qualitative Comparison of Different Domain Adaption Methods.}}\hspace{3pt}
In Figure \ref{fig:comp1} and Figure \ref{fig:comp2}, we visualize segmentation results of different methods for qualitative comparisons. We can observe that DPL and DPL-Dual generate most visually pleasurable results which are similar to the ground truth among all methods.

\end{document}